\documentclass[10pt,journal,compsoc]{IEEEtran}

\usepackage{cite}
\usepackage{times}
\usepackage{epsfig}
\usepackage{graphicx}
\usepackage{amsmath}
\usepackage{amssymb}
\usepackage{multirow}
\usepackage{amsthm}
\usepackage{subfigure}
\usepackage{color}
\usepackage{csquotes}
\usepackage{soul}
\usepackage{enumerate}
\usepackage[normalem]{ulem}
\usepackage{algorithm}
\usepackage{algorithmicx}
\usepackage{algpseudocode}
\usepackage{booktabs}
\usepackage{array}
\usepackage{rotating}
\usepackage{arydshln}
\usepackage{ragged2e}
\usepackage{hyperref}
\hypersetup{breaklinks=true,colorlinks}

\newcommand\eg{\emph{e.g.}} 
\newcommand\ie{\emph{i.e.}} 
 
\newcommand\etc{\emph{etc.}}

\begin{document}
%
\title{Dense Uncertainty Estimation
}
%
%
%
%

       
\author{Jing~Zhang,
       Yuchao~Dai, 
       Mochu Xiang,
       Deng-Ping~Fan,
       Peyman Moghadam \\
       Mingyi He,
       Christian Walder, 
       Kaihao Zhang,
       Mehrtash Harandi
       and
       Nick~Barnes
\IEEEcompsocitemizethanks{\IEEEcompsocthanksitem Jing Zhang, Kaihao Zhang and Nick Barnes are with School of Computing, the Australian National University. (Email: zjnwpu@gmail.com, kaihao.zhang@anu.edu.au, nick.barnes@anu.edu.au)
\IEEEcompsocthanksitem Yuchao Dai, Mochu Xiang and Mingyi He are with School of Electronics and Information, Northwestern Polytechnical University, China. (Email: daiyuchao@gmail.com, xiangmochu@mail.nwpu.edu.cn, myhe@nwpu.edu.cn)
\IEEEcompsocthanksitem Deng-Ping Fan is with the CS, Nankai University, China. (Email: dengpfan@gmail.com)
\IEEEcompsocthanksitem Peyman Moghadam and Christian Walder are with CSIRO, Australia. (Email: peyman.moghadam@data61.csiro.au, christian.walder@data61.csiro.au)
\IEEEcompsocthanksitem Mehrtash Harandi is with Monash University, Australia. (Email: mehrtash.harandi@monash.edu)


}
}

%
%

\markboth{Journal of \LaTeX\ Class Files,~Vol.~14, No.~8, August~2015}%
{Shell \MakeLowercase{\textit{et al.}}: Bare Demo of IEEEtran.cls for Computer Society Journals}
%



\IEEEtitleabstractindextext{%
\begin{abstract}
\justifying
Deep neural networks
can be roughly divided into deterministic neural networks
and stochastic neural networks.
The former is usually trained to achieve
a mapping from input space to
output space via maximum likelihood estimation for the weights, which leads to deterministic predictions during testing. In this way, a specific weights set is estimated while ignoring any uncertainty that may occur
in the proper weight space. The latter introduces randomness into the framework, either by assuming a prior distribution over model parameters (\ie~Bayesian Neural Networks) or including latent variables (\ie~generative models) to explore the contribution of latent variables for model predictions, leading to stochastic predictions during testing. Different from the
former that achieves point estimation, the latter aims to estimate the
prediction distribution,
making it possible to estimate uncertainty, representing model ignorance about its predictions.
We claim that
conventional deterministic neural network based dense prediction tasks are prone to overfitting, leading to over-confident predictions, which is undesirable for decision making.
In this paper, we investigate
stochastic neural networks
and uncertainty estimation techniques to achieve both accurate deterministic prediction and reliable uncertainty estimation. Specifically, we work on two types of uncertainty estimations solutions, namely ensemble based methods and generative model based methods, and explain their pros and cons while using them in fully/semi/weakly-supervised framework. Due to the close connection between uncertainty estimation and model calibration, we also introduce how uncertainty estimation can be used for deep model calibration to achieve well-calibrated models, namely dense model calibration.
Code and data are available at \url{https://github.com/JingZhang617/UncertaintyEstimation}.

\end{abstract}

}

\maketitle

\IEEEdisplaynontitleabstractindextext

\IEEEpeerreviewmaketitle

\IEEEraisesectionheading{\section{Introduction}\label{sec:introduction}}

%
%
%
%

\IEEEPARstart{D} {eep} models are prone to becoming over-confident \cite{on_calibration}, where \cite{on_calibration} explains that a deep neural network is capable of fitting any type
of noise.
Many different regularizations \cite{Vapnik1998} have been proposed to prevent neural networks from overfitting, such as L1 or L2 regularization \cite{bn_normalization}, early stopping \cite{early_stopping2001} and dropout \cite{srivastava2015highway}. L1 or L2 regularization \cite{bn_normalization} try to add constraints to the weights set, thus to limits weights' flexibility.
The early stopping technique \cite{early_stopping2001} is based on the observation that the model we get at the end is not the best we could have possibly created. To this end, we may obtain a
proper model before the model starts overfitting. Dropout \cite{srivastava2015highway} is another
popular method of regularization. The main idea behind it is to assign a probability (dropout ratio) to each unit of network of being temporarily ignored in the learning stage. Although those solutions have
proven effective in various tasks in preventing the
model from being over-confident, there exists no explicit output from those models to explain the degree of model calibration \cite{on_calibration}, or how much the model is regularized before we resort to the ground truth of testing sets.


Given the training dataset $D=\{x_i,y_i\}_{i=1}^N$, the objective of machine learning methods is to minimize the expected loss function, or in practice the empirical loss function (also known as empirical risk minimization \cite{empirical_risk}) as:
\begin{equation}
\begin{aligned}
\min_\theta &\mathbb{E}_{x,y}[\mathcal{L}(f(x;\theta),y)]=\int\mathcal{L}(f(x;\theta),y)d p(x,y)\\
&\approx\frac{1}{N}\sum_{i=1}^N\mathcal{L}(f(x_i;\theta),y_i), \quad (x_i,y_i)\sim p(x,y),
\end{aligned}
\label{uncertainty_formulation}
\end{equation}
where $x,y$ are input and output variables from the sampled training dataset $D$, $\theta$ is the learned model parameter set, $p(x,y)$ is the joint data distribution, $(x_i,y_i)$ represents a
sampled pair from the joint data distribution $p(x,y)$, and $\mathcal{L}(.,.)$ is the loss function. Under the maximum likelihood framework, we maximize the likelihood to achieve the minimized $\mathcal{L}(.,.)$. Given training dataset $D$, the predictive distribution $p(y|x)$ is defined as:
\begin{equation}
    \label{predictive_distribution}
    p(y|x) = \int p(y|x,\theta)p(\theta)d\theta,
\end{equation}
where $p(y|x,\theta)$ is the likelihood, $p(\theta)$ is the approximate posterior for model parameters $\theta$.
For a regression task, $p(y|x,\theta)$ is usually defined as Gaussian likelihood, which is $p(y|x,\theta)=\mathcal{N}(f(x,\theta),\Sigma))$, where $f(x,\theta)$ is the model output, $\Sigma$ is the diagonal covariance matrix, indicating the inherent noise within the label. For a classification task, $p(y|x,\theta)$ is defined as Softmax likelihood, leading to $p(y|x,\theta)=Softmax(f(x,\theta)/exp(\sigma^2))$ \cite{kendall2017uncertainties}, where $\sigma^2$ is the noise level, and $exp(\sigma^2)$ serves as the temperature as in model calibration \cite{on_calibration}, where we have $exp(\sigma^2)=1$ for confident predictions and $exp(\sigma^2)>1$ for uncertain
predictions. 

Eq.~\ref{predictive_distribution} shows that the randomness or uncertainty of $y$ comes from both inherent noise $\Sigma$ (or $\sigma^2$ for classification) and $p(\theta)$, which is an approximation of the true posterior distribution of model parameters $p(\theta|D)$.
\cite{KIUREGHIAN2009105,kendall2017uncertainties} define the former as aleatoric uncertainty (or data uncertainty) and the latter as epistemic uncertainty (or model uncertainty), where \enquote{uncertainty} is a term to explain the randomness of model predictions. Aleatoric uncertainty comes from the intrinsic randomness of the task \cite{KIUREGHIAN2009105}, which cannot not be reduced, and
is usually independent of the choice of the learner or the learned model. Epistemic uncertainty, on the other hand, comes from the limited knowledge about which model generated the provided dataset. In this case, the aleatoric uncertainty cannot be reduced with a
larger training dataset, while the epistemic uncertainty can be explained away with a
larger and more
diverse training dataset, leading to a comprehensive understanding about the model that generated our dataset.
The goal of uncertainty estimation \cite{uncertainty_survey_jakob,Smith2018UnderstandingMO, malinin2021uncertainty_autoregressive,predictive_prior,dbu-robustness,Uncertainty_Estimation_Deterministic,Carvalho_2020_CVPR,Postels_2019_ICCV} is then to produce a measure of confidence for model predictions, either by learning from the data directly \cite{simple_scalable_uncertainty,kendall2017uncertainties},
or by acquiring from a diverse set of predictions \cite{simple_scalable_uncertainty,Gal2016Dropout}.

A systematic way to deal with uncertainty is via Bayesian statistics \cite{Bayesian_Uncertainty,bayesian_conv, generalized_bnn, deup_arxiv,canyoutrust,modern_bnn,kendall2017uncertainties,uncertainty_decomposition}. Bayesian Neural Networks
aim to learn a distribution over each of the network parameters
by placing a prior probability distribution over network weights. Calculating the exact Bayesian posterior is computational intractable. More effort has been put into developing approximations of Bayesian Neural Network that can work in practice. In the following, we will introduce the each type of uncertainty and induce their computation details within both Bayesian Neural Networks and non-Bayesian Neural Networks.

\subsection{Aleatoric Uncertainty}
\label{sec_aleatoric_uncertainty}
As discussed above, the aleatoric uncertainty \cite{kendall2017uncertainties,random_forest_uncertainty,uncertainty_decomposition} captures the observation noise $\Sigma$ inherent in the task,
which can be categorised as image independent uncertainty
(homoscedastic uncertainty that does not vary with the input)
and image conditional uncertainty
hetroscedastic uncertainty that various with the input)
\cite{kendall2017uncertainties}. For the former, the noise is totally random, and for the latter, it is assumed that the noise is related to the context of the input image $x$.
As we want to perform variational inference over the noise level $\Sigma$, non-Bayesian Neural Networks can be used with a deterministic model parameter set $\theta$. In this following, we introduce the aleatoric uncertainty for both regression tasks and classification tasks. For all the presentation below, we define $x$ and $y$ as the input and output pair, and $\theta$ the model parameters.

\noindent\textbf{Aleatoric uncertainty for regression tasks:} For a regression task, the observed label $y$ is assumed to be corrupted with the observational noise $\epsilon$:
\begin{equation}
\label{regression_noisy_label}
    y=f(x,\theta)+\epsilon,
\end{equation}
where the observational noise $\epsilon$ can be either image-independent or image-dependent. In the following, we only discuss the image dependent noise, or more specifically,
the pixel-level noise $\epsilon=\epsilon(x)\sim\mathcal{N}(0,\Sigma)$, where $\Sigma$ is the diagonal covariance matrix representing the inherent noise.

With the above noise corruption process, we define the Gaussian likelihood for the regression task as $p(y|x,\theta)=\mathcal{N}(f(x,\theta),\Sigma)$, where $\Sigma=diag(\sigma^2(x))$. Given the Gaussian likelihood:
\begin{equation}
\label{regression_gaussian_likelihood}
p(y|x,\theta) = \frac{1}{\sqrt{2\pi\sigma^2(x)}}exp(\frac{-[y-f(x,\theta)]^2}{2\sigma^2(x)}),
\end{equation}
the log-likelihood can be obtained as:
\begin{equation}
\label{regression_gaussian_log_likelihood}
\log(p(y|x,\theta)) = -\frac{[y-f(x,\theta)]^2}{2\sigma^2(x)}-\frac{1}{2}\log(\sigma^2(x))-\frac{1}{2}\log(2\pi).
\end{equation}

Then the loss function with noise corruption model as in Eq.~\ref{regression_noisy_label} is achieved as \cite{nicta_uncertainty,kendall2017uncertainties}:
\begin{equation}
    \label{loss_aleatoric_uncertainty}
    \mathcal{L}(\theta)=\frac{1}{N}\sum_{i=1}^N(\frac{1}{2\sigma^2(x_i)}\mathcal{L}_2+\frac{1}{2}log(\sigma^2(x_i))),
\end{equation}
where the constant $\frac{1}{2}\log(2\pi)$ is removed. $\mathcal{L}_2$ is L2 loss, and $\sigma(x)$ is a noise estimation module, representing the pixel-level aleatoric uncertainty. According to Eq.~\ref{regression_noisy_label}, we need to estimate both mean prediction $f(x,\theta)$ and variance $\sigma^2(x)$ indicating the noise level, representing the aleatoric uncertainty. \cite{nicta_uncertainty,kendall2017uncertainties} achieve this with a multi-head network, where the head of the network is split to produce both task related prediction $f(x,\theta)$ and the aleatoric uncertainty $\sigma^2(x)$.

To train with the
loss function in Eq.~\ref{loss_aleatoric_uncertainty}, we only need supervision $y$ for the $\mathcal{L}_2$ term, and the noise estimation module does not need any supervision.
In practice, usually the log variance is predicted as $s=\log\sigma^2(x)$ to achieve stable training \cite{kendall2017uncertainties}, leading to:
\begin{equation}
    \label{loss_aleatoric_uncertainty_log}
    \mathcal{L}(\theta)=\frac{1}{N}\sum_{i=1}^N(\frac{1}{2}exp(-s_i)\mathcal{L}_{ce}+\frac{1}{2}s_i),
\end{equation}
where the uncertainty representation $s$ serves as both weight for the original loss function $\mathcal{L}_2$ and regularizer to achieve accurate prediction and reliable uncertainty estimation. 

\noindent\textbf{Aleatoric uncertainty for classification tasks:} For a classification task with Softmax likelihood, noise-corruption is achieved by squashing the model prediction with uncertainty related variable \cite{Kendall_2018_CVPR}:
\begin{equation}
    \label{classification_likelihood}
    p(y|x,\theta)=Softmax(\frac{f(x,\theta)}{\exp(\sigma^2(x))}),
\end{equation}
where $\exp(\sigma^2(x))$ can also be explained as temperature as in \cite{on_calibration}.

Let's define $T=\exp(\sigma^2(x))$, the log-likelihood can then be obtained as:
\begin{equation}
\label{classification_log_likelihood}
    \log(p(y|x,\theta))=\frac{1}{T}f_\theta(x)-\log(\exp(\frac{1}{T}f(x,\theta))+1).
\end{equation}

We also have:
\begin{equation}
\label{loss_approximation1}
    (\exp(f(x,\theta))+1)^\frac{1}{T}\approx\frac{1}{T}(\exp(\frac{1}{T}f(x,\theta))+1),
\end{equation}
which becomes equal when $T=\exp(\sigma^2)\rightarrow 1$, indicating the uncertainty is not very high, which is consistent with our basic understanding about aleatoric uncertainty, which should be not too large and not too small.

With the approximation of $\exp(\frac{1}{T}f(x,\theta))+1$ as in Eq.~\ref{loss_approximation1}, we take it back to Eq.~\ref{classification_log_likelihood} and obtain the approximated likelihood as:
\begin{equation}
\label{approximated_classification_log_likelihood}
    \log(p(y|x,\theta))=-\log T - \frac{1}{T}\mathcal{L}_{ce},
\end{equation}
where $\mathcal{L}_{ce}$ is the cross-entropy loss. The corresponding loss function is then:
\begin{equation}
\label{approximated_classification_log_likelihood_loss}
    \mathcal{L}(\theta)=\frac{1}{N}\sum_{i=1}^N(\frac{1}{T_i}\mathcal{L}_{ce} + \log T_i),
\end{equation}

\noindent\textbf{Aleatoric uncertainty within BNN:}
The aleatoric uncertainty $\sigma^2(x)$ is induced from a non-Bayesian Neural Network (BNN) with fixed model parameters.
\cite{uncertainty_decomposition} defines the aleatoric uncertainty within the BNN as mean entropy or the Bayesian Active Learning by Disagreement Objective (BALD) \cite{bald_2011} of model prediction:
\begin{equation}
    \label{aleatoric_bnn_entropy}
    U_a=\mathbb{E}_{p(\theta|D)}[H(y|x,\theta)],
\end{equation}
where $H(.)$ is the entropy. Eq.~\ref{aleatoric_bnn_entropy} integrates out the model parameters $\theta$, which only captures the prediction randomness caused by the inherent noise. Note that the aleatoric uncertainty in Eq.~\ref{loss_aleatoric_uncertainty_log} or Eq.~\ref{approximated_classification_log_likelihood_loss} is obtained via a dual-head network, which can
produce model prediction and uncertainty simultaneously, while aleatoric uncertainty in Eq.~\ref{aleatoric_bnn_entropy} is within the BNN, which can be computed directly from the prediction itself by integrating out the contribution of model parameters $\theta$. Alternatively, \cite{deup_arxiv} define aleatoric uncertainty as the uncertainty based on the best model given the training dataset $D$. In this way, the aleatoric uncertainty with the best model can be obtained via:
\begin{equation}
    U_a = H(y|x,\theta^*),
\end{equation}
where $\theta^*$ is the model parameter that yields the best performance (or the smallest loss).

\subsection{Epistemic Uncertainty}
Aleatoric uncertainty can be directly estimated through the auxiliary head $\sigma^2(x)$ or by the mean conditional prediction entropy as in Eq.~\ref{aleatoric_bnn_entropy}. Epistemic uncertainty, on the contrary, is usually estimated as the residual of predictive uncertainty and aleatoric uncertainty. The predictive uncertainty is defined as the total uncertainty representing prediction randomness from both the inherent noise and the model weights space $\theta\sim p(\theta|D)$. Specifically, the predictive uncertainty is usually defined as: $U_p=H(y|x)$, which is obtained as the entropy of mean model prediction with model parameters sampled from its posterior distribution. With both predictive uncertainty $U_p$ and aleatoric uncertainty $U_a$, the epistemic uncertainty is then defined as:
\begin{equation}
\begin{aligned}
\label{epistemic_uncertainty_entropy}
    U_e&=U_p-U_a\\
    &=H(y|x)-\mathbb{E}_{p(\theta|D)}[H(y|x,\theta)],
\end{aligned}
\end{equation}
where $H(y|x)-\mathbb{E}_{p(\theta|D)}[H(y|x,\theta)]$ is the mutual information of model prediction and parameters. \cite{uncertainty_decomposition} then defines
the epistemic uncertainty as the mutual information of model prediction and model parameters conditioned on the input image $x$, which is: $U_e=I(y,\theta|x)$.

Eq.~\ref{epistemic_uncertainty_entropy} can be used in both regression tasks
and classification tasks
\cite{uncertainty_decomposition}. \cite{kendall2017uncertainties} defines an alternative solution for epistemic uncertainty estimation within the BNN for regression tasks, which uses variance instead of entropy for uncertainty measure\footnote{Entropy measures the average amount of information contained in a distribution, and is reduced to a function of variance when the distribution is Gaussian} as:
\begin{equation}
    \label{regression_epistemic}
    U_e=\mathbb{E}_{p(\theta|D)}[p(y|x,\theta)^2] - (\mathbb{E}_{p(\theta|D)}[p(y|x,\theta)])^2.
\end{equation}
With epistemic uncertainty $U_e$ in Eq.~\ref{regression_epistemic}, the predictive uncertainty can then be defined as:
\begin{equation}
    \label{regression_predictive}
    U_p=\mathbb{E}_{p(\theta|D)}[p(y|x,\theta)^2] - (\mathbb{E}_{p(\theta|D)}[p(y|x,\theta)])^2 + \mathbb{E}_{p(\theta|D)}[\sigma^2(x)],
\end{equation}
where $\sigma^2(x)$ is obtained via the dual-head structure \cite{nicta_uncertainty,kendall2017uncertainties}.

\section{Uncertainty Approximation}
Before the deep learning era, the dual-head based aleatoric uncertainty \cite{nicta_uncertainty}, or disagreement based epistemic uncertainty via Gaussian processes \cite{gaussian_process_regression96} could
be used directly for
uncertainty modeling.
Most of the recent techniques for epistemic uncertainty estimation \cite{deup_arxiv,canyoutrust,modern_bnn,kendall2017uncertainties,uncertainty_decomposition} are based on Bayesian Neural Networks (BNN), where a pre-defined prior distribution $p(\theta|D)$ is set as a
constraint
to regularize the distribution of the model parameters and achieve stochastic prediction. In this way, the epistemic uncertainty is obtained via the posterior distribution of network weights, which captures the set of plausible model parameters given the training dataset $D$. 
According to the
Bayesian rule, the posterior over model parameters $p(\theta|x,y)$ (or $p(\theta|D)$) can be achieved by $p(\theta|x,y)=p(y|x,\theta)p(\theta)/p(y|x)$. As the marginal likelihood $p(y|x)$ cannot be evaluated analytically, $p(\theta|x,y)$ is then not available in closed-form, making it difficult to estimate aleatoric uncertainty, epistemic uncertainty or predictive uncertainty as in Eq.~\ref{aleatoric_bnn_entropy},  Eq.~\ref{epistemic_uncertainty_entropy}, Eq.~\ref{regression_epistemic} and Eq.~\ref{regression_predictive}. In this way, approximations are required to achieve Bayesian posterior inference, including Variational Inference (VI) \cite{Jordan99anintroduction,Graphical_Models_VI} and Markov Chain Monte Carlo (MCMC) \cite{mcmc_sampling}. The former approximate the intractable posterior distribution $p(\theta|D)$ with an auxiliary tractable distribution $p_\phi(\theta)$ by minimizing the Kullback–Leibler divergence between $p_\phi(\theta)$ and $p(\theta|D)$, where $\phi$ is the variational parameters set. The latter is a
sampling based posterior approximation, which draws a correlated sequence of $\theta_t$ from $p(\theta|D)$ (where $t$ indicates the iteration of sampling), and uses it to approximate the posterior predictive distribution as a Monte Carlo average. 

In this paper, we divide the uncertainty estimation techniques by the uncertainty they focus on modeling, either on aleatoric uncertainty or epistemic uncertainty. We then discuss two types of methods, namely ensemble based methods, which construct model ensemble for effective posterior approximation (mainly for epistemic uncertainty estimation), and generative model based methods for inherent noise modeling (specifically for aleatoric uncertainty estimation). We also introduce the Bayesian latent variable model for both types of uncertainty modeling.

\subsection{Ensemble based Posterior Approximation}
Ensemble methods \cite{Gal2016Dropout,emsemble,Durasov21_maskensemble, simple_scalable_uncertainty,ensemble_accurate_uncertainty,reverse_kl,Malinin2020Ensemble,hyperparameter_ensemble} are machine learning techniques that combine several base models in order to produce one optimal predictive model.

\noindent\textbf{MC-dropout:}
Monte Carlo Dropout (MC Dropout) \cite{Gal2016Dropout} has been shown to be
an effective variational inference, where dropout is used before every weight layer during both training and testing stages. \cite{Gal2016Dropout} showed that networks with dropout following every weight layer are equivalent to deep Gaussian Process \cite{deep_gaussian_process} marginalized over its covariance functions. Later, \cite{kendall2017uncertainties} found that the dropout following each layer strategy is too strong a regulariser in practice, which may lead to very slow learning. Instead, they proposed to drop out the deepest half of the encoder and decoder units \cite{kendall2017bayesian}. With the relaxed MC Dropout, the predictive distribution in Eq.~\ref{predictive_distribution} can be approximated using MC integration as:
\begin{equation}
    \label{likelihood_classfication}
    p(y|x)\approx \frac{1}{T}\sum_{t=1}^T p(y_t|x,\theta_t),
\end{equation}
where $T$ indicates the $T$ iterations of sampling from the approximate posterior distribution $q(\theta|\gamma)$ ($\gamma$ represents our ignorance about the model based on the training dataset $D$.), and $\theta_t$ is the model parameter set of the $t^{th}$ iteration of sampling.
With MC-dropout as posterior approximation, the aleatoric uncertainty in Eq.~\ref{aleatoric_bnn_entropy} can be achieved as:
\begin{equation}
    \label{aleatoric_mc_dropout}
    U_a = \frac{1}{T}\sum_{t=1}^T H(y|x,\theta_t).
\end{equation}
Similarly, the predictive uncertainty and epistemic uncertainty in Eq.~\ref{epistemic_uncertainty_entropy} can be obtained as:
\begin{equation}
    \label{predictive_mc_dropout}
    U_p = H(\frac{1}{T}\sum_{t=1}^T p(y|x,\theta_t)),
\end{equation}
and epistemic uncertainty as $U_e = u_p-U_a$. For regression task, the epistemic uncertainty and predictive uncertainty in Eq.~\ref{regression_epistemic} and Eq.~\ref{regression_predictive} can then be re-wrote as:
\begin{equation}
    \label{epistemic_regression_dropout}
    U_e=\frac{1}{T}\sum_{t=1}^T p(y|x,\theta_t)^2 -(\frac{1}{T}\sum_{t=1}^T p(y|x,\theta_t))^2,
\end{equation}
and:
\begin{equation}
    \label{predictive_regression_dropout}
    U_p=\frac{1}{T}\sum_{t=1}^T p(y|x,\theta_t)^2 -(\frac{1}{T}\sum_{t=1}^T p(y|x,\theta_t))^2 + \frac{1}{T}\sum_{t=1}^T \sigma^2(x).
\end{equation}

The main advantage of MC dropout \cite{Gal2016Dropout,kendall2017bayesian} is that it imposes Bernoulli distribution over the network weights, without requiring any additional model parameters.


\noindent\textbf{Deep Ensemble:}
Different from MC-dropout \cite{Gal2016Dropout} that is
based on the same base network parameters. Deep ensemble solutions \cite{simple_scalable_uncertainty, snapshot_ensembles, snapshot_uncertainty} intend to generate multiple models to approximate the posterior distribution of the true model parameters. \cite{simple_scalable_uncertainty} found that random initialization of the network parameters, along with random shuffling of the data points, was sufficient to obtain good performance in practice. Specially, they train $M$ different networks with random initialization and get their final prediction as a mean of above $M$ ensembles.
Differently, \cite{NIPS2016_6501} proposed to train different networks with the same training dataset to account for ensembles. Both above methods need re-train the same \cite{simple_scalable_uncertainty} or different \cite{NIPS2016_6501} networks several times to obtain ensembles. \cite{adaptive_curri} proposed an ensemble generation structure with shared
encoder and several different decoders (the \enquote{Multi-head solution}). In this case, prediction from each decoder represents one ensemble. Compared with training the entire network several times, this strategy is much more convenient. Similar to above idea, pseudo multi-task learning \cite{pseudo_multitask} was proposed to achieve one task in different ways, thus forming a pseudo multi-task learning framework, and prediction of each pseudo task can be treated as one ensemble.

\noindent\textbf{Snapshots Ensemble:}
The success of deep ensemble solutions \cite{simple_scalable_uncertainty} based on
multiple copies of network from
multiple
random initializations of the same network, which is computationally expensive in practice. \cite{snapshot_ensembles} introduce snapshot ensembles, where the multiple predictions are obtained from models of various epochs, leading to a cheap uncertainty estimation solution.
However, although \cite{snapshot_ensembles} is easy to implement with good uncertainty estimation, \cite{snapshot_uncertainty} argued that such techniques tend to introduce biased estimates for instances whose predictions are supposed to be highly confident. Thus they develop an uncertainty estimation algorithm that selectively estimates the uncertainty of highly confident points using earlier snapshots of the trained model inspired by \cite{snapshot_ensembles}. They use one single forward pass
in the testing stage to estimate uncertainty without any other hyper-parameters.


\subsection{Generative Model for Aleatory Modeling}
With an
extra latent variable $z$, the generative models \cite{NIPS2014_5423_gan, VAE1, cvae, ABP_aaai} are capable of generating stochastic predictions, making it possible to evaluate the
uncertainty of model prediction via a generative model based framework \cite{adversarial_distillation,henning_approximating_predicyive,hypergan_icml2019,bhm2019uncertainty}. Given the latent variable $z$, the predictive distribution $p(y|x)$ is defined as:
\begin{equation}
    \label{predictive_distribution_generative_model}
    p(y|x) = \int p(y|x,z,\theta)p(\theta)p(z)d\theta dz,
\end{equation}
where $p(y|x,z,\theta)$ is the likelihood, which is $p(y|x,z,\theta) = \mathcal{N}(f(x,z,\theta),\Sigma)$, similar to the deterministic neural network. For a non-Bayesian Neural Network, the model parameter $\theta$ is estimated
via point estimation by minimizing the regularized negative log-likelihood. In this way, the randomness of model prediction has its origin in both $z$ and $\Sigma$, which captures the aleatoric uncertainty of model prediction.
In this paper, we discuss four generative models \cite{cvae,NIPS2014_5423_gan,ABP_aaai} and apply them for uncertainty estimation.

\noindent\textbf{Conditional Variational Autoencoders (CVAE):} 
A CVAE \cite{cvae} is a conditional directed graph model, which includes three variables, the input $x$ or conditional variable that modulates the prior on the
Gaussian latent variable $z$ and generates the output prediction $y$. Two main modules are included in a conventional CVAE based framework: a generator model $p(y|x,z,\theta)$, which generates prediction $y$ with input $x$ and $z$ as input, and an inference model $p(z|x,y,\theta)$, which infers the latent variable $z$ with input $x$ and output $y$ as input.
Learning a CVAEs framework involves approximation of the true posterior distribution of $z$ with an
inference model $q(z|x,y,\phi)$, where $\phi$ is network parameter set of the inference model.

The parameter sets of CVAEs
(including both generator model parameters $\theta$ and inference model parameters $\phi$) 
can be estimated in Stochastic Variational Bayes (SGVB) \cite{VAE1} framework by maximizing the expected variational lower bound (ELBO) as:
\begin{equation}
    \label{vea_lower_bound}
\begin{aligned}
    L(\theta,\phi;x) &= \mathbb{E}_{q(z|x,y,\phi)}[\log(p(y|x,z,\theta))]\\&-D_{KL}(q(z|x,y,\phi)||p(z|x,\theta)),
\end{aligned}
\end{equation}
where the first term is the expected log-likelihood and the second term measures the information lost using $q(z|x,y),\phi$ to approximate the true posterior distribution of latent variable $z$.

\noindent\textbf{Generative Adversarial Networks (GAN):} Similar to the CVAE based model, two different modules (a generator and a discriminator) play the minimax game in GAN based framework as shown below:
\begin{equation}
\label{gan_loss}
\begin{aligned}
    \underset{G}{min} \, \underset{D}{max} \, V(D,G) &= E_{(x,y)\in p_{data}(x,y)}[\log D(y|x)]\\ &+ E_{z\in q(z)}[\log(1-D(G(x,z))],
\end{aligned}
\end{equation}
where $G$ and $D$ are the generator model and discriminator model respectively, $p_{data}(x,y)$ is the joint distribution of training data, $q(z)$ is the prior distribution of the latent variable $z$, which is usually defined as $q(z) = \mathcal{N}(0,\mathbf{I})$. Specifically, the discriminator is designed with a cross-entropy loss to distinguish ground truth $y$ (with supervision as $1$) and model prediction $f(x,z,\theta)$ (with supervision as $0$).

\noindent\textbf{Alternating Back-propagation:} With the CVAE \cite{cvae} based framework, an extra inference model $q(z|x,y,\phi)$ is used to approximate the true posterior distribution of the latent variable. For the GAN \cite{NIPS2014_5423_gan} based framework, the discrinimator is designed to estimate the quality of the model prediction. Alternating back-propagation \cite{ABP_aaai} is another type of generative model, which samples the latent variable directly from it's true posterior distribution via Langevin Dynamic based MCMC \cite{mcmc_langevin}.

Alternating Back-Propagation \cite{ABP_aaai} was introduced for learning the generator network model. It updates the latent variable and network parameters in an EM-manner. Firstly, given the network prediction with the
current parameter set, it infers the latent variable by
Langevin dynamics based MCMC, which they call 
\enquote{Inferential back-propagation} \cite{ABP_aaai}. Secondly, given the updated latent variable, the network parameter set is updated with gradient descent. They call it 
\enquote{Learning back-propagation} \cite{ABP_aaai}.
Following the previous variable definitions,
given the training example $(x,y)$, we intend to infer $z$ and learn the network parameter $\theta$ to minimize the reconstruction error as well as a regularization term that corresponds to the prior on $z$.

As a non-linear generalization of factor analysis, the conditional generative model aims to generalize the mapping from the continuous latent variable $z$ to the prediction $y$ conditioned on the input image $x$. Similar to
traditional factor analysis, we define our generative model as:
\begin{eqnarray}
    && z \sim q(z)=\mathcal{N}(0,\mathbf{I}), \label{eq:abp_1}\\
    && y = f(x,z,\theta) + \epsilon, \epsilon \sim \mathcal{N}(0,\Sigma), 
    \label{eq:abp_3} 
\end{eqnarray}
where $q(z)$ is the prior distribution of $z$, $\Sigma = \text{diag}(\sigma^2)$ is the inherent label noise. The conditional distribution of $y$ given $x$ is $p(y|x,\theta) = \int q(z) p(y|x,z,\theta) dz$ with the latent variable $z$ integrated out. We define the observed-data log-likelihood as $L(\theta)=\sum_{i=1}^N \log p(y_i|x_i,\theta)$, where the
gradient of $p(y|x,\theta)$ is defined as:
\begin{equation}
\label{update_omega}
\begin{aligned}
    \frac{\partial}{\partial \theta}\log p(y|x,\theta)&= 
    \frac{1}{p(y|x,\theta)}\frac{\partial}{\partial \theta}  p(y|x,\theta)\\
    &=\text{E}_{p(z|x,y,\theta)} \left[\frac{\partial}{\partial \theta}\log p(y,z|x,\theta)\right].
\end{aligned}   
\end{equation}

The expectation term $\text{E}_{p(z|x,y,\theta)}$ can be approximated by drawing samples from $p(z|x,y,\theta)$, and then computing
the latent variable $z$. Following ABP \cite{ABP_aaai}, we use Langevin Dynamics based MCMC (a gradient-based Monte Carlo method) to sample $z$, which iterates:
\begin{equation}
\begin{aligned}
    z_{t+1}=z_{t}+ \frac{s^2}{2}\left[ \frac{\partial}{\partial z}\log p(y,z_{t}|x,\theta)\right]+s \mathcal{N}(0,\mathbf{I}),
    \label{langevin_dynamics}
\end{aligned}
\end{equation}
with 
\begin{equation}
\frac{\partial}{\partial z}\log p(y,z|x,\theta) = \frac{1}{\sigma^2}(y-f(x,z,\theta))\frac{\partial}{\partial z}f(x,z,\theta) - z,
\end{equation}
where $t$ is the
time step for Langevin sampling, and $s$ is the step size. Eq.~\ref{langevin_dynamics} provides an efficient way of updating the latent variable $z$ by directly sampling it from the true posterior distribution $p(z|x,y,\theta)$.

\noindent\textbf{Energy-based Model:} Energy-based models (EBM) \cite{ACKLEY1985147,contrastive_hinton,DBM_hinton,coopnets,LeCun06atutorial,implicit_ebm,xie_generative_covnet}, which are usually represented as a deep neural networks, aim to learn an energy function, which assigns low energy to in-distribution samples and vice versa.
In this way, the energy function is designed to estimate the compatibility of the input variable $x$ and the output variable $y$.

Let's define the energy-based model as $p(y|x,\theta)$, which represents a distribution of prediction $y$ given an image $x$ by:
\begin{equation}
    p(y|x,\theta) = \frac{p(y,x,\theta)}{\int p(y,x,\theta)dy} = \frac{1}{Z(x;\theta)}\exp[-U(y,x,\theta)], \label{eq:ebm}
\end{equation}
where the energy function $U(y,x,\theta)$, parameterized by a bottom-up neural network with parameters $\theta$, maps the input/output pair to a scalar.
$Z(x;\theta)=\int \exp[-U(y,x,\theta)]dy$ is the normalizing constant. When the energy function
$U$ is learned and an image $x$ is given, the model prediction $y$ can be achieved by Langevin sampling \cite{mcmc_langevin} $y \sim p(y|x,\theta)$, which makes use of the gradient of the energy function and iterates the following step:
\begin{equation}
    y_{\tau+1} = y_{\tau} - \frac{\delta^2}{2} \frac{\partial U(y_{\tau},x,\theta)}{\partial y} + \delta \Delta_{\tau},  \Delta_{\tau} \sim N(0,\Sigma),
\label{equ:ebm_Langevin}
\end{equation}
where $\tau$ indexes the Langevin time steps, and $\delta$ is the step size. Langevin dynamics \cite{mcmc_langevin} is equivalent to a stochastic gradient descent algorithm that seeks to find the minimum of the objective function defined by $U(y,x,\theta)$. The Gaussian noise term $\Delta_{\tau}$ is a Brownian motion that prevents gradient descent from
being trapped by local minima of $U(y,x,\theta)$. 

Two main issues exist in the
above energy based model. Firstly, we need to set a start point $y_0$ for the Langevin sampling process as in Eq.~\ref{equ:ebm_Langevin}. Secondly, we should decide whether to fix the start point (cold start) or iteratively update it (warm start) while training the energy function $U(y,x,\theta)$. Following the cold start solution, we can train a deterministic network and fixed it while training the energy function $U(y,x,\theta)$, or we can define $y_0$ as random noise and learn $U(y,x,\theta)$ to gradually update it. Although random initialization is easy to define, its quite slow during training \cite{coopnets}. For the warm start, the initial prediction (start point) generator can be a deterministic network or latent variable model, which produces initial prediction, and then the energy function $U(y,x,\theta)$ updates the initial prediction by running several steps of Langevin sampling via Eq.~\ref{equ:ebm_Langevin}. The latent variable model as warm start solution has been discussed in \cite{coopnets}. 


For both cold start and warm start, we need to update the task related generator ($y_0$ generator) and the energy function $U(y,x,\theta)$. The $y_0$ generator can be learned following the deterministic learning pipeline or latent variable learning pipeline depending on its framework. The energy function $U(y,x,\theta)$ can be updated via MLE:
\begin{equation}
    \Delta{\theta} \approx \frac{1}{N} \sum_{i=1}^{N} \frac{\partial}{\partial \theta}U(\tilde{y}_i,x_i,\theta) -  \frac{1}{N} \sum_{i=1}^{N} \frac{\partial}{\partial \theta} U(y_i,x_i,\theta),
\label{equ:ebm_update}
\end{equation}
where $\tilde{y}$ is the updated prediction
from the start point $y_0$ with several steps of Langevin sampling.

\noindent\textbf{Uncertainty computation:} As we have no distribution assumption for model parameters within the generative models, we cannot estimate epistemic uncertainty within generative models. We can only estimate the aleatoric uncertainty (which is the total uncertainty or predictive uncertainty in this circumstance) by sampling from the distribution of the latent variable, and define the mean entropy of the predictions as aleatoric uncertainty.


\subsection{Bayesian Latent Variable Model}
\label{blvm_sec}
As discussed, the Bayesian Neural Network is effective in modeling the epistemic uncertainty, and the generative model makes it easier to estimate the aleatoric uncertainty. To achieve both types uncertainty estimation, a Bayesian latent variable model (BLVM) can be designed following \cite{uncertainty_decomposition,Non-Identifiability_BLVM}. Within the Bayesian latent variable model, the predictive distribution is defined as:
\begin{equation}
    \label{predictive_distribution_bayesian_lvm}
    p(y|x) = \int p(y|x,z,\theta)p(\theta)p(z)d\theta dz,
\end{equation}
where $p(y|x,z,\theta)=\mathcal{N}(f(x,z,\theta),\Sigma))$ is the likelihood as in Eq.~\ref{predictive_distribution_generative_model}.
Different from Eq.~\ref{predictive_distribution_generative_model} where $p(\theta)$ is constant, indicating a non-Bayesian Neural Network. Within the BLVM, $p(\theta)$ is the approximate posterior distribution of $p(\theta|D)$. In this way, the randomness of $y$ comes from three parts: 1) the approximated posterior distribution $p(\theta)$; 2) the latent variable $z$ and 3) the inherent label noise $\Sigma$, where $p(\theta|D)$ contributes to epistemic uncertainty and the other two contribute to aleatoric uncertainty.


Given the Bayesian latent variable model in Eq.~\ref{predictive_distribution_bayesian_lvm}, the total uncertainty (predictive uncertainty) is quantified as $U_p=H(y|x)$. \cite{uncertainty_decomposition} defines the other two types of uncertainty as:
\begin{equation}
\label{decomp_aleatoric_general}
    U_a = \mathbf{E}_{p(\theta)}[H(y|x,\theta)],
\end{equation}
and 
\begin{equation}
\label{decomp_epistemic_general}
    U_e = U_p-U_a=H(y|x)-\mathbf{E}_{p(\theta)}[H(y|x,\theta)]=I(y,\theta|x).
\end{equation}
$H(y|x,\theta)$ is the entropy of $p(y|x,\theta)$, which is defined as:
$p(y|x,\theta)=\int p(y|x,\theta,z)p(z)dz$. For a specific model parameter $\theta$, we obtain model prediction $p(y|x,\theta)$, and compute its entropy. Then, the expectation of $p(y|x,\theta)$ under $p(\theta)$ is used to quantify uncertainty come from $z$ and inherent noise $\Sigma$, which is the aleatoric uncertainty. Epistemic uncertainty is then the residual of predictive uncertainty and aleatoric uncertainty, which is the mutual information of $y$ and $\theta$ conditioned on the input $x$.

\cite{uncertainty_decomposition} introduces variance as another type of uncertainty measure, which defines the total uncertainty as variance of the predictive distribution, which is $U_p=\sigma^2(y|x)$, where $\sigma^2(.)$ computes the variances of a probability distribution. Based on the law of total variance:
\begin{equation}
\label{decomp_predictive}
    \sigma^2(y|x) = \sigma^2_{p(\theta)}(\mathbf{E}[y|x,\theta]) + \mathbf{E}_{p(\theta)}[\sigma^2(y|x,\theta)],
\end{equation}
where $\mathbf{E}[y|x,\theta]$ and $\sigma^2(y|x,\theta)$ are the mean and variances of prediction according to $p(y|x,\theta)$. Specifically, we fixed $\theta$, and sample $z$ to obtain the mean and variance of $p(y|x,\theta)$. $\sigma^2_{p(\theta)}(\mathbf{E}[y|x,\theta])$ is then the variance of the mean when $\theta\sim p(\theta)$, which ignores the contribution of $z$ and inherent noise $\Sigma$, making it the epistemic uncertainty, which is:
\begin{equation}
    \label{blvm_epistemic_variance}
    U_e=\sigma^2_{p(\theta)}(\mathbf{E}[y|x,\theta]).
\end{equation}
Further, $\mathbf{E}_{p(\theta)}[\sigma^2(y|x,\theta)]$ is the mean variance when $\theta\sim p(\theta)$, which represents uncertainty from $z$ and $\Sigma$, and is defined as aleatoric uncertainty, which is:
\begin{equation}
    \label{blvm_aleatoric_variance}
    U_a = \mathbf{E}_{p(\theta)}[\sigma^2(y|x,\theta)].
\end{equation}

In practice, to compute aleatoric uncertainty $U_a$, with fixed $\theta$, we sample from the latent space multiple times and obtain the mean prediction $\mathbf{E}[y|x,\theta]$ and the variance $\sigma^2(y|x,\theta)$ of the multiple predictions. Then, with $\theta\sim p(\theta)$, we define the
mean of above variance ($\mathbf{E}_{p(\theta)}[\sigma^2(y|x,\theta)]$) as aleatoric uncertainty,
and define variance of multiple $\mathbf{E}[y|x,\theta]$ with $\theta\sim p(\theta)$
as the epistemic uncertainty.

\section{Bayesian Latent Variable Model for Efficient Dense Uncertainty Estimation}
We analyse existing uncertainty estimation solutions and
introduce an efficient uncertainty estimation methods.

\begin{figure}[tp]
   \begin{center}
   \begin{tabular}{c@{ }}
   {\includegraphics[width=0.95\linewidth]{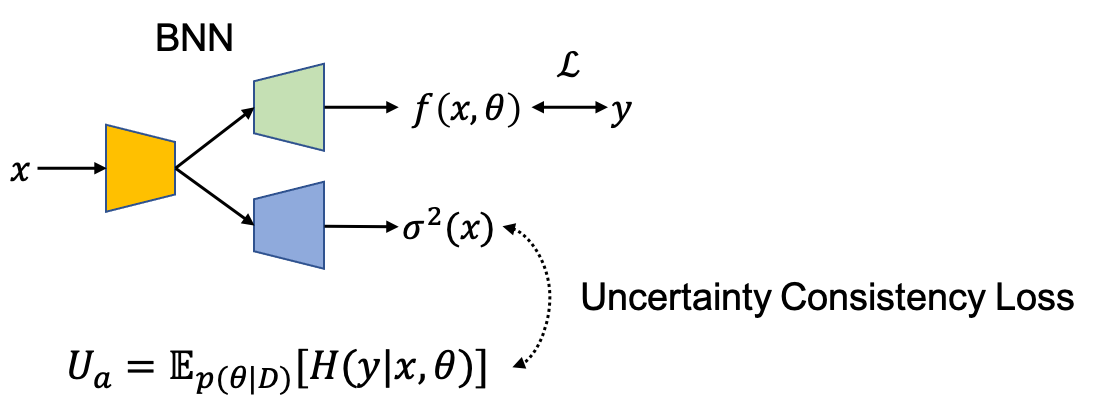}}\\
   \end{tabular}
   \end{center}
   \caption{Uncertainty consistency loss within BNN to avoid trivial solution of aleatoric uncertainty from the dual-head structure.
   }
\label{fig:avoid_trivial_solution_aleatoric}
\end{figure}

\begin{figure}[tp]
   \begin{center}
   \begin{tabular}{c@{ }}
   {\includegraphics[width=0.95\linewidth]{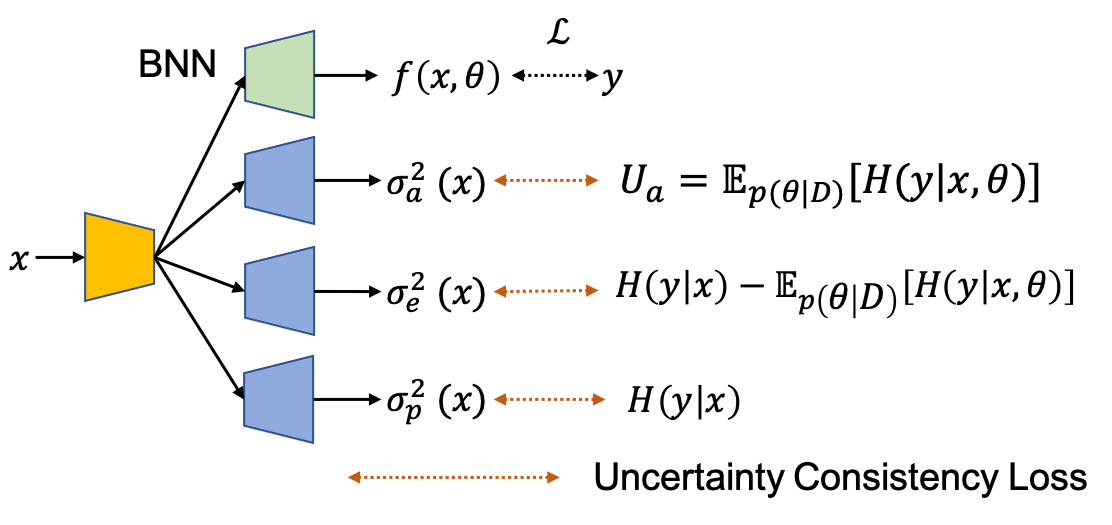}}\\
   \end{tabular}
   \end{center}
   \caption{Approximating uncertainty with multi-head uncertainty consistency loss within BNN.
   }
\label{fig:uncertainty_approximation}
\end{figure}

\subsection{Trivial Solution for Aleatoric Uncertainty}
As discussed in Section \ref{sec_aleatoric_uncertainty}, the loss function with noise corruption model for regression task is achieved as \cite{nicta_uncertainty,kendall2017uncertainties}:
\begin{equation}
    \label{loss_aleatoric_uncertainty_sum}
    \mathcal{L}(\theta)=\frac{1}{N}\sum_{i=1}^N(\frac{1}{2\sigma^2(x_i)}\mathcal{L}_2+\frac{1}{2}log(\sigma^2(x_i))),
\end{equation}
and for classification task is achieved as:
\begin{equation}
\label{approximated_classification_log_likelihood_loss_sum}
    \mathcal{L}(\theta)=\frac{1}{N}\sum_{i=1}^N(\frac{1}{T_i}\mathcal{L}_{ce} + \log T_i),
\end{equation}
where $T=\exp(\sigma^2(x))$, representing the temperature as in model calibration \cite{on_calibration}.
As there is
no extra supervision or constraints for the aleatoric uncertainty from the dual-head framework, we observe one trivial solution of aleatoric uncertainty for both the regression tasks and classification tasks. For the former, the $\sigma^2(x)=1$ will lead to the original loss function, $\mathcal{L}_2$ in particular, resulting in
image-independent uncertainty. For the latter, $\sigma^2(x)=0$ will also lead to the original cross-entropy loss $\mathcal{L}_{ce}$. To avoid the trivial solution, we can relate the two definitions of aleatoric uncertainty from both the dual-head structure \cite{nicta_uncertainty,kendall2017uncertainties} and mean entropy of prediction within the Bayesian Neural Network (BNN) as shown in Fig.~\ref{fig:avoid_trivial_solution_aleatoric}. Specifically, we define uncertainty consistency loss within a BNN, which is a cross-entropy loss or a symmetric loss function, \ie~SSIM \cite{ssim_raw,ssim_cvpr2017} to regularize the aleatoric uncertainty from dual head to be similar to the sampling based uncertainty as in Eq.~\ref{aleatoric_bnn_entropy}. The main advantage of this strategy is that, the trained dual-head based aleatoric uncertainty estimation module can achieve sampling-free uncertainty estimation during testing.

\subsection{Efficient Uncertainty Estimation within BLVM}
\cite{uncertainty_decomposition} define three origins of uncertainty within BNN: 1) inherent noise $\Sigma$; 2) latent variable $z$ and 3) the approximated posterior distribution $p(\theta)$, where the first two are claimed as the
origins of aleatoric uncertainty and the last one is related to epistemic uncertainty. For both types of uncertainty estimation and the total uncertainty, the expectation term indicates that the
sampling process is usually required to estimate the three types of uncertainties, which is less efficient during testing.

Inspired by the dual-head structure for aleatoric uncertainty estimation, we introduce uncertainty approximation with two extra heads for epistemic uncertainty estimation and predictive uncertainty estimation as shown in Fig.~\ref{fig:uncertainty_approximation}. In this way, we avoid the tedious sampling process, leading to efficient uncertainty estimation during testing.


\subsection{Uncertainty Computation}
\section{Experimental Results}
\subsection{Setup}
\noindent\textbf{Dataset:}
We perform experiments on two binary segmentation tasks, namely salient object detection \cite{cpd_sal,scrn_sal,wei2020f3net,wang2020progressive,wei2020label} and camouflaged object detection \cite{troscianko2017quantifying, pike2018quantifying, tankus2001convexity,xue2016camouflage, li2017foreground,fan2020camouflaged}, and two regression tasks, namely monocular depth estimation \cite{midas_tpami,Bhat_2021_CVPR,dpt} and image deblurring \cite{nah2017deep,shen2019human,rim2020real}, to verify the effectiveness of each uncertainty estimation method. Salient object detection aims to localize the region that attract human attention. Camouflaged object detection models, on the contrary, are designed to identify the camouflaged objects hiding in the environment, which usually share similar pattern as their surroundings. Monocular depth estimation is a well-studied task, aiming to estimate depth from a single view RGB image. Image deblurring techniques try to recover a sharp image from its blurred version.

For salient object detection, we used the DUTS dataset~\cite{imagesaliency} for training. Testing images include 1) DUTS testing dataset,
2) DUT \cite{Manifold-Ranking:CVPR-2013}, 3) HKU-IS \cite{MDF:CVPR-2015} and 4) PASCAL-S \cite{pascal_s_dataset}.
For camouflaged object detection, the benchmark training dataset is the combination of 3,040 images from COD10K training dataset \cite{fan2020camouflaged} and 1,000 images from CAMO training dataset \cite{le2019anabranch}. We then test our model on four benchmark testing datasets, namely CAMO testing dataset \cite{le2019anabranch}, COD10K testing dataset \cite{fan2020camouflaged}, CHAMELEMON dataset \cite{Chameleon2018} and NC4K dataset \cite{yunqiu2021ranking}. 

For monocular depth estimation, the typical training and testing strategy is that the training and testing images should come from the same dataset as in \cite{Bhat_2021_CVPR}. \cite{midas_tpami} presents a mixed training dataset to achieve zero-shot cross-dataset transfer. Following the same training dataset setting, \cite{dpt} also train with the mixed large training dataset, and fine-tune on the other dataset for performance evaluation. In this paper, we adopt both settings, 1) following \cite{Bhat_2021_CVPR}, we train and test on NYUv2 \cite{nyuv2} and KITTI dataset \cite{kitti}, with the backbone network initialized with parameters for image classification on ImageNet; 2) following \cite{midas_tpami}, the backbone is initialized with parameters trained for monocular depth estimation with the mixed large training dataset, and then fine-tune further on NYUv2 \cite{nyuv2} and KITTI dataset \cite{kitti} for performance evaluation.

For image deblurring, the training samples are 2103 pairs of blurry images and the corresponding ground truth sharp images from the GoPro training dataset \cite{nah2017deep}. The testing images include 1) 1111 pairs from the GoPro testing dataset. 2) 2025 pairs from the HIDE testing dataset \cite{shen2019human}. 3) 4937 pairs from the RealBlur-J dataset \cite{rim2020real}. The blurry images from GoPro and HIDE datasets are synthesized via averaging continuous frames. The blurry images from RealBlur-J testing dataset are captured in the real world.

\noindent\textbf{Metrics:}
As an uncertainty estimation network, we aim to produce both task related prediction and the corresponding uncertainty map representing model awareness about it's prediction. For the former, the task related metric can be used, \ie we use mean F-measure and mean absolute error (MAE) for both the salient object detection and camouflaged object detection task. For monocular depth estimation, following the conventional solution \cite{midas_tpami,dpt}, we use the percentage of the pixel with $\delta=max(\frac{d_i}{d^*_i},\frac{d^*_i}{d_i}>1.25)$ to evaluate model performance. For image deblurring, the candidate metrics are
PSNR, SSIM \cite{ssim_raw} and LPIPS \cite{zhang2018unreasonable}.
In this paper, we use the two widely used metrics, which are PSNR and SSIM \cite{ssim_raw}.

For uncertainty estimation, we use the metric from \cite{evaluation_uncertainty}. In \cite{evaluation_uncertainty}, they define two conditional probabilities:
1) $p(accurate|certainty)$: the probability that the model is accurate given that it is confident; and 2) $p(uncertainty|inaccurate)$: the probability that the model is uncertainty given that it is inaccurate.
The basic assumption is that for certain pixels (or regions), the model should have accurate predictions and for inaccurate predictions, the corresponding certain level should be low.

To implement above metric, \cite{evaluation_uncertainty} introduce three scores, namely $p(accurate|certainty)$, $p(uncertainty|inaccurate)$ and Patch accuracy vs patch uncertainty (PAvPU), which has been used in \cite{probabilistic_instance_eccv_workshop} for uncertainty evaluation with the instance segmentation task.
For each patch $s_k$, \cite{evaluation_uncertainty} computes the patch accuracy $a_k$ (pixel accuracy metric defined in \cite{FCN}). Then, \cite{evaluation_uncertainty} decides whether
the patch is accurate based on certain threshold $h_a$, \eg prediction of $s_k$ is accurate if $a_k > h_a$.
Similarly, the uncertainty of patch $s_k$ is achieved if its uncertainty $u_k>h_u$, where $h_u$ is a confidence related threshold.
Further, they define four metric related variables:
1) $n_{ac}$ is the number of patches which are accurate and certain; 
2) $n_{au}$ is the number of patches which are accurate and uncertain;
3) $n_{ic}$ is the number of patches which are inaccurate and certain; and
4) $n_{iu}$ is the number of patches which are inaccurate and uncertain. Finally, three uncertainty evaluation scores are generated as:
\begin{equation}
    p(accurate|certainty) = \frac{n_{ac}}{(n_{ac}+n_{ic})},
\end{equation}
\begin{equation}
    p(uncertain|inaccurate) = \frac{n_{iu}}{(n_{ic}+n_{iu})},
\end{equation}
and
\begin{equation}
    PAvPU = \frac{(n_{ac}+n_{iu})}{(n_{ac}+n_{au}+n_{ic}+n_{iu})}.
\end{equation}
A model with higher above scores is a better performer.

Instead of defining a
uniform patch as in \cite{evaluation_uncertainty}, we over-segment image to super-pixels to achieve uniform accuracy/uncertainty with in a same super-pixel considering the semantic consistency attribute. Further, the three above scores are based on handcrafted hyper-parameters, namely accuracy related threshold $h_a$ and uncertainty related threshold $h_u$. Inspired by the expected calibration error \cite{on_calibration}, we use binning
instead of hard-thresholding. Specifically, we use threshold in the range of $[0,1]$ with 10 bins, and the final $n_{ac}$, $n_{au}$, $n_{ic}$, $n_{iu}$ are all 10-dimensional vectors, as well as the $p(accurate|certainty)$, $p(uncertain|inaccurate)$ and $PAvPU$. We then define mean $p(accurate|certainty)$, $p(uncertain|inaccurate)$ and $PAvPU$ for uncertainty performance evaluation. 

Note that, the pixel accuracy \cite{FCN} $a_k$ is defined as follows.
Given $n_{ij}$ as number of pixels of class $i$ predicted as class $j$, and $t_{i}=\sum_j n_{ij}$ is the total number of pixels of class $i$, the pixel accuracy is defined as:
\begin{equation}
    pixel\_acc = \sum_i n_{ii}/\sum_i t_i
\end{equation}
For binary segmentation:
\begin{equation}
    pixel\_acc = \frac{(n_{ff}+n_{bb})}{H*W}
\end{equation}
where $H$ and $W$ are image height and width.







\subsection{Implementation Details}
\label{implementation_details_sec}
We design both ensemble based models, generative model based models and Bayesian latent variable model based models for the four classification and regression tasks.

\noindent\textbf{Task related generator:} We trained the salient object detection, camouflaged object detection and monocular depth estimation using PyTorch with a maximum of 30 epochs. ResNet50 \cite{ResHe2015} is chosen as backbone, and we choose the decoder from \cite{midas_tpami} as our decoders for the three tasks, which gradually aggregates higher level features with lower level features with residual connections. For both salient object detection and camouflaged object detection, we adopt the structure-aware loss function \cite{wei2020f3net}. 
For monocular depth estimation, following \cite{Alhashim2018densedepth}, we apply the loss function as the weighted sum of point-wise L1 loss, gradient loss and SSIM loss on inverse-depth predictions.
Different from above tasks, which are
usually built upon backbone networks \cite{ResHe2015,VGG}, image deblurring methods do not rely on existing backbone networks.
Among these methods, the multi-scale networks DeepDeblur \cite{nah2017deep} is one of the most popular deblurring methods, and the multi-stage network MPRNet \cite{zamir2021multi} is the state-of-the-art method. Therefore, we conduct uncertainty estimation experiments based on them. Models are trained based on the PyTorch framework with a maximum of 200 epochs. The loss functions are the same as the original papers.

\noindent\textbf{Uncertainty estimation models:} Given the backbone features $\{s_i\}_{i=1}^4$ for the chosen ResNet50 \cite{ResHe2015} backbone\footnote{We use backbone network for salient object detection, camouflaged object detection and monocular depth estimation.}, to achieve MC-dropout \cite{Gal2016Dropout} (\enquote{MD}), we add a dropout of rate 0.3 before $\{s_i\}_{i=1}^4$, and then feed the features after dropout to the decoder. For the deep ensemble \cite{simple_scalable_uncertainty} models (\enquote{DE}), we attach five decoders of the same structure after the backbone feature to generate five different predictions. For the snapshots ensembles \cite{snapshot_ensembles} (\enquote{SE}), we save the model after five epochs, and generate prediction with each snapshot model.

For the CVAE \cite{cvae} based framework (\enquote{CVAE}), we use two extra encoders to generate the prior and posterior distribution of the latent variable, which share the same network structures as in \cite{Zhang_2020_CVPR_UCNet}. For the GAN \cite{NIPS2014_5423_gan} based framework (\enquote{GAN}), we design a fully convolutional discriminator as in \cite{aixuan_cod_sod21}. For the ABP \cite{ABP_aaai} based framework (\enquote{ABP}), we update the latent variable via Langevin Dynamics \cite{mcmc_langevin} as shown in Eq.~\ref{langevin_dynamics}. For the EBM \cite{LeCun06atutorial} based framework (\enquote{EBM}), we design the same energy-based model as in \cite{jing2021learning_ebm}, and we then generate multiple predictions via Eq.~\ref{equ:ebm_Langevin}.

For the Bayesian latent variable model, we simply apply MC-dropout \cite{Gal2016Dropout} to the CVAE, GAN and ABP based framework, leading to  a Bayesian model \enquote{BCVAE}, \enquote{BGAN}, \enquote{BABP} respectively. As an extension, we extend the designed energy-based model to the latent variable based EBM, where the prediction from the three latent variable model serve as the start point for the energy function based Langevin Dynamics as in Eq.~\ref{equ:ebm_Langevin}, leading to \enquote{ECVAE}, \enquote{EGAN}, \enquote{EABP} respectively

For all the generative models, empirically, we set the dimension of the latent space as $K=8$, which is introduced to the network by concatenating with the highest level backbone feature, \ie $s_4$ in this paper. 
\subsection{Task Related Uncertainty Analysis}
\noindent\textbf{Uncertainty for Camouflaged Object Detection:}
Camouflaged object detection \cite{fan2020camouflaged,le2019anabranch} aims to localize the camouflaged objects within an given input RGB image. Due to the complex attributes that makes the object
camouflaged, \ie~color, texture, and \etc~, and also the similar pattern of camouflaged object compared with its surrounding,
there exists both task related ambiguity and labeling level ambiguity for camouflaged object detection \cite{fan2021ugtr}, where the former can be modeled as epistemic uncertainty and the latter can be represented as aleatoric uncertainty.

\noindent\textbf{Uncertainty for Salient Object Detection:} Saliency is defined as the
attribute that make the object distinct from its surrounding \cite{Itti,Koch1985ShiftsIS}. Many factors can lead something to be \enquote{salient}, including the stimulus itself that makes the item distinct, \ie~color, texture, direction of movement and \etc~, and the internal cognitive state of the observer, leading to his/her understanding of saliency. The \enquote{subjective nature} \cite{Zhang_2020_CVPR_UCNet,jing2021learning_ebm} of saliency leads to the ambiguity in both task understanding and labeling, which can be explained as epistemic uncertainty and aleatoric uncertainty respectively.

\begin{table}[t!]
  \centering
  \scriptsize
  \renewcommand{\arraystretch}{1.2}
  \renewcommand{\tabcolsep}{1.55mm}
  \caption{Ensemble based solutions for \textbf{camouflaged object detection}. $\uparrow$ indicates the higher the score the better, and vice versa for $\downarrow$.}
  \begin{tabular}{l|cc|cc|cc|cc}
  \toprule
   Method &\multicolumn{2}{c|}{CAMO~\cite{le2019anabranch}}&\multicolumn{2}{c|}{CHAMELEON~\cite{Chameleon2018}}&\multicolumn{2}{c|}{COD10K~\cite{fan2020camouflaged}}&\multicolumn{2}{c}{NC4K~\cite{yunqiu2021ranking}} \\
    &$F_{\beta}\uparrow$&$\mathcal{M}\downarrow$&$F_{\beta}\uparrow$&$\mathcal{M}\downarrow$
    &$F_{\beta}\uparrow$&$\mathcal{M}\downarrow$
    &$F_{\beta}\uparrow$&$\mathcal{M}\downarrow$\\
  \hline
  Base & .757 & .079 & .848 & .029 & .731 & .035 & .803 & .048 \\
  MD & .767 & .080 & .842 & .028 & .731 & .035 & .803 & .048 \\
  DE & .729 & .088 & .846 & .030 & .718 & .037 & .796 & .051 \\
   \bottomrule
  \end{tabular}
  \label{tab:ablation_cod_ensemble}
\end{table}

\begin{figure}[tp]
   \begin{center}
   \begin{tabular}{c@{ }c@{ }c@{ }c@{ }c@{ }c@{ }c@{ }c@{ }}
   {\includegraphics[width=0.11\linewidth]{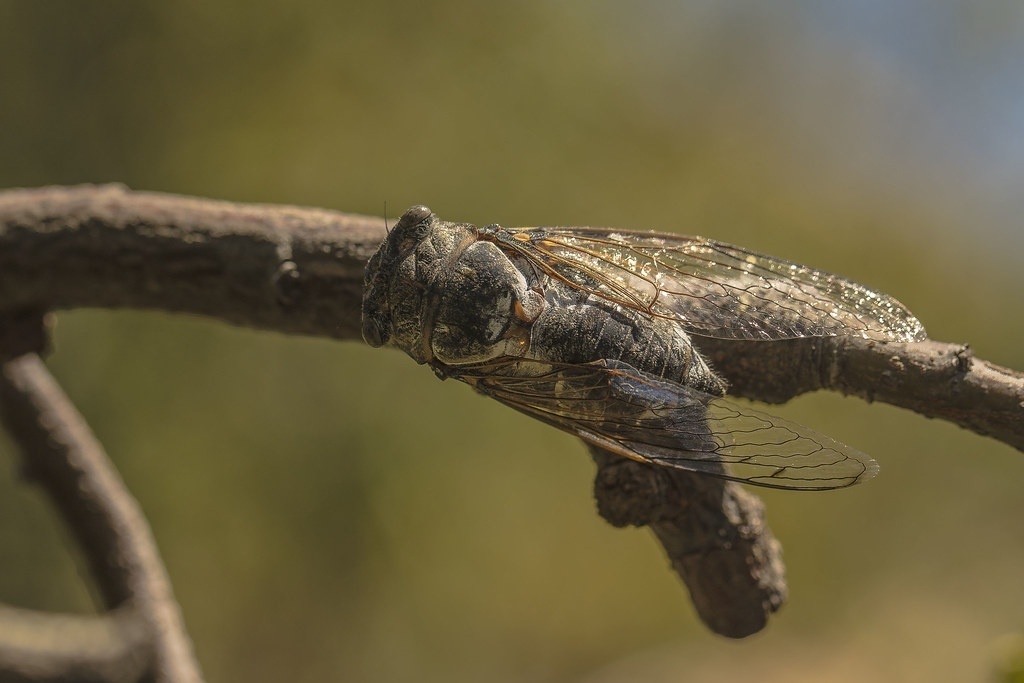}} &
   {\includegraphics[width=0.11\linewidth]{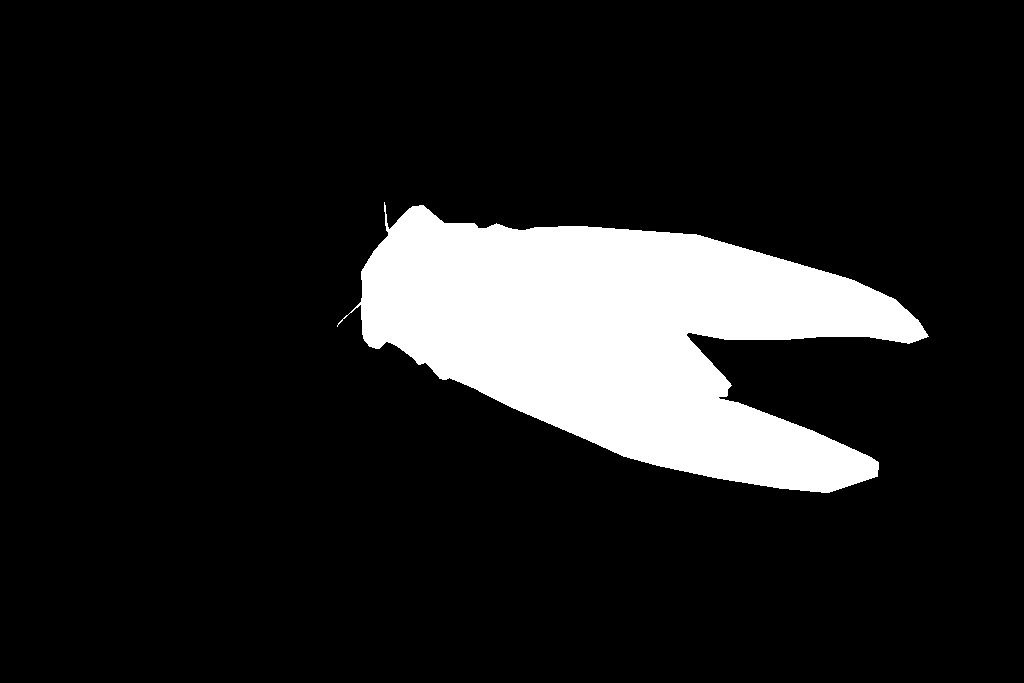}} &
   {\includegraphics[width=0.11\linewidth]{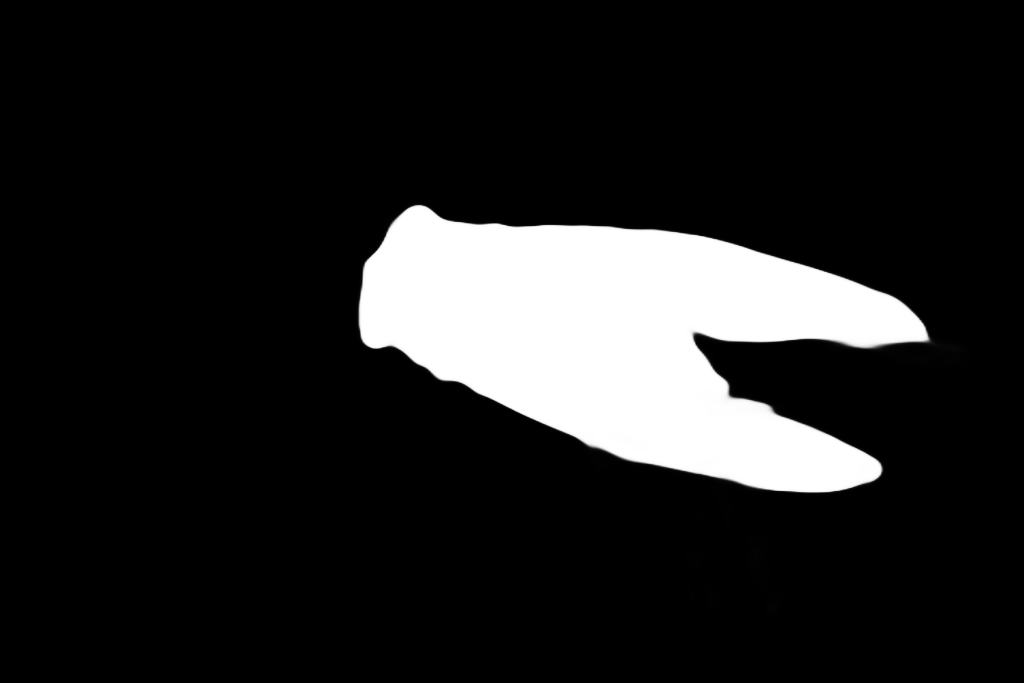}} &
   {\includegraphics[width=0.11\linewidth]{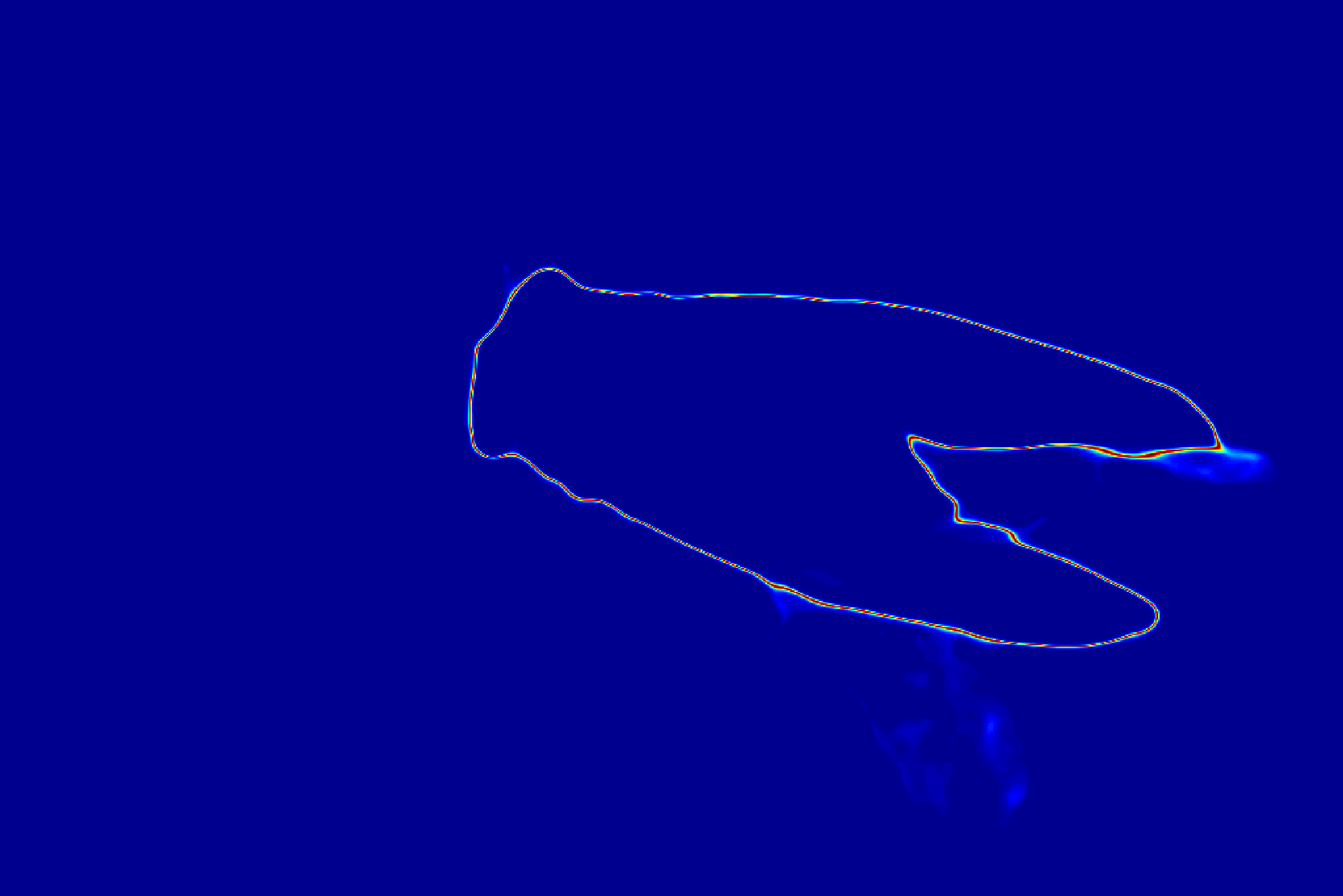}} &
   {\includegraphics[width=0.11\linewidth]{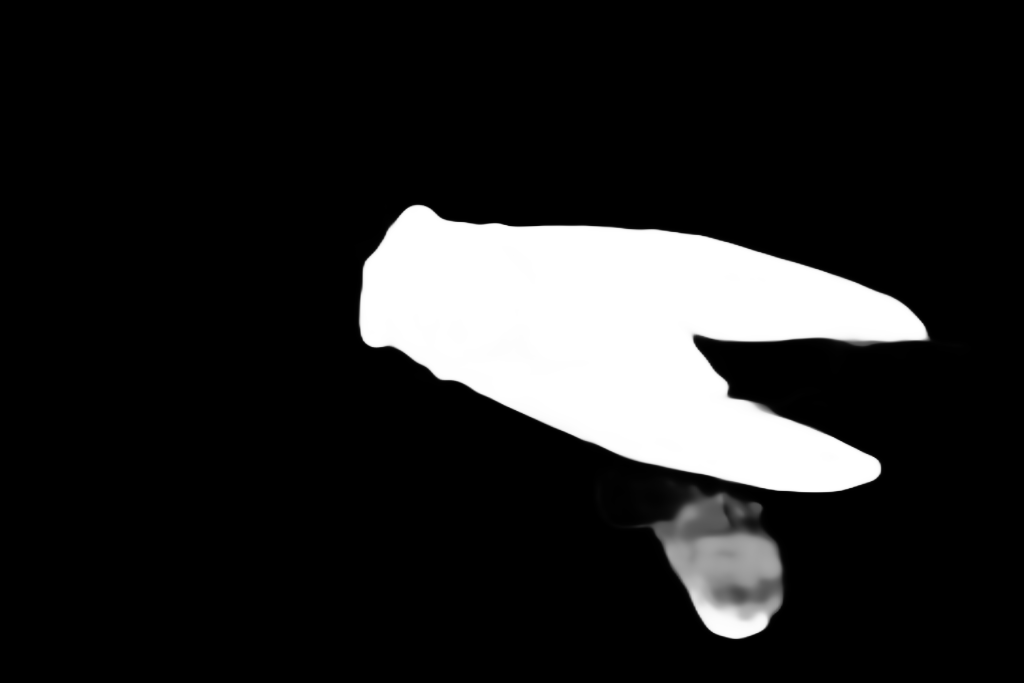}}&
   {\includegraphics[width=0.11\linewidth]{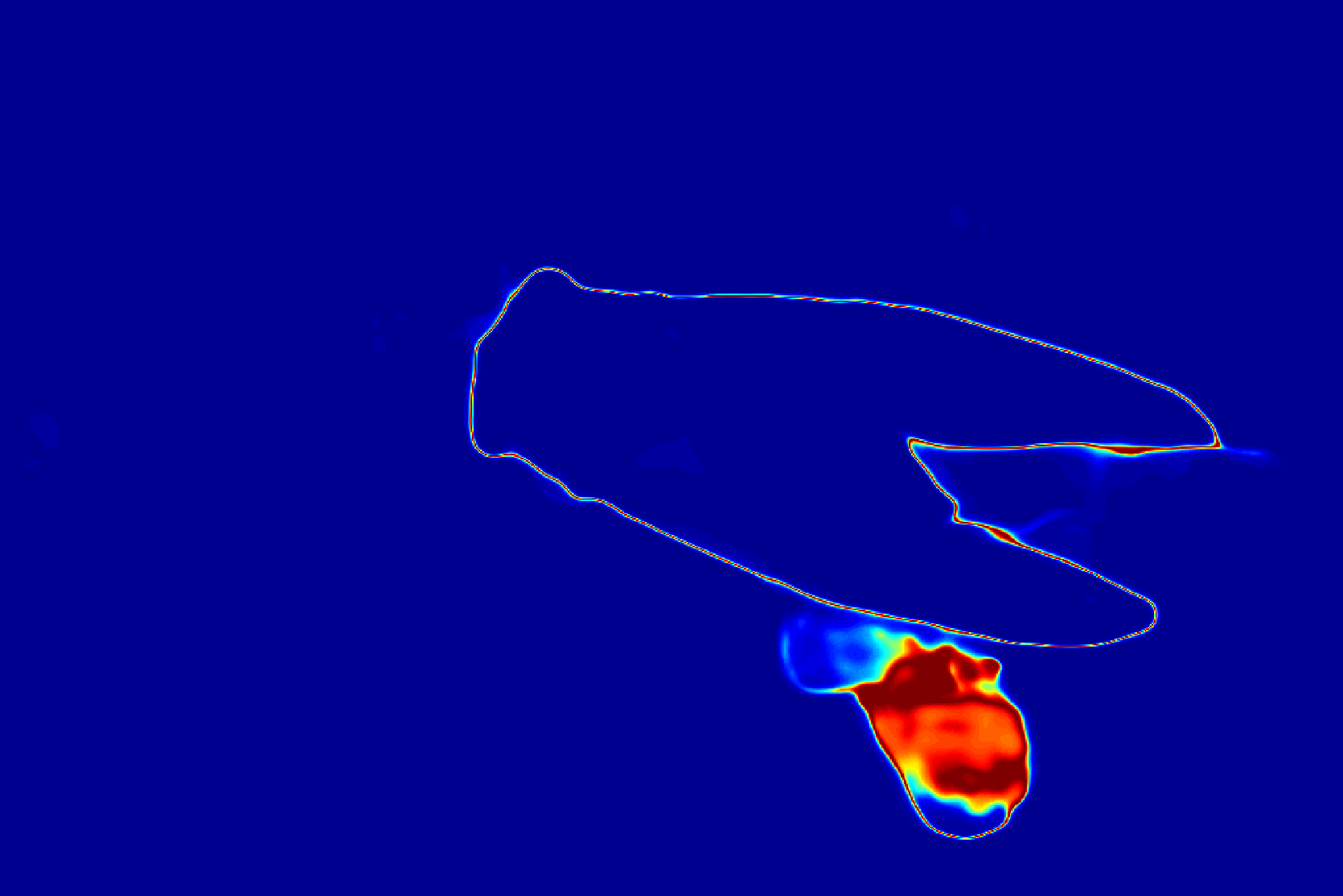}}&
   {\includegraphics[width=0.11\linewidth]{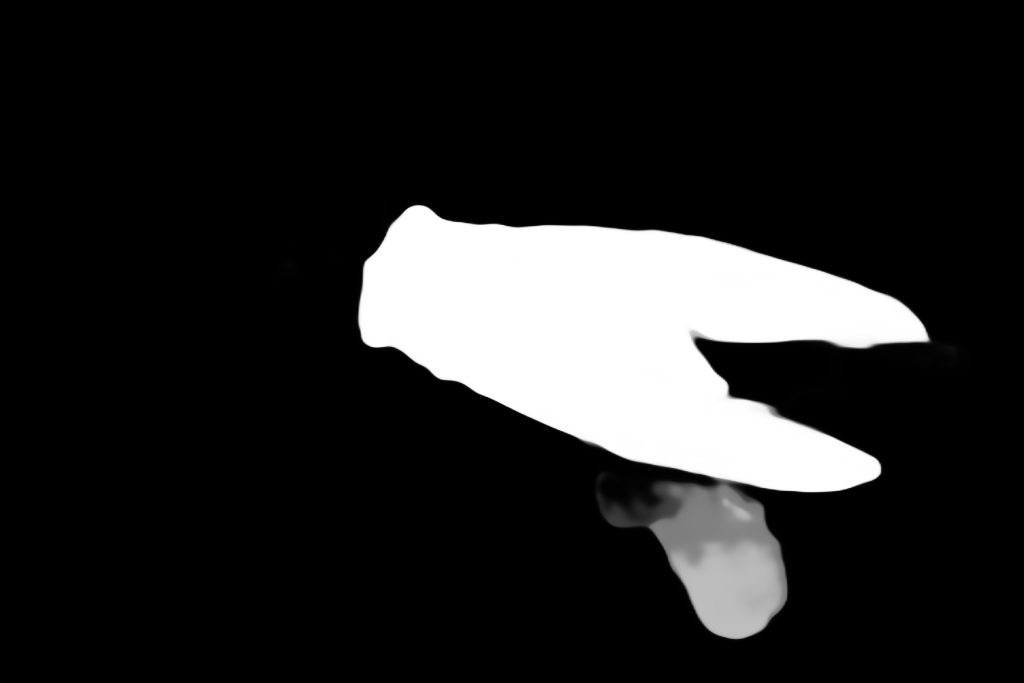}}&
   {\includegraphics[width=0.11\linewidth]{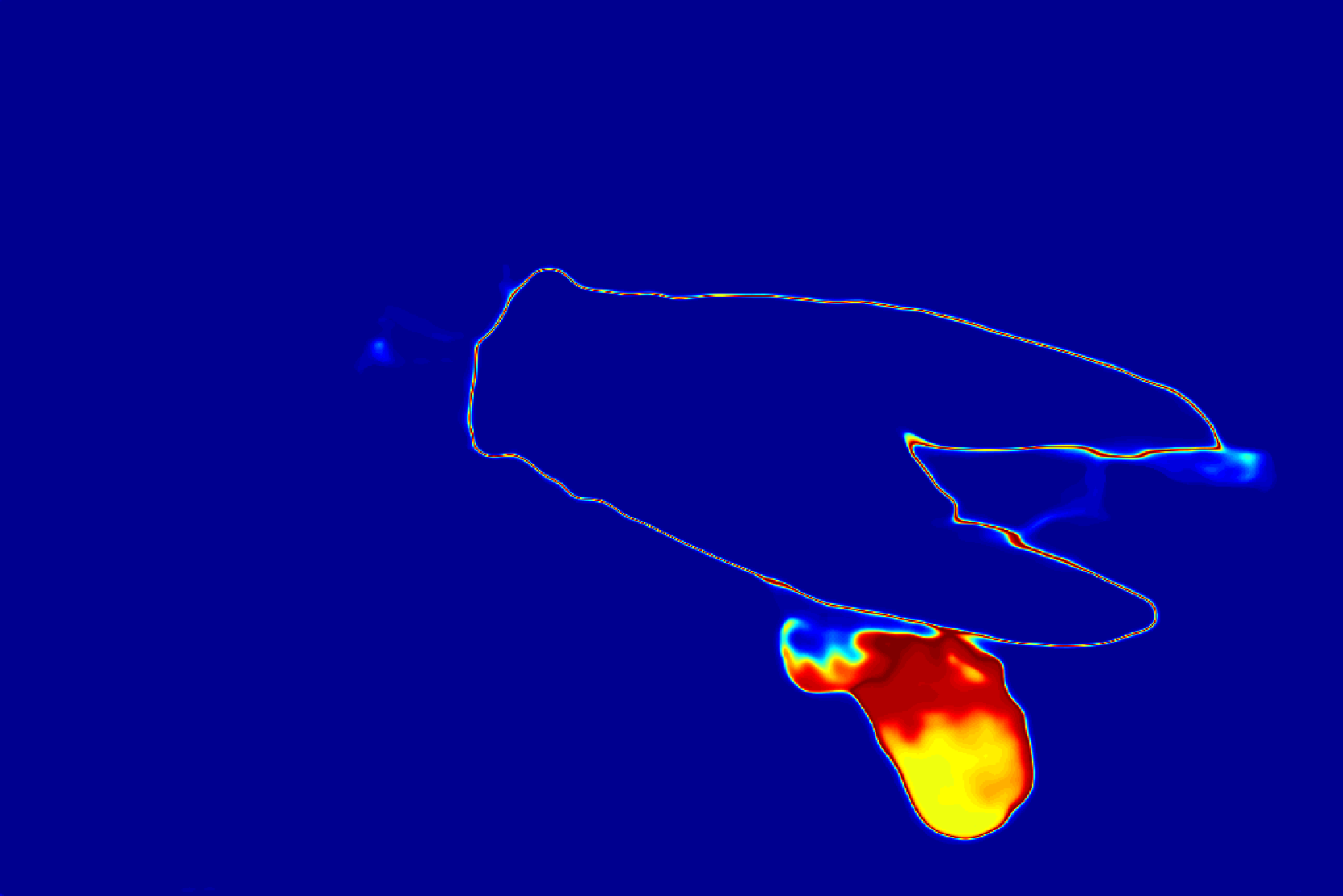}} \\
      {\includegraphics[width=0.11\linewidth]{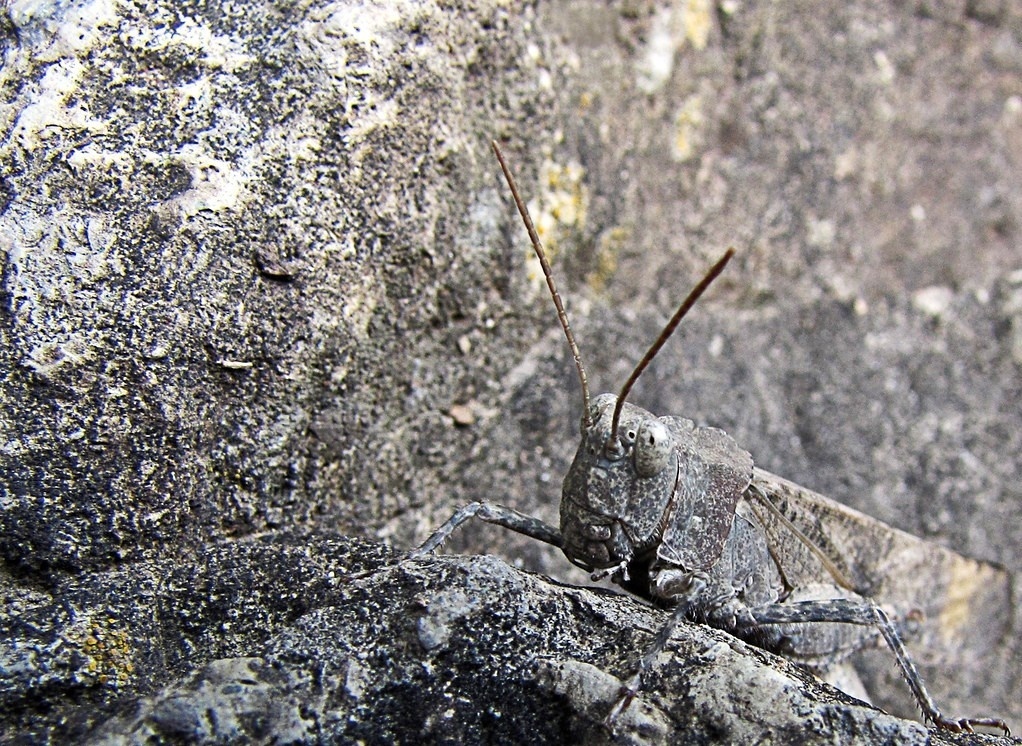}} &
   {\includegraphics[width=0.11\linewidth]{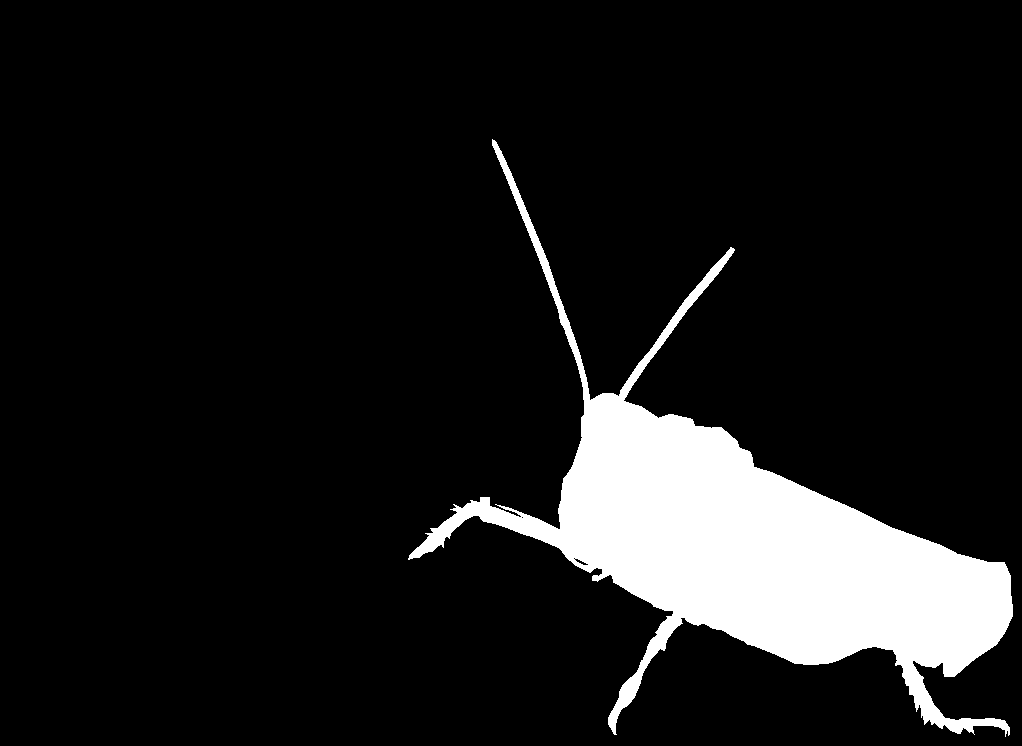}} &
   {\includegraphics[width=0.11\linewidth]{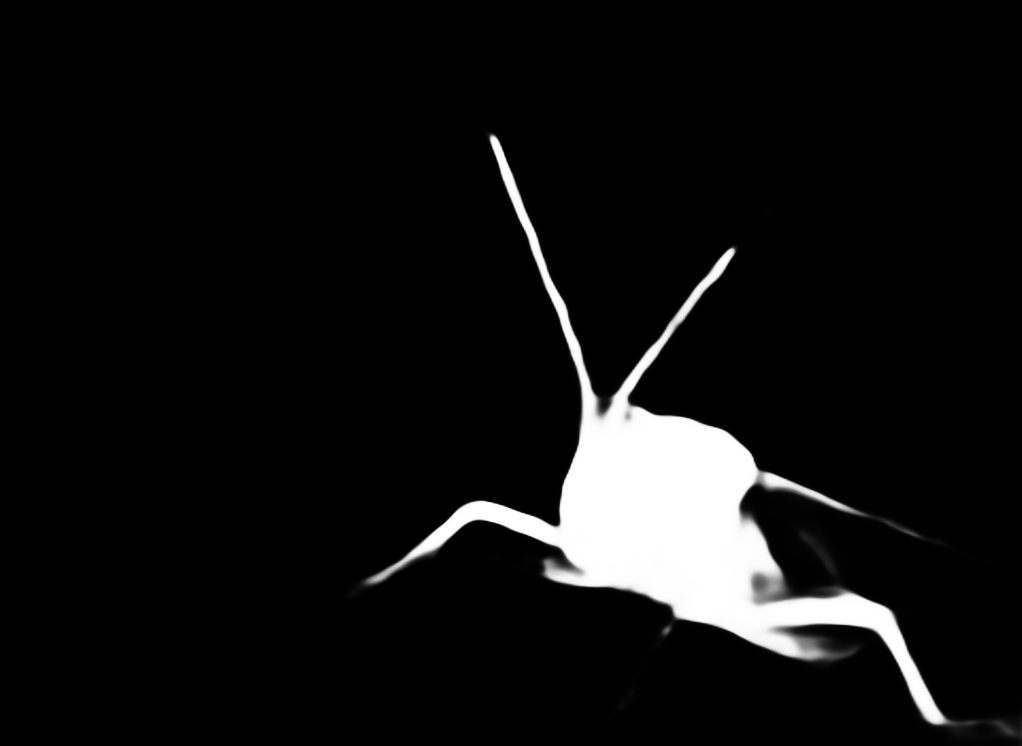}} &
   {\includegraphics[width=0.11\linewidth]{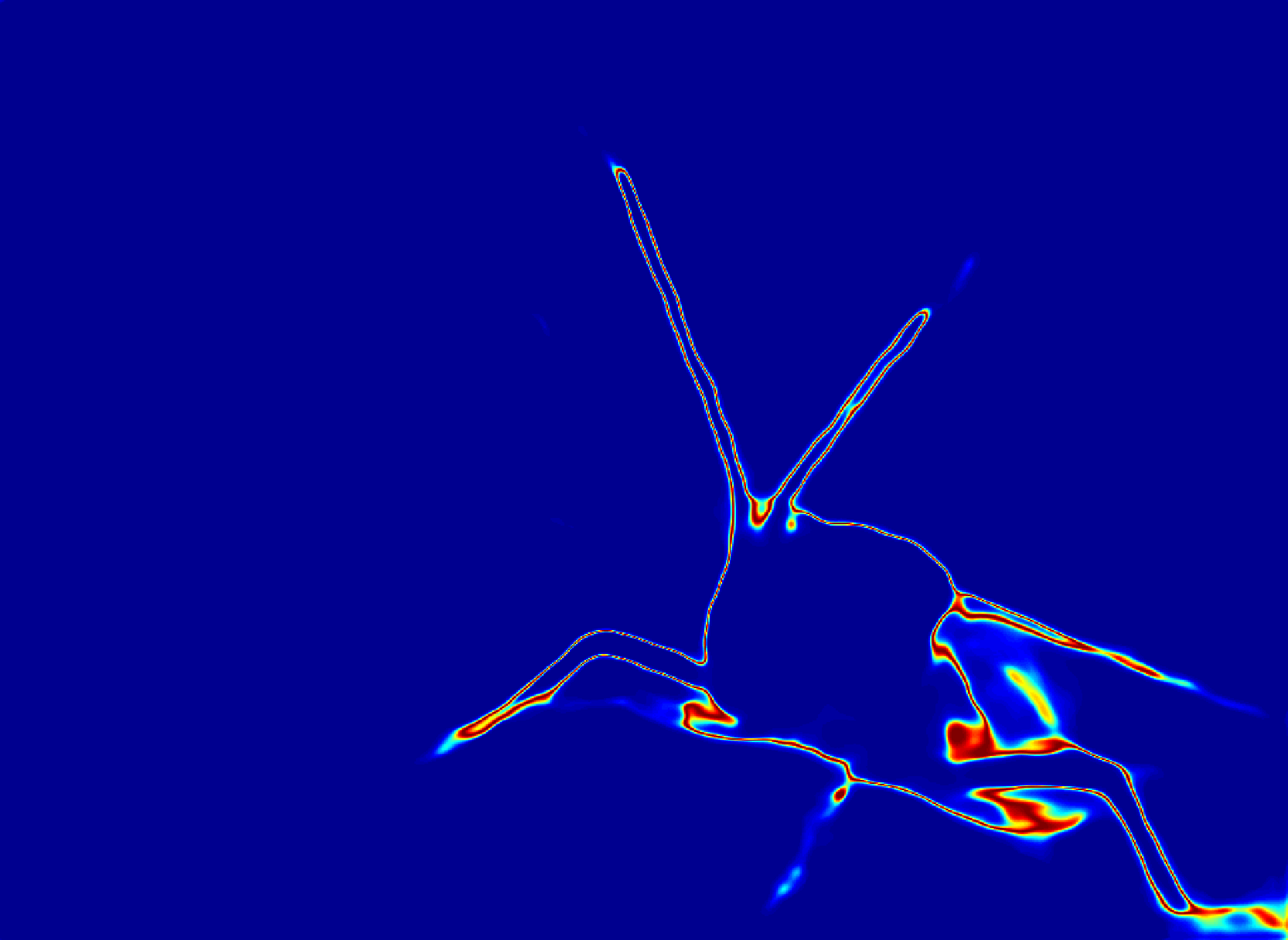}} &
   {\includegraphics[width=0.11\linewidth]{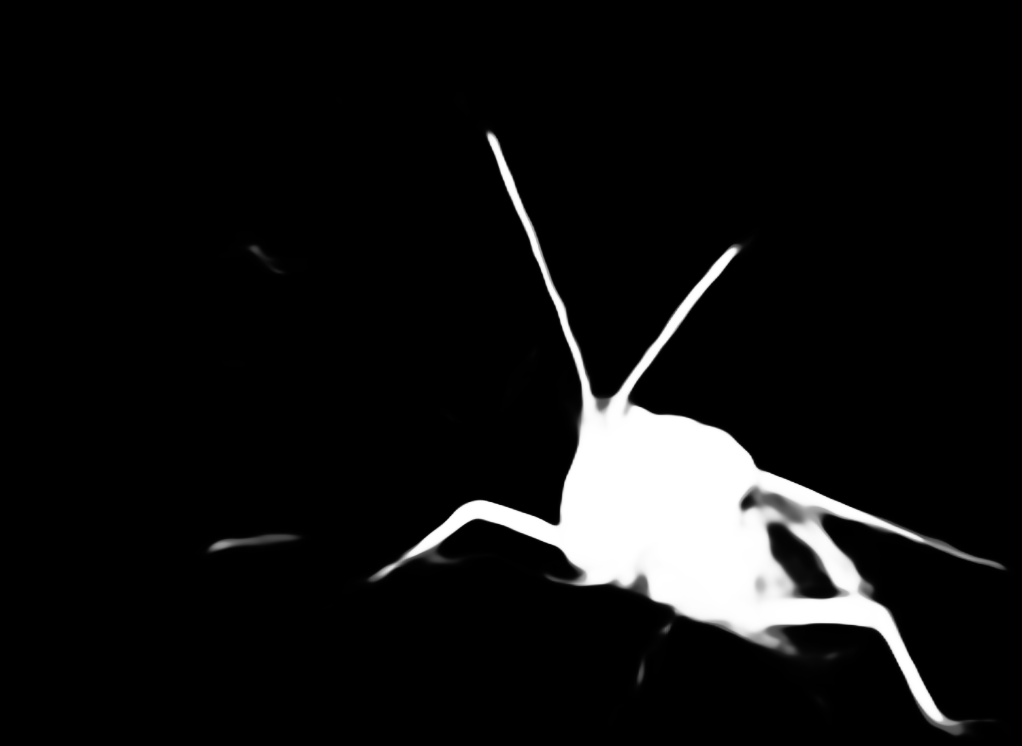}}&
   {\includegraphics[width=0.11\linewidth]{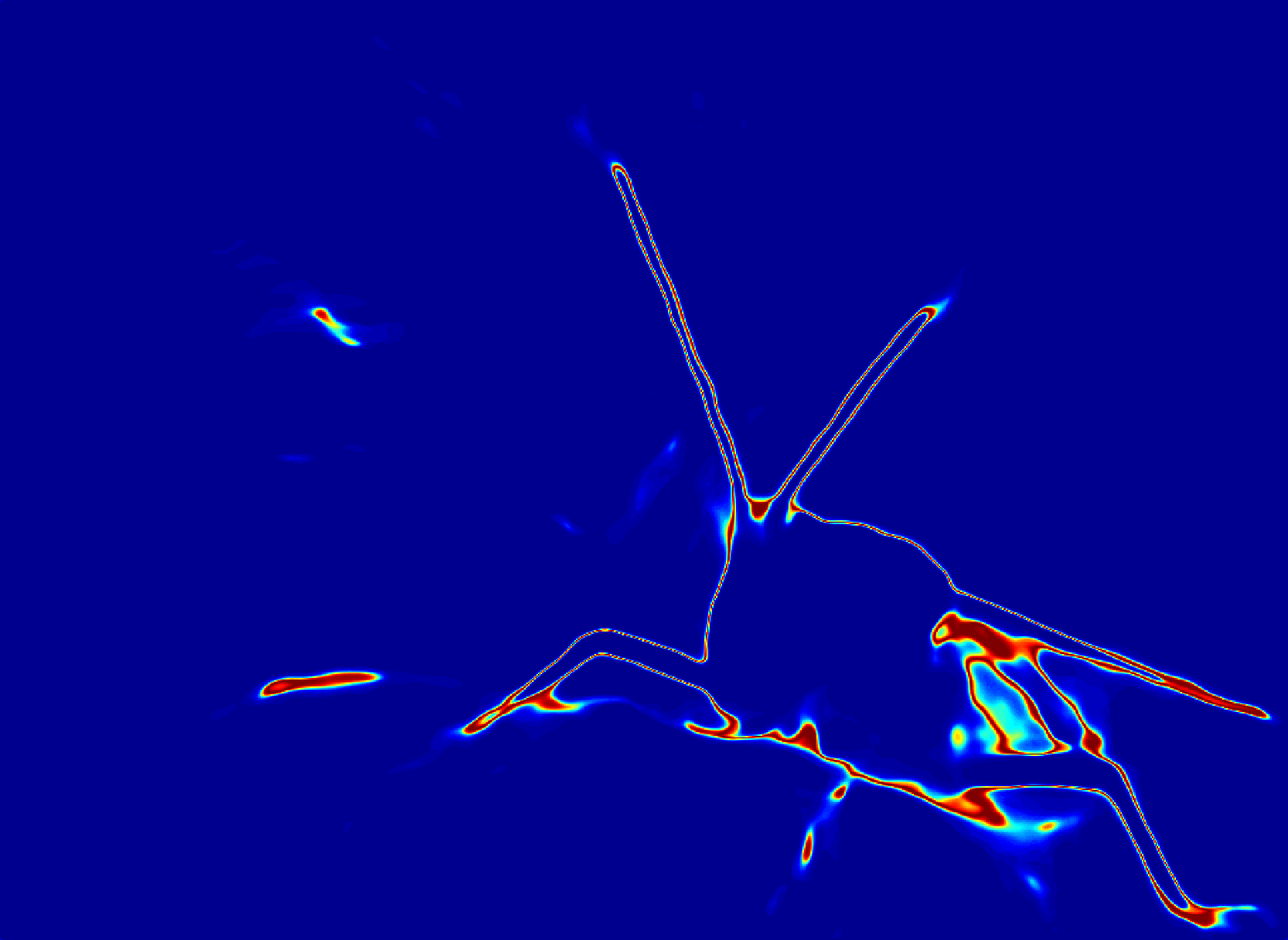}}&
   {\includegraphics[width=0.11\linewidth]{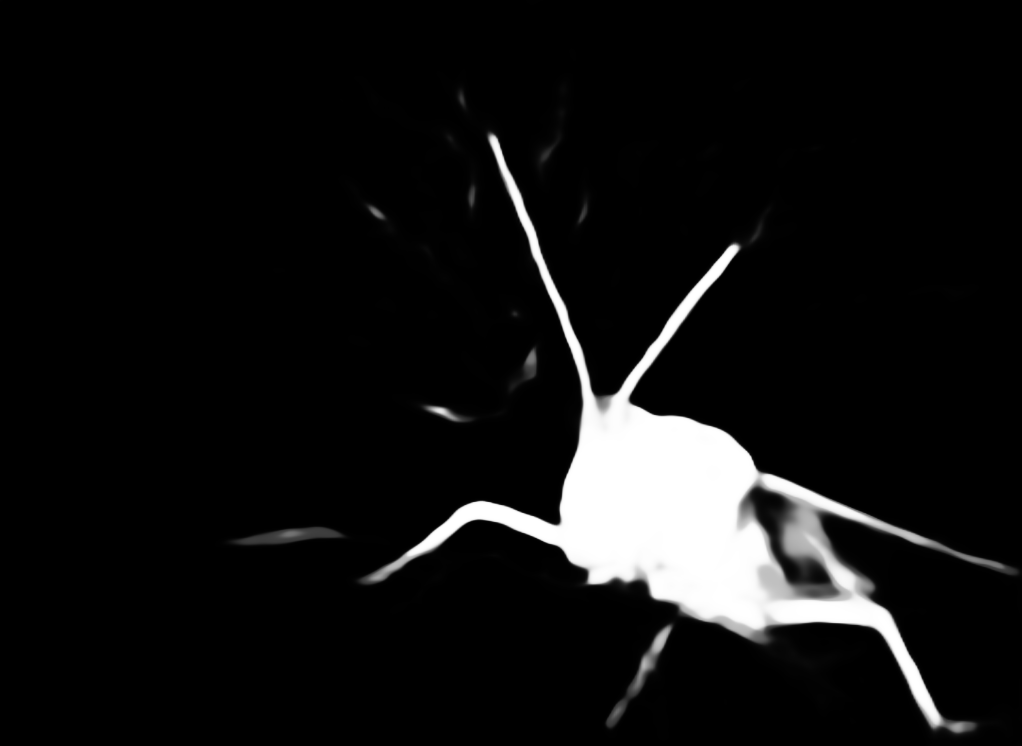}}&
   {\includegraphics[width=0.11\linewidth]{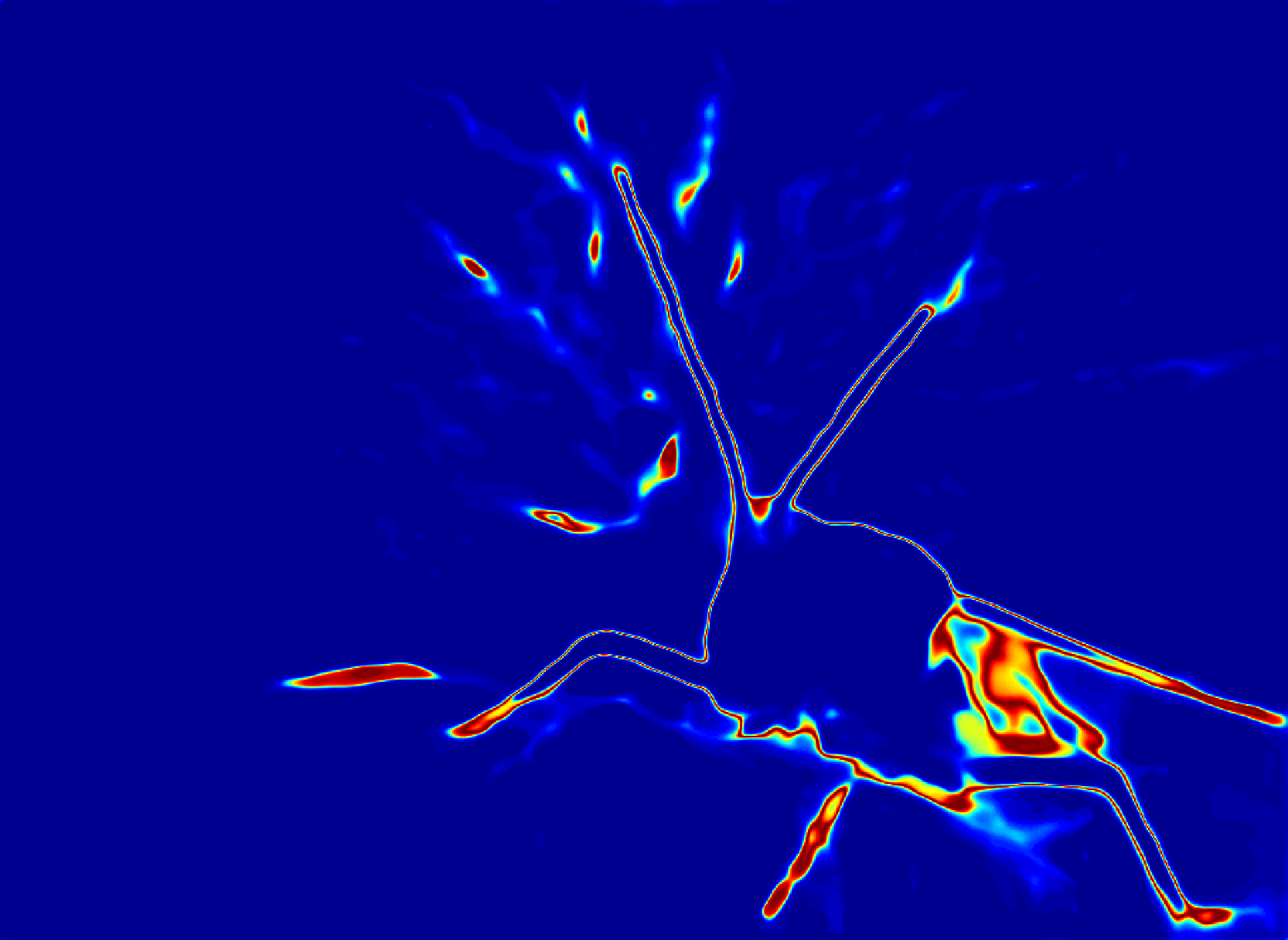}} \\
      {\includegraphics[width=0.11\linewidth]{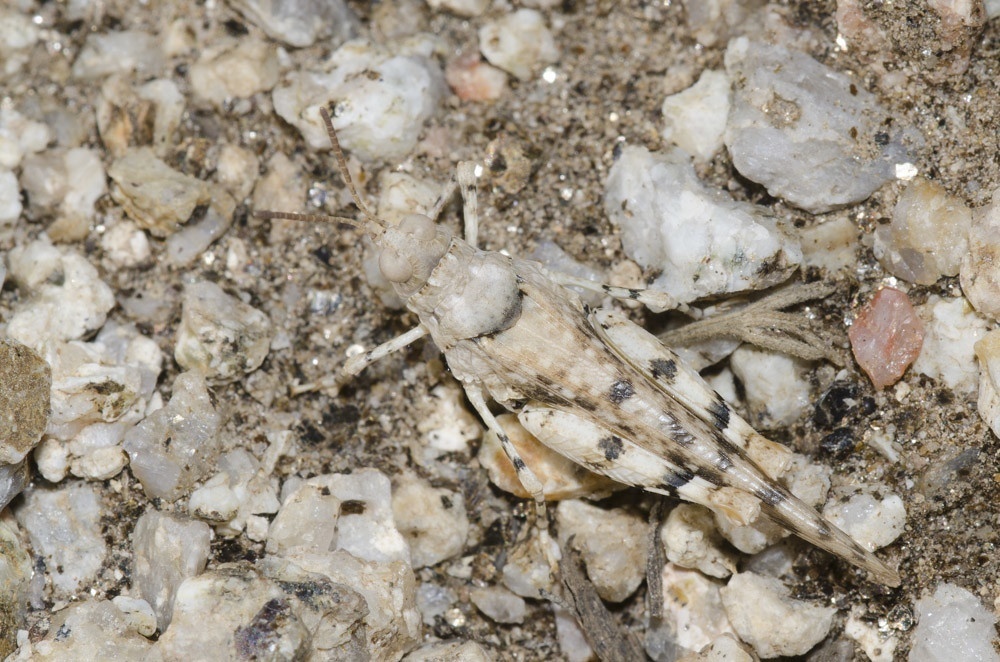}} &
   {\includegraphics[width=0.11\linewidth]{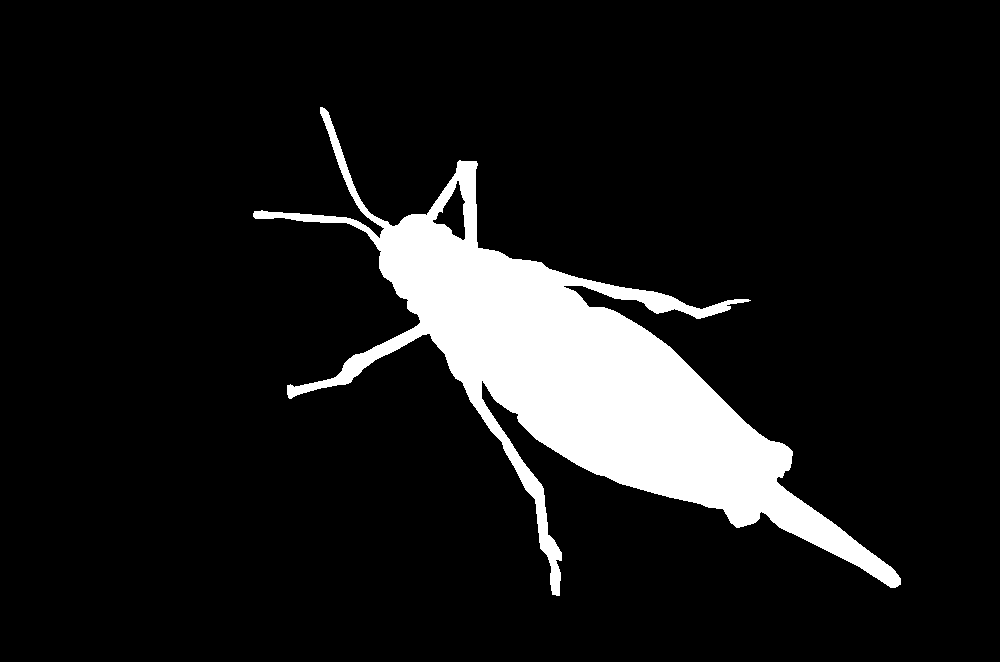}} &
   {\includegraphics[width=0.11\linewidth]{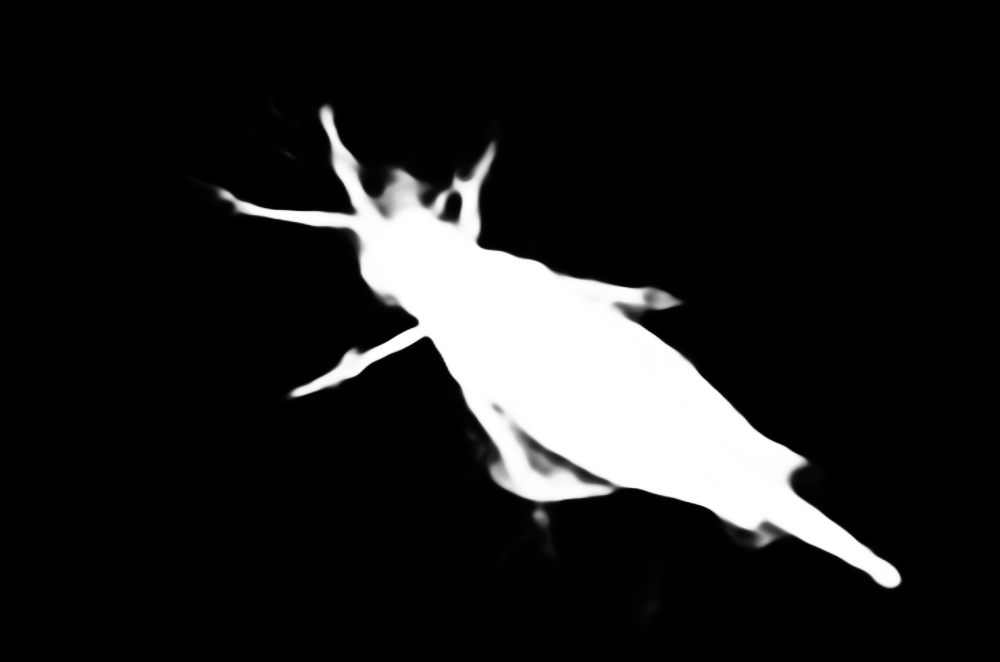}} &
   {\includegraphics[width=0.11\linewidth]{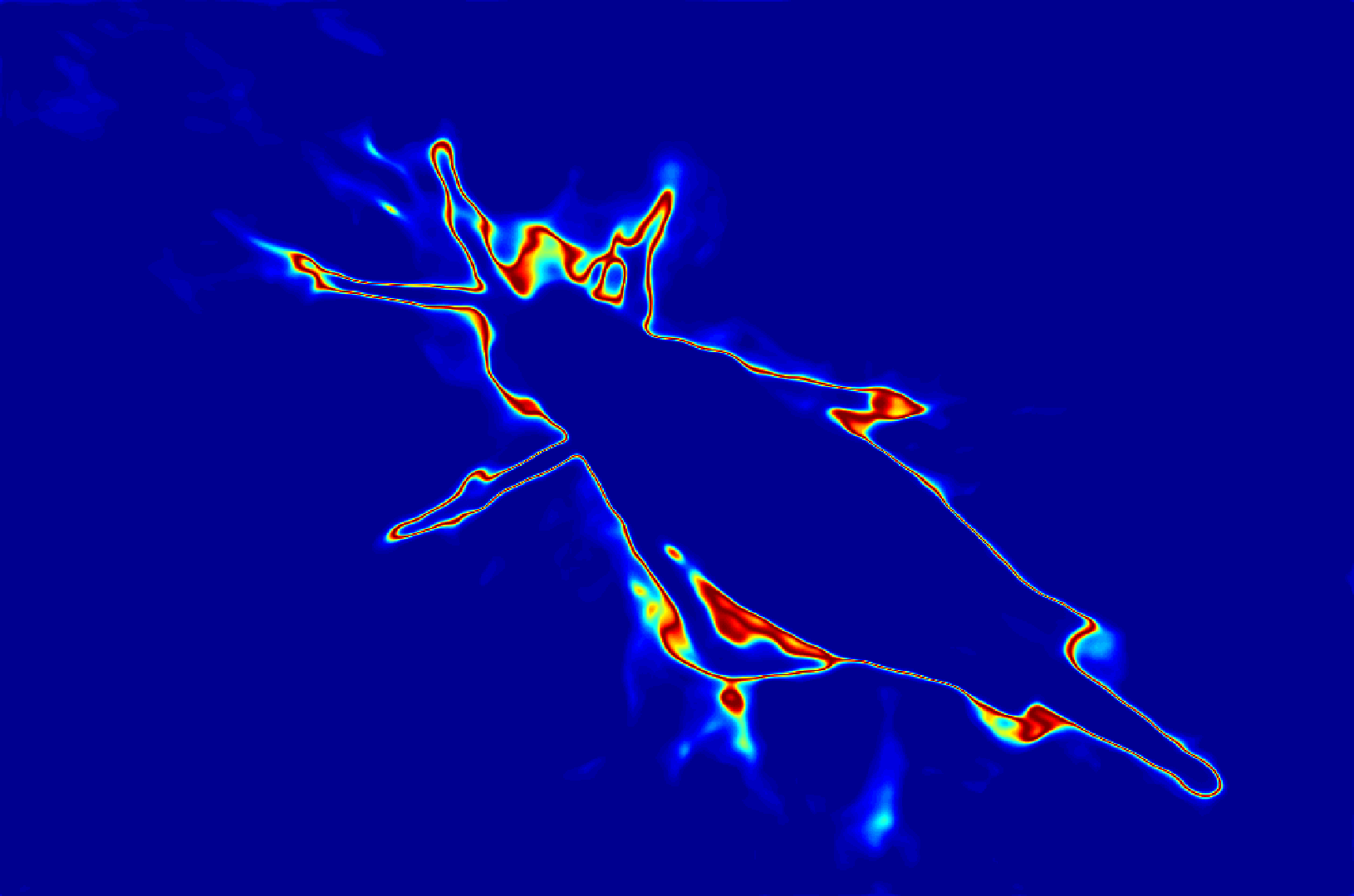}} &
   {\includegraphics[width=0.11\linewidth]{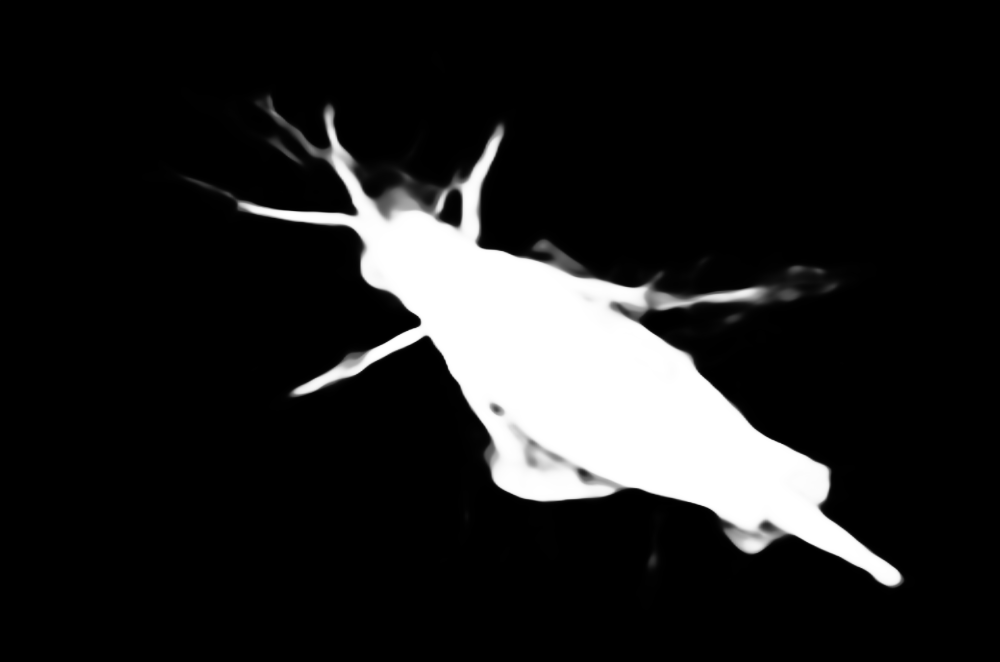}}&
   {\includegraphics[width=0.11\linewidth]{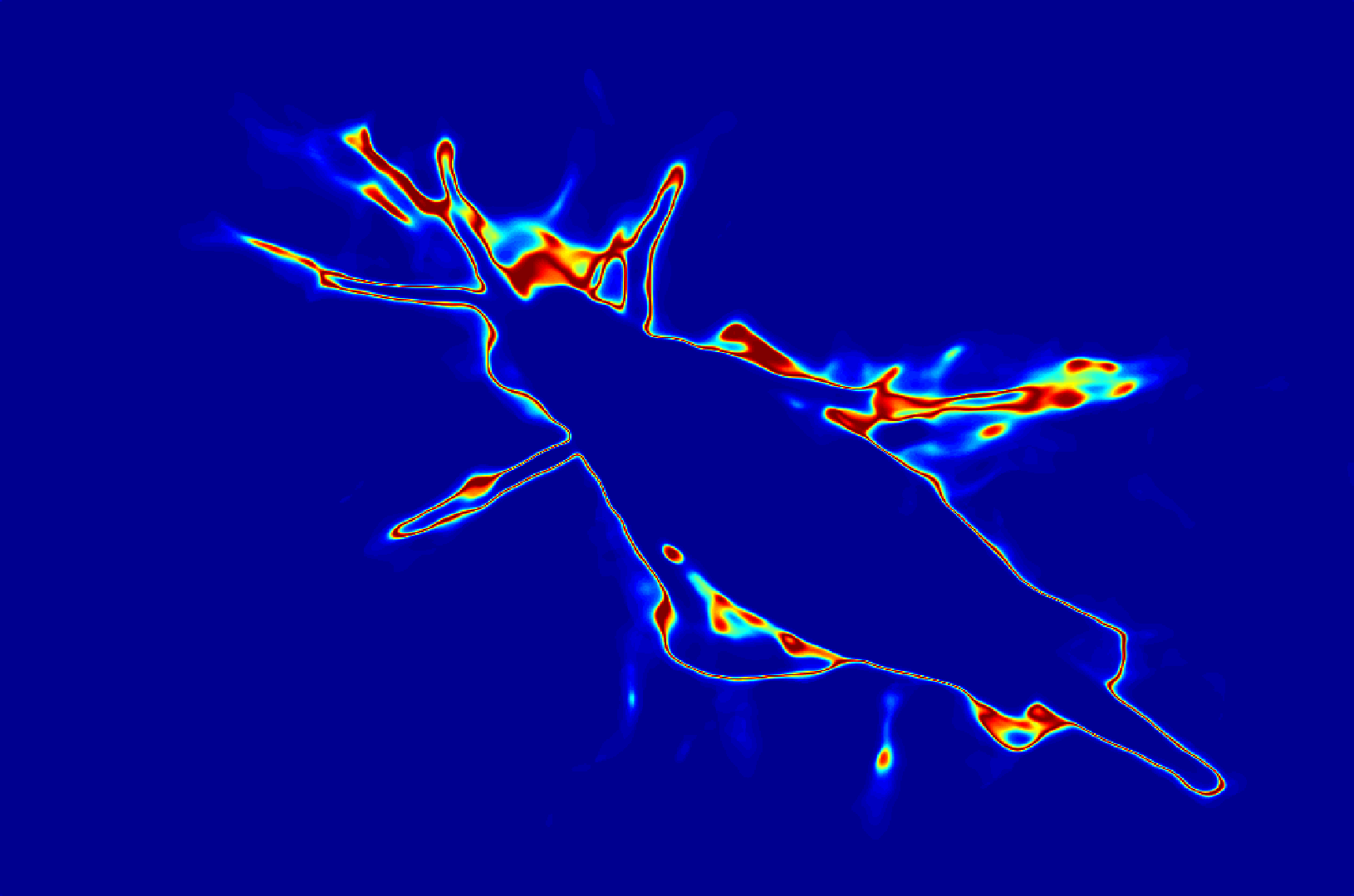}}&
   {\includegraphics[width=0.11\linewidth]{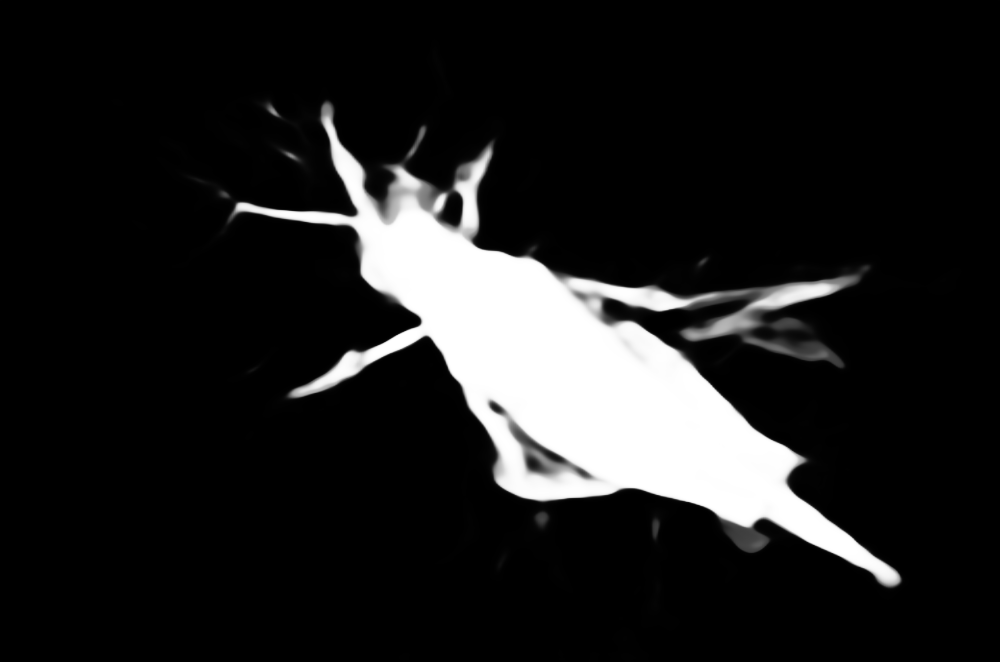}}&
   {\includegraphics[width=0.11\linewidth]{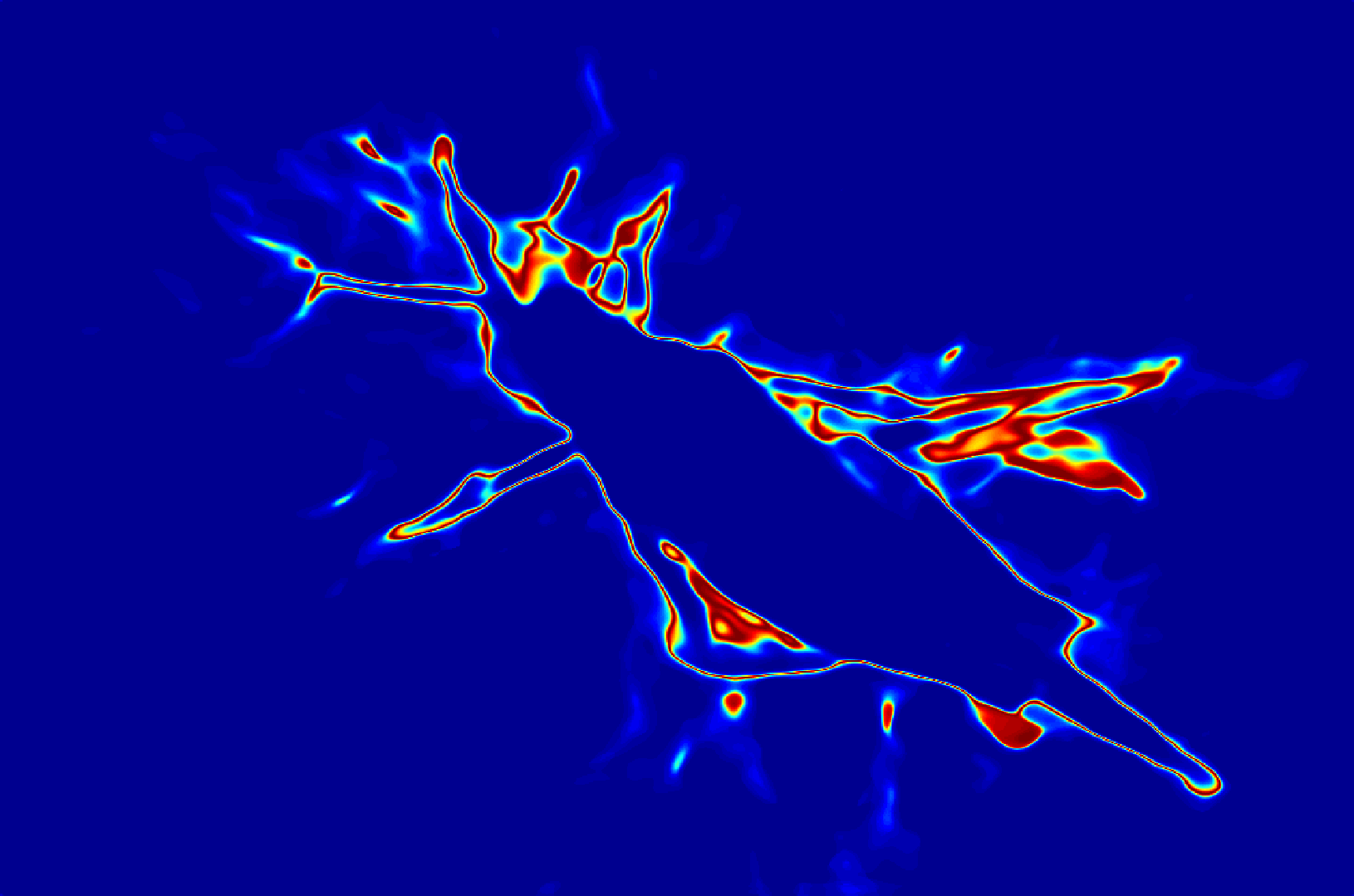}} \\
      {\includegraphics[width=0.11\linewidth]{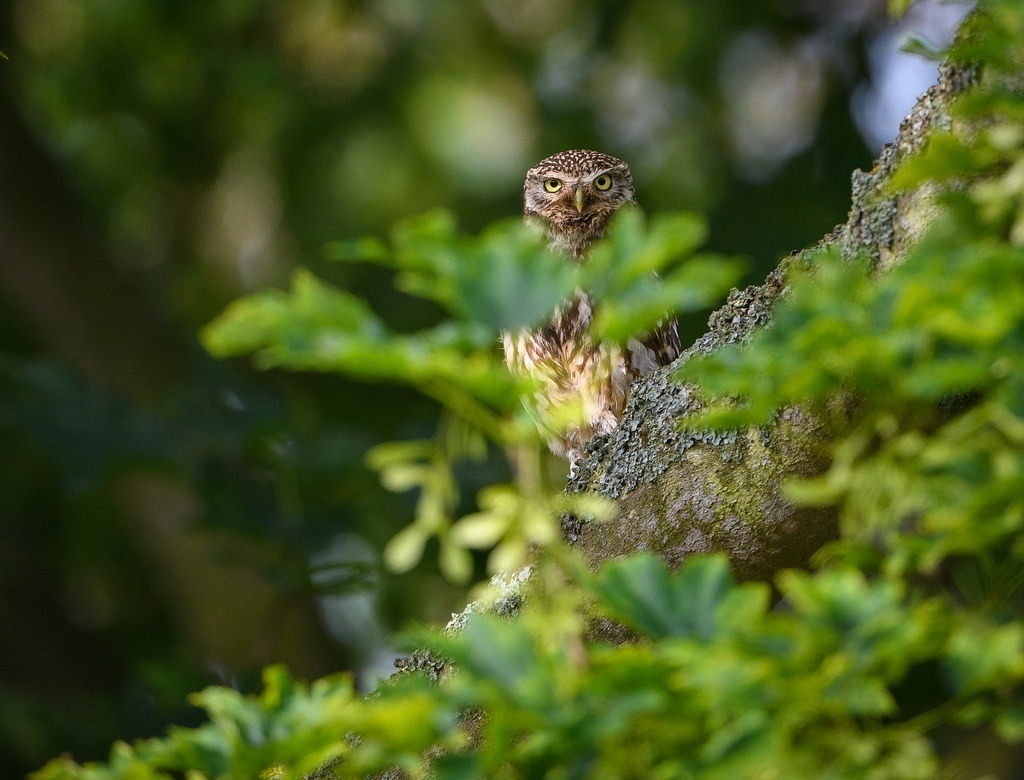}} &
   {\includegraphics[width=0.11\linewidth]{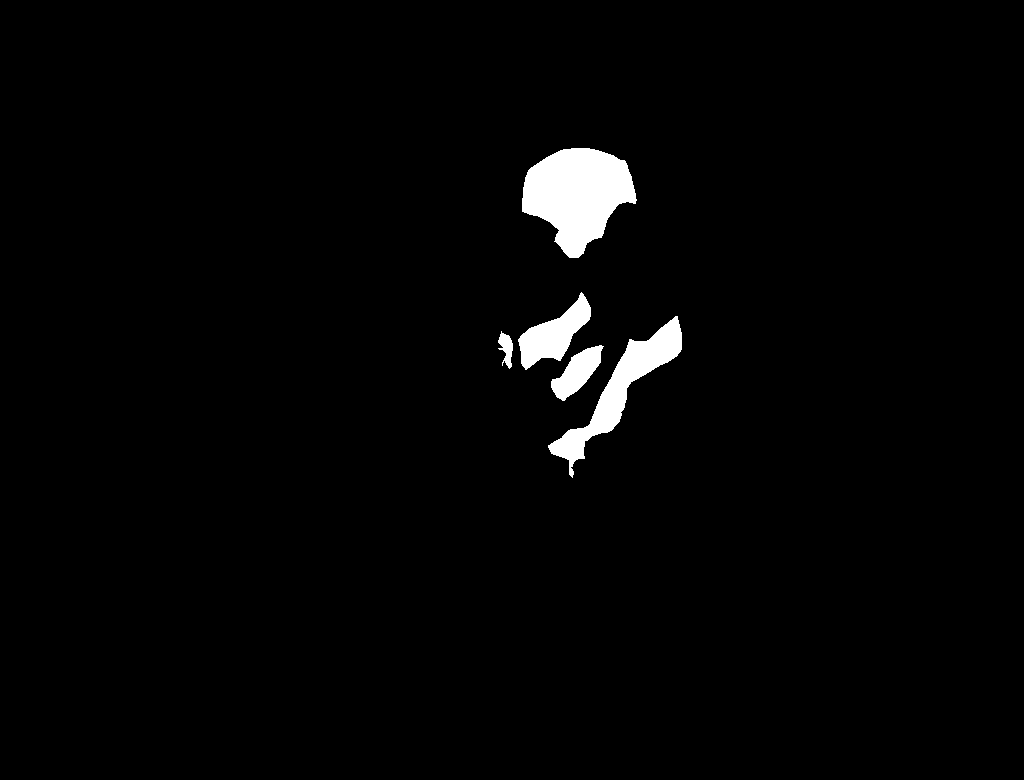}} &
   {\includegraphics[width=0.11\linewidth]{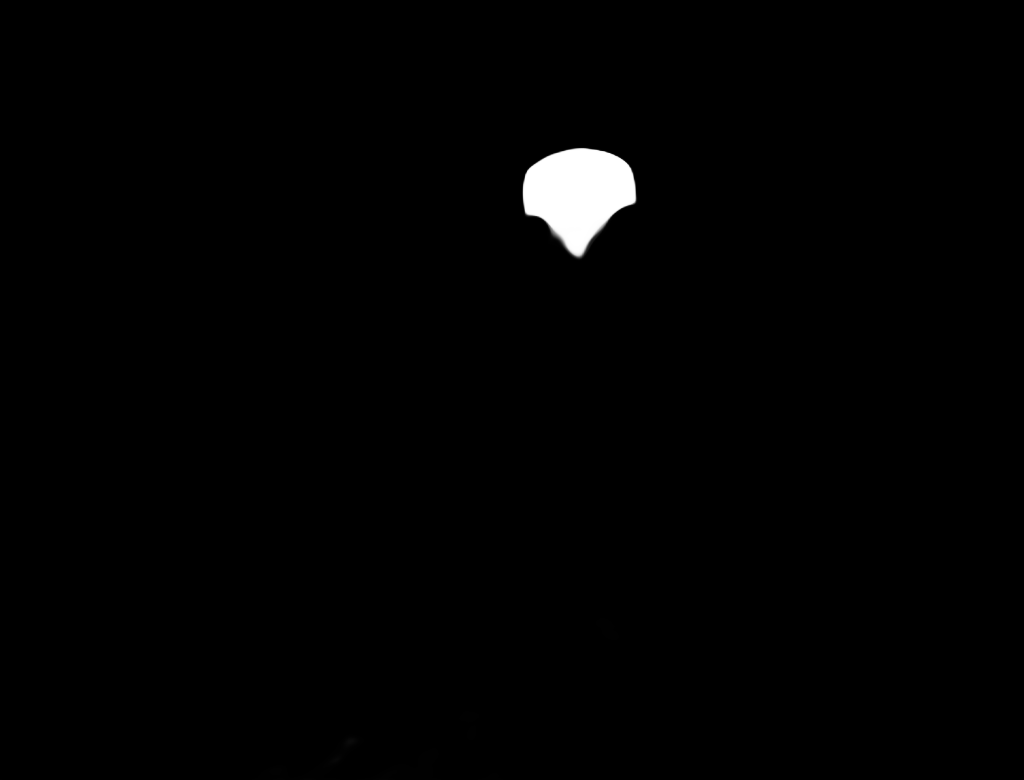}} &
   {\includegraphics[width=0.11\linewidth]{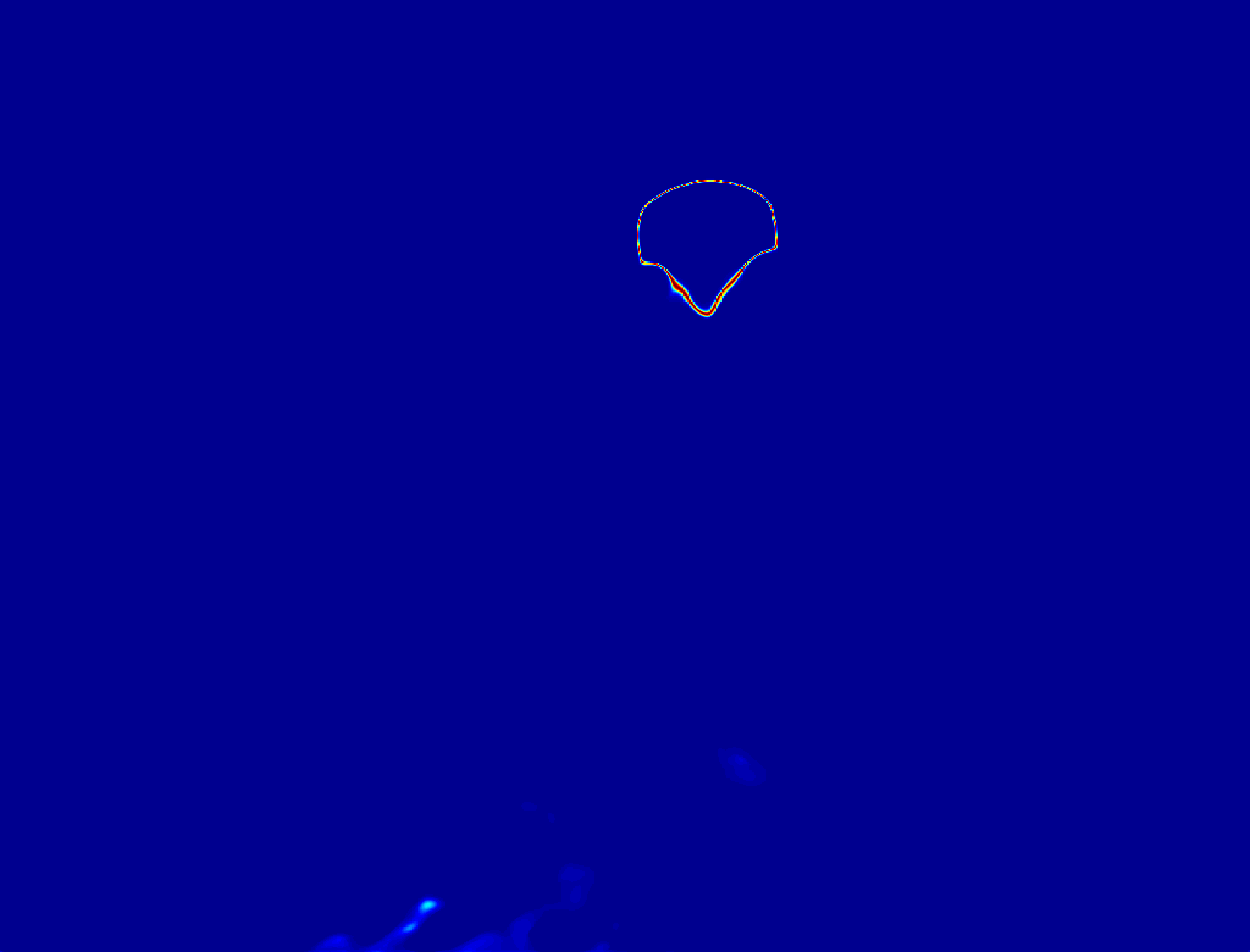}} &
   {\includegraphics[width=0.11\linewidth]{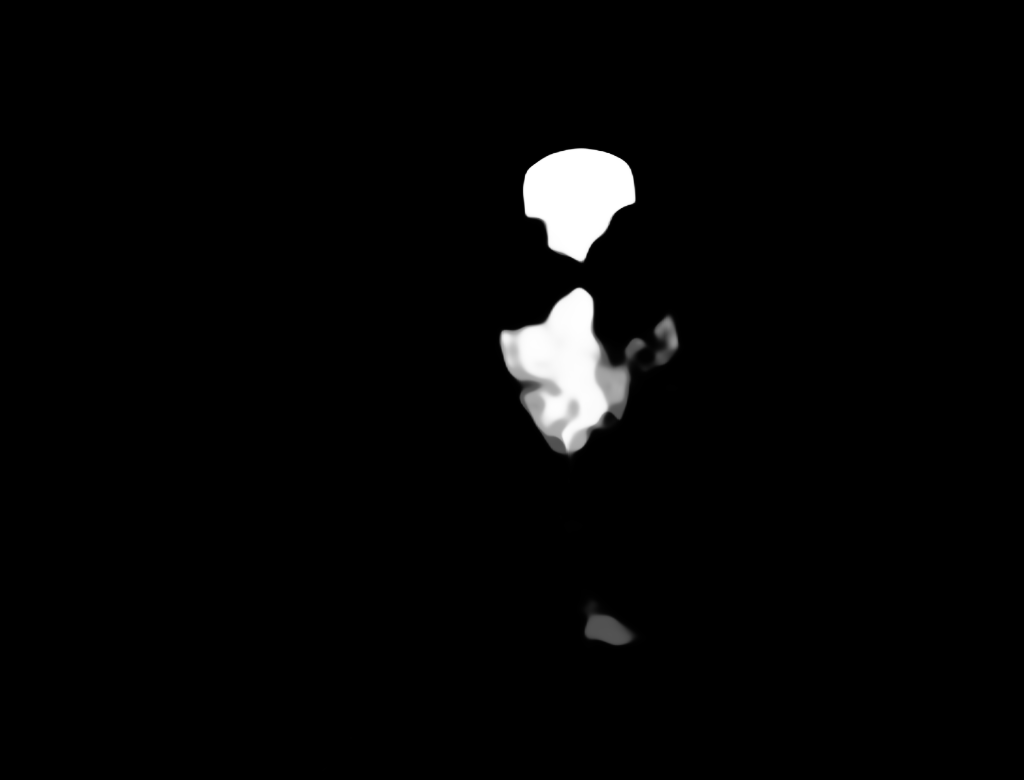}}&
   {\includegraphics[width=0.11\linewidth]{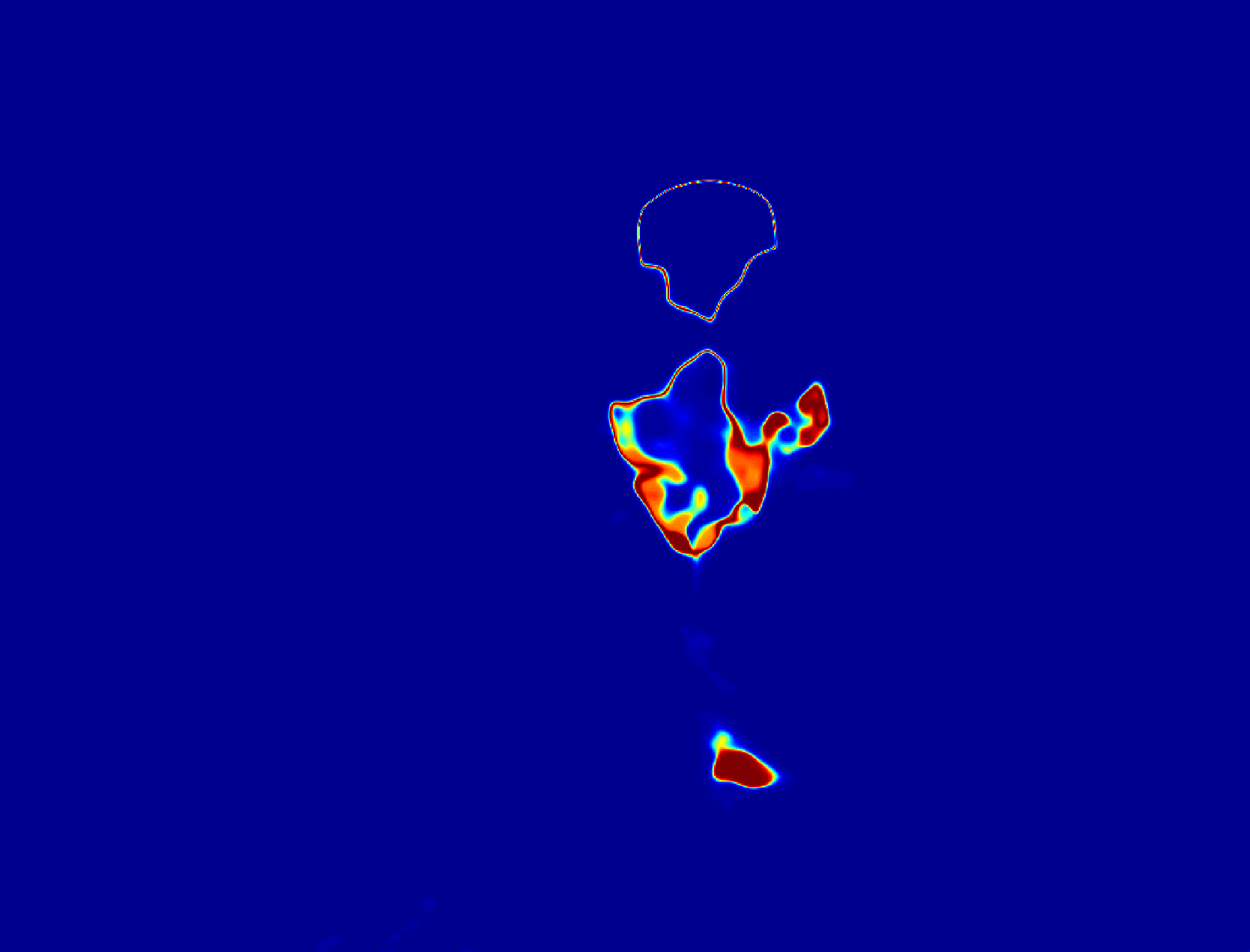}}&
   {\includegraphics[width=0.11\linewidth]{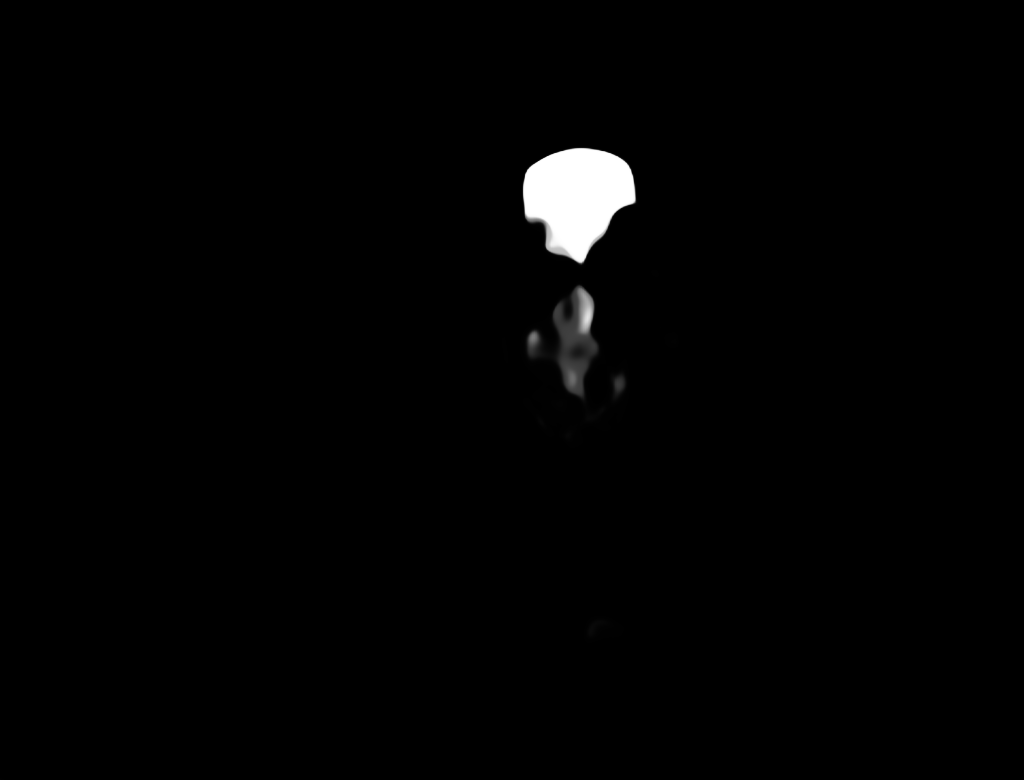}}&
   {\includegraphics[width=0.11\linewidth]{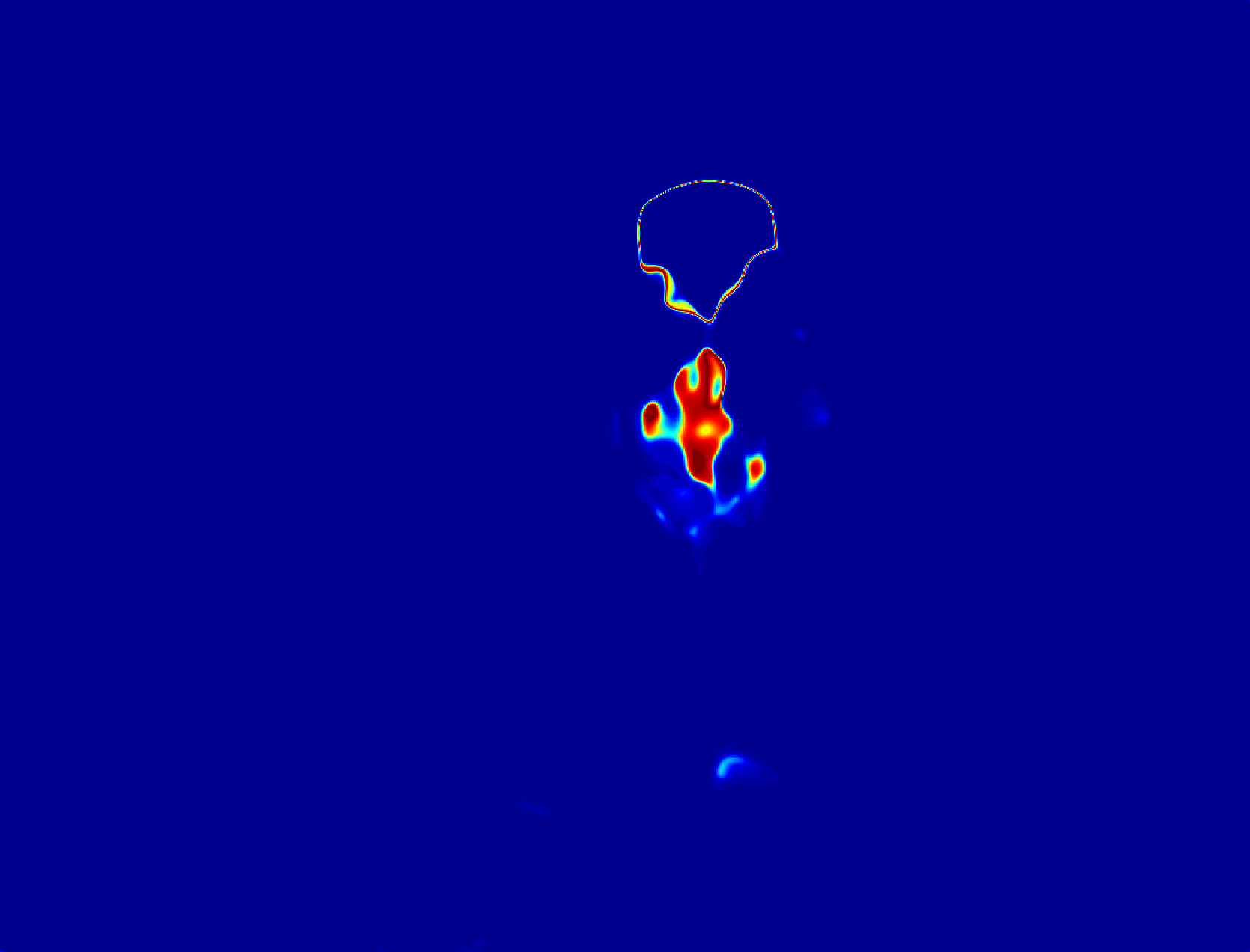}} \\
      {\includegraphics[width=0.11\linewidth]{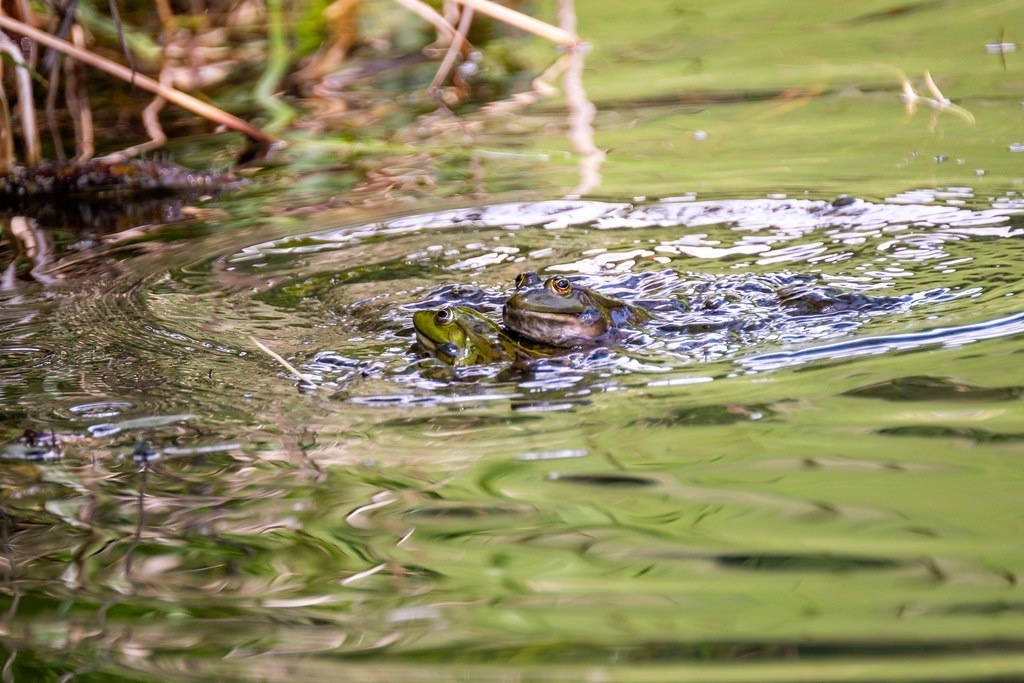}} &
   {\includegraphics[width=0.11\linewidth]{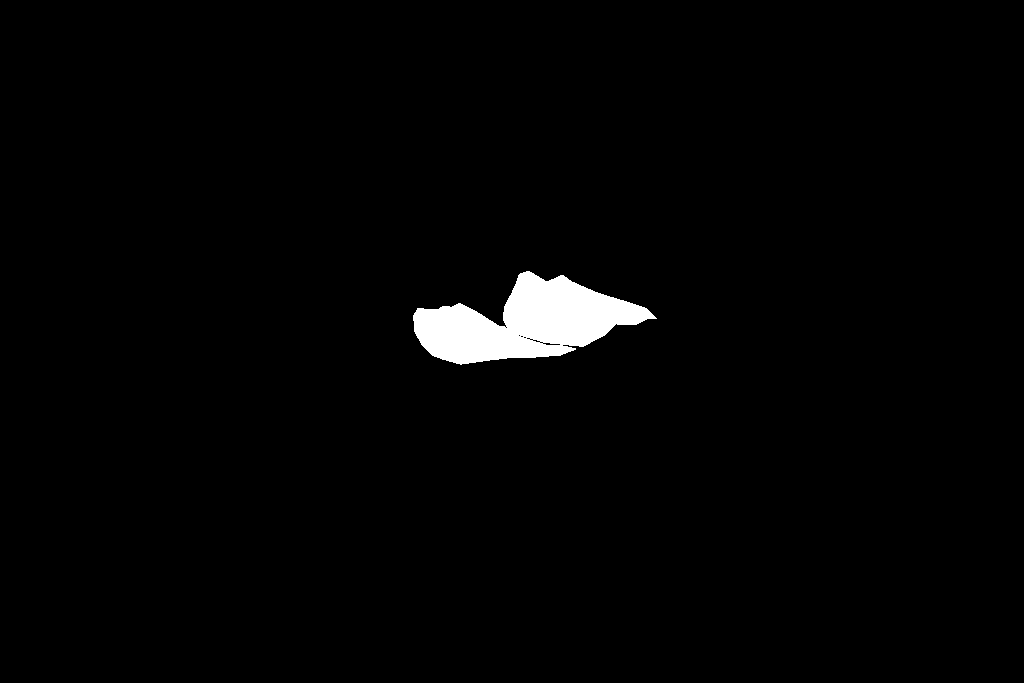}} &
   {\includegraphics[width=0.11\linewidth]{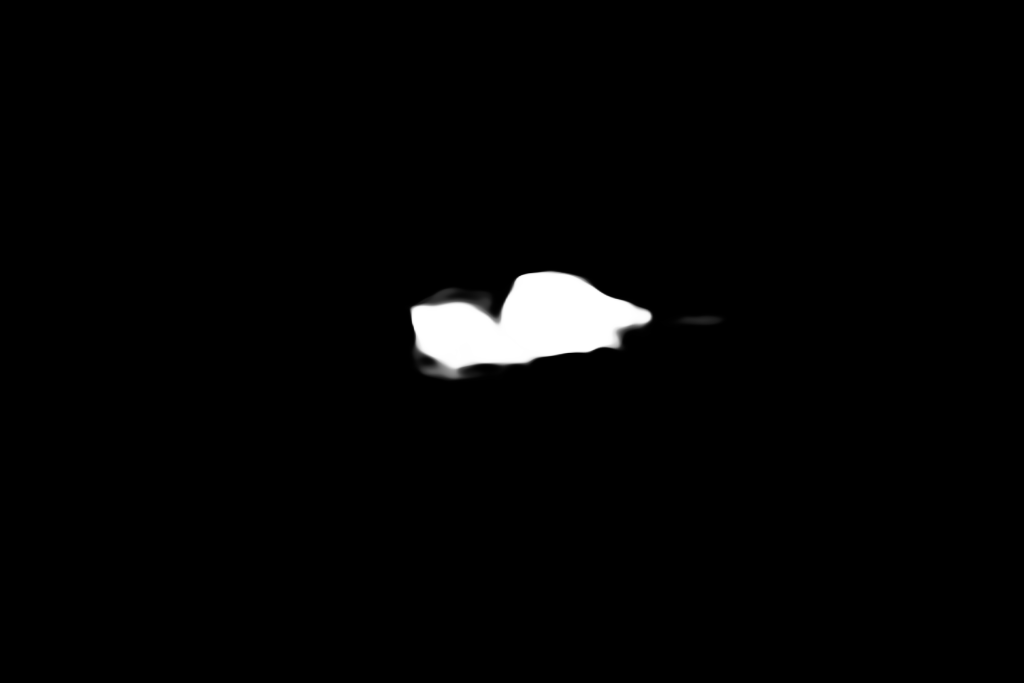}} &
   {\includegraphics[width=0.11\linewidth]{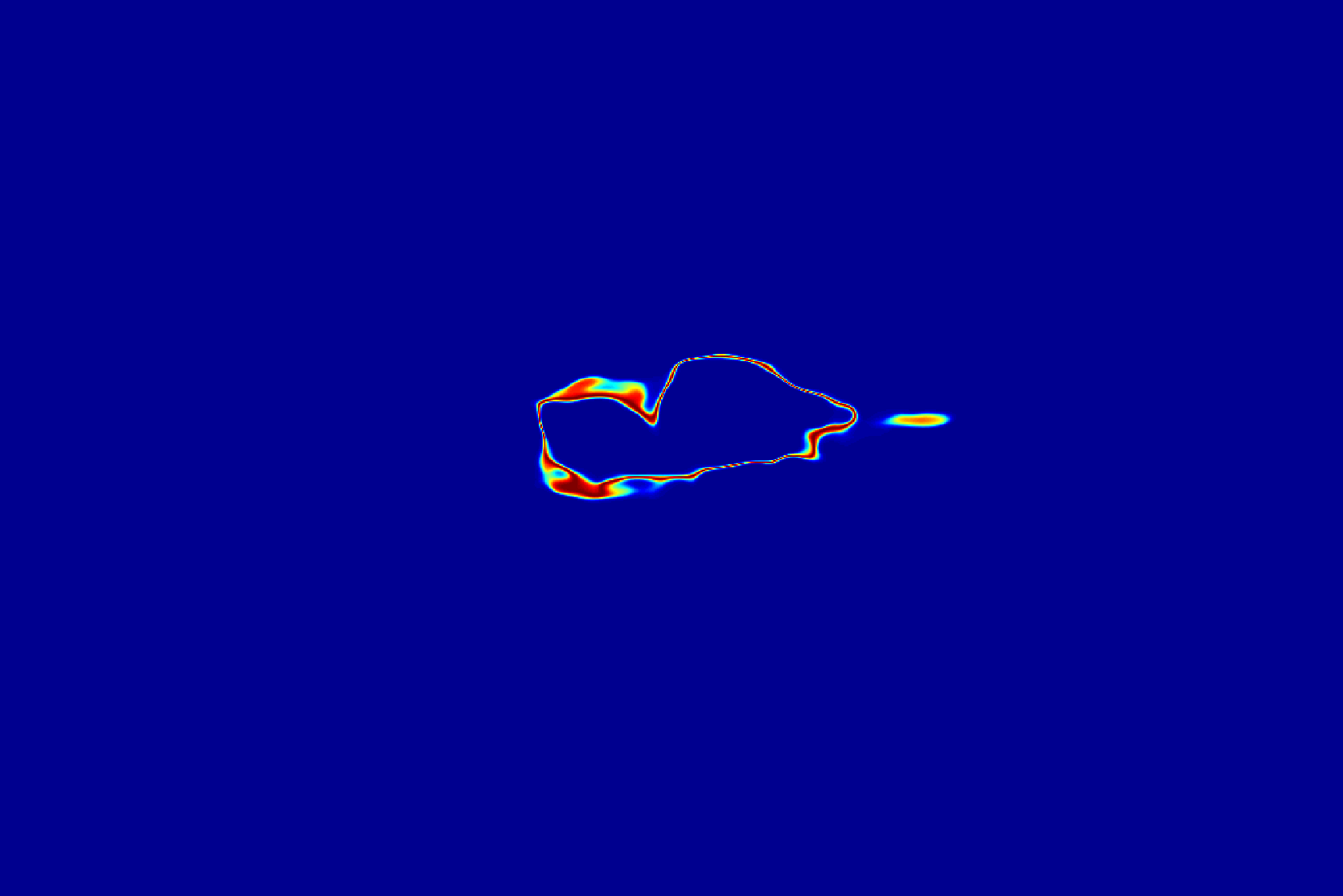}} &
   {\includegraphics[width=0.11\linewidth]{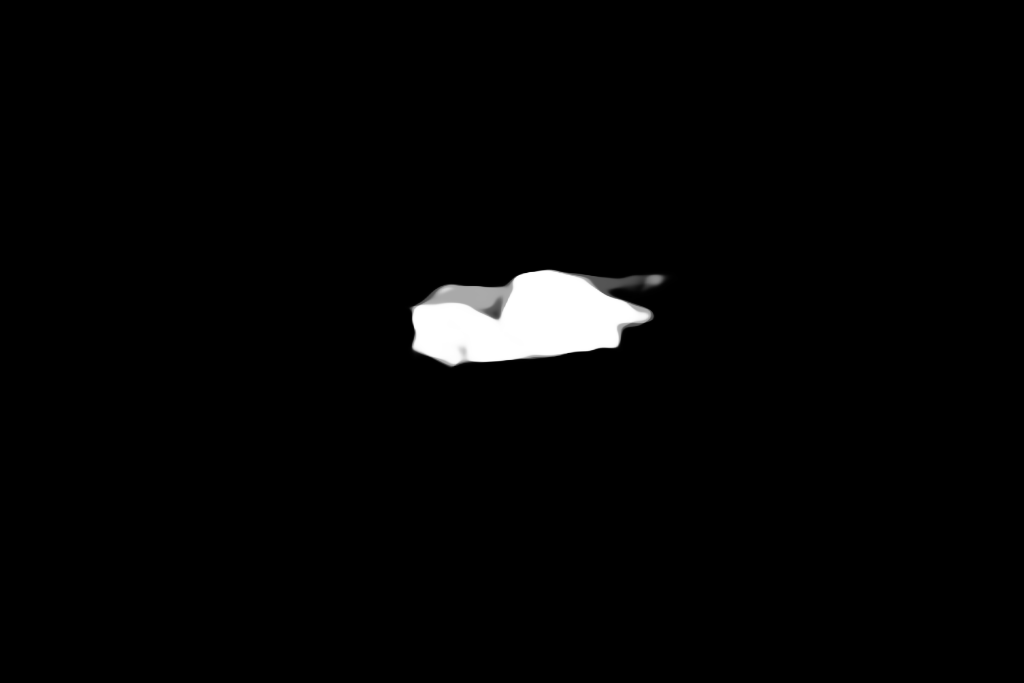}}&
   {\includegraphics[width=0.11\linewidth]{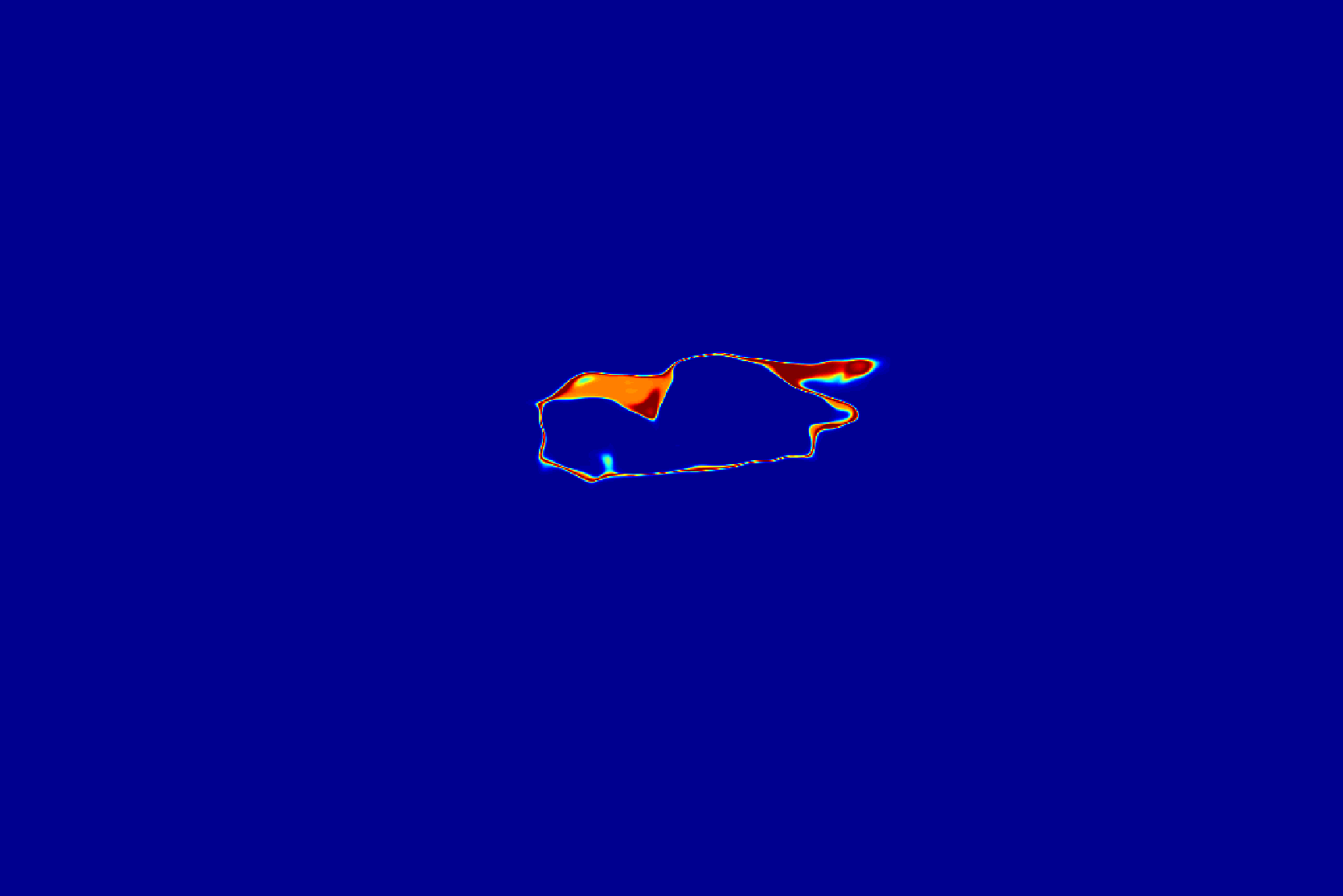}}&
   {\includegraphics[width=0.11\linewidth]{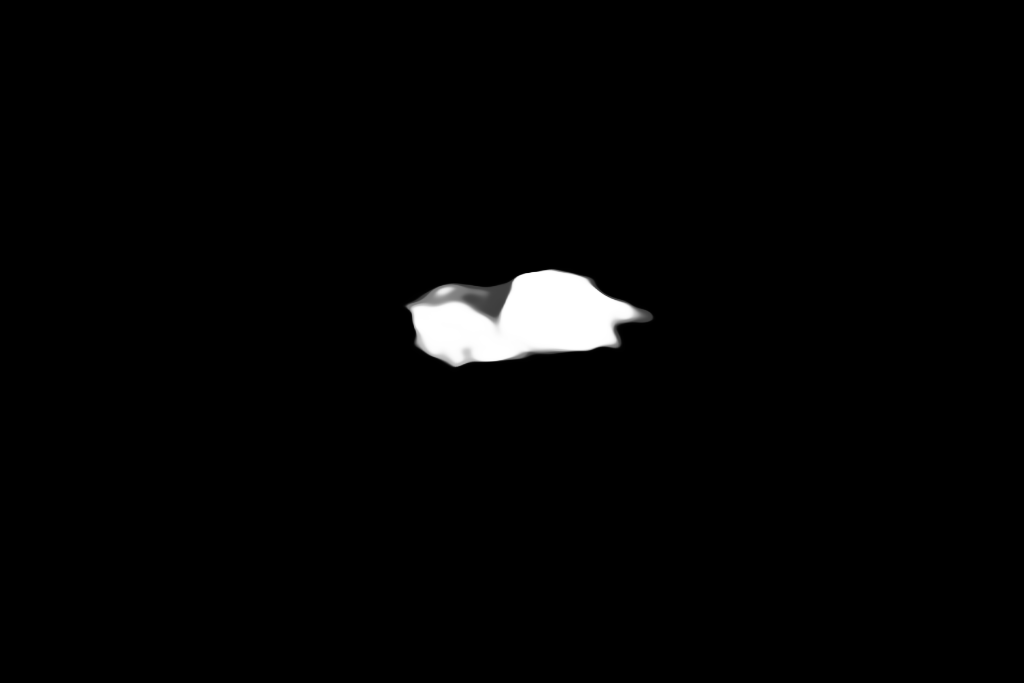}}&
   {\includegraphics[width=0.11\linewidth]{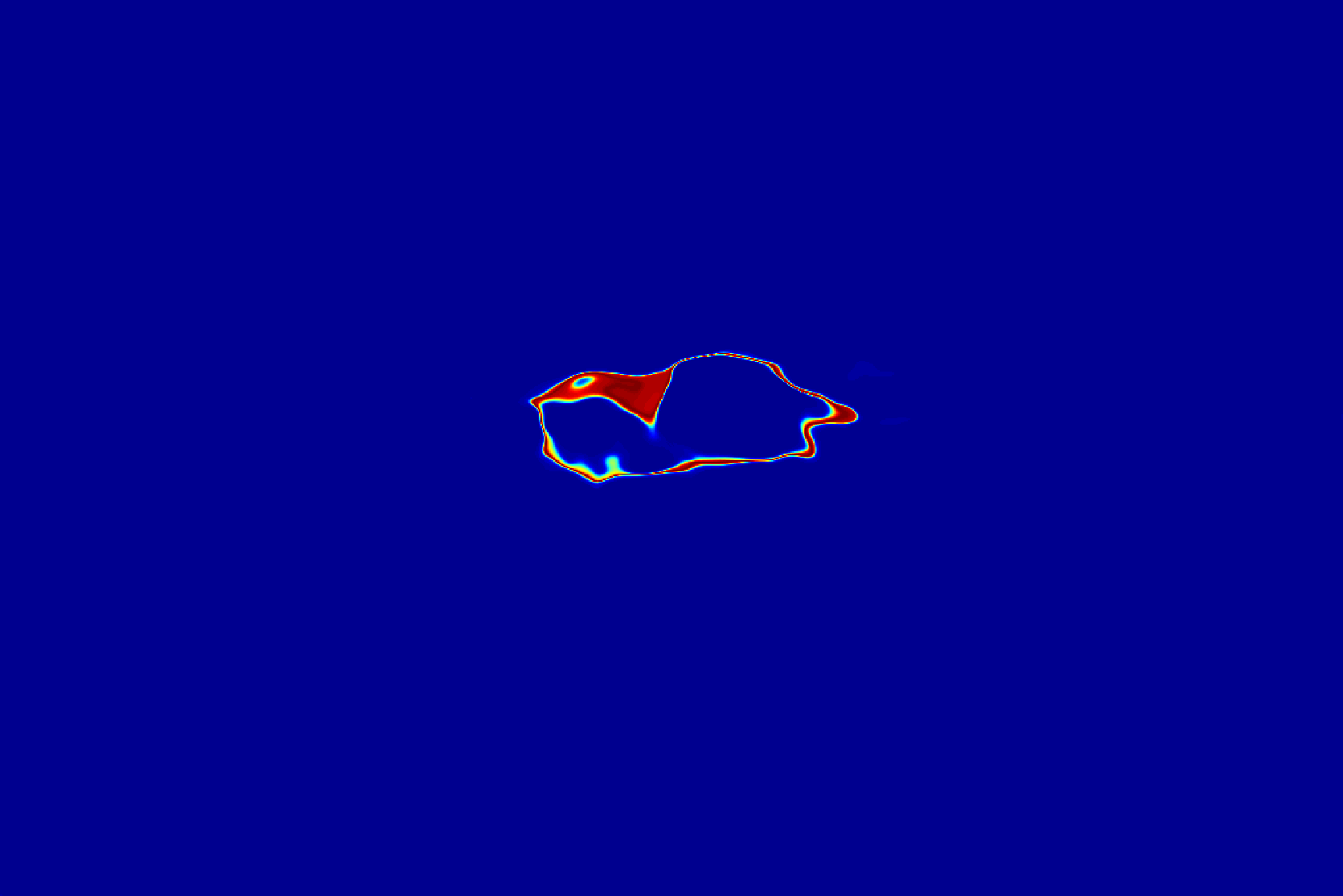}} \\
   \footnotesize{Image}&\footnotesize{GT}&\footnotesize{MD}&\footnotesize{$U_p$}&\footnotesize{DE}&\footnotesize{$U_p$}&\footnotesize{SE}&\footnotesize{$U_p$}\\
   \end{tabular}
   \end{center}
   \caption{\footnotesize{Predictive uncertainty of ensemble based solutions for \textbf{camouflaged object detection}.}
   }
\label{fig:predictive_ensemble_cod}
\end{figure}

\begin{figure}[tp]
   \begin{center}
   \begin{tabular}{c@{ }c@{ }c@{ }c@{ }c@{ }c@{ }c@{ }c@{ }}
   {\includegraphics[width=0.11\linewidth]{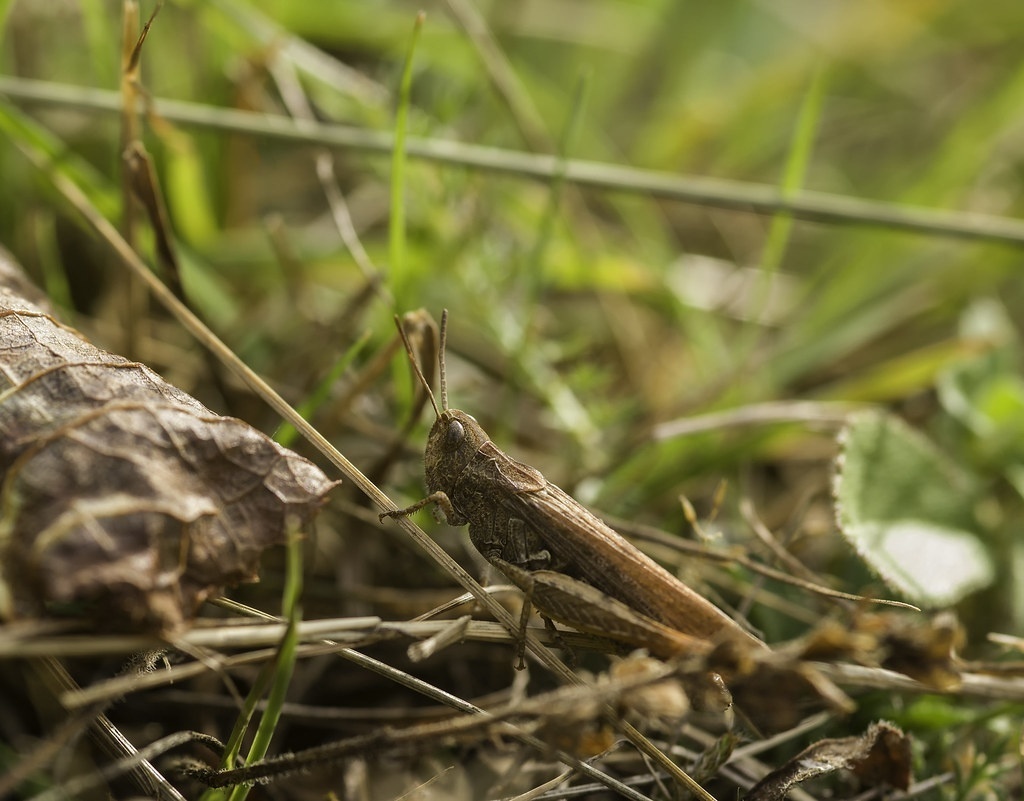}} &
   {\includegraphics[width=0.11\linewidth]{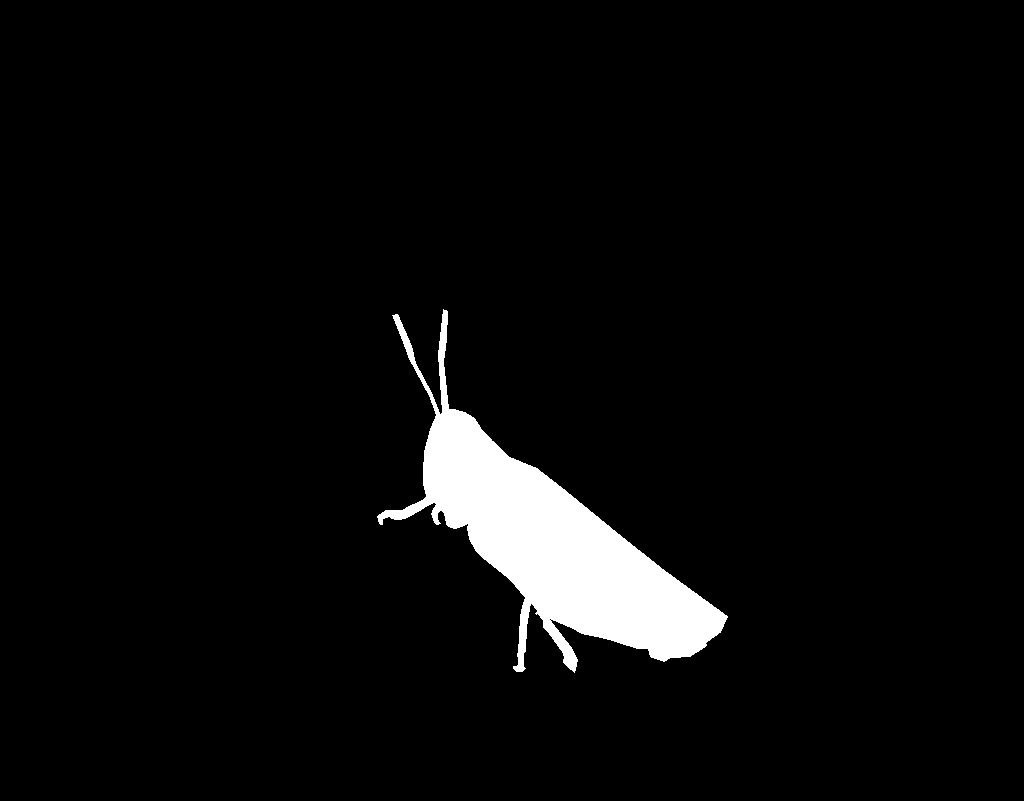}} &
   {\includegraphics[width=0.11\linewidth]{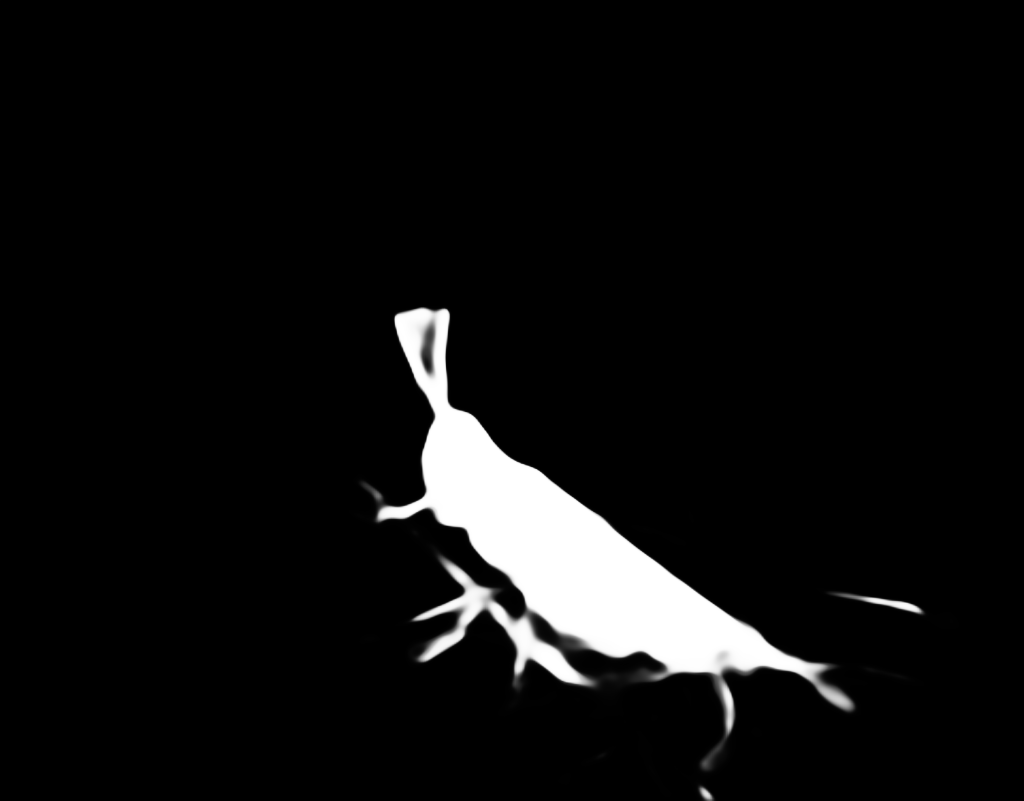}} &
   {\includegraphics[width=0.11\linewidth]{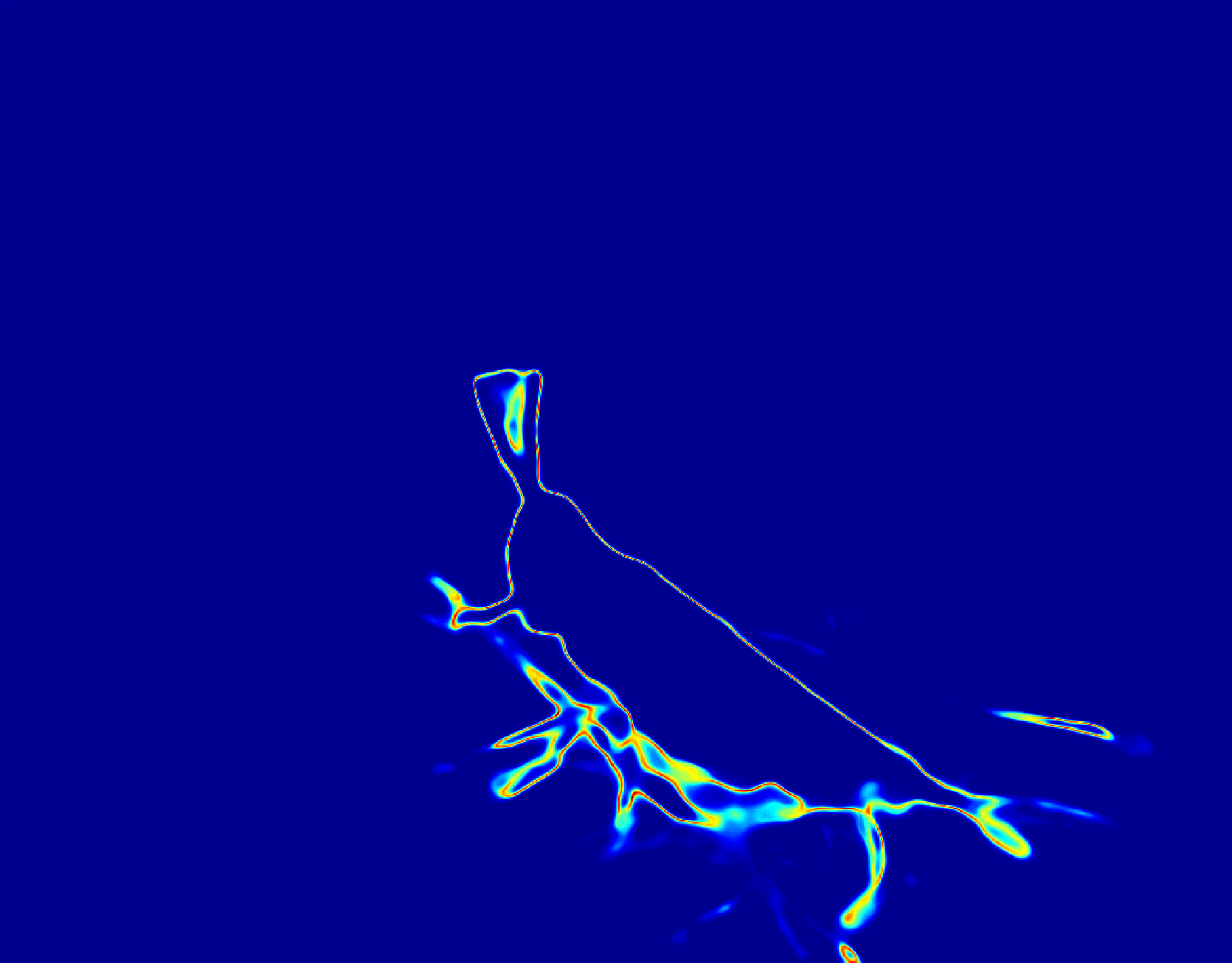}} &
   {\includegraphics[width=0.11\linewidth]{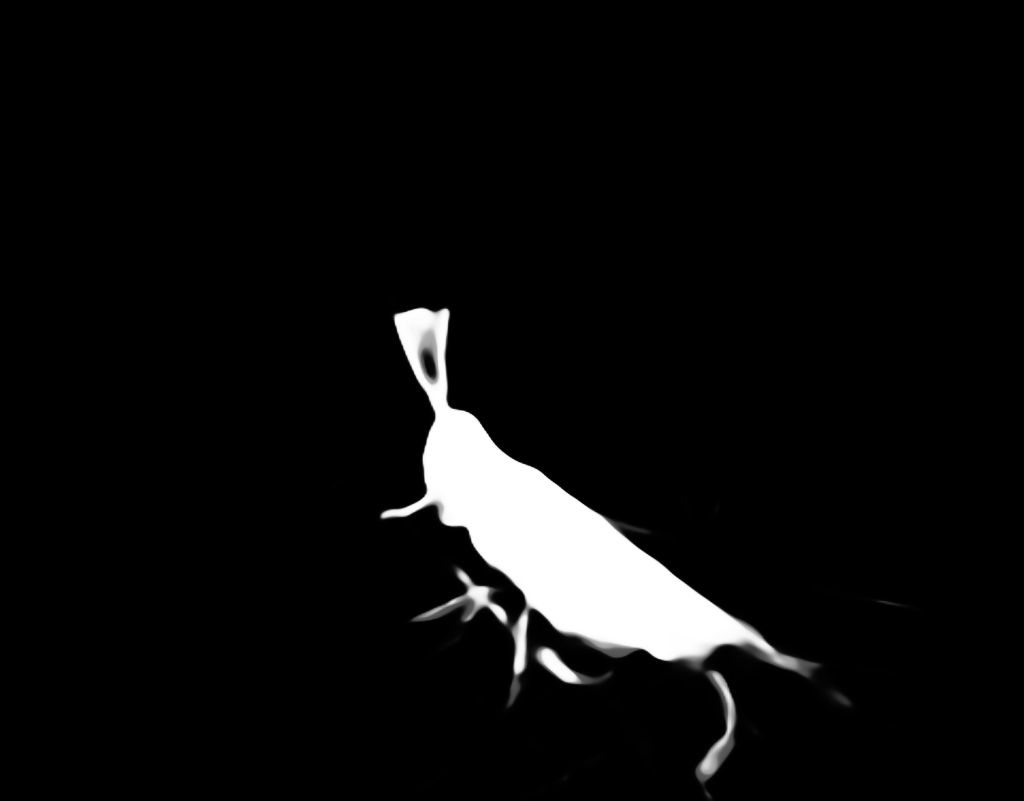}}&
   {\includegraphics[width=0.11\linewidth]{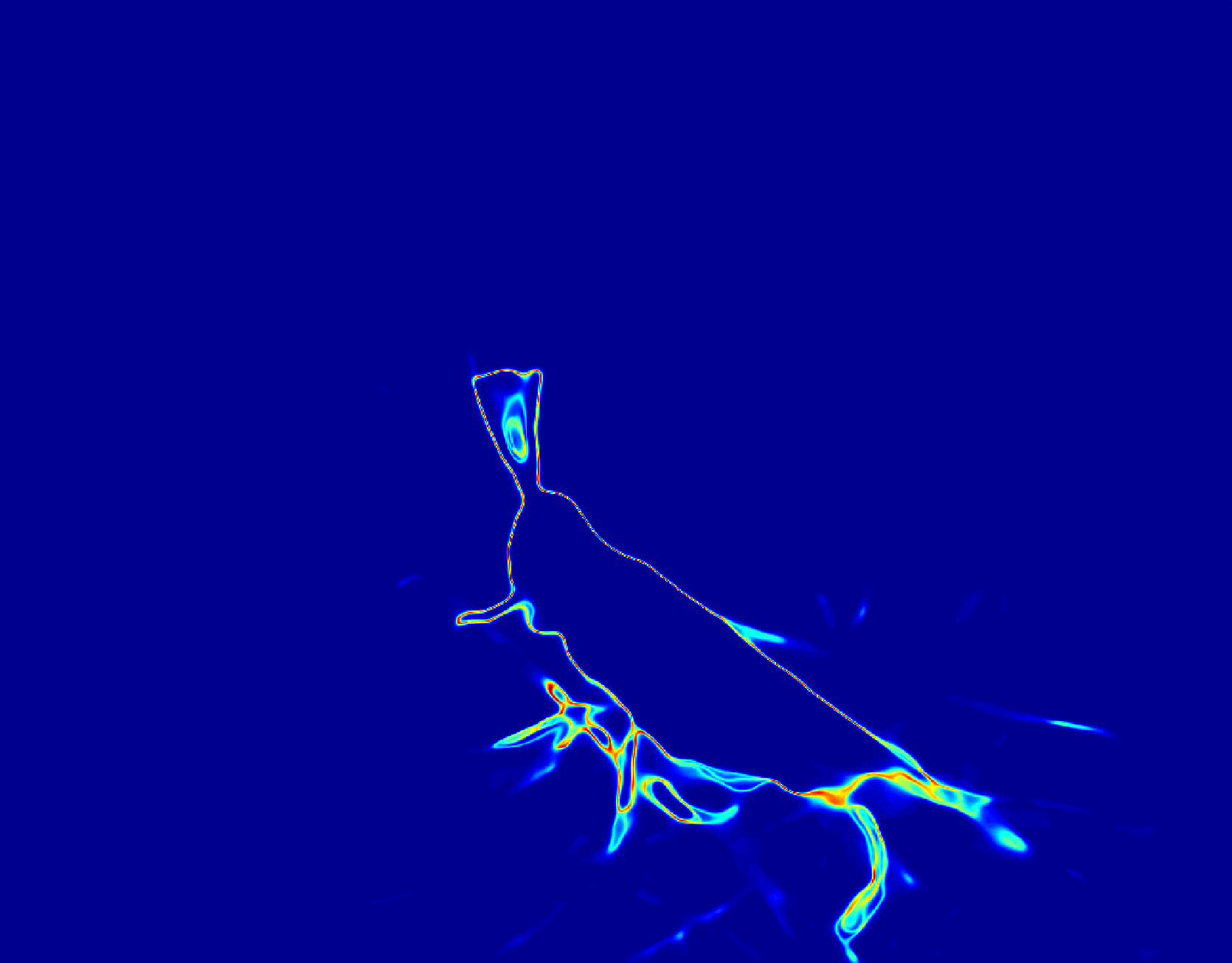}}&
   {\includegraphics[width=0.11\linewidth]{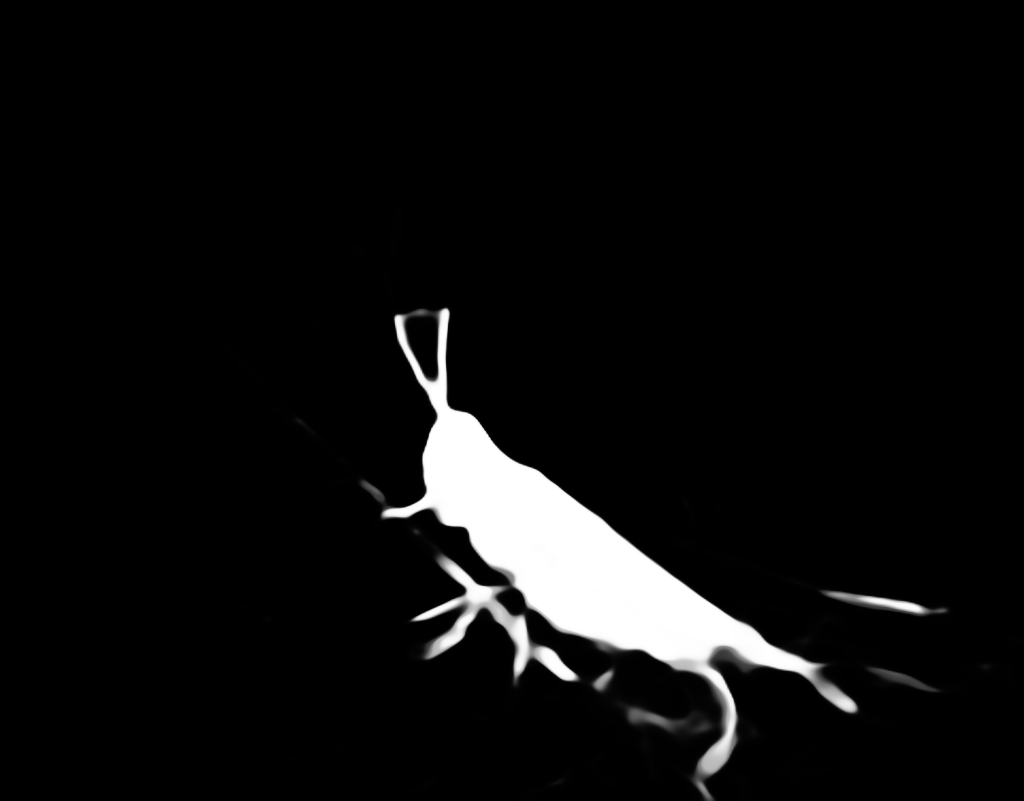}}&
   {\includegraphics[width=0.11\linewidth]{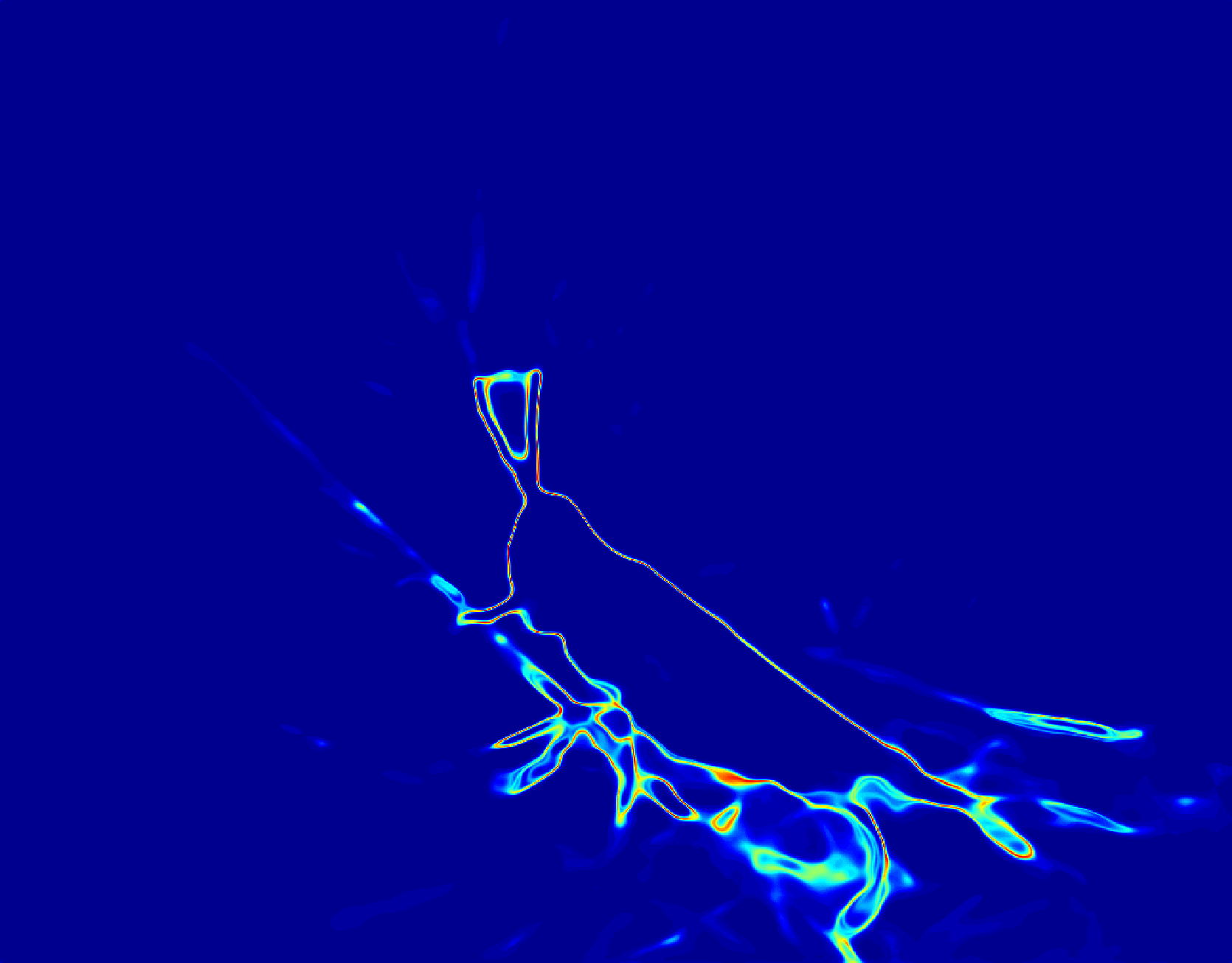}} \\
      {\includegraphics[width=0.11\linewidth]{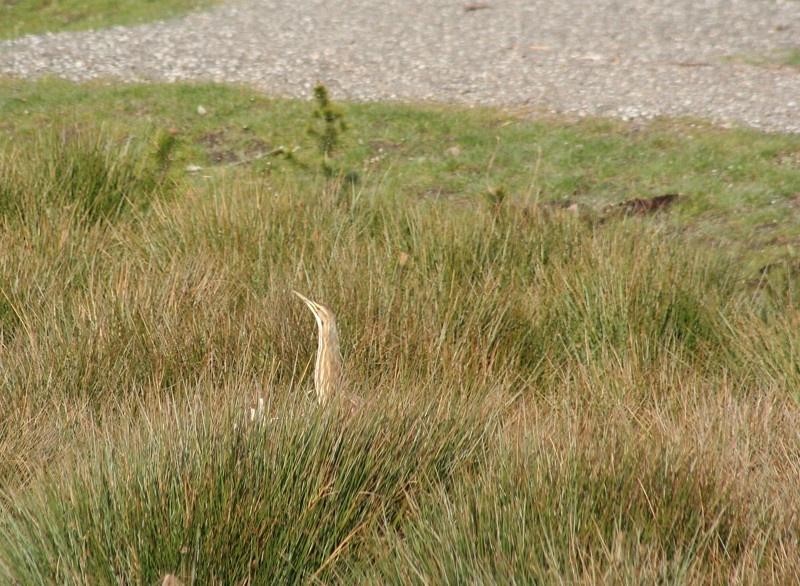}} &
   {\includegraphics[width=0.11\linewidth]{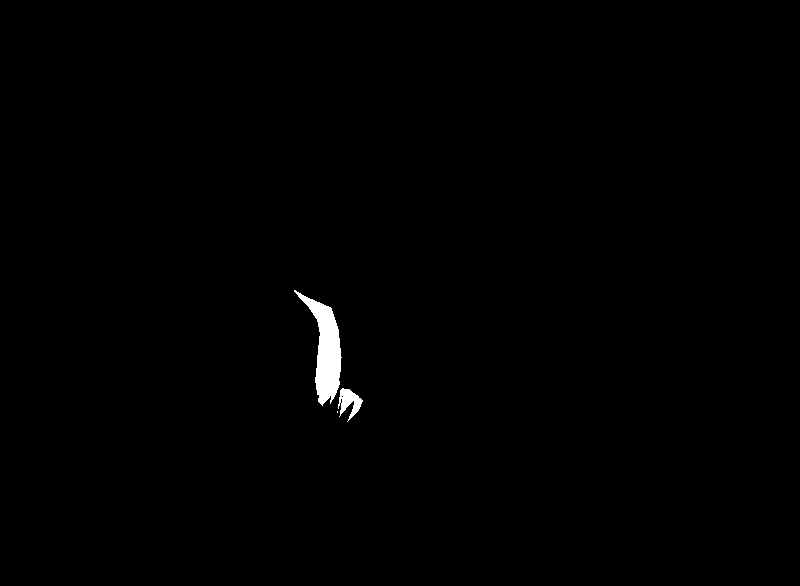}} &
   {\includegraphics[width=0.11\linewidth]{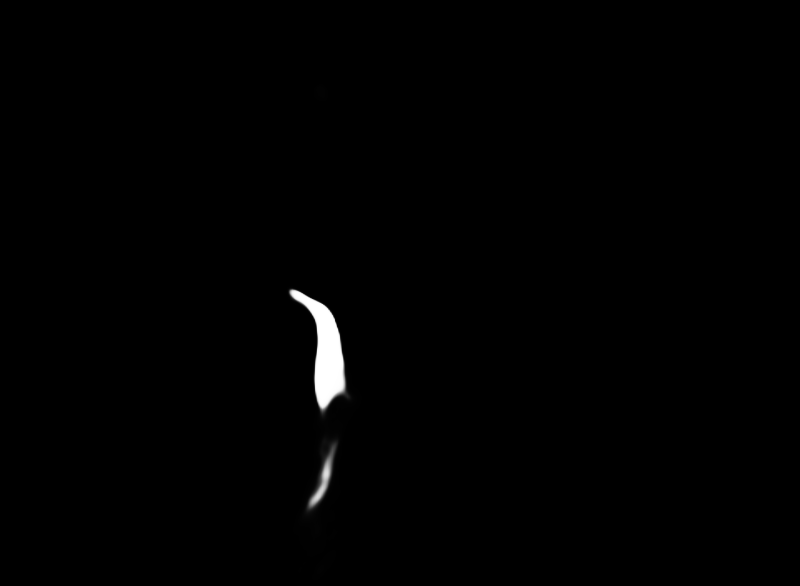}} &
   {\includegraphics[width=0.11\linewidth]{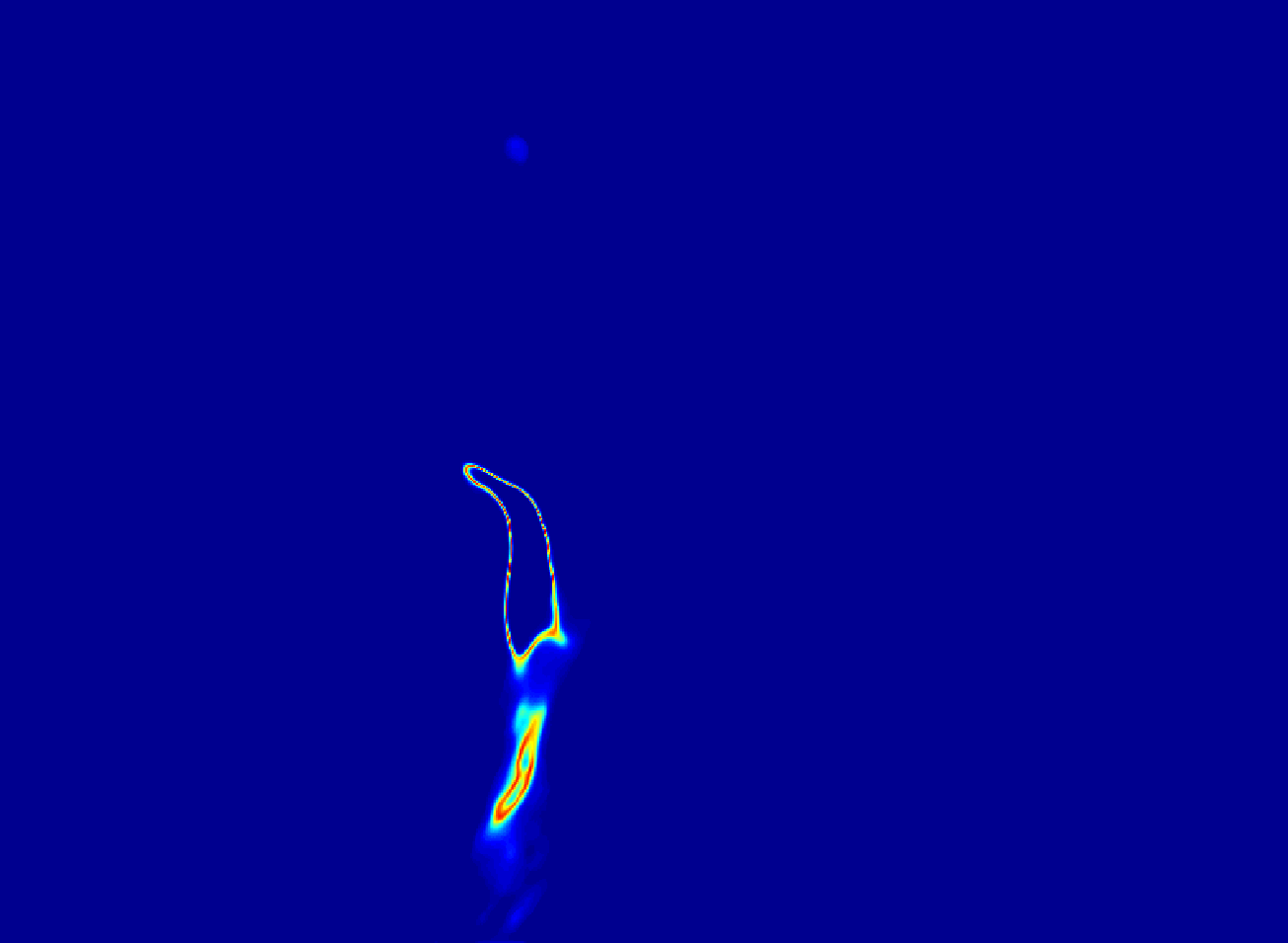}} &
   {\includegraphics[width=0.11\linewidth]{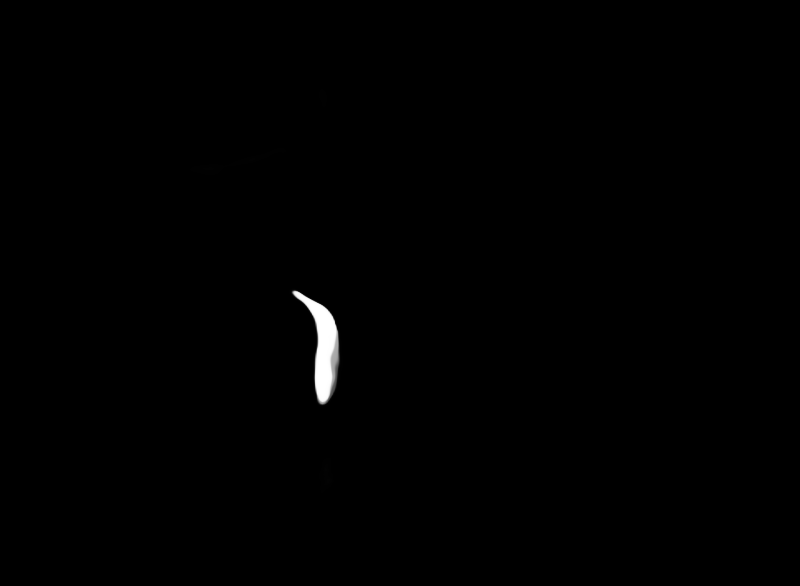}}&
   {\includegraphics[width=0.11\linewidth]{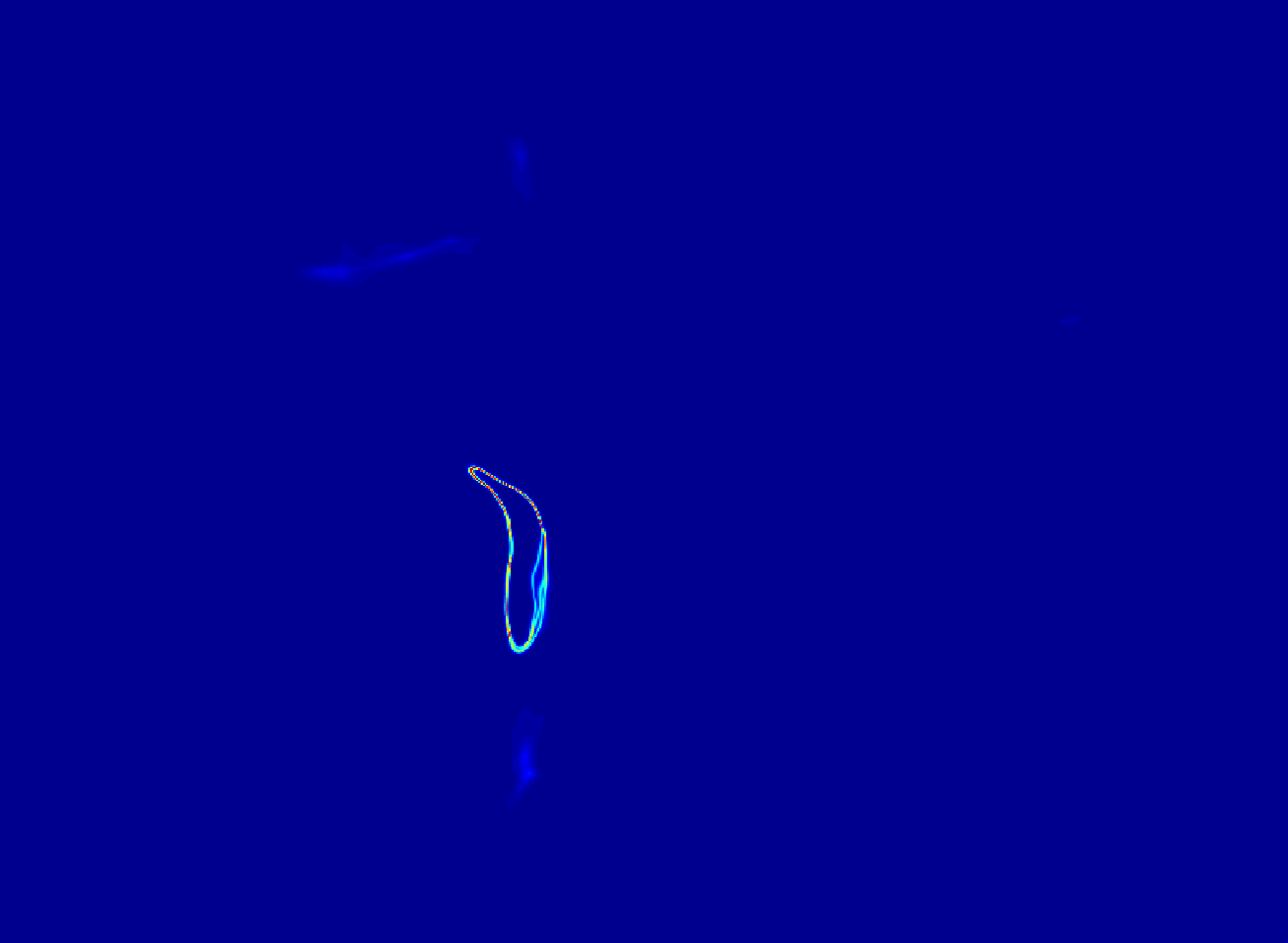}}&
   {\includegraphics[width=0.11\linewidth]{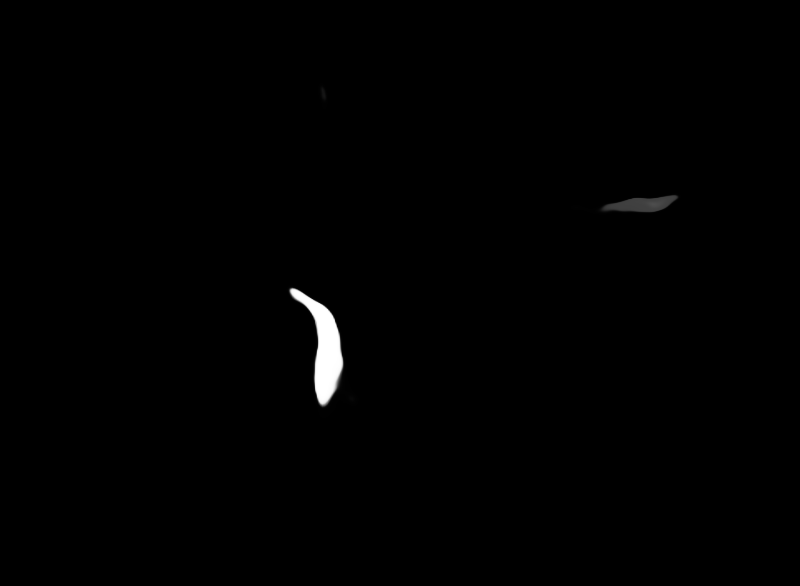}}&
   {\includegraphics[width=0.11\linewidth]{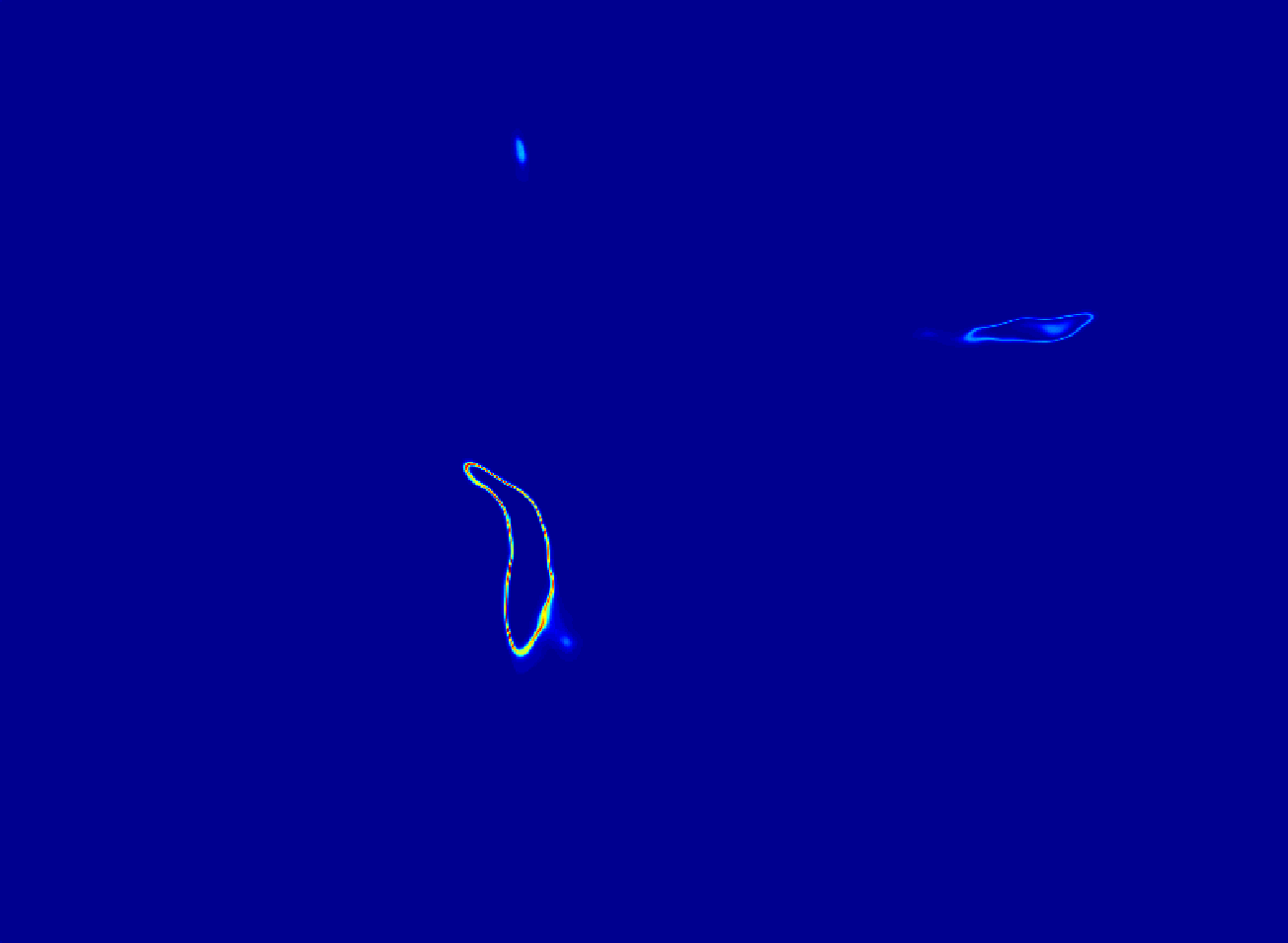}} \\
      {\includegraphics[width=0.11\linewidth]{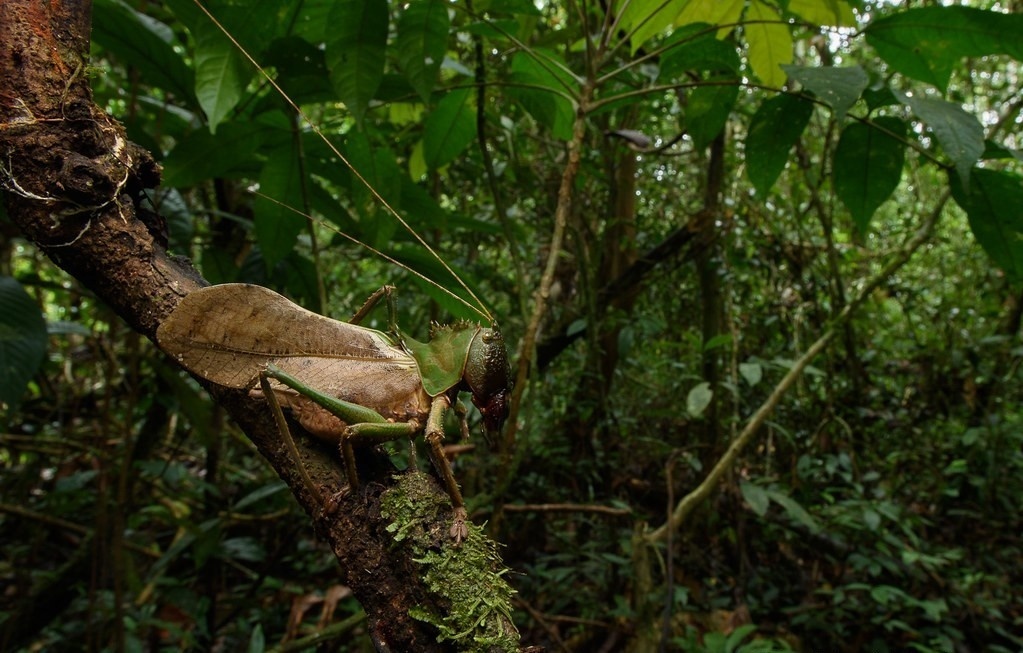}} &
   {\includegraphics[width=0.11\linewidth]{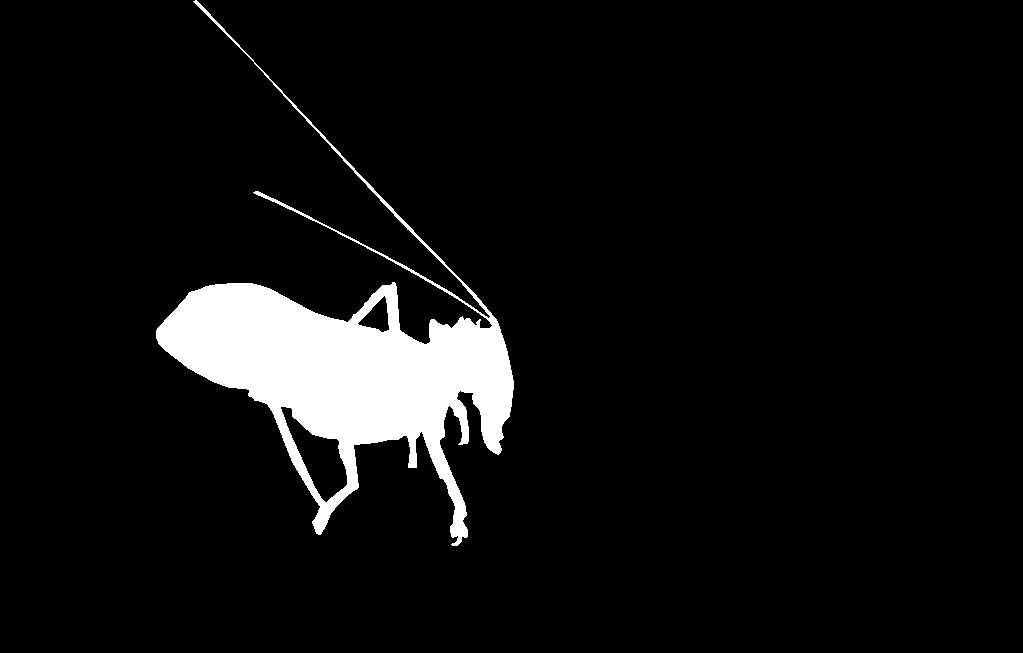}} &
   {\includegraphics[width=0.11\linewidth]{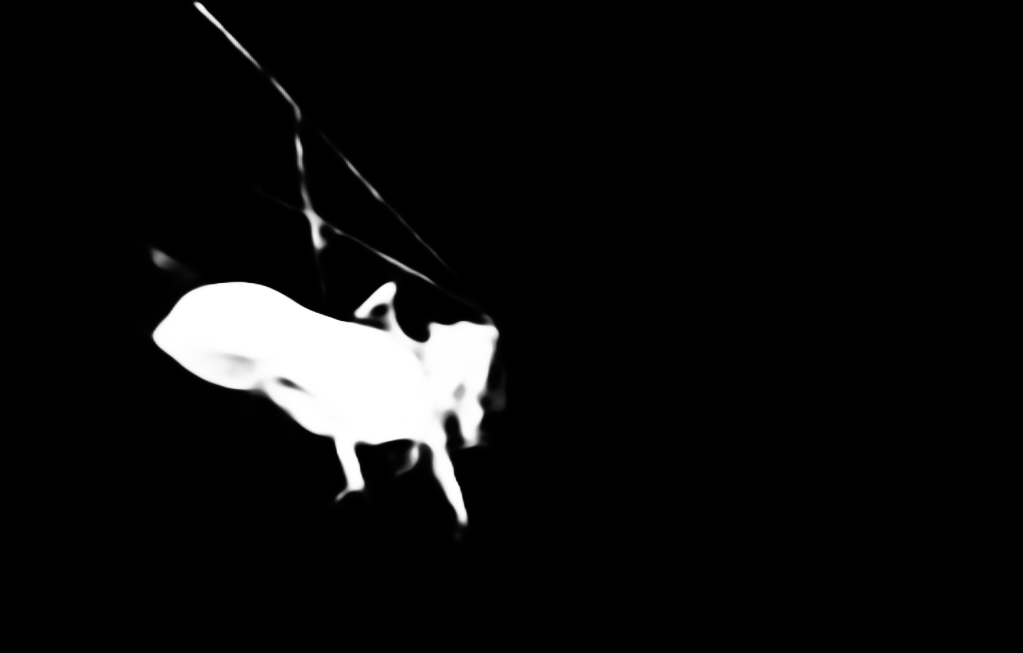}} &
   {\includegraphics[width=0.11\linewidth]{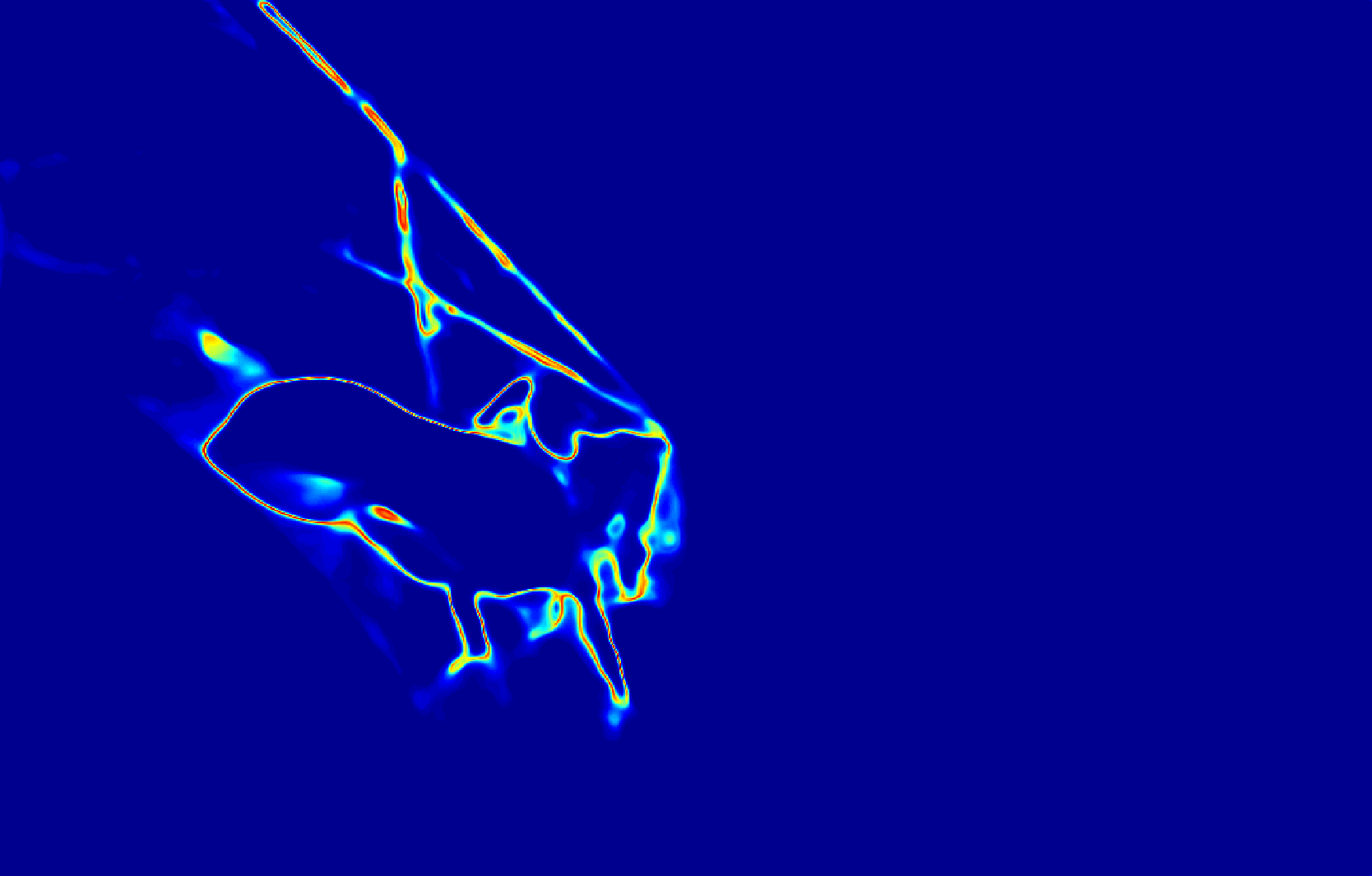}} &
   {\includegraphics[width=0.11\linewidth]{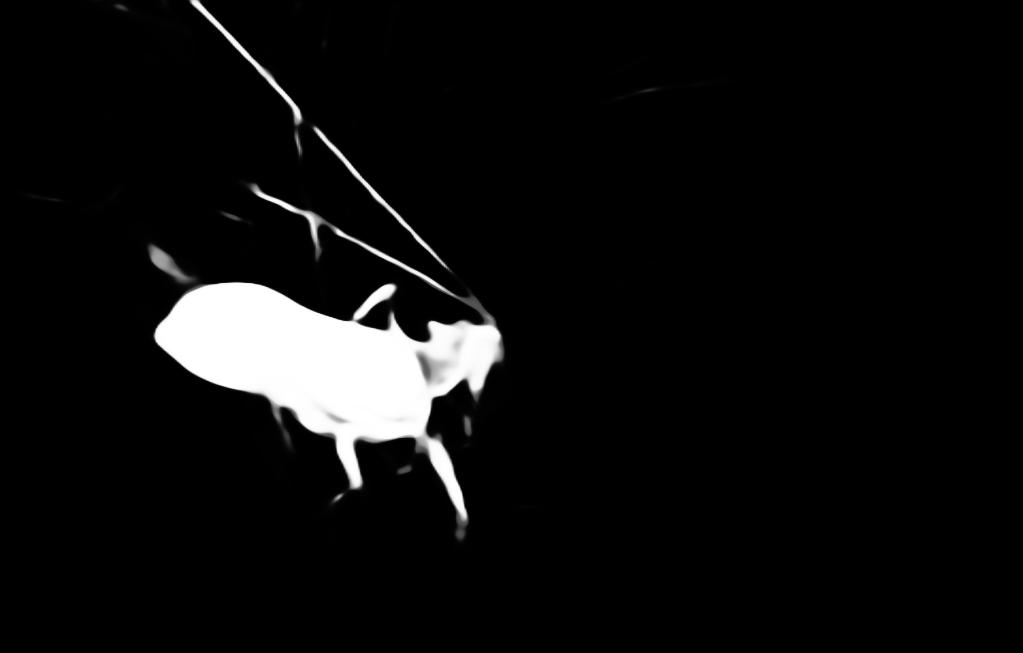}}&
   {\includegraphics[width=0.11\linewidth]{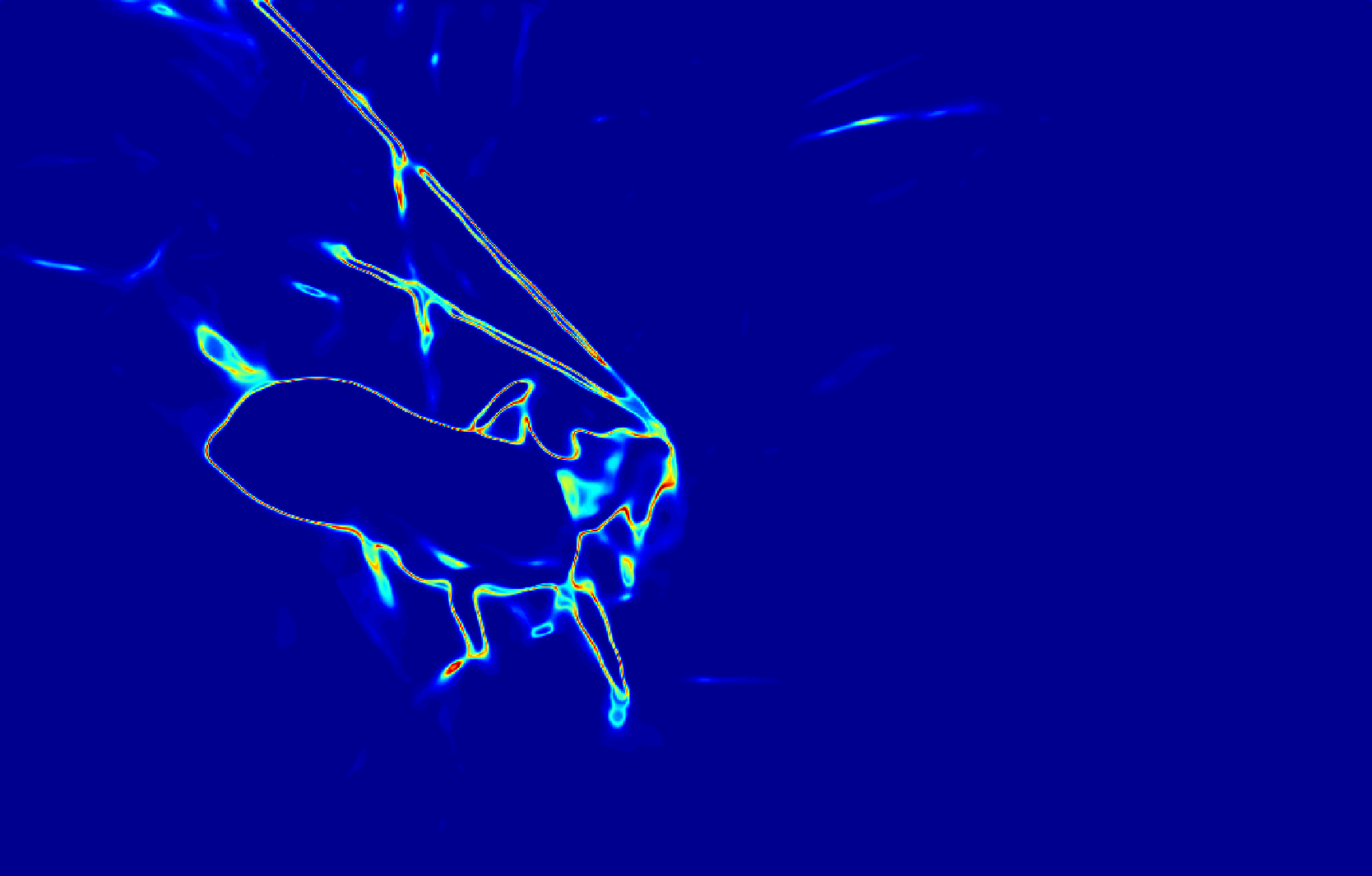}}&
   {\includegraphics[width=0.11\linewidth]{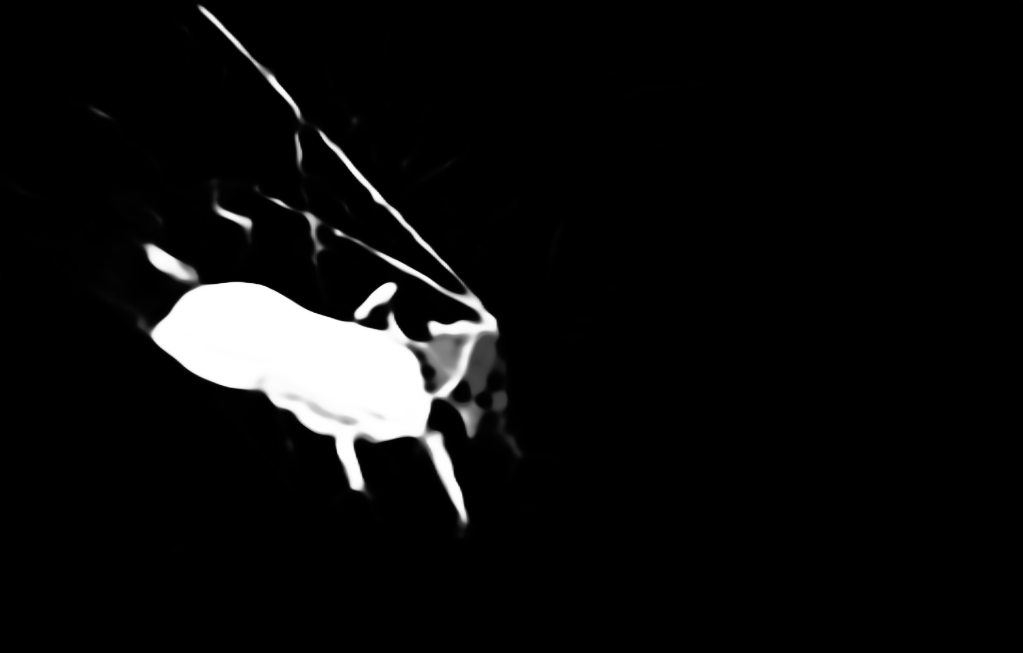}}&
   {\includegraphics[width=0.11\linewidth]{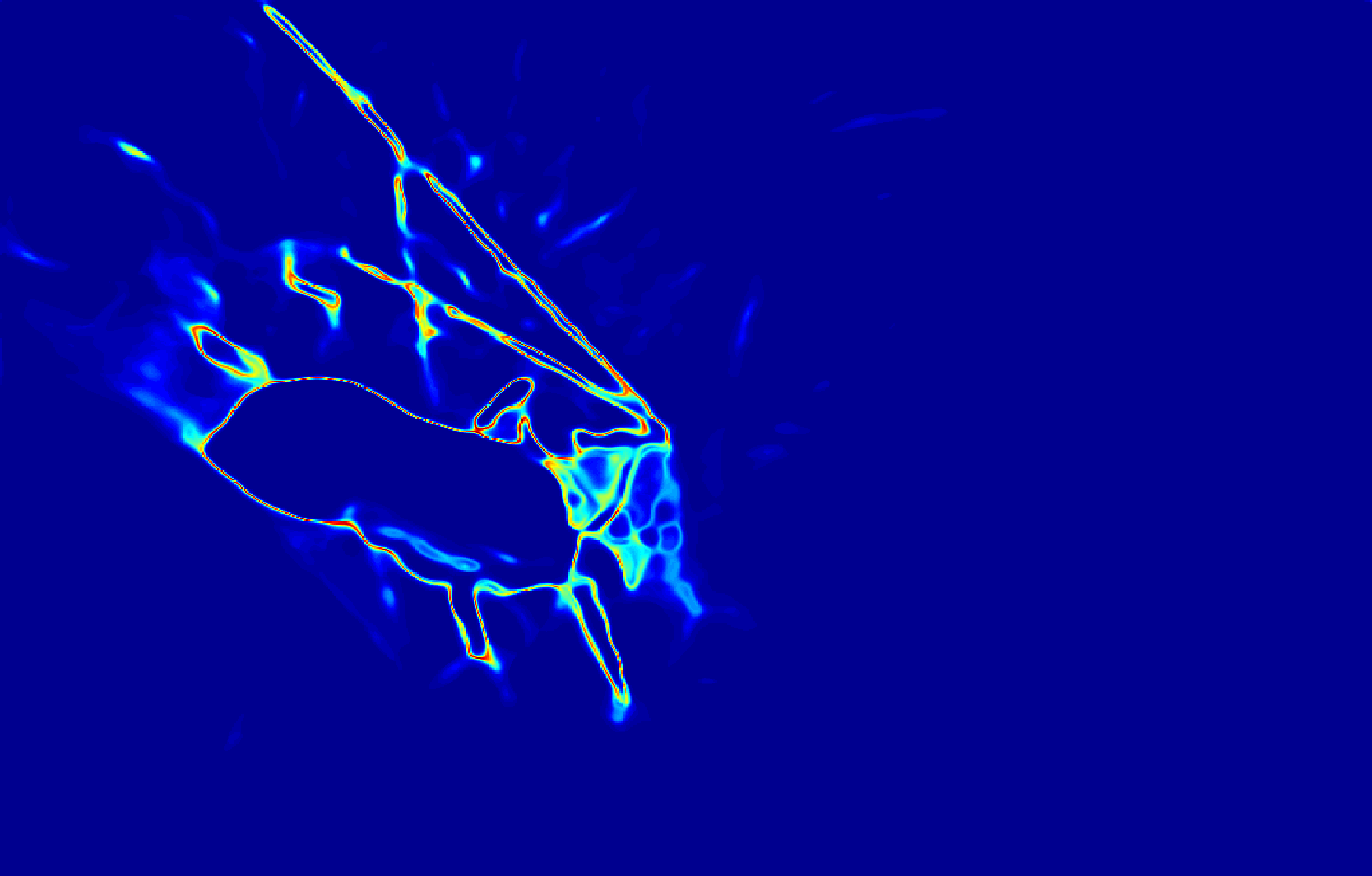}} \\
      {\includegraphics[width=0.11\linewidth]{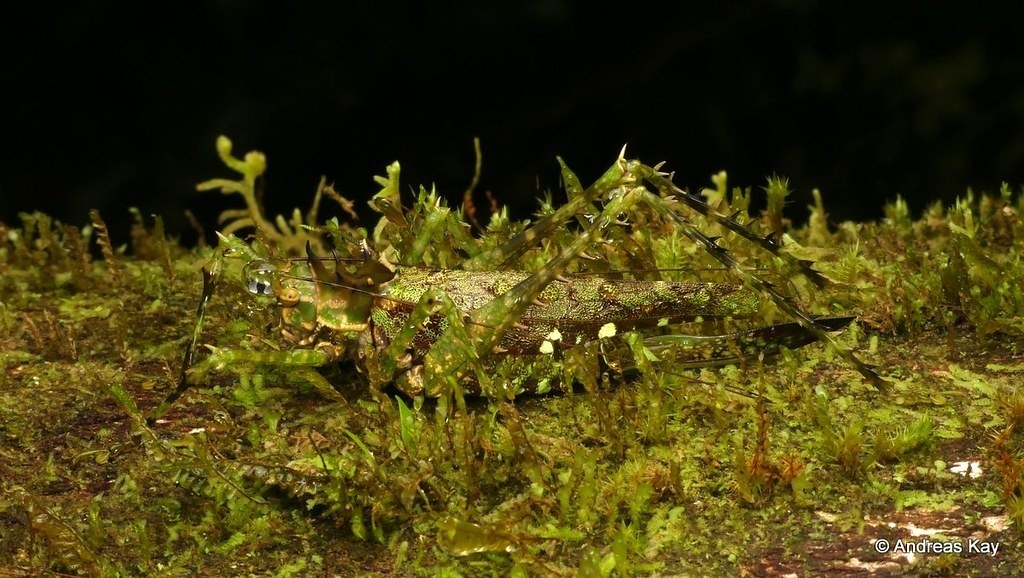}} &
   {\includegraphics[width=0.11\linewidth]{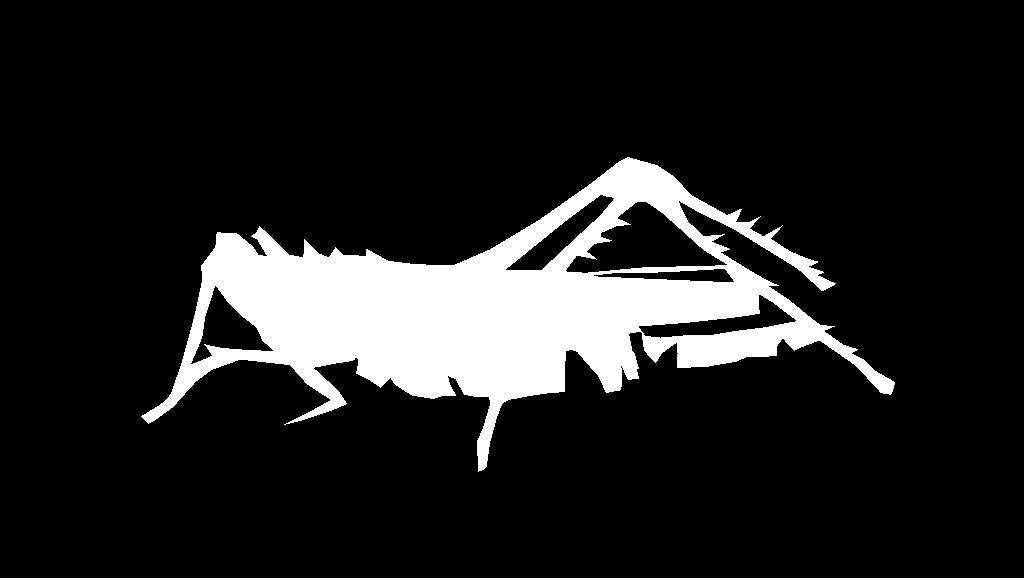}} &
   {\includegraphics[width=0.11\linewidth]{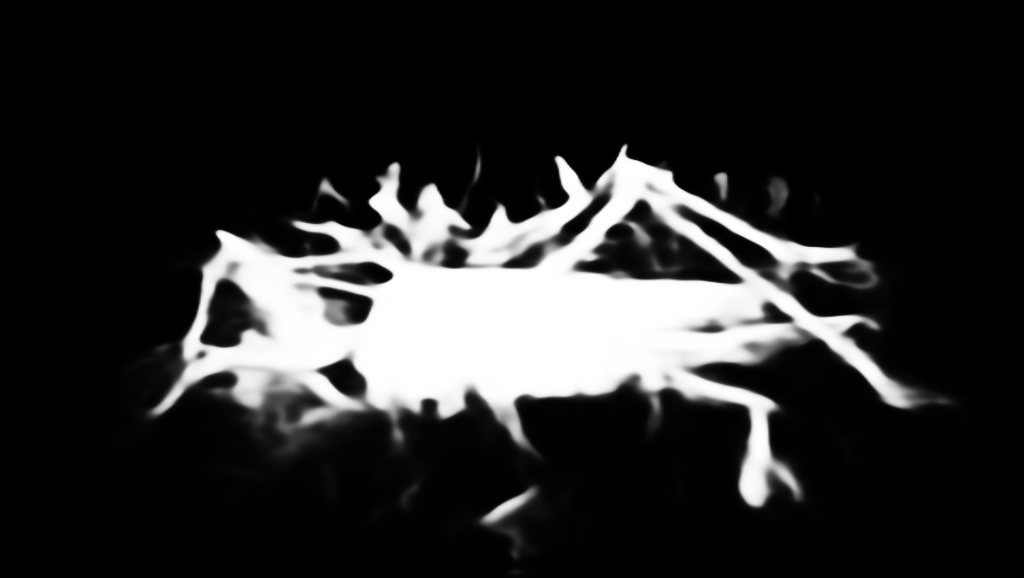}} &
   {\includegraphics[width=0.11\linewidth]{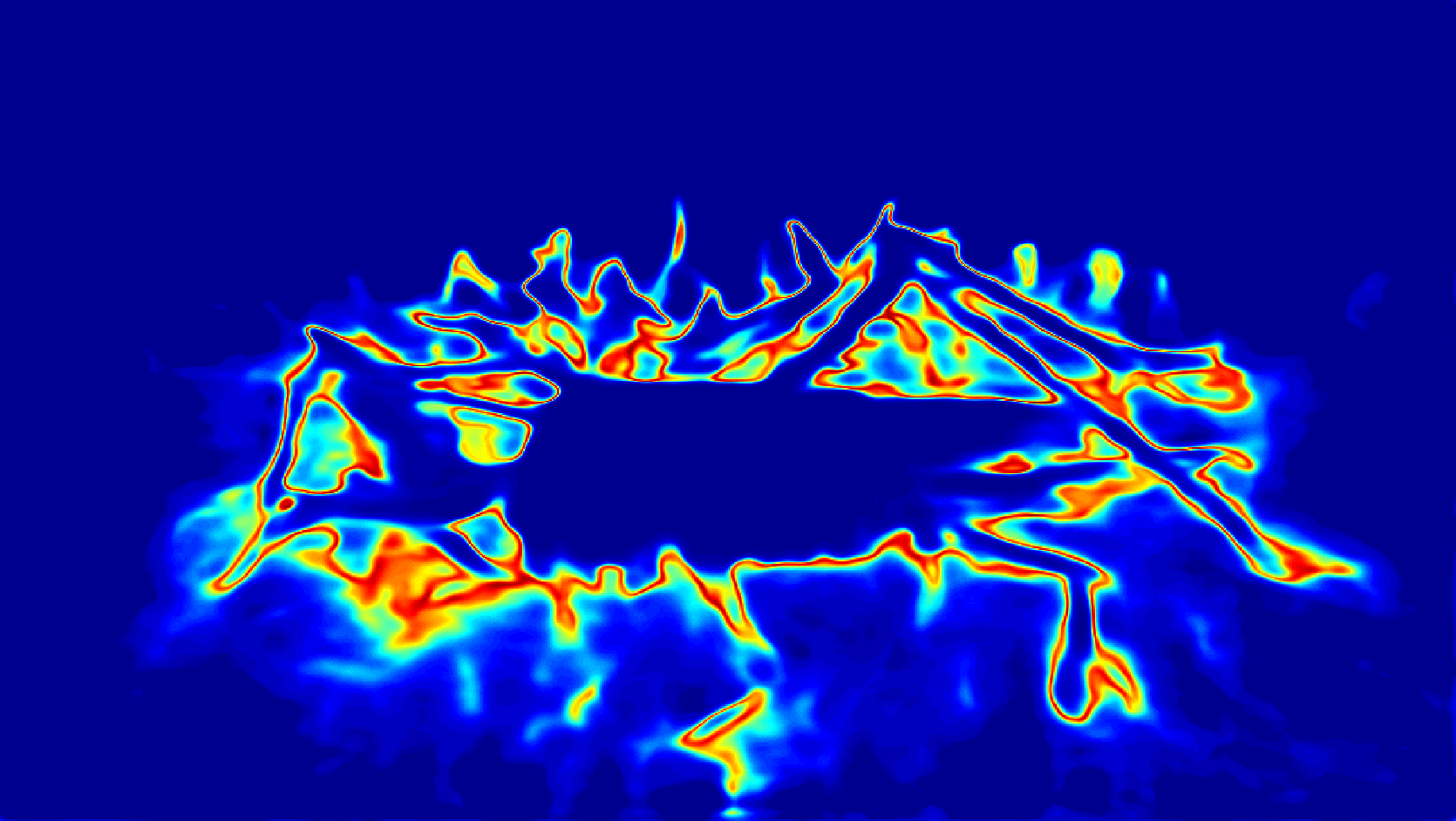}} &
   {\includegraphics[width=0.11\linewidth]{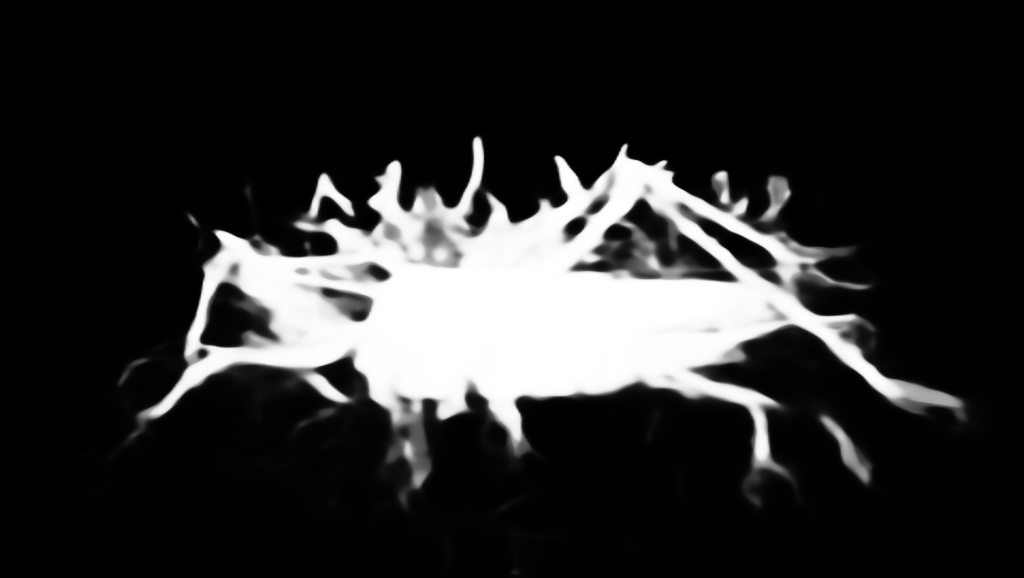}}&
   {\includegraphics[width=0.11\linewidth]{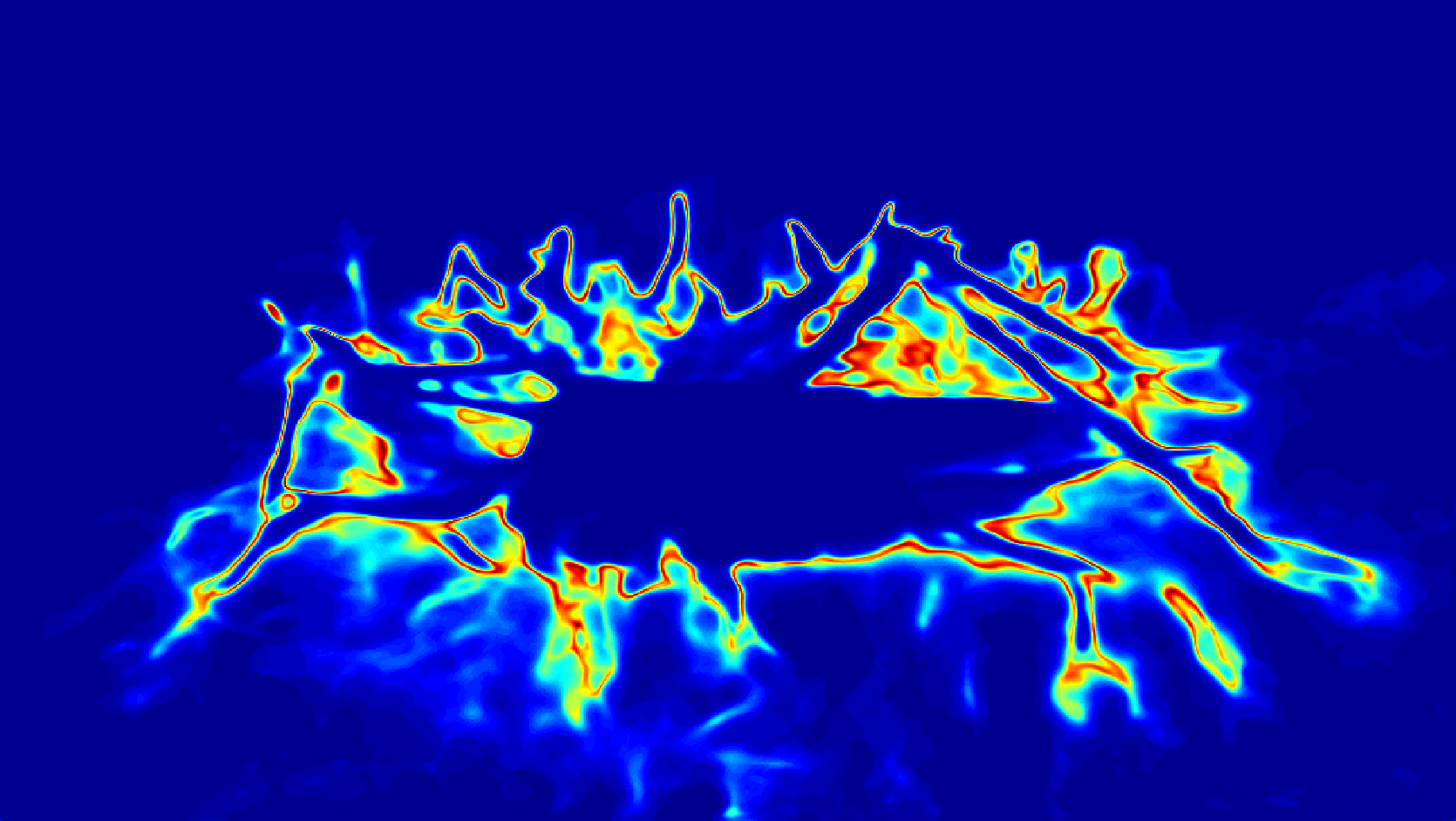}}&
   {\includegraphics[width=0.11\linewidth]{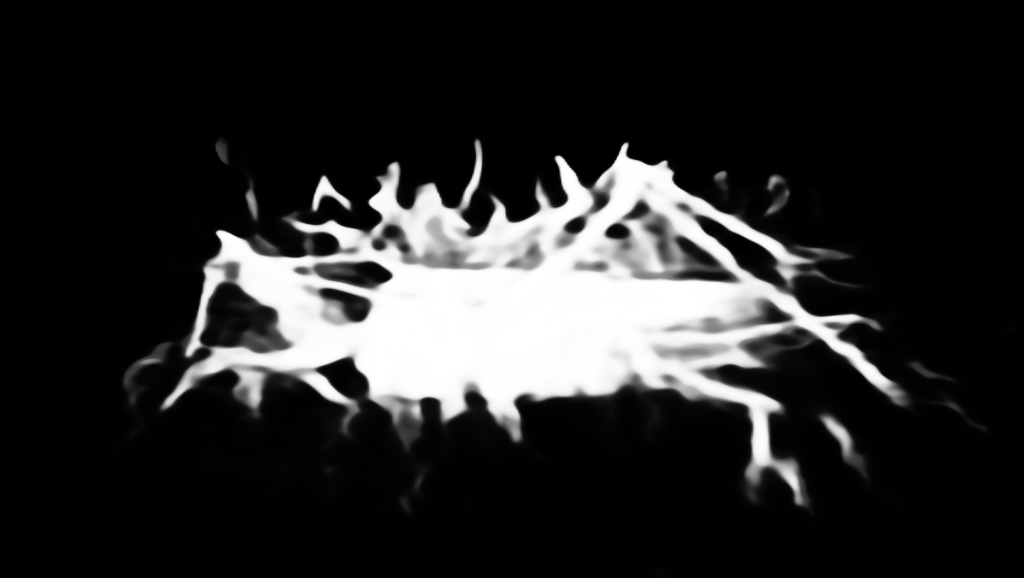}}&
   {\includegraphics[width=0.11\linewidth]{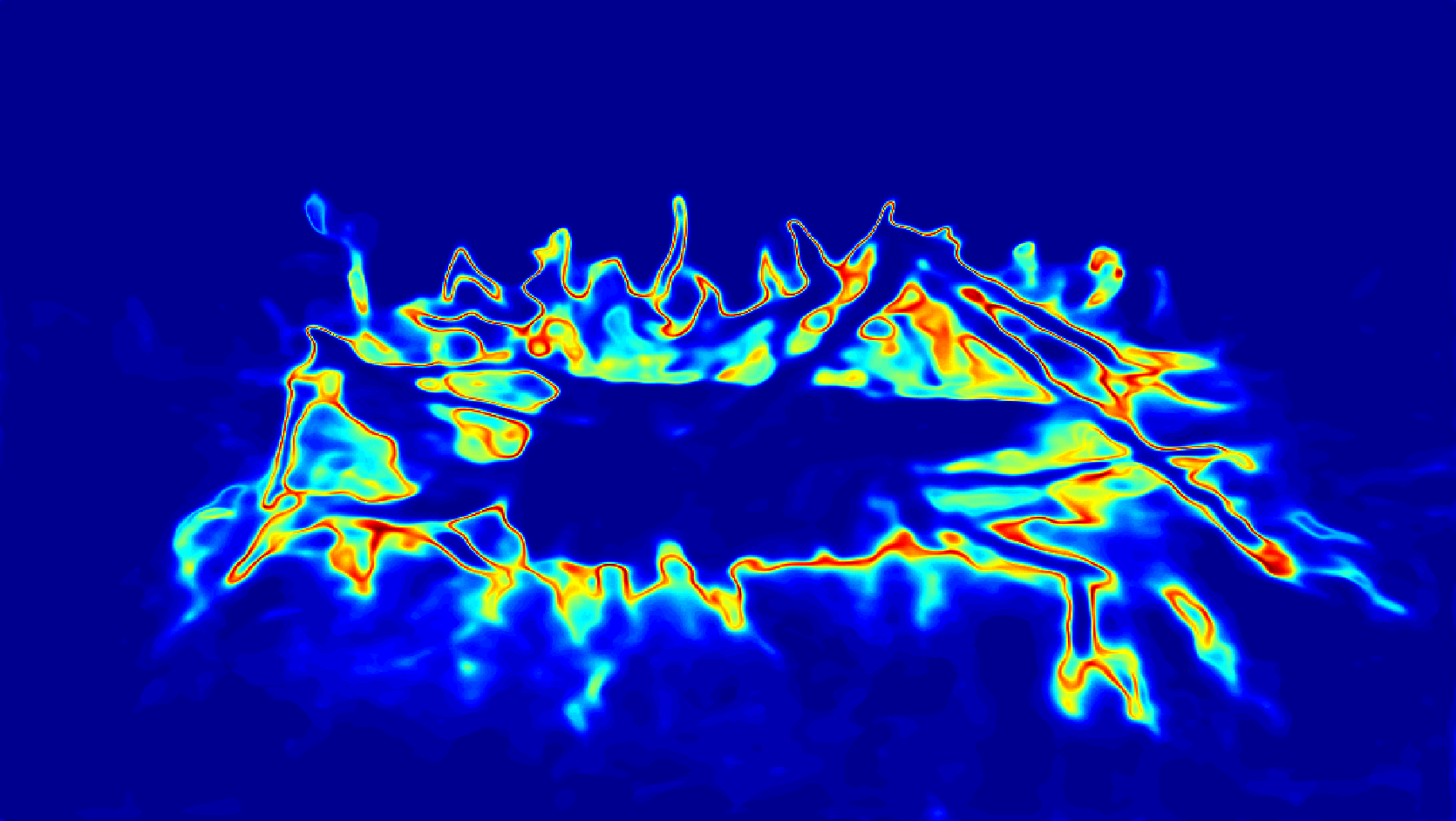}} \\
      {\includegraphics[width=0.11\linewidth]{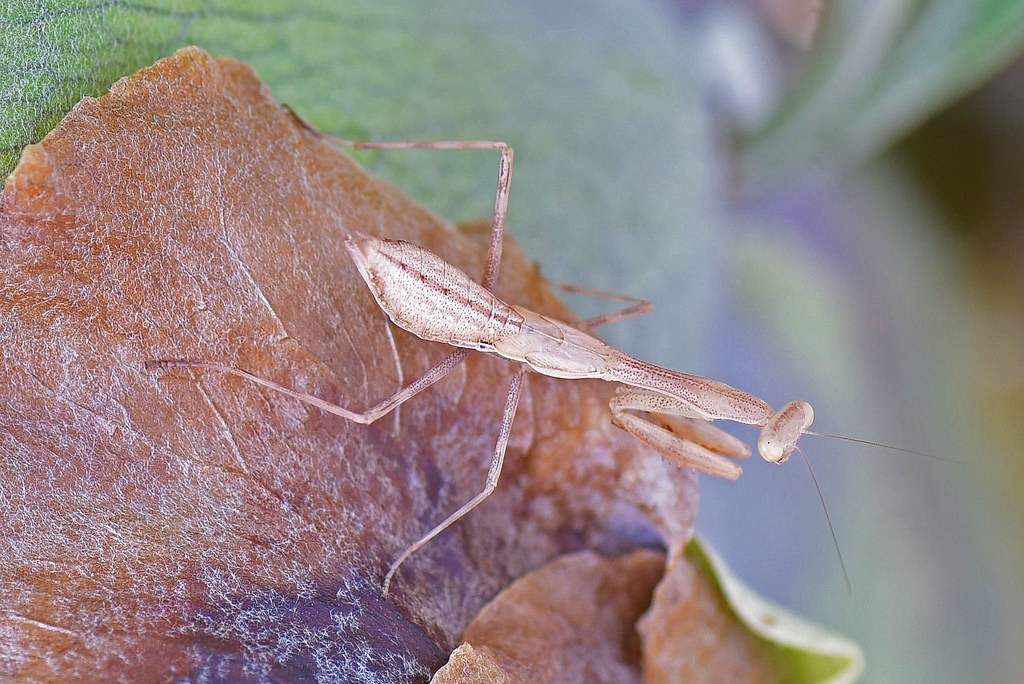}} &
   {\includegraphics[width=0.11\linewidth]{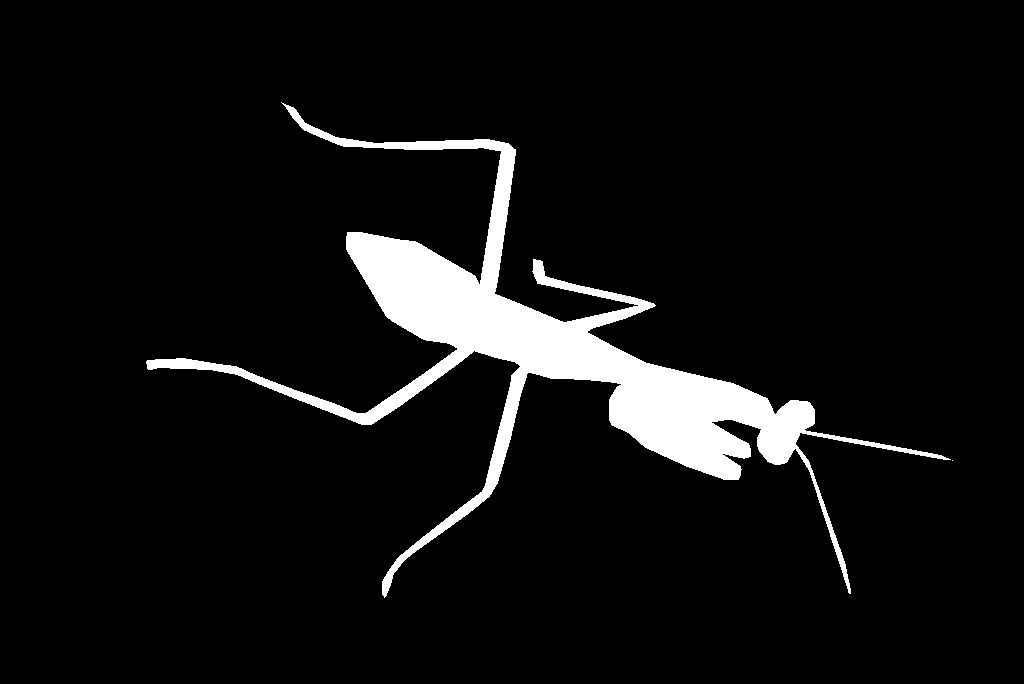}} &
   {\includegraphics[width=0.11\linewidth]{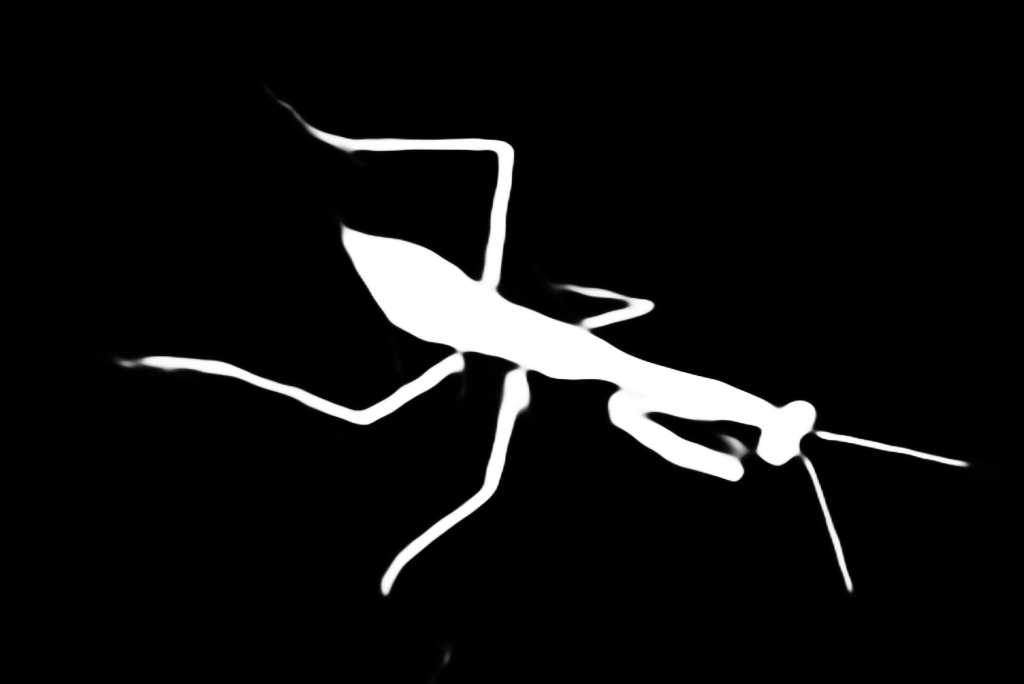}} &
   {\includegraphics[width=0.11\linewidth]{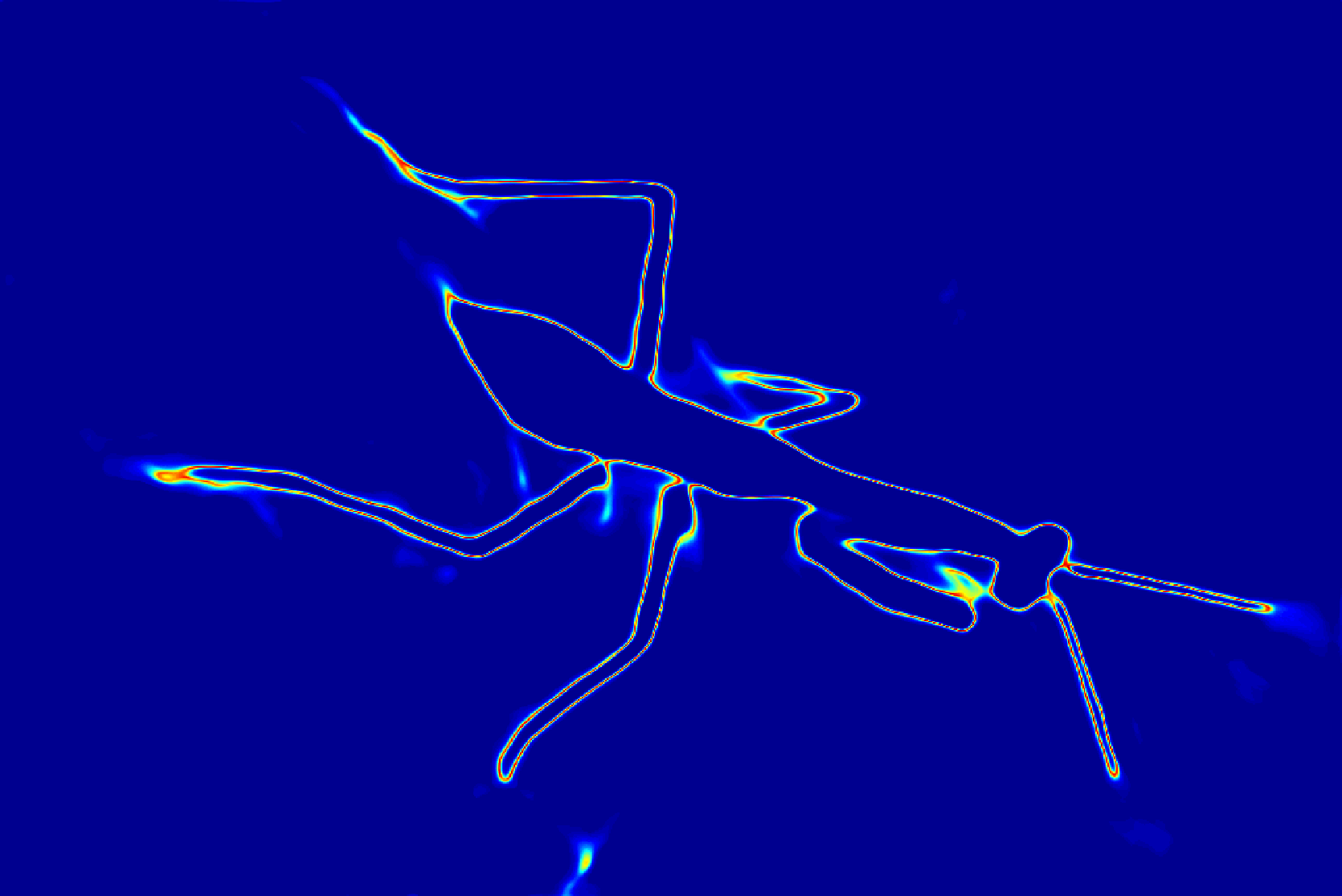}} &
   {\includegraphics[width=0.11\linewidth]{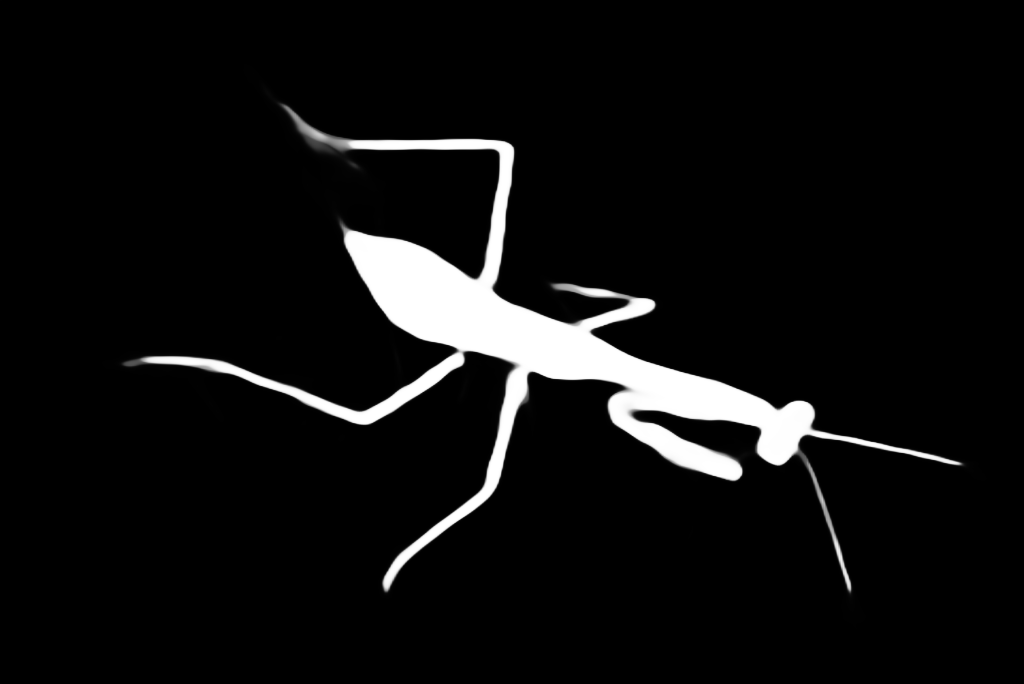}}&
   {\includegraphics[width=0.11\linewidth]{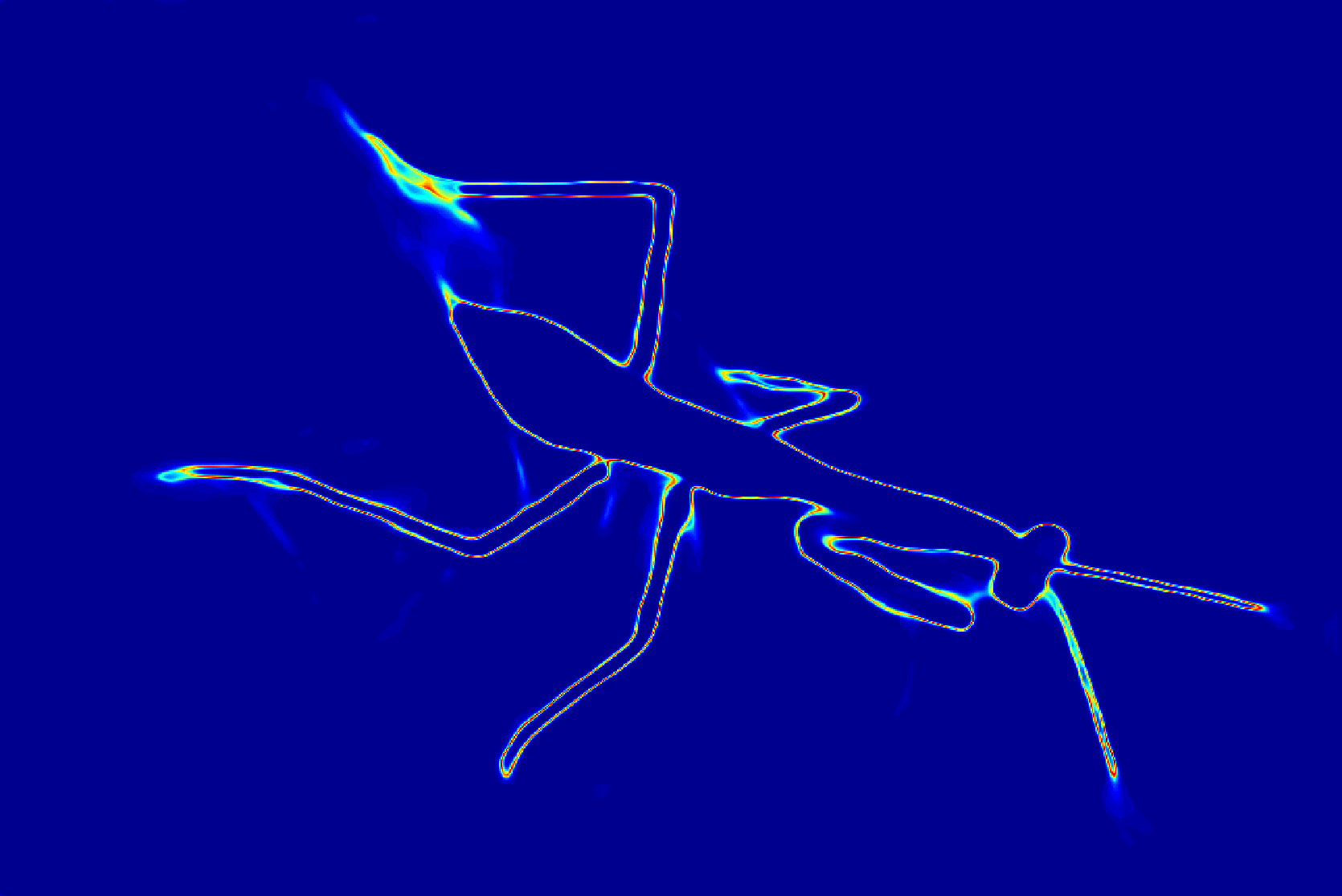}}&
   {\includegraphics[width=0.11\linewidth]{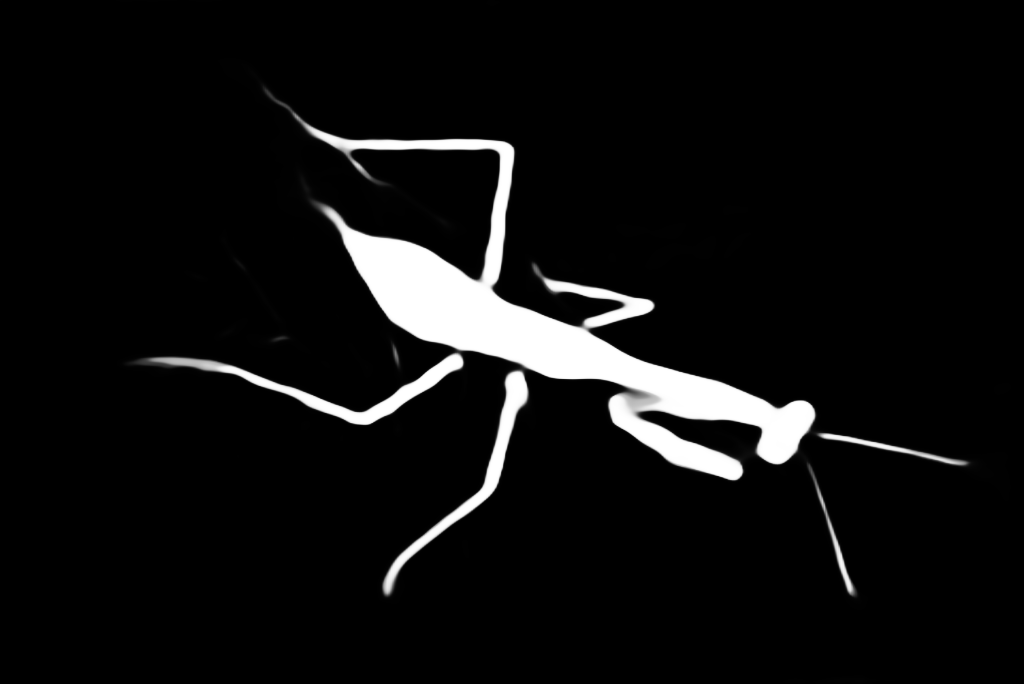}}&
   {\includegraphics[width=0.11\linewidth]{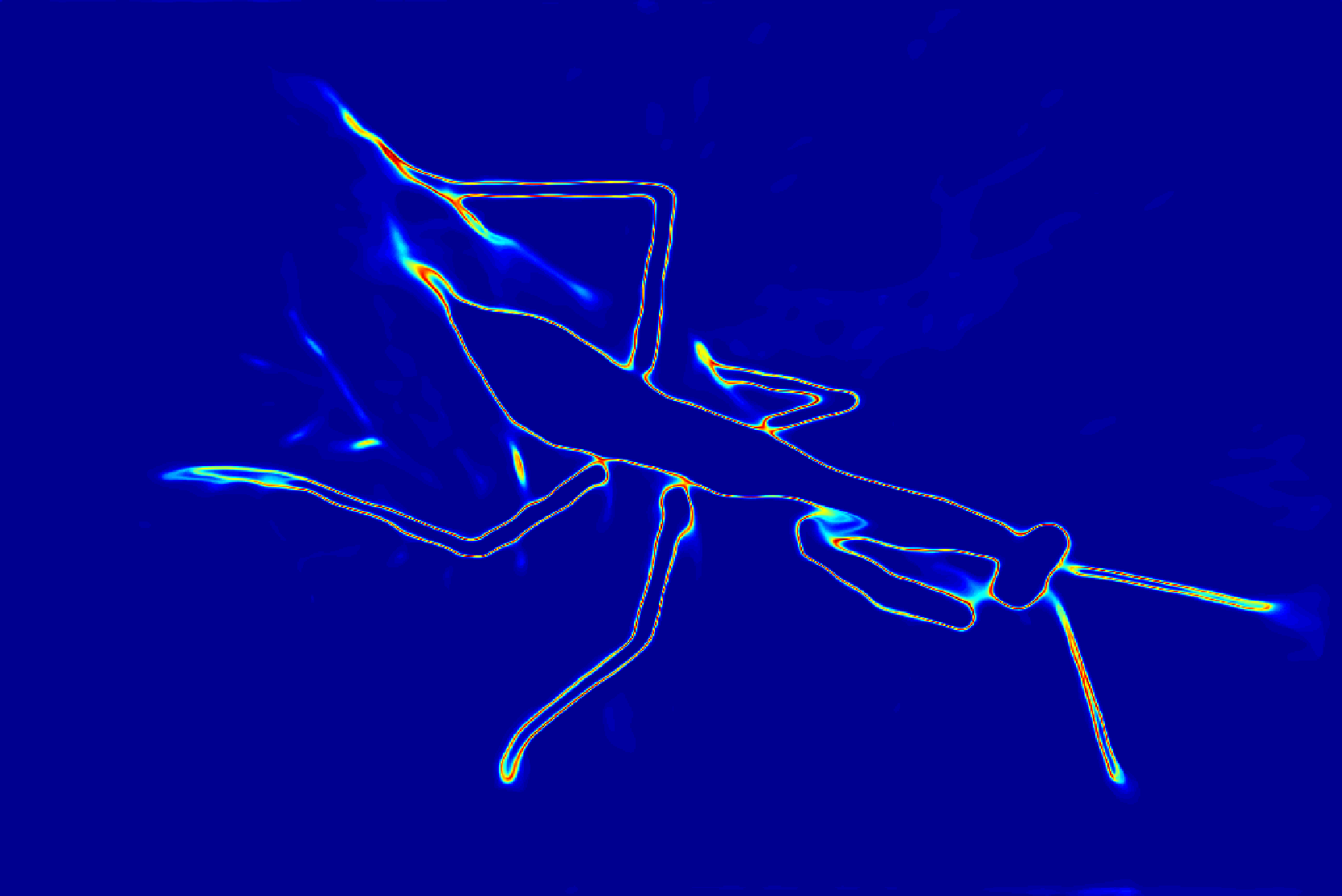}} \\
   \footnotesize{Image}&\footnotesize{GT}&\footnotesize{MD}&\footnotesize{$U_a$}&\footnotesize{DE}&\footnotesize{$U_a$}&\footnotesize{SE}&\footnotesize{$U_a$}\\
   \end{tabular}
   \end{center}
   \caption{\footnotesize{Aleatoric uncertainty of ensemble based solutions for \textbf{camouflaged object detection}.}
   }
\label{fig:aleatoric_ensemble_cod}
\end{figure}

\begin{figure}[tp]
   \begin{center}
   \begin{tabular}{c@{ }c@{ }c@{ }c@{ }c@{ }c@{ }c@{ }c@{ }}
   {\includegraphics[width=0.11\linewidth]{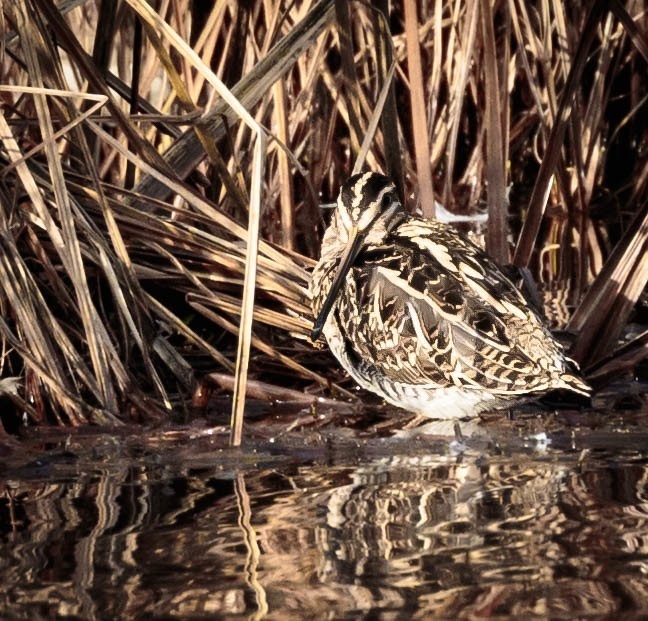}} &
   {\includegraphics[width=0.11\linewidth]{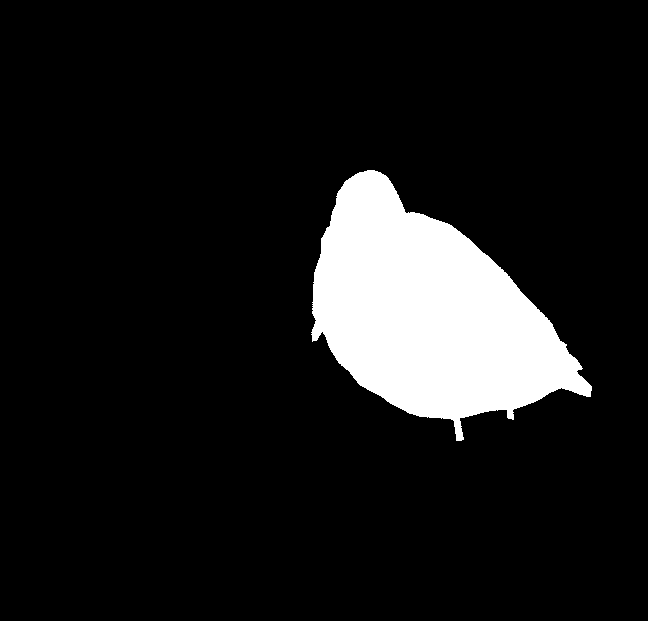}} &
   {\includegraphics[width=0.11\linewidth]{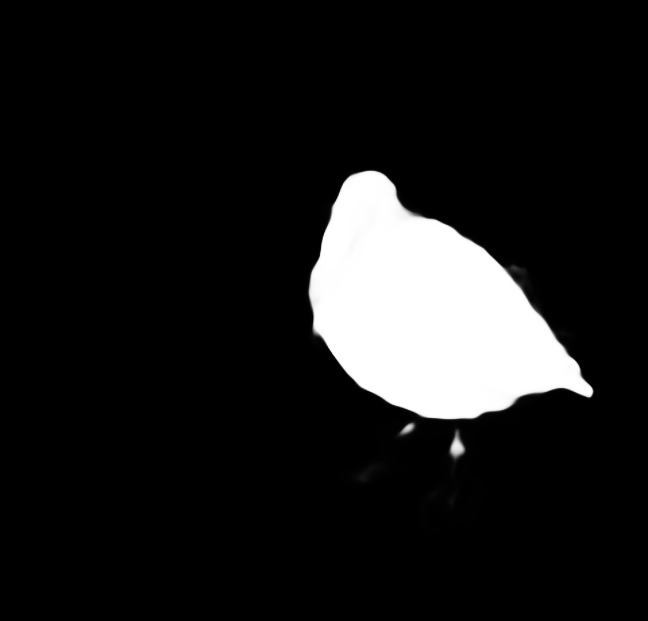}} &
   {\includegraphics[width=0.11\linewidth]{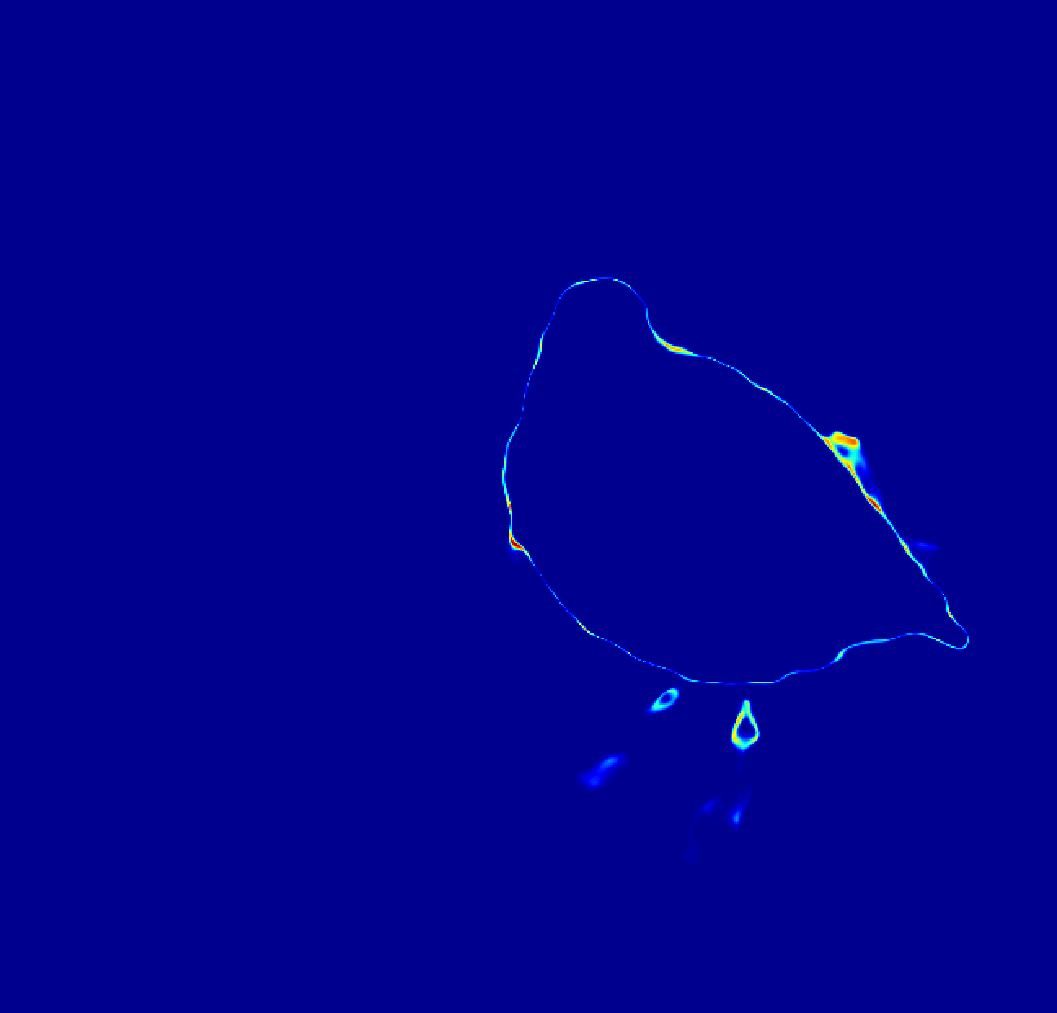}} &
   {\includegraphics[width=0.11\linewidth]{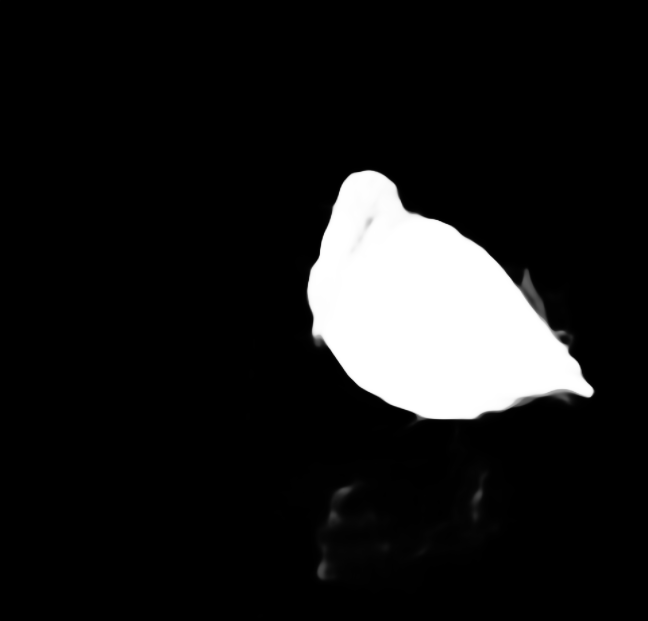}}&
   {\includegraphics[width=0.11\linewidth]{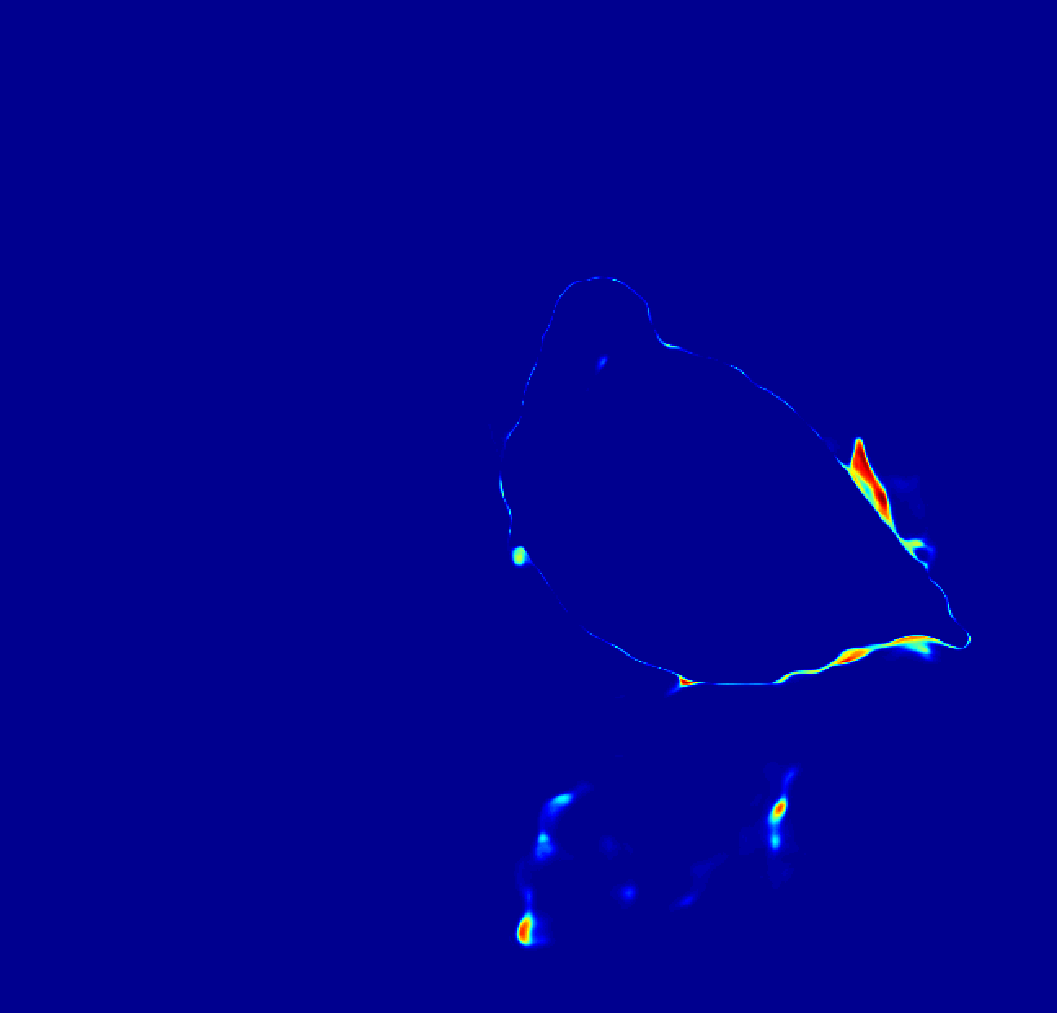}}&
   {\includegraphics[width=0.11\linewidth]{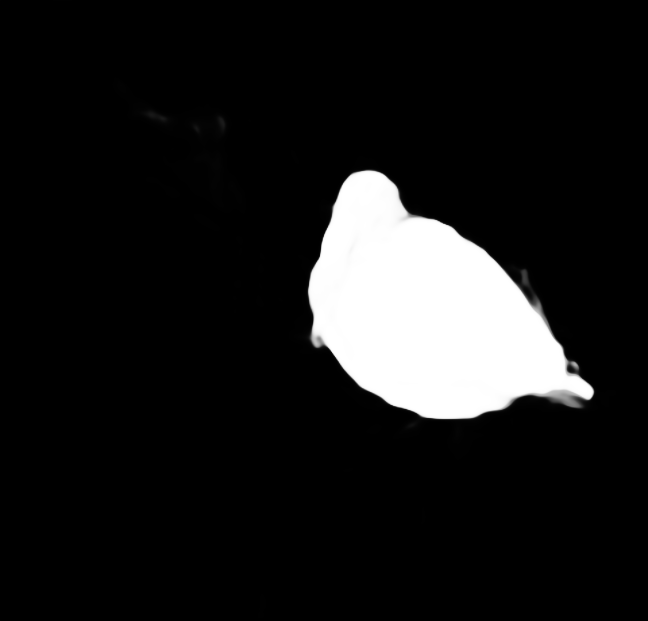}}&
   {\includegraphics[width=0.11\linewidth]{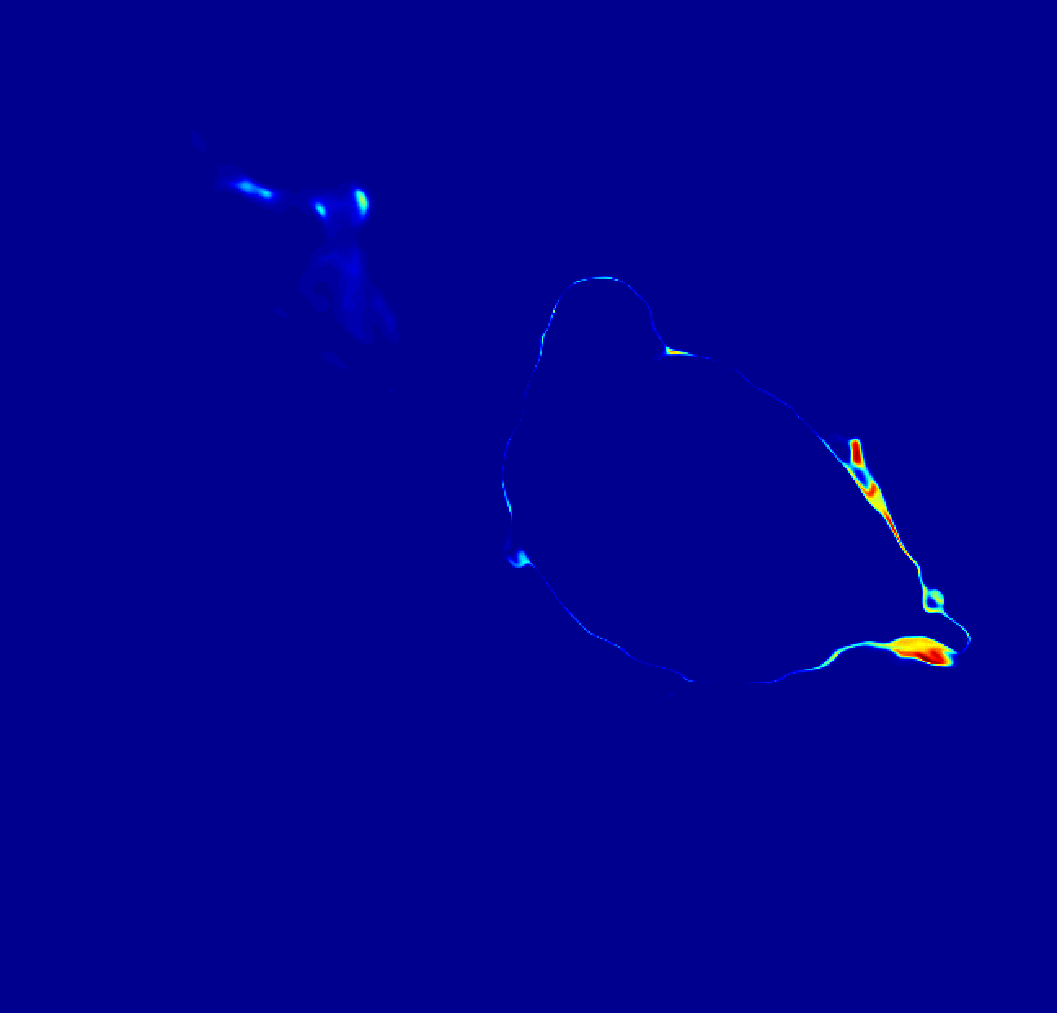}} \\
      {\includegraphics[width=0.11\linewidth]{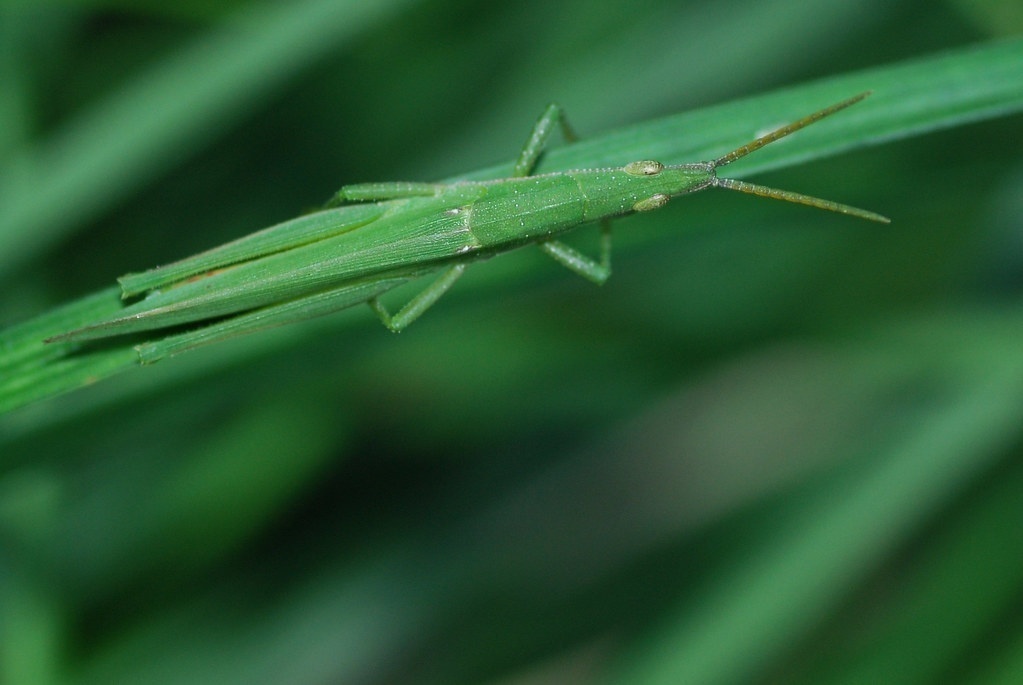}} &
   {\includegraphics[width=0.11\linewidth]{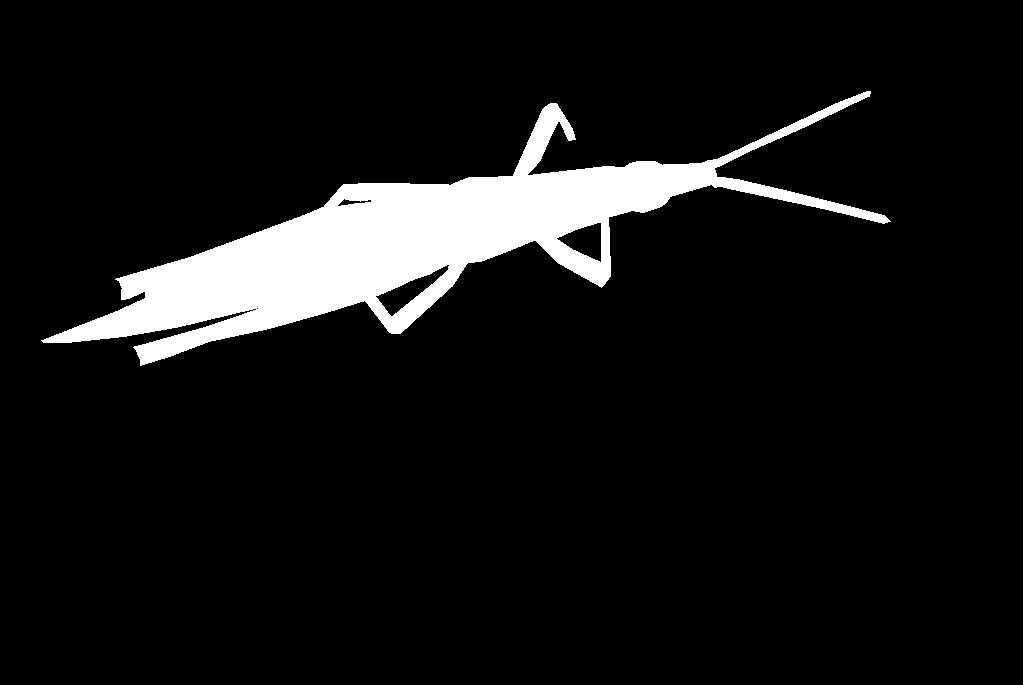}} &
   {\includegraphics[width=0.11\linewidth]{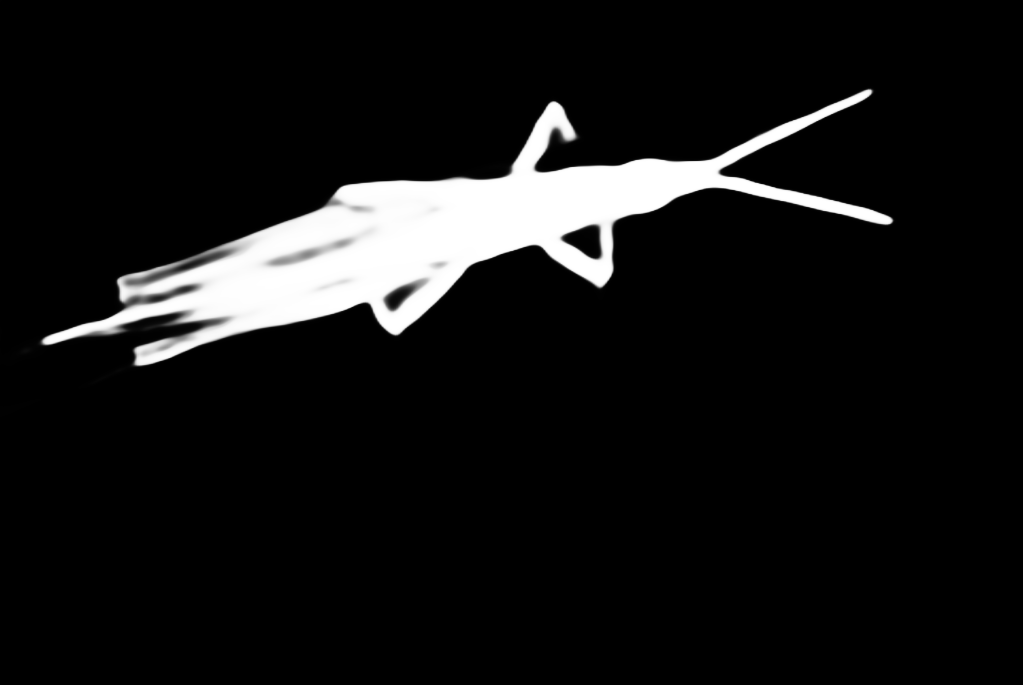}} &
   {\includegraphics[width=0.11\linewidth]{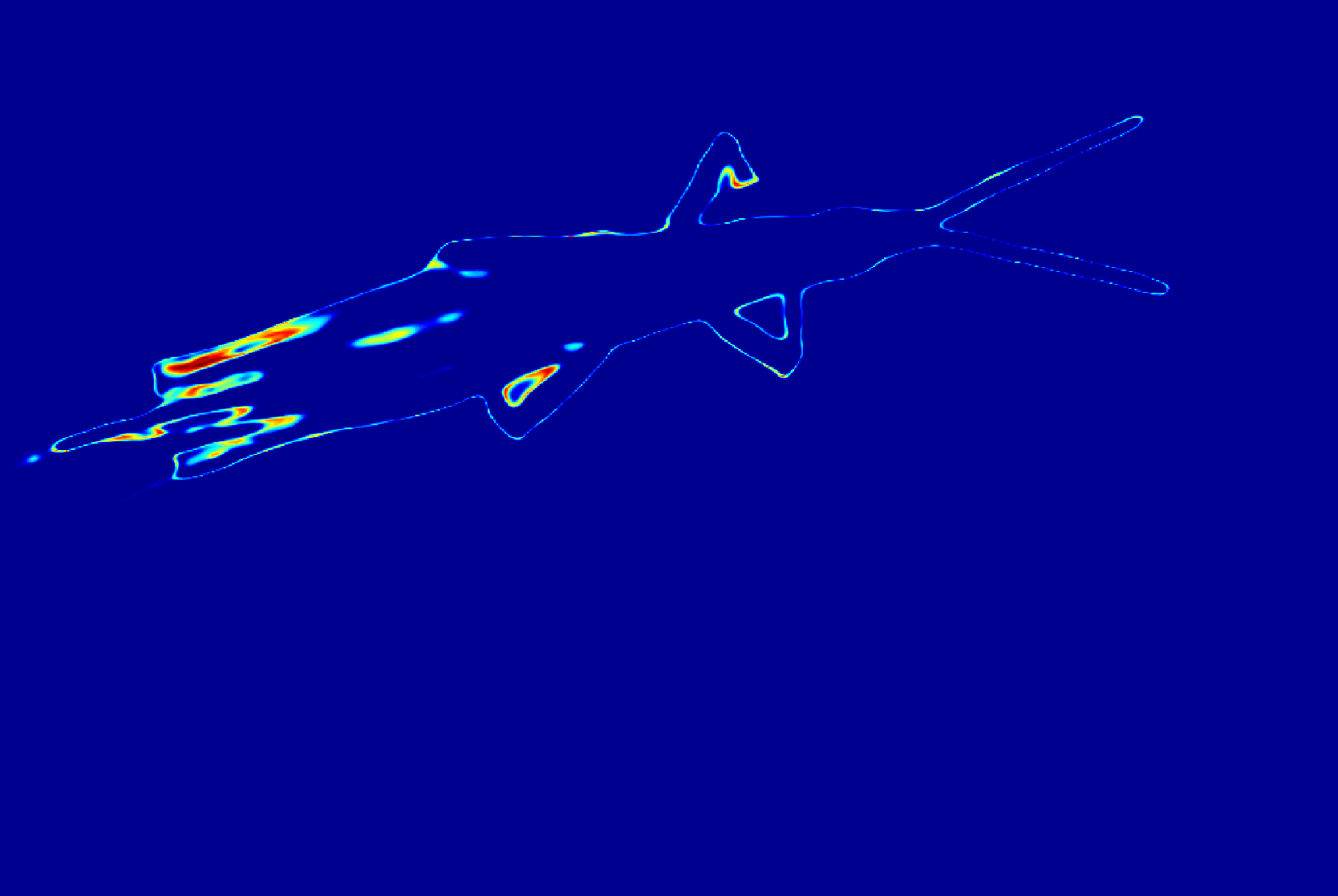}} &
   {\includegraphics[width=0.11\linewidth]{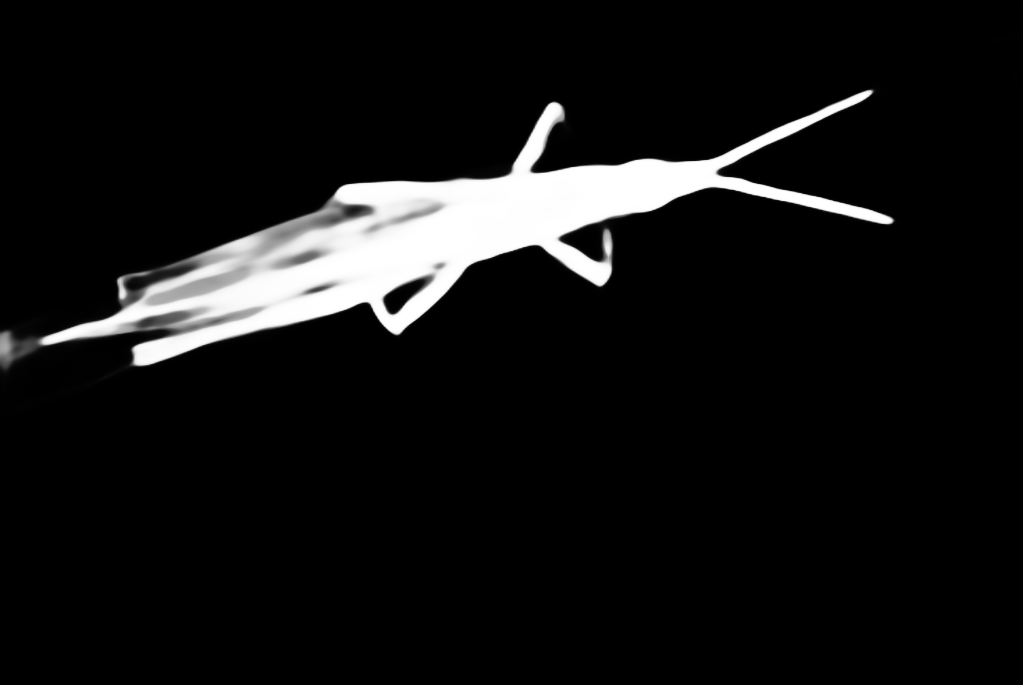}}&
   {\includegraphics[width=0.11\linewidth]{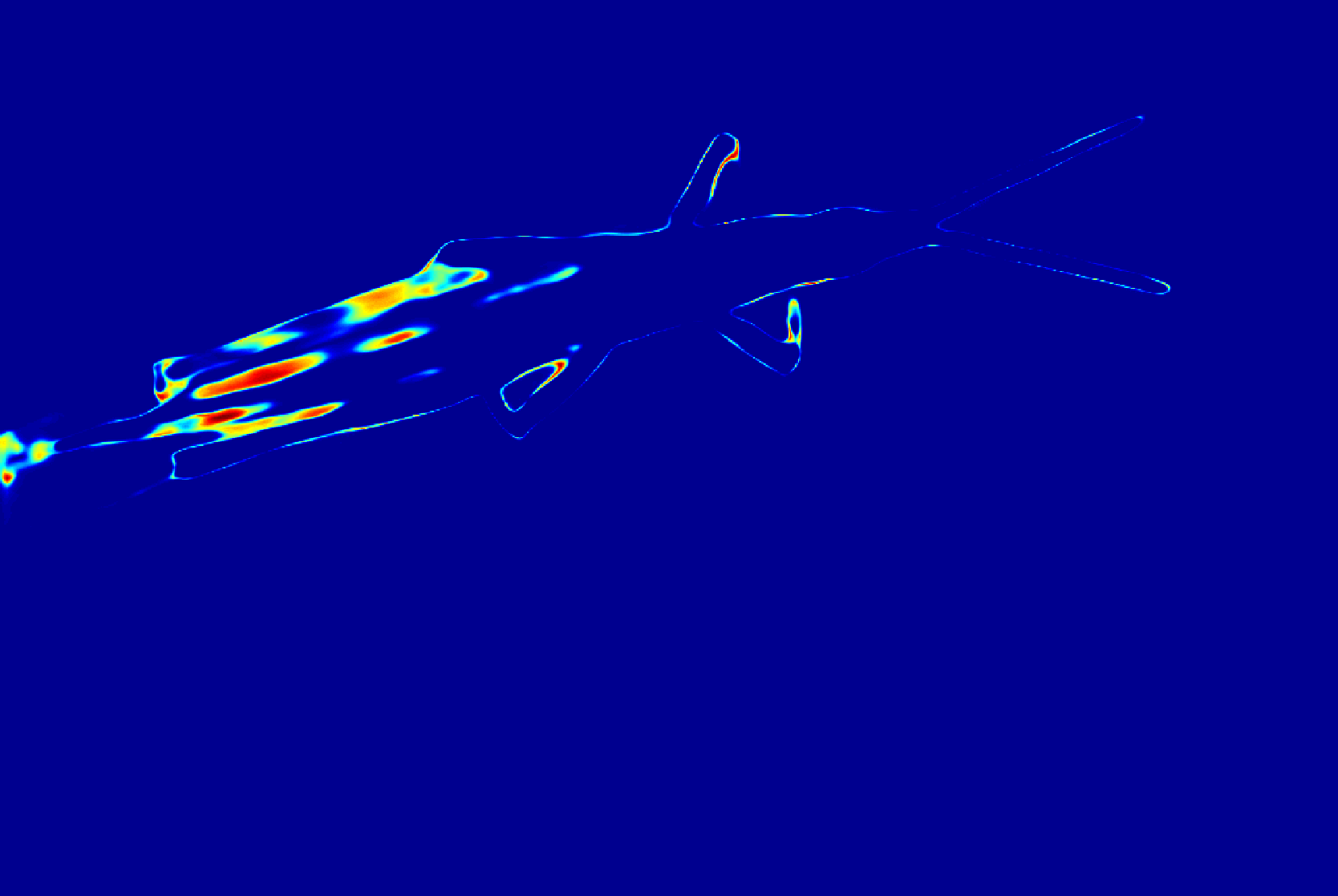}}&
   {\includegraphics[width=0.11\linewidth]{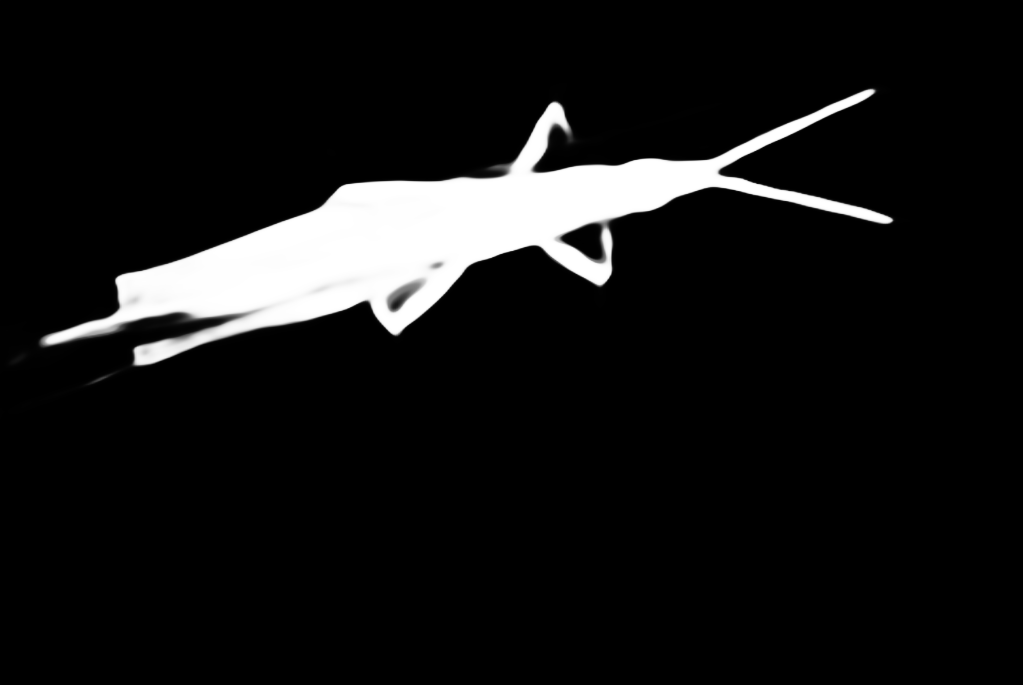}}&
   {\includegraphics[width=0.11\linewidth]{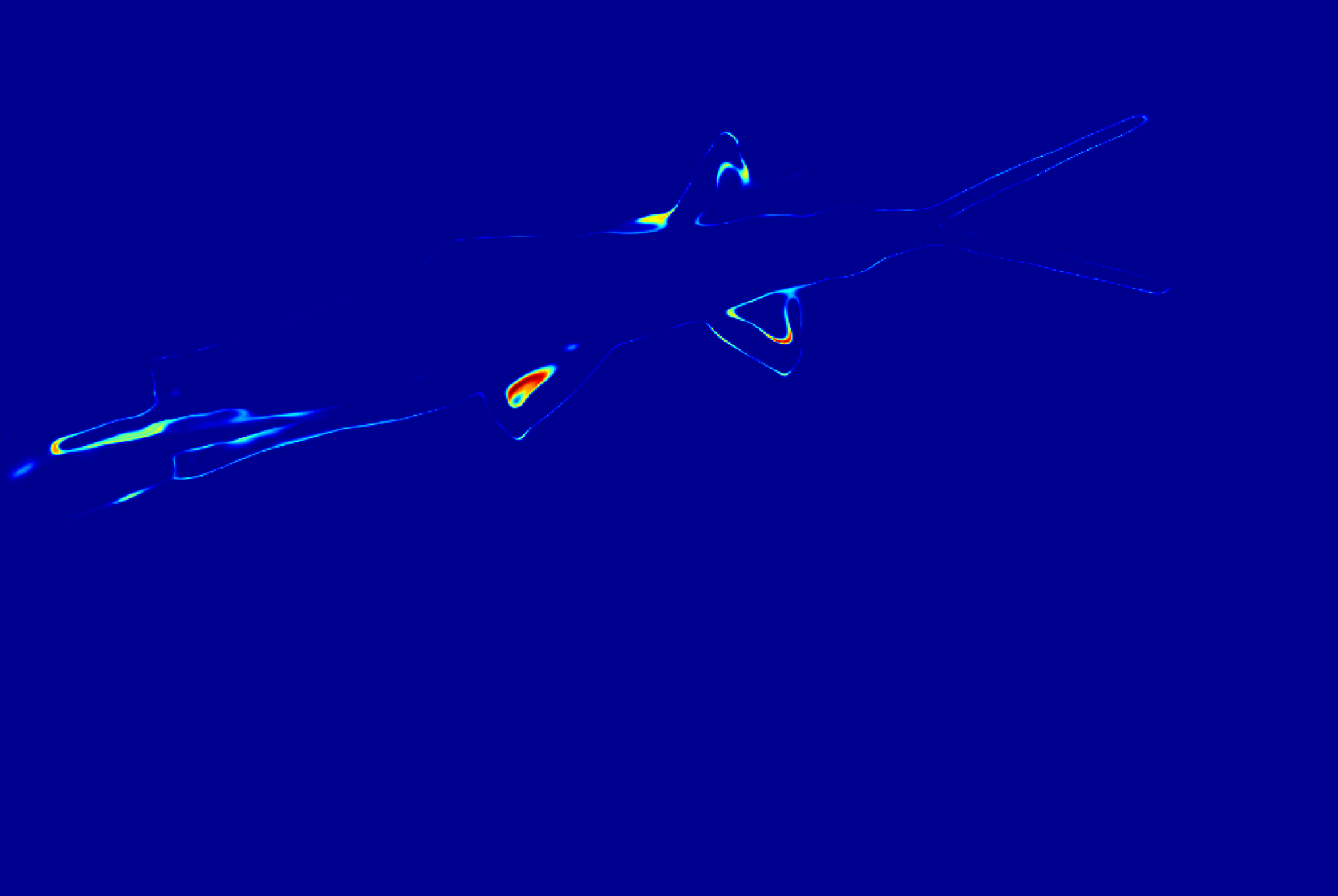}} \\
      {\includegraphics[width=0.11\linewidth]{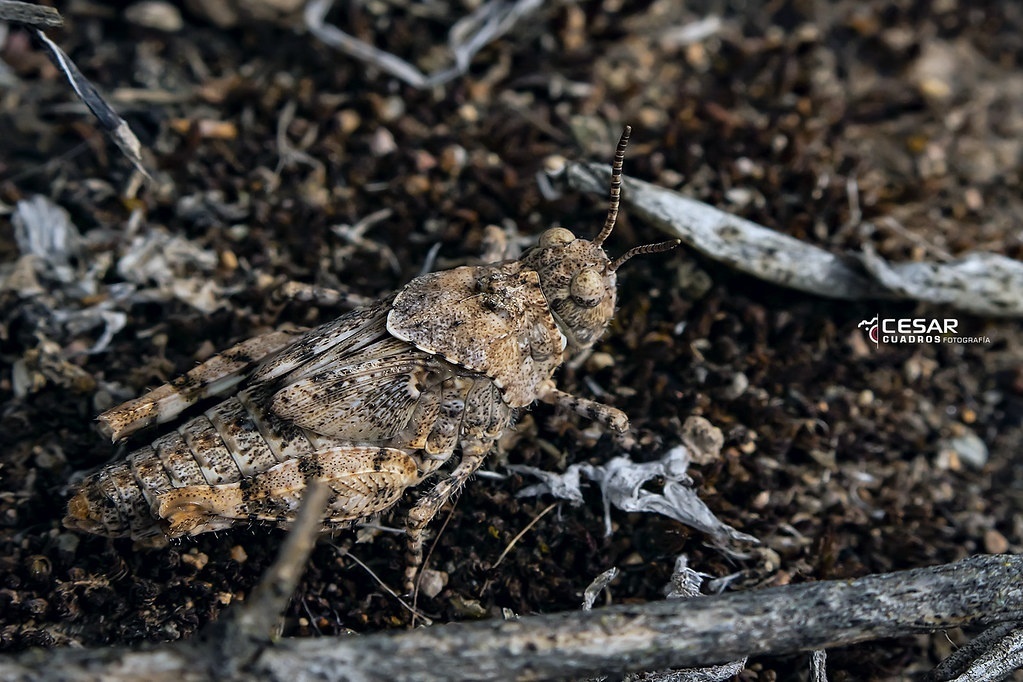}} &
   {\includegraphics[width=0.11\linewidth]{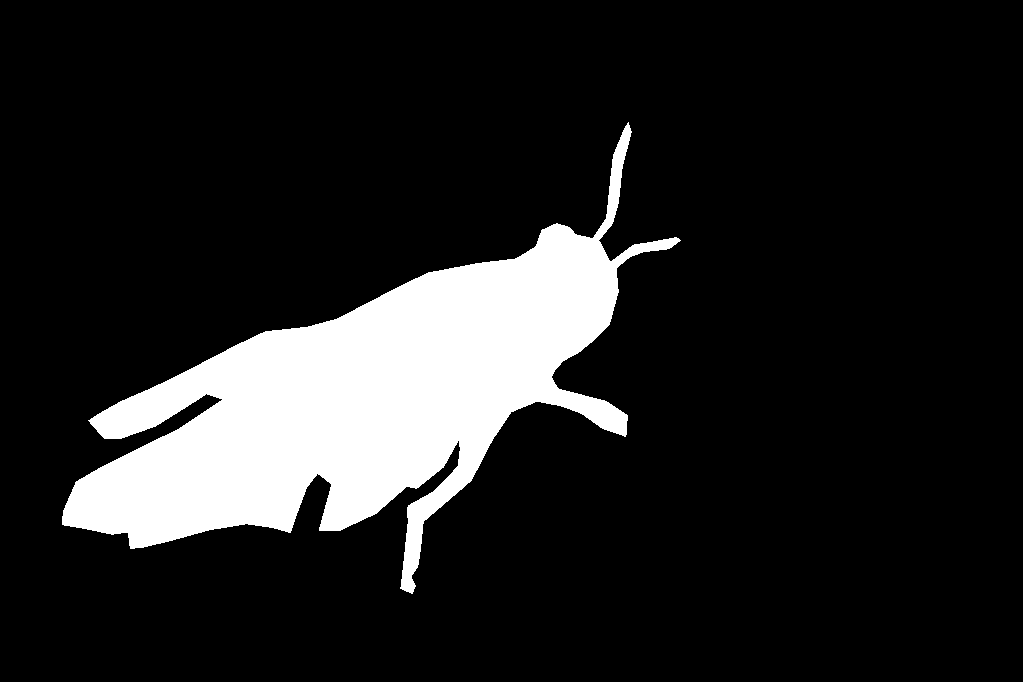}} &
   {\includegraphics[width=0.11\linewidth]{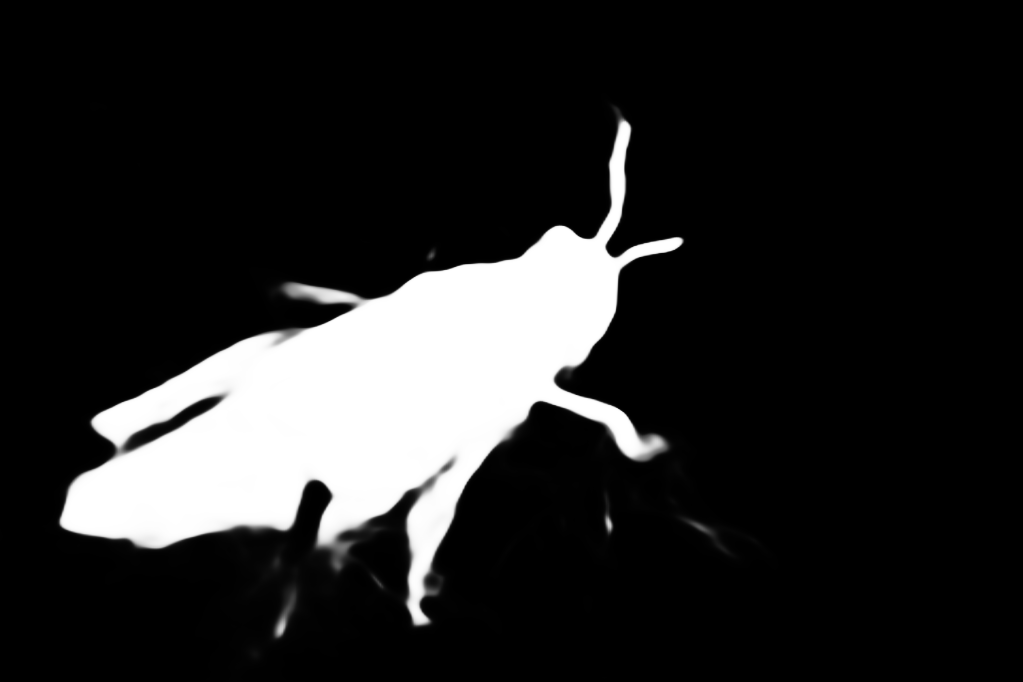}} &
   {\includegraphics[width=0.11\linewidth]{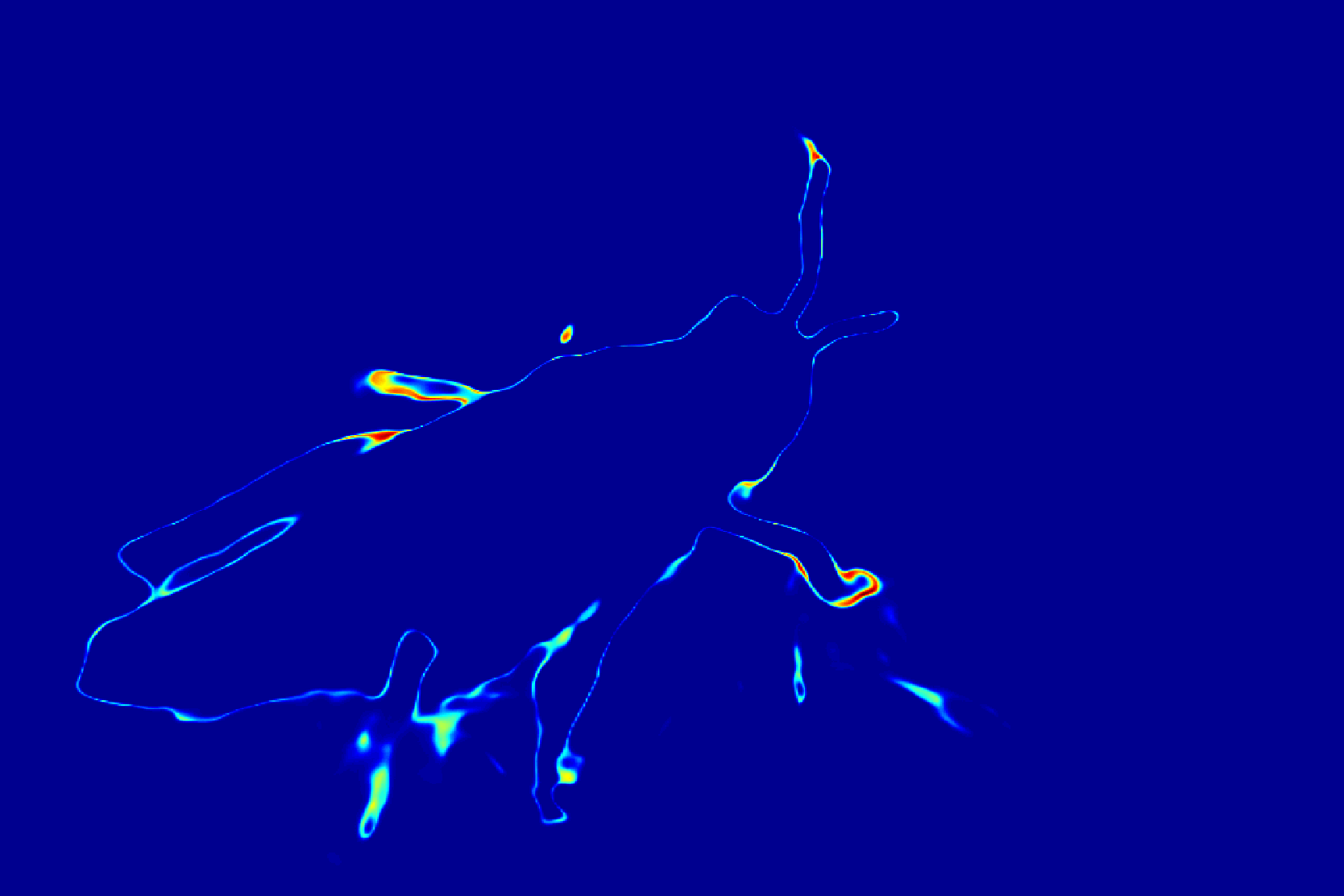}} &
   {\includegraphics[width=0.11\linewidth]{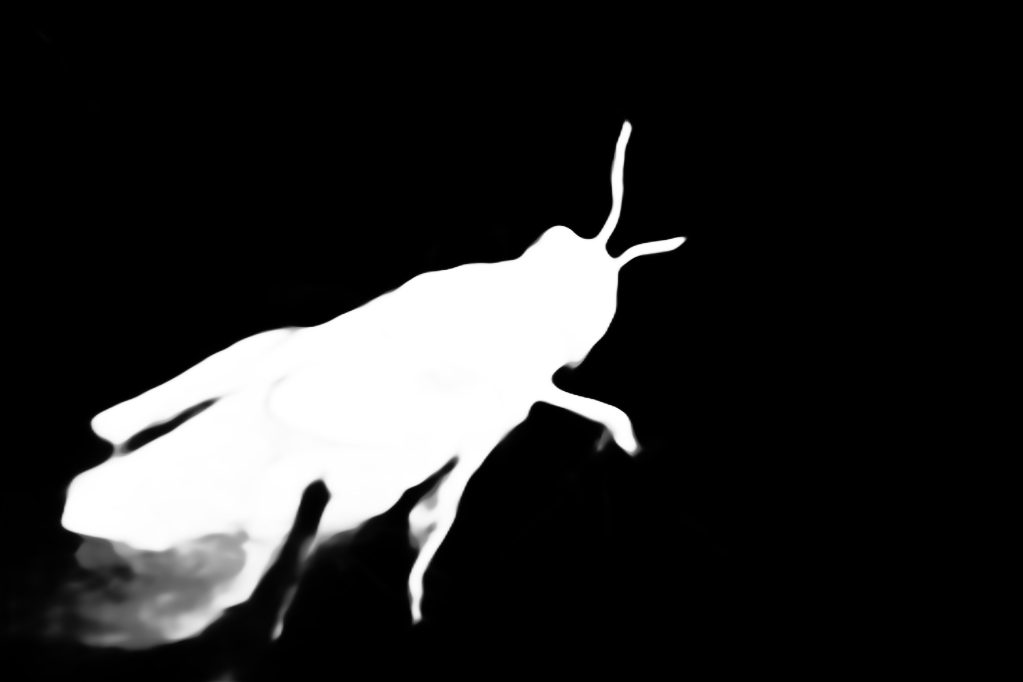}}&
   {\includegraphics[width=0.11\linewidth]{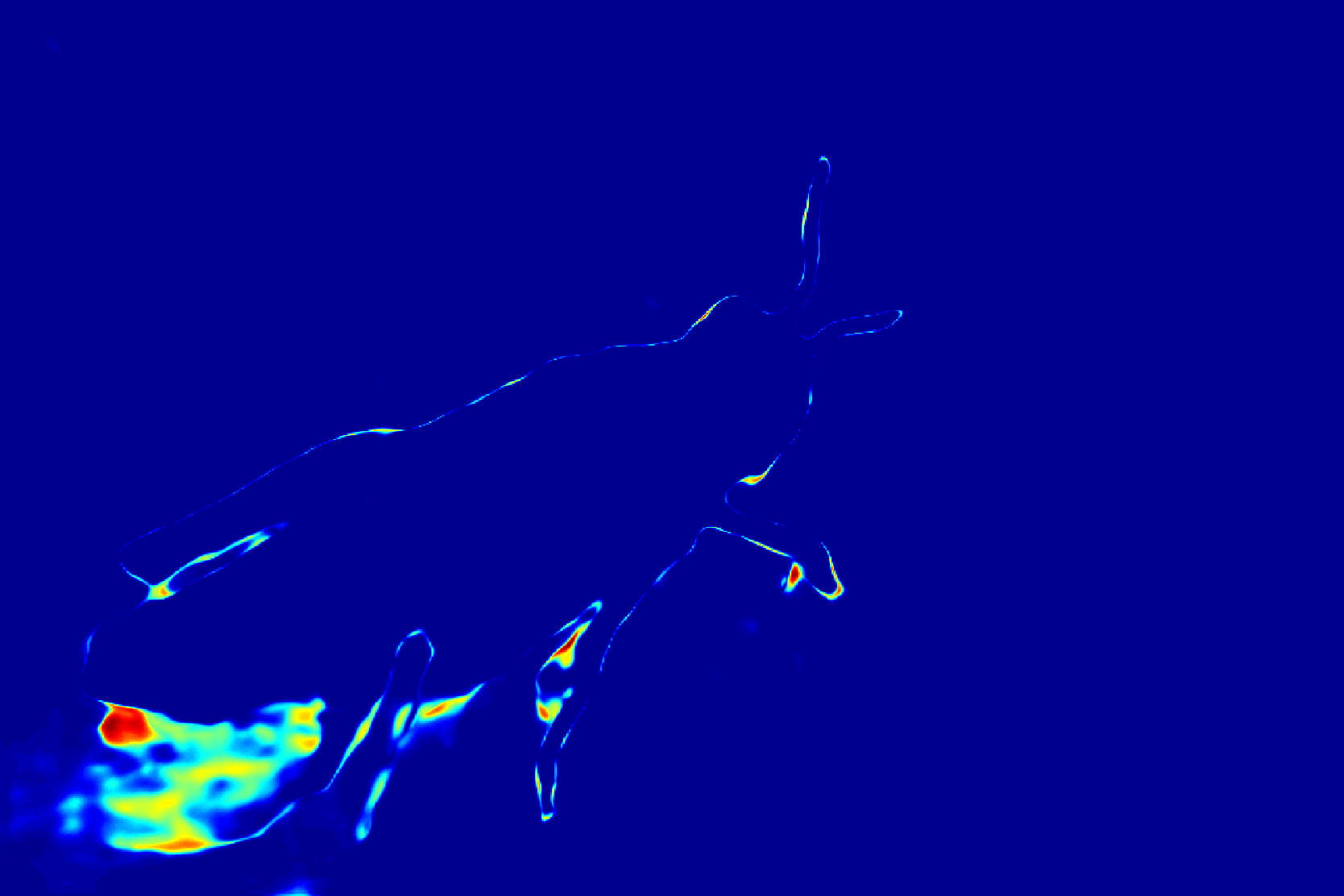}}&
   {\includegraphics[width=0.11\linewidth]{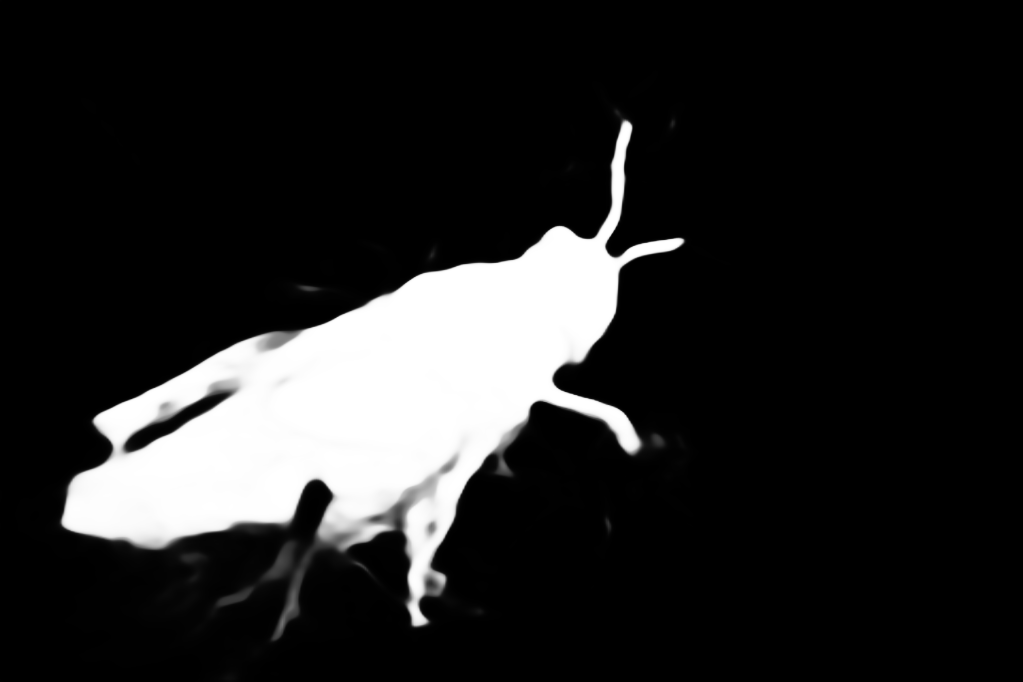}}&
   {\includegraphics[width=0.11\linewidth]{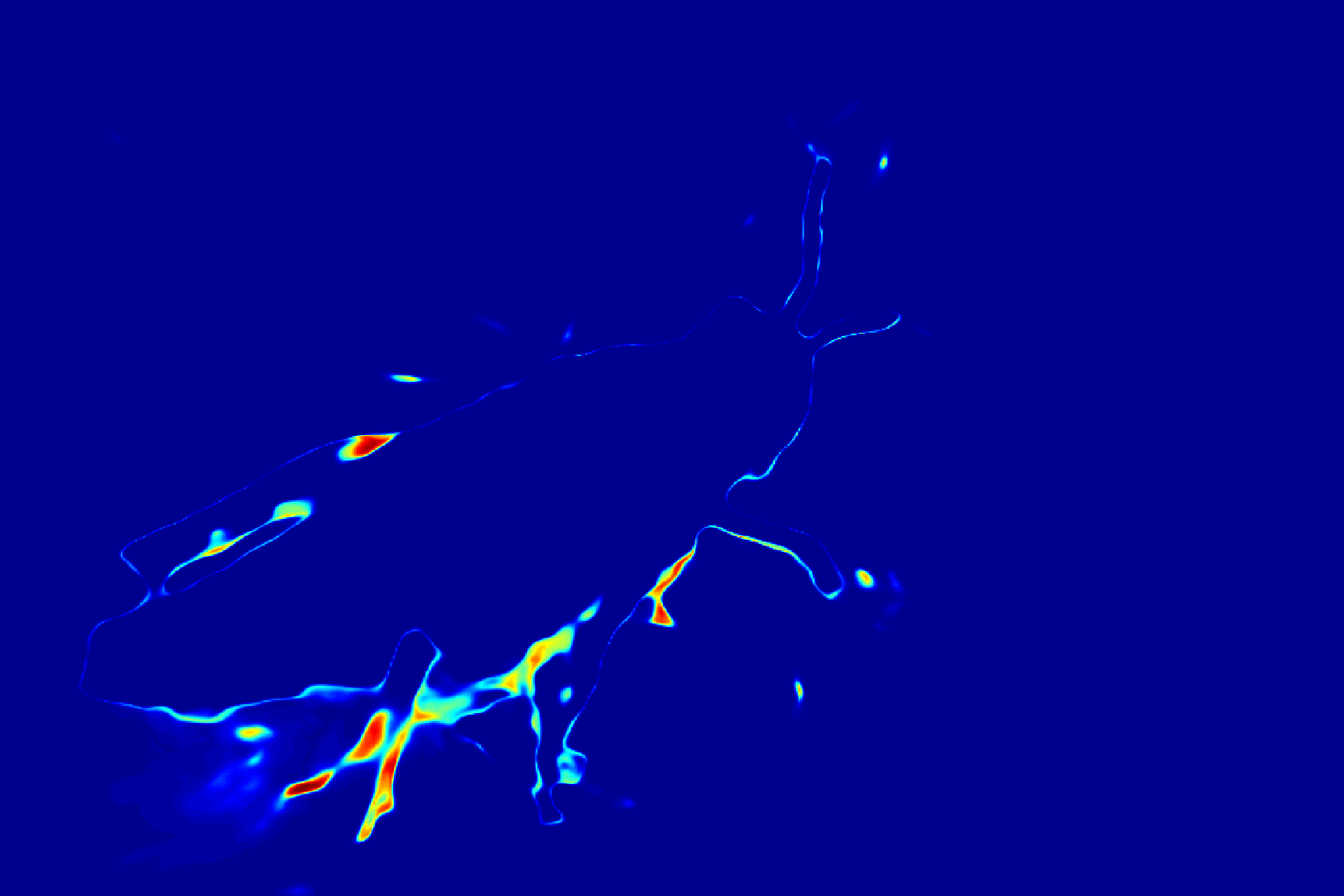}} \\
      {\includegraphics[width=0.11\linewidth]{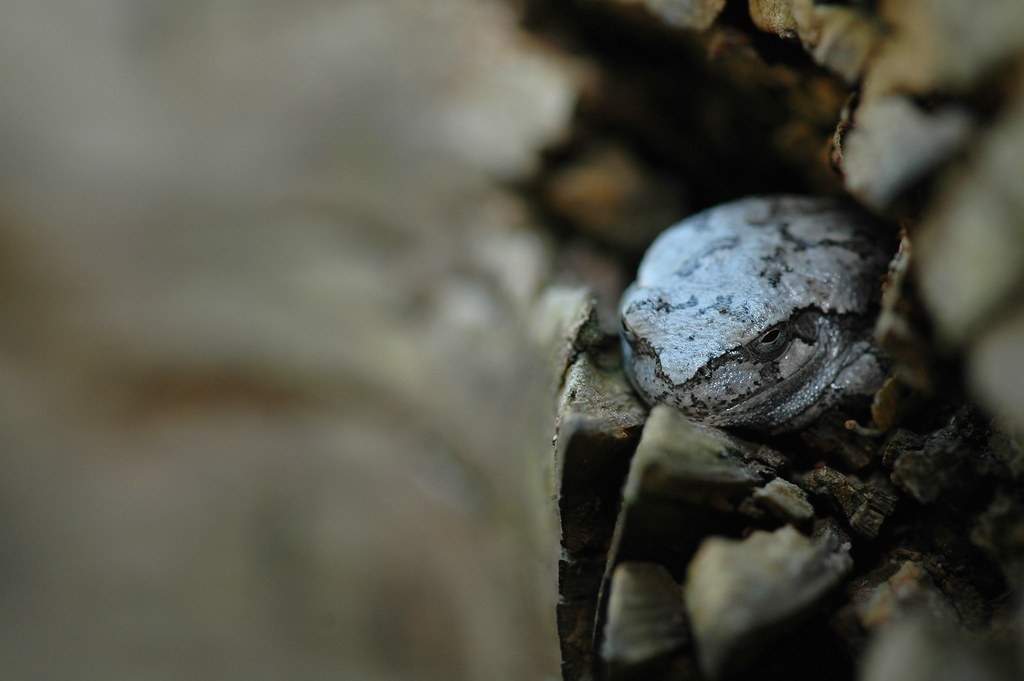}} &
   {\includegraphics[width=0.11\linewidth]{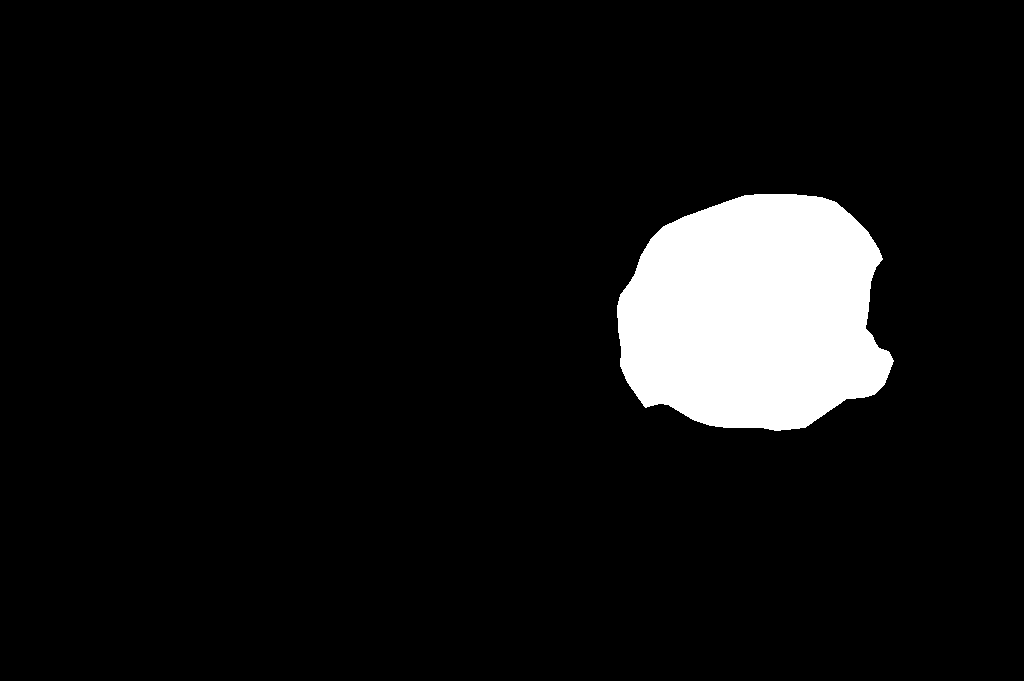}} &
   {\includegraphics[width=0.11\linewidth]{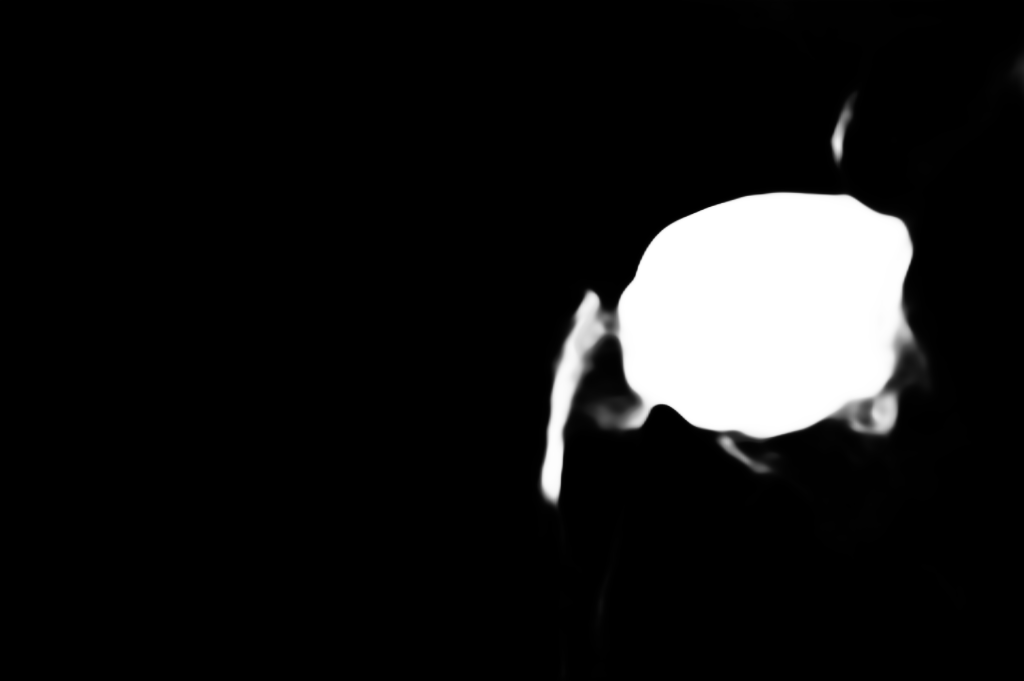}} &
   {\includegraphics[width=0.11\linewidth]{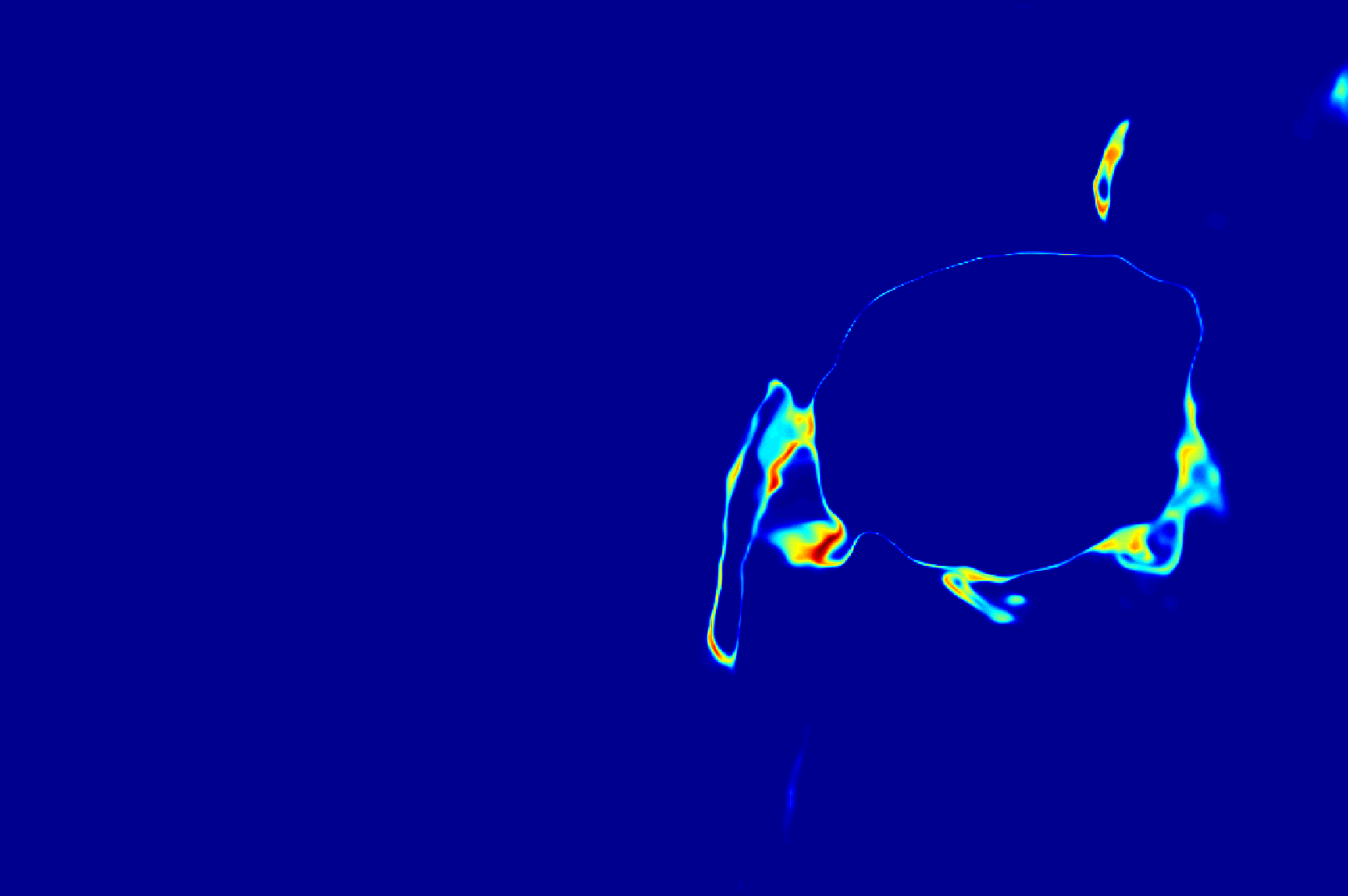}} &
   {\includegraphics[width=0.11\linewidth]{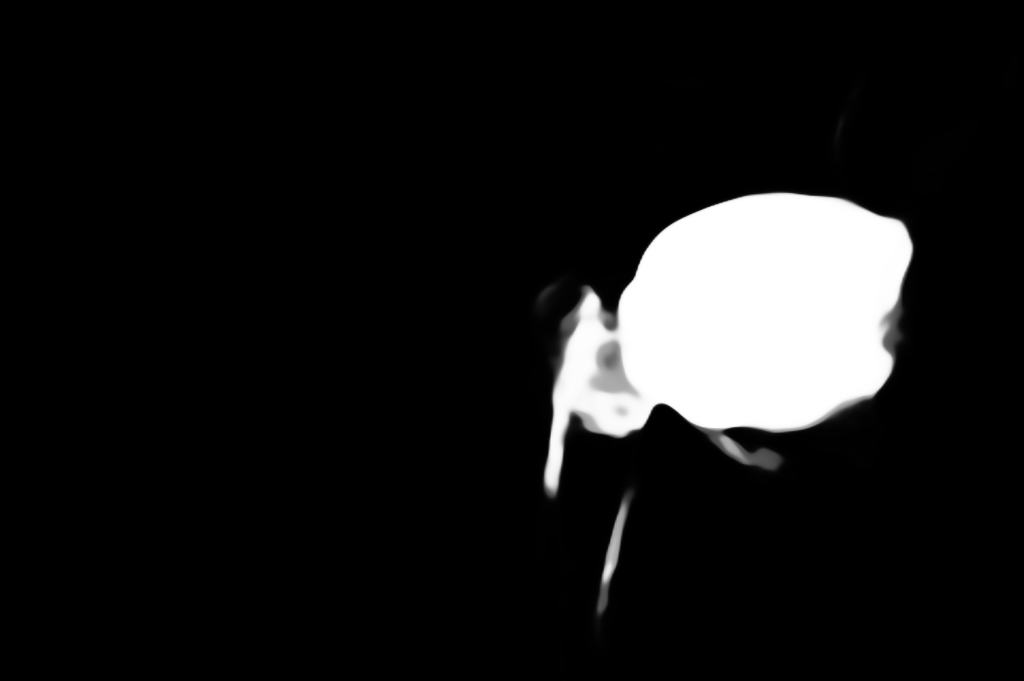}}&
   {\includegraphics[width=0.11\linewidth]{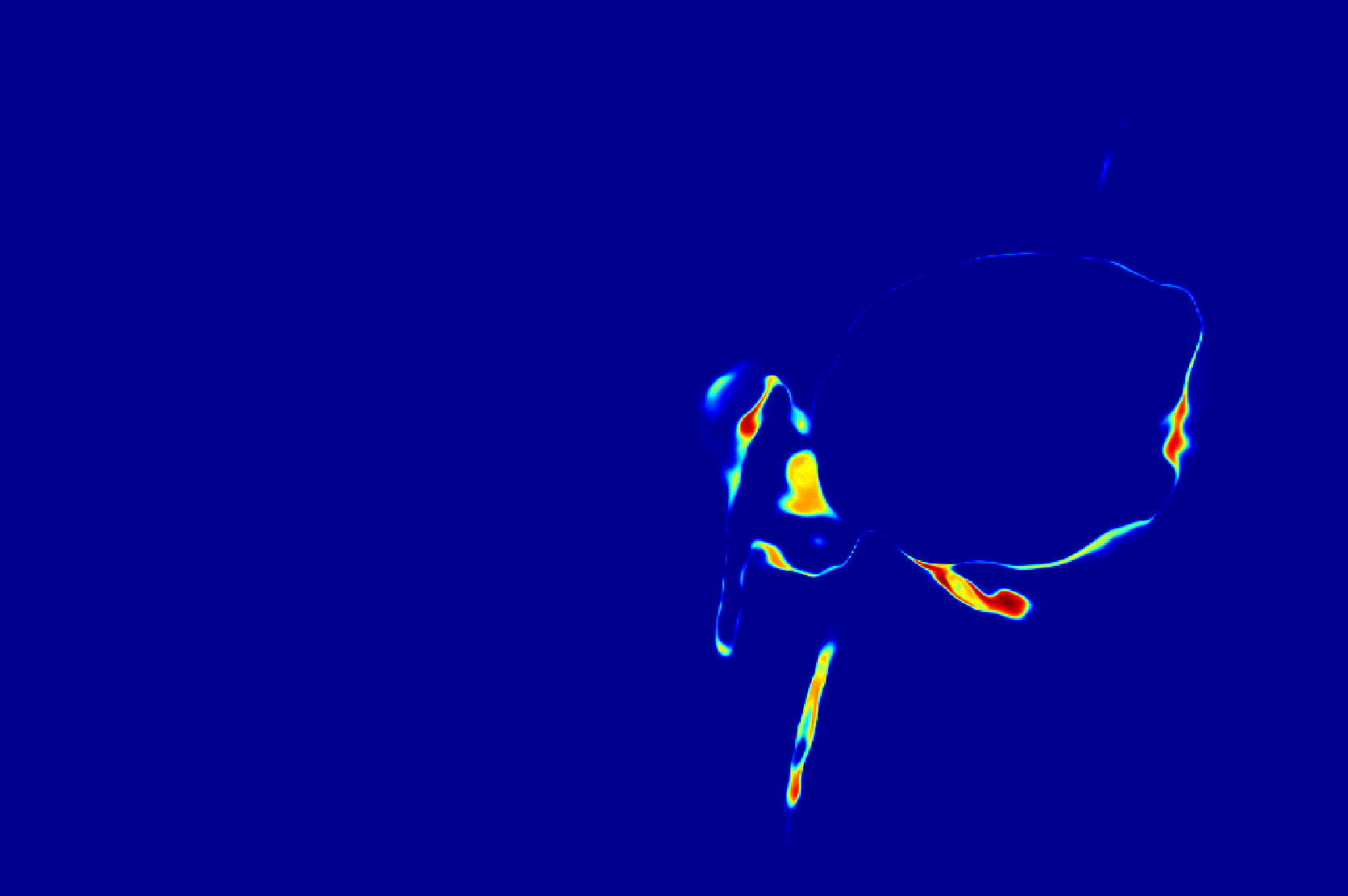}}&
   {\includegraphics[width=0.11\linewidth]{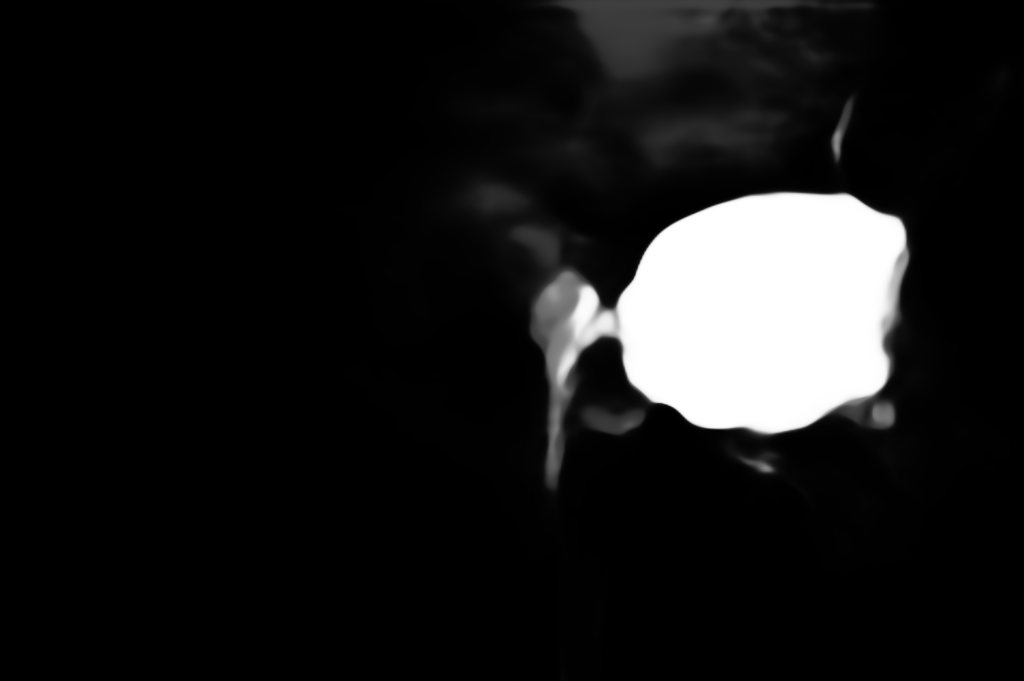}}&
   {\includegraphics[width=0.11\linewidth]{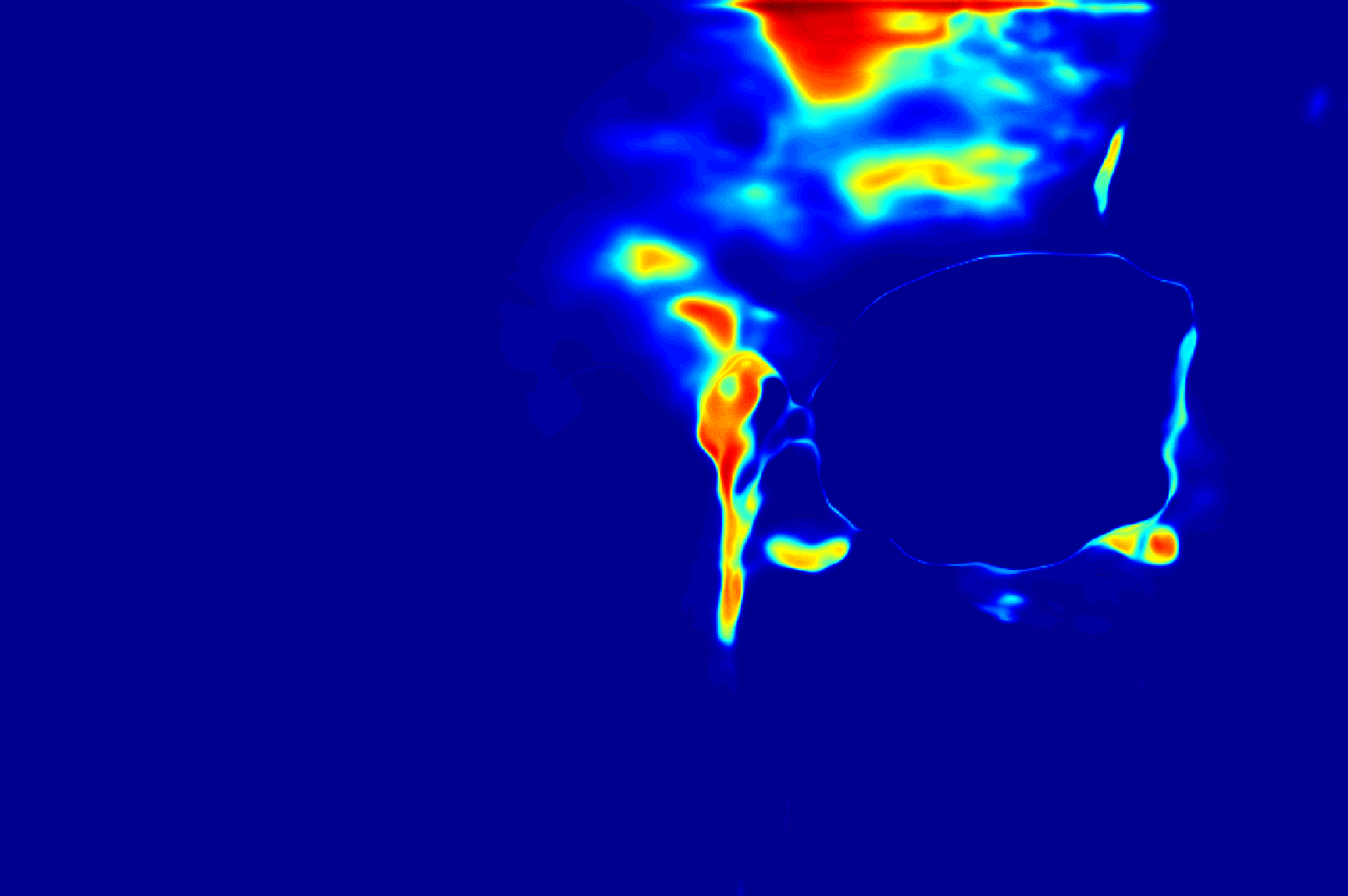}} \\
      {\includegraphics[width=0.11\linewidth]{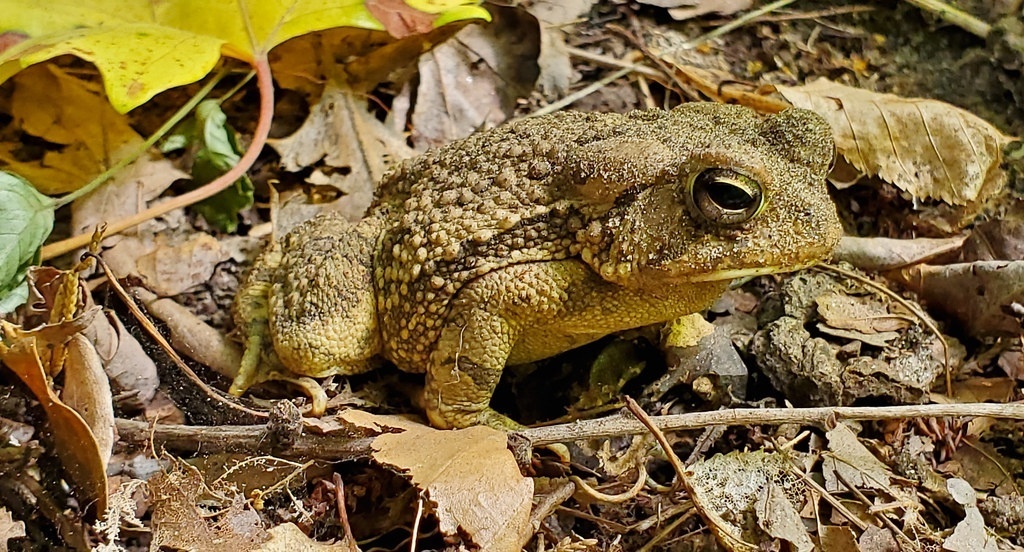}} &
   {\includegraphics[width=0.11\linewidth]{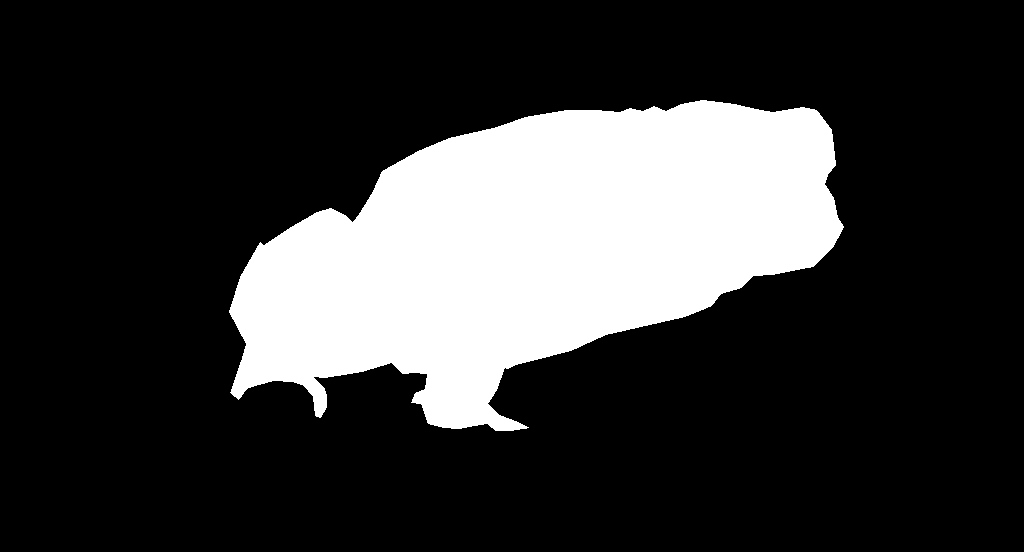}} &
   {\includegraphics[width=0.11\linewidth]{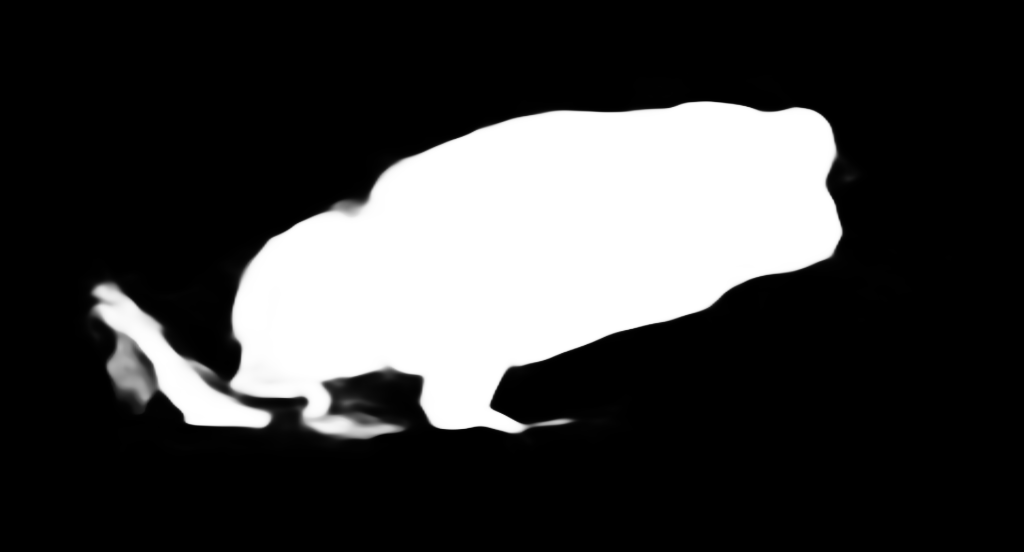}} &
   {\includegraphics[width=0.11\linewidth]{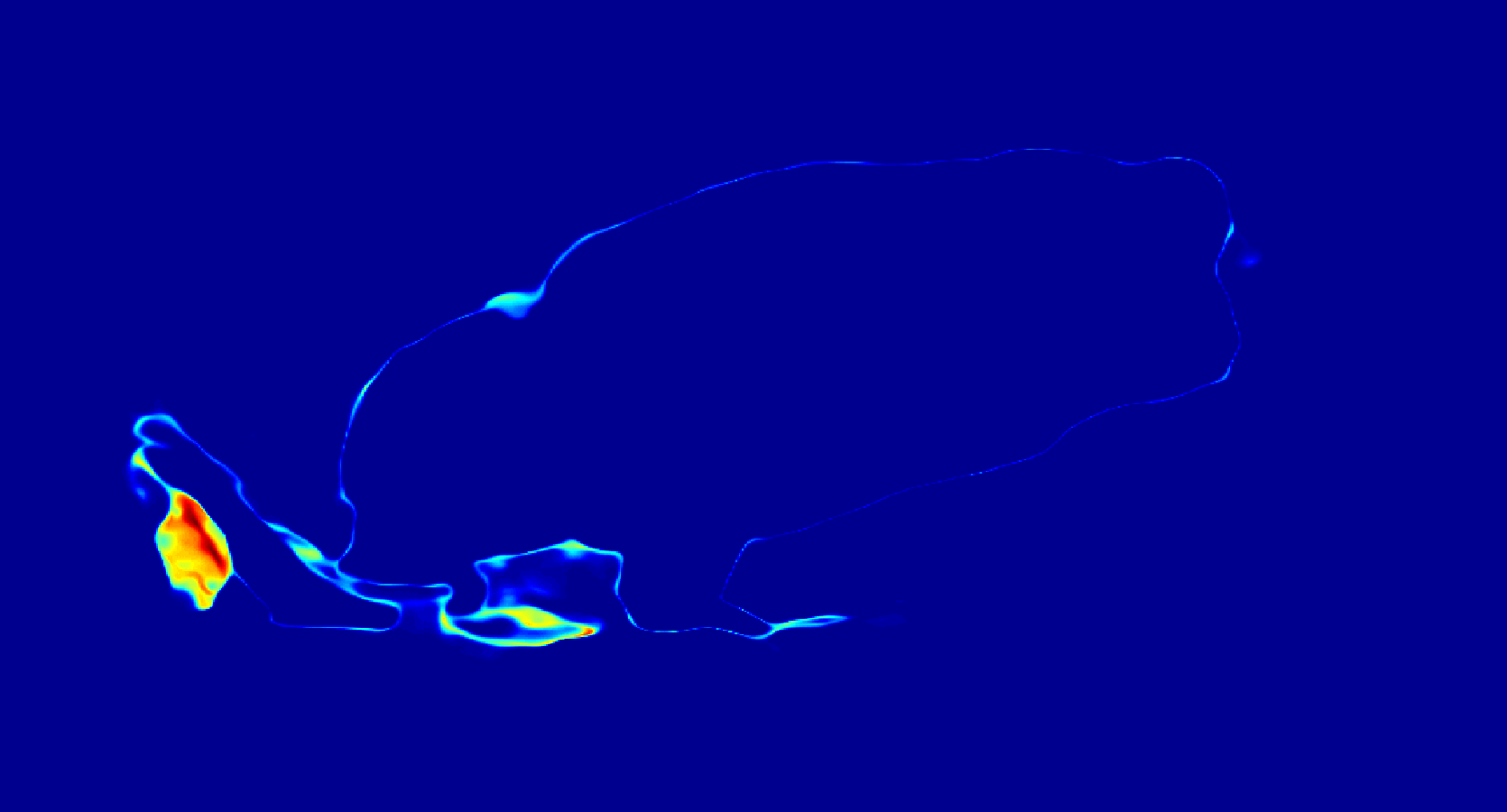}} &
   {\includegraphics[width=0.11\linewidth]{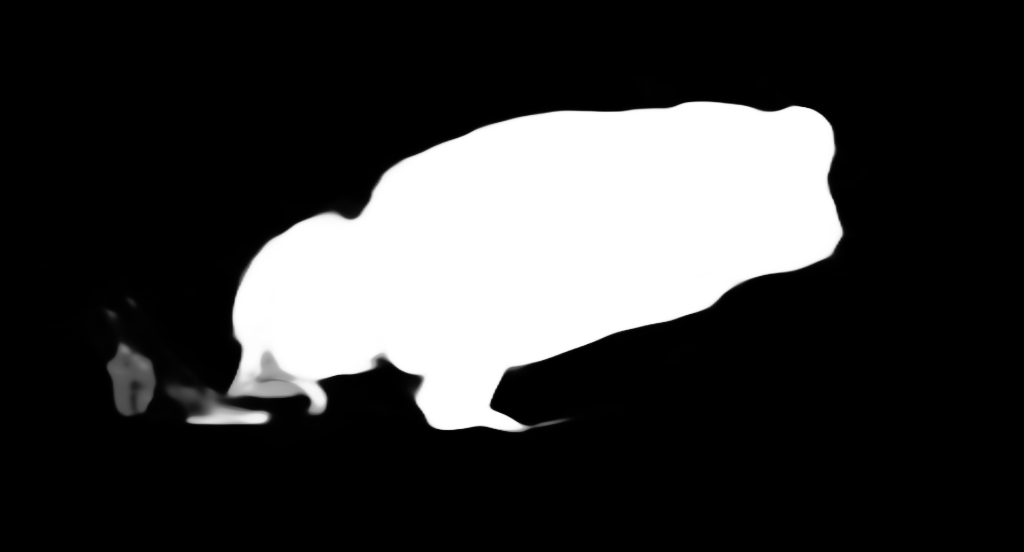}}&
   {\includegraphics[width=0.11\linewidth]{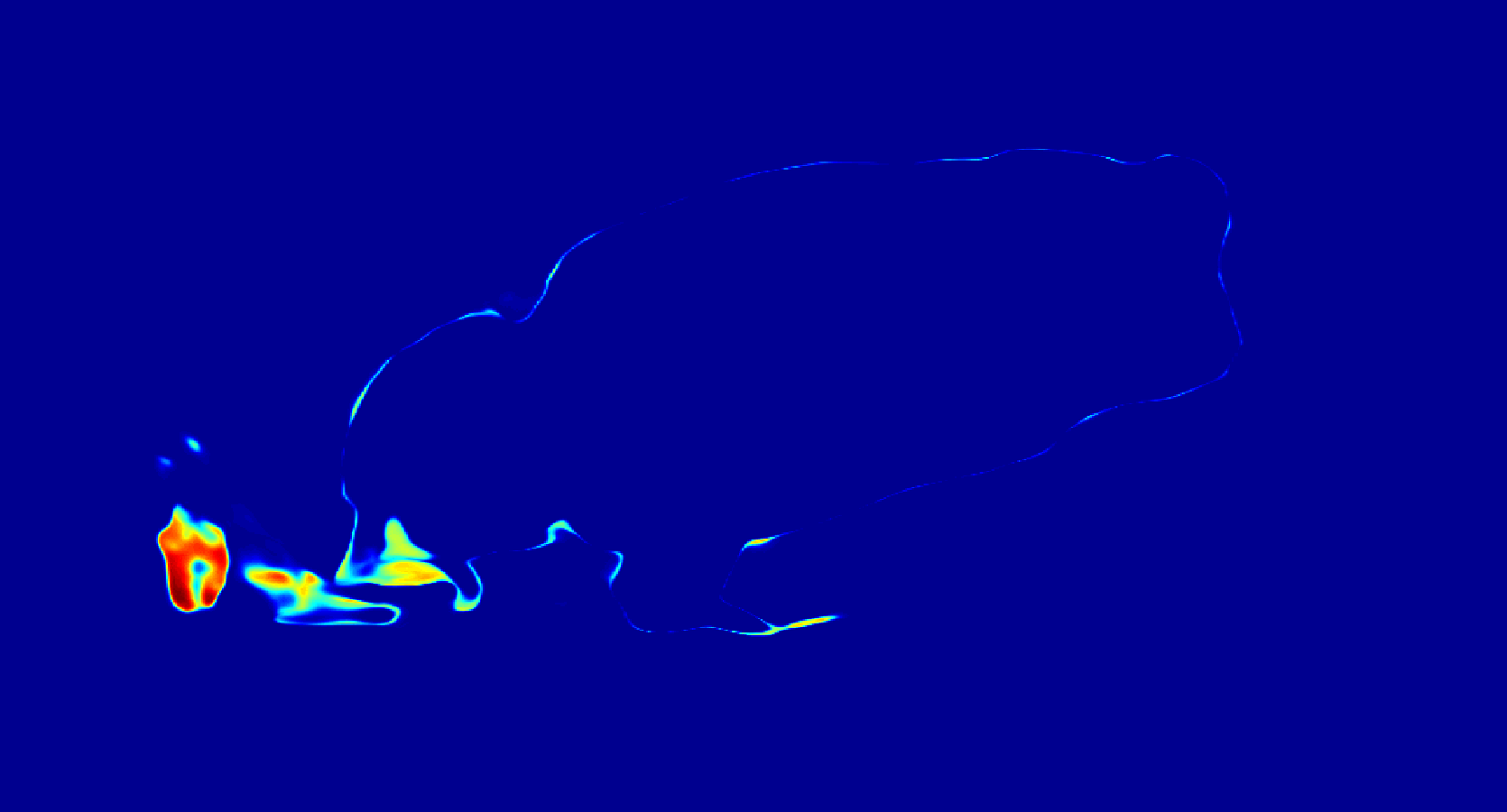}}&
   {\includegraphics[width=0.11\linewidth]{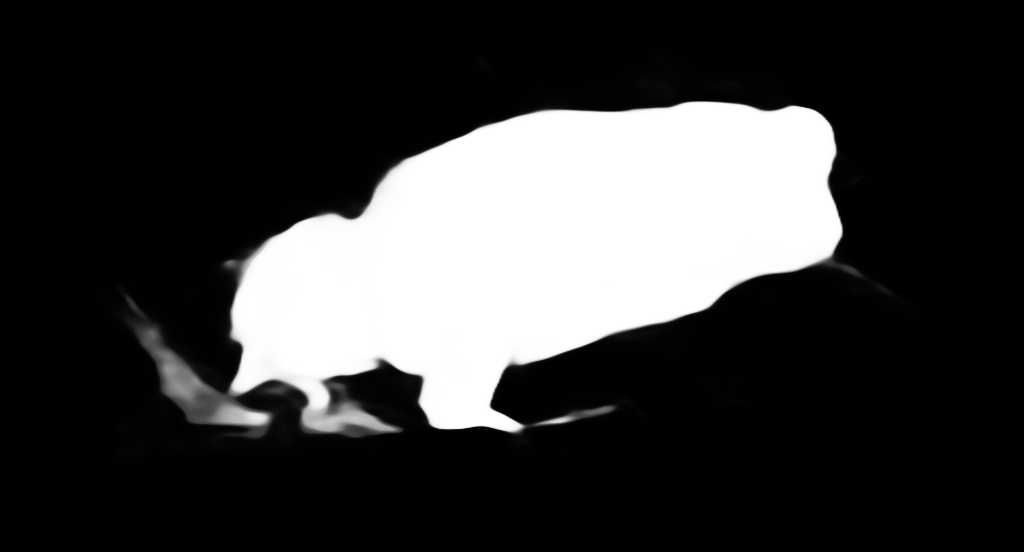}}&
   {\includegraphics[width=0.11\linewidth]{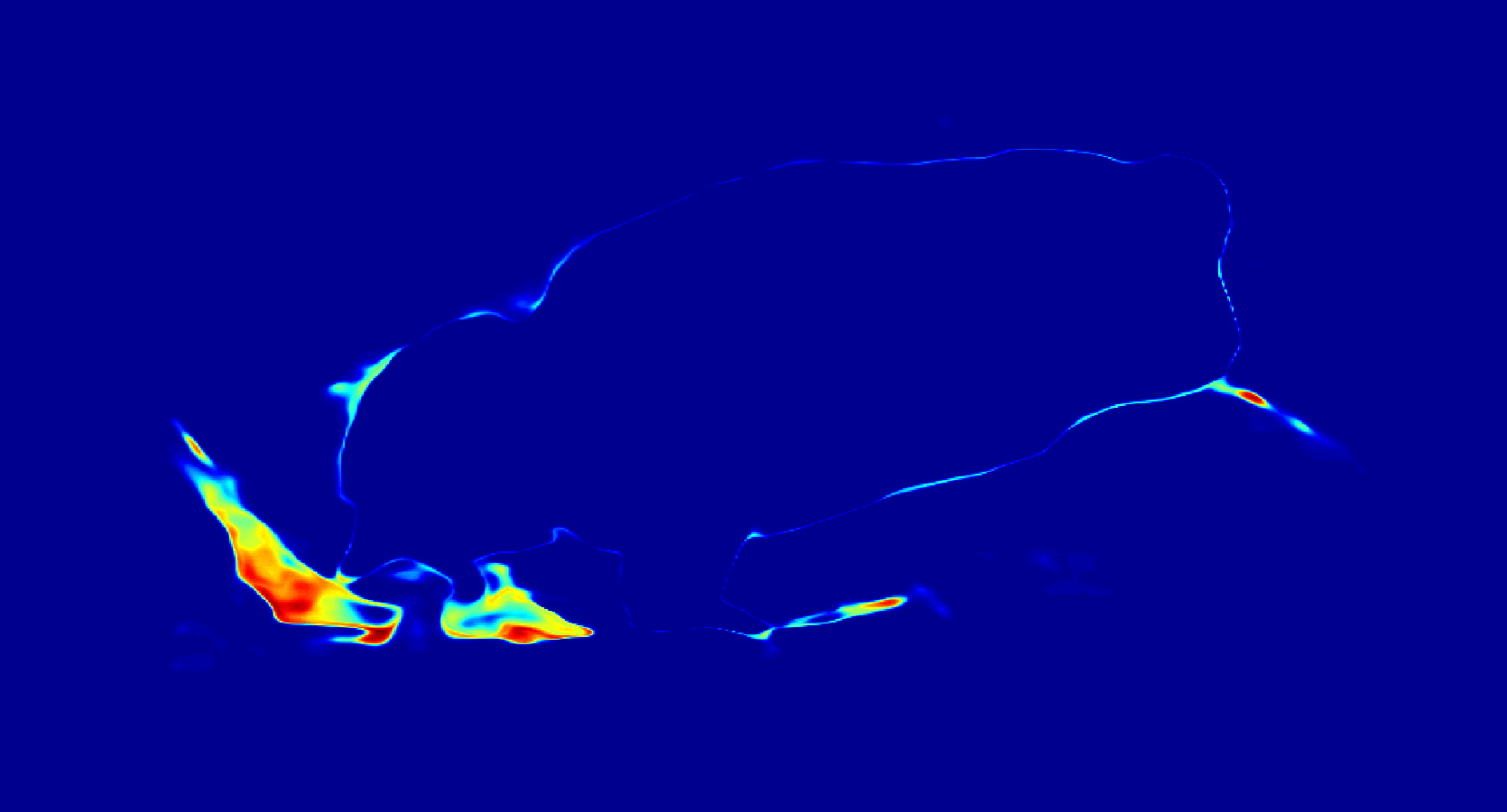}} \\
   \footnotesize{Image}&\footnotesize{GT}&\footnotesize{MD}&\footnotesize{$U_e$}&\footnotesize{DE}&\footnotesize{$U_e$}&\footnotesize{SE}&\footnotesize{$U_e$}\\
   \end{tabular}
   \end{center}
   \caption{\footnotesize{Epistemic uncertainty of ensemble based solutions for \textbf{camouflaged object detection}.}
   }
\label{fig:epistemic_ensemble_cod}
\end{figure}

\begin{figure}[tp]
   \begin{center}
   \begin{tabular}{c@{ }c@{ }c@{ }c@{ }c@{ }c@{ }c@{ }c@{ }}
   {\includegraphics[width=0.11\linewidth]{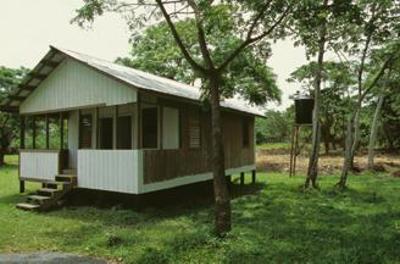}} &
   {\includegraphics[width=0.11\linewidth]{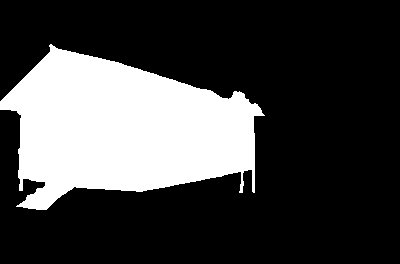}} &
   {\includegraphics[width=0.11\linewidth]{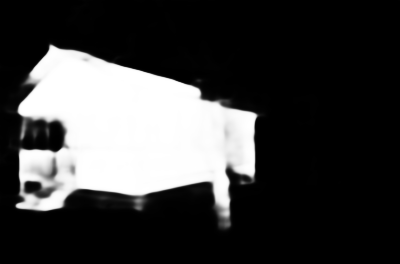}} &
   {\includegraphics[width=0.11\linewidth]{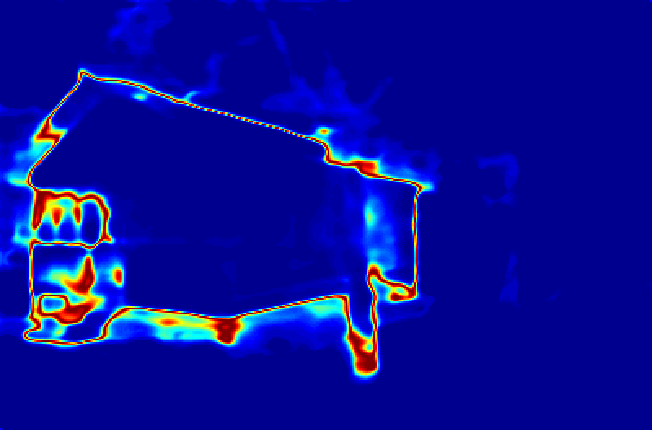}} &
   {\includegraphics[width=0.11\linewidth]{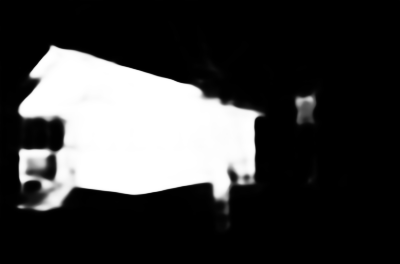}}&
   {\includegraphics[width=0.11\linewidth]{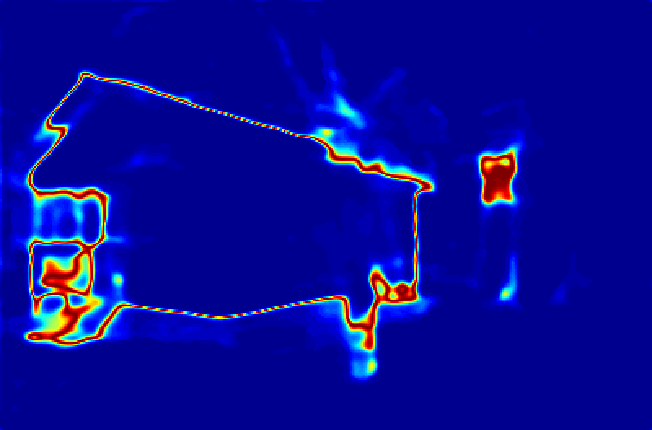}}&
   {\includegraphics[width=0.11\linewidth]{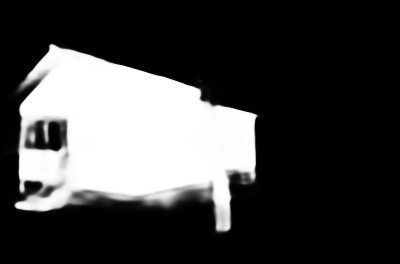}}&
   {\includegraphics[width=0.11\linewidth]{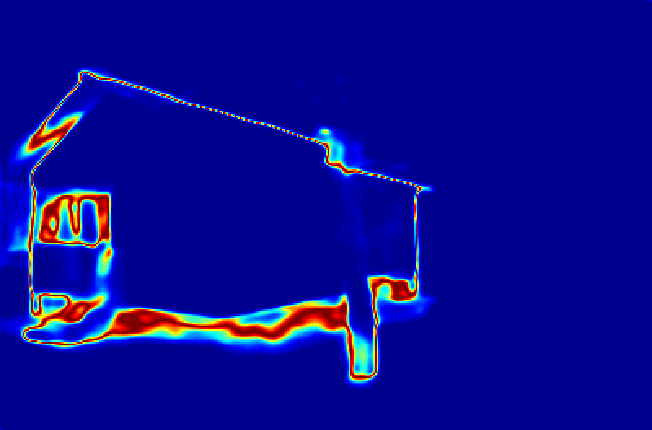}} \\
      {\includegraphics[width=0.11\linewidth]{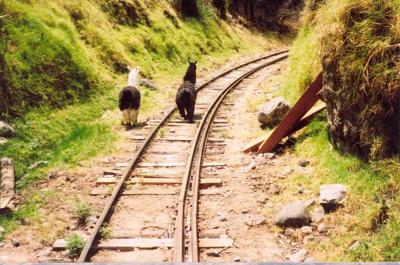}} &
   {\includegraphics[width=0.11\linewidth]{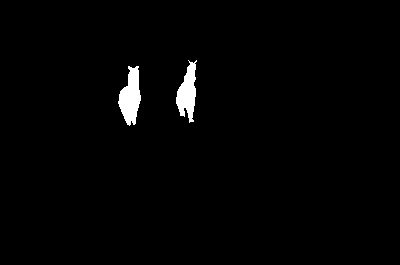}} &
   {\includegraphics[width=0.11\linewidth]{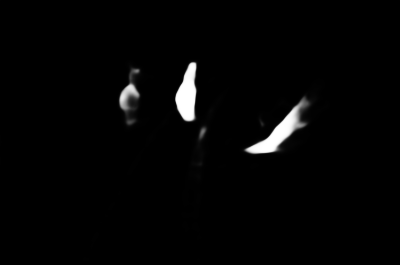}} &
   {\includegraphics[width=0.11\linewidth]{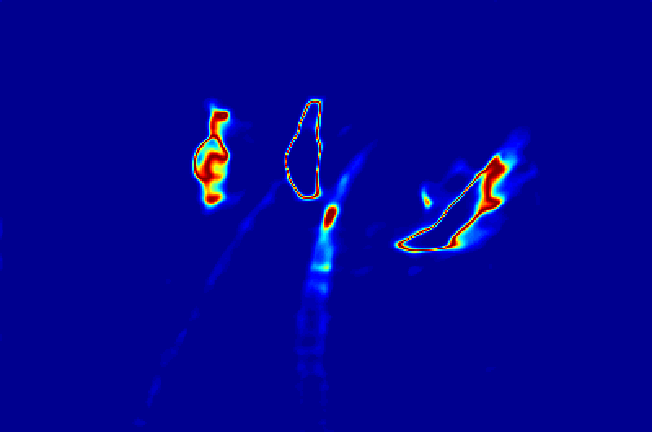}} &
   {\includegraphics[width=0.11\linewidth]{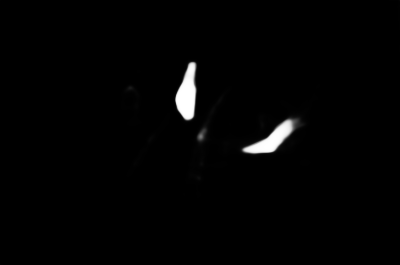}}&
   {\includegraphics[width=0.11\linewidth]{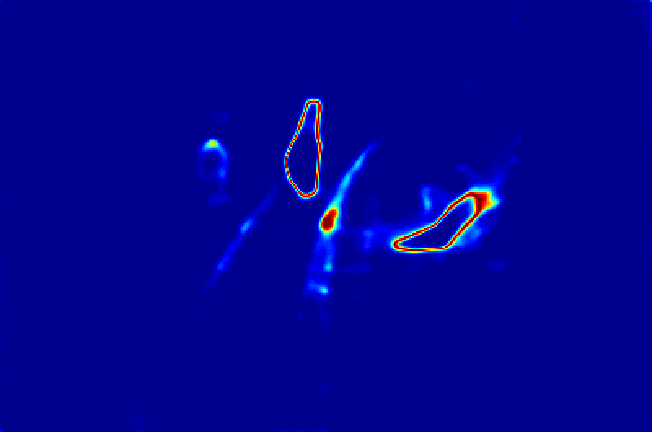}}&
   {\includegraphics[width=0.11\linewidth]{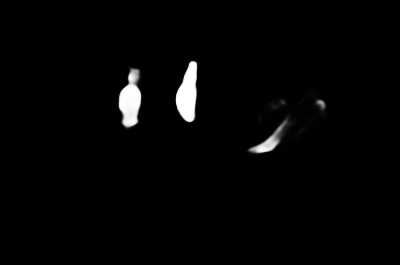}}&
   {\includegraphics[width=0.11\linewidth]{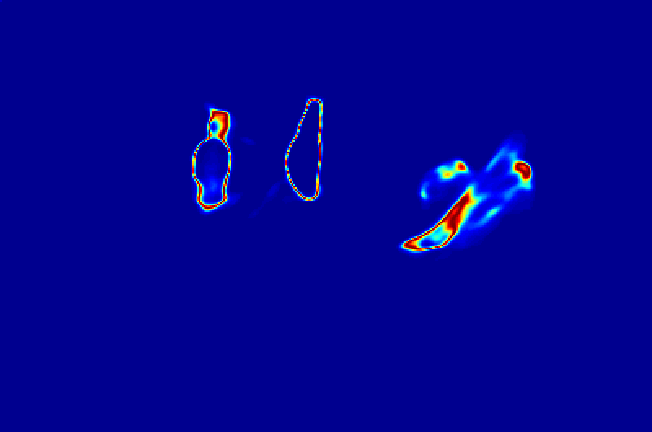}} \\
      {\includegraphics[width=0.11\linewidth]{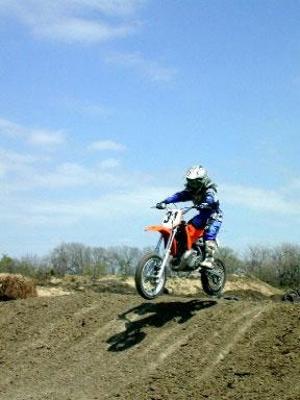}} &
   {\includegraphics[width=0.11\linewidth]{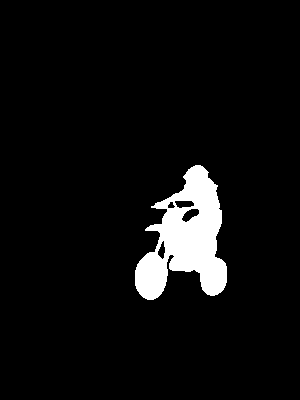}} &
   {\includegraphics[width=0.11\linewidth]{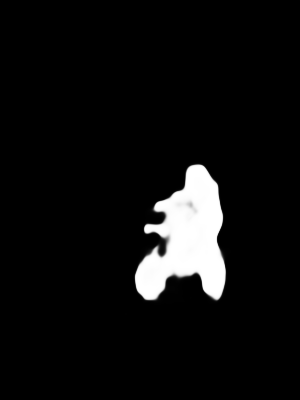}} &
   {\includegraphics[width=0.11\linewidth]{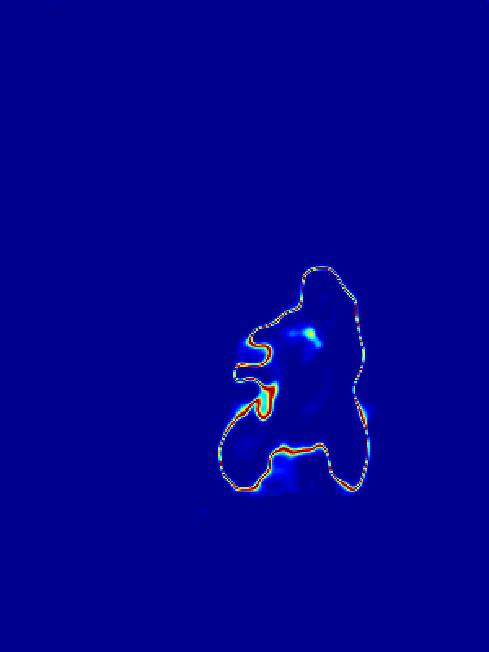}} &
   {\includegraphics[width=0.11\linewidth]{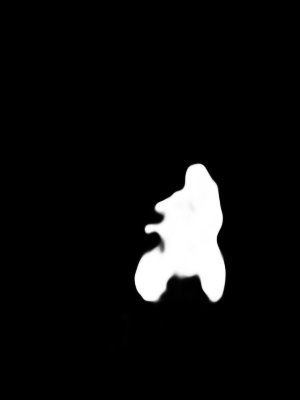}}&
   {\includegraphics[width=0.11\linewidth]{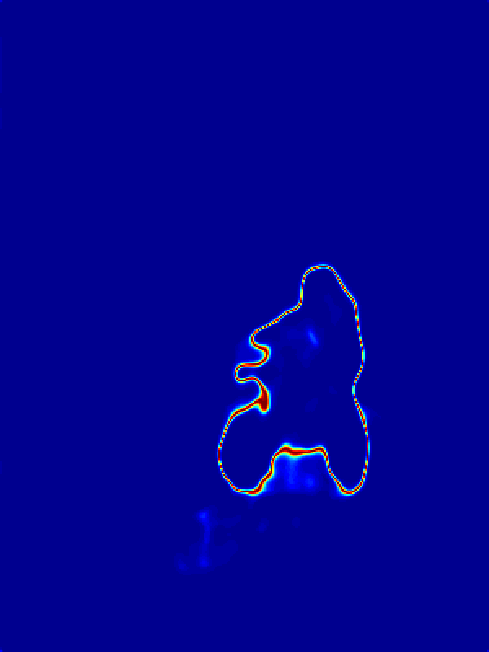}}&
   {\includegraphics[width=0.11\linewidth]{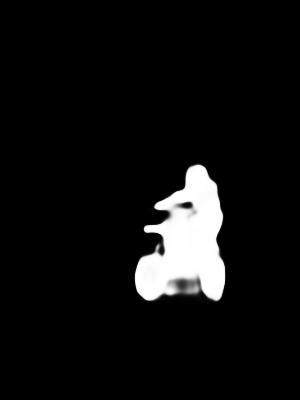}}&
   {\includegraphics[width=0.11\linewidth]{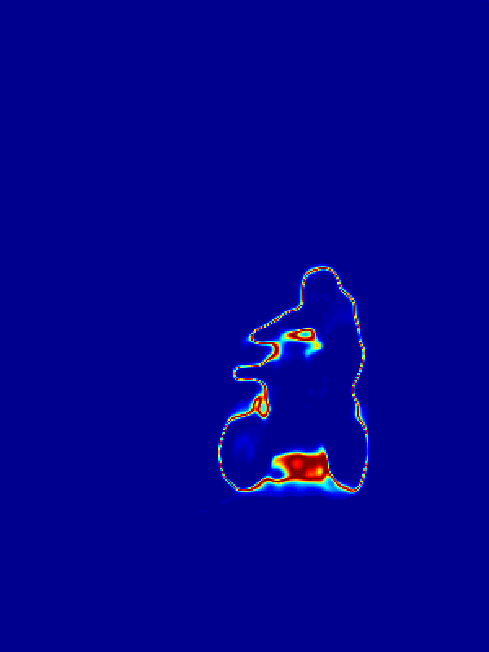}} \\
      {\includegraphics[width=0.11\linewidth]{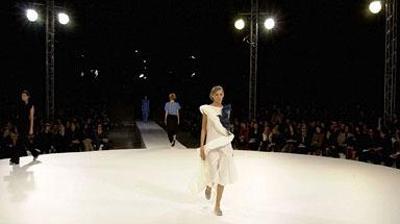}} &
   {\includegraphics[width=0.11\linewidth]{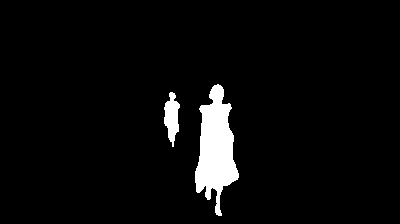}} &
   {\includegraphics[width=0.11\linewidth]{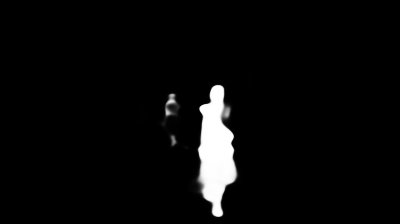}} &
   {\includegraphics[width=0.11\linewidth]{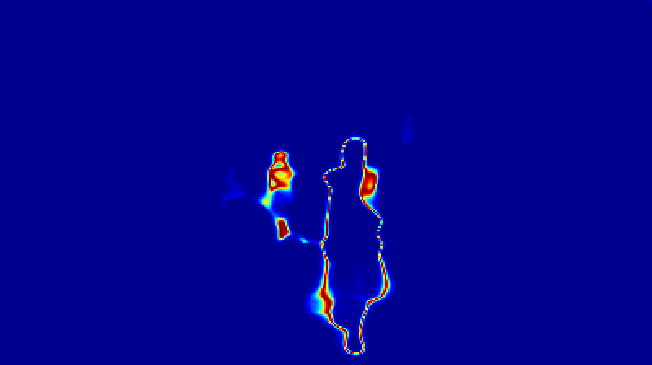}} &
   {\includegraphics[width=0.11\linewidth]{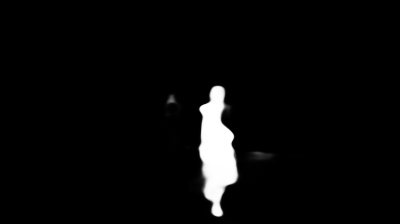}}&
   {\includegraphics[width=0.11\linewidth]{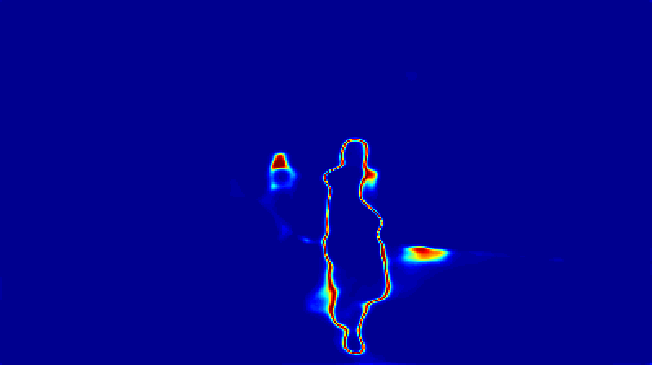}}&
   {\includegraphics[width=0.11\linewidth]{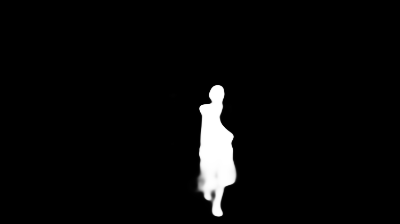}}&
   {\includegraphics[width=0.11\linewidth]{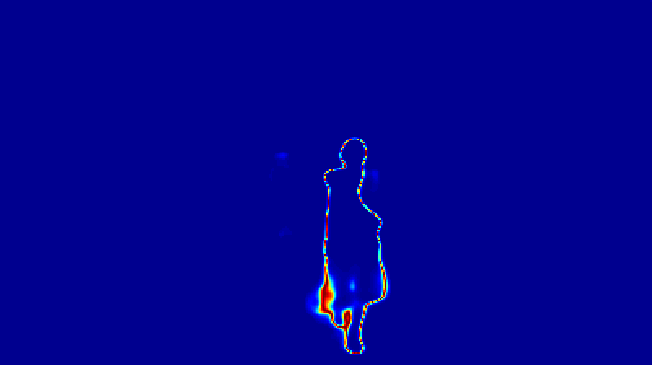}} \\
      {\includegraphics[width=0.11\linewidth]{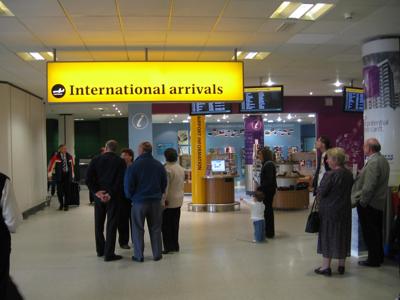}} &
   {\includegraphics[width=0.11\linewidth]{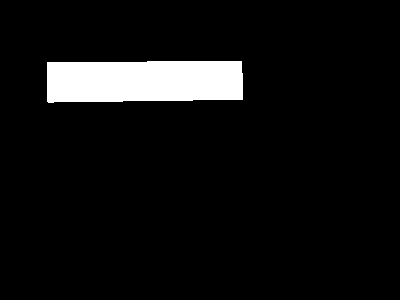}} &
   {\includegraphics[width=0.11\linewidth]{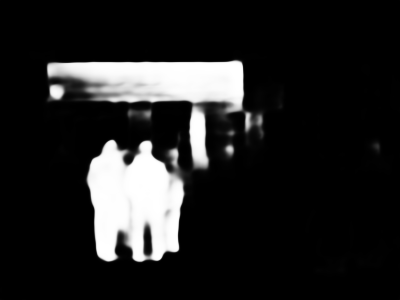}} &
   {\includegraphics[width=0.11\linewidth]{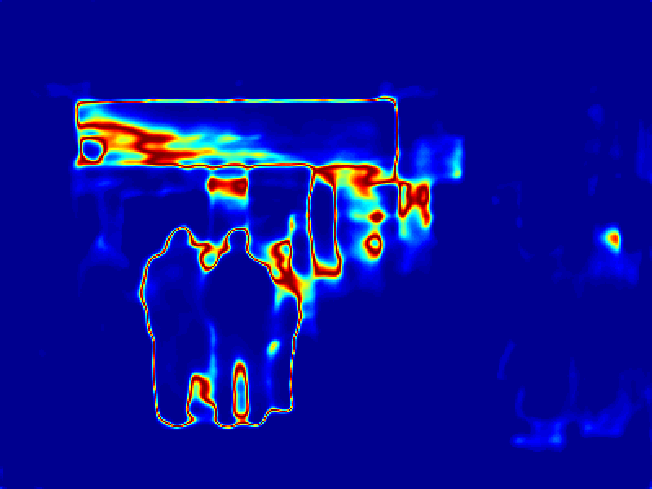}} &
   {\includegraphics[width=0.11\linewidth]{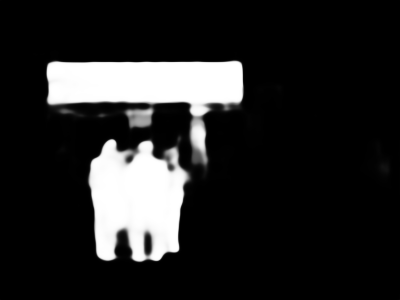}}&
   {\includegraphics[width=0.11\linewidth]{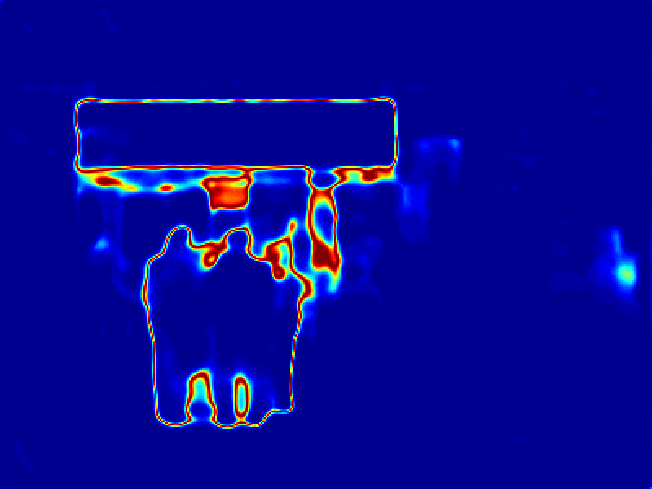}}&
   {\includegraphics[width=0.11\linewidth]{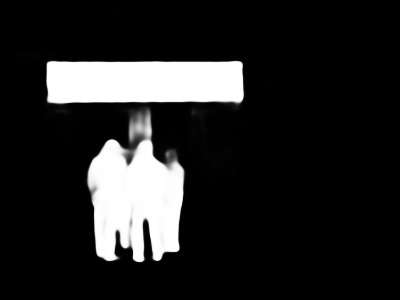}}&
   {\includegraphics[width=0.11\linewidth]{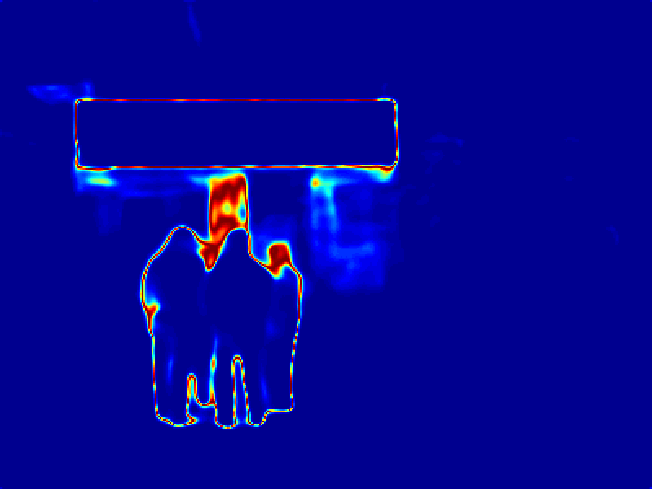}} \\
   \footnotesize{Image}&\footnotesize{GT}&\footnotesize{MD}&\footnotesize{$U_p$}&\footnotesize{DE}&\footnotesize{$U_p$}&\footnotesize{SE}&\footnotesize{$U_p$}\\
   \end{tabular}
   \end{center}
   \caption{\footnotesize{Predictive uncertainty of ensemble based solutions for \textbf{salient object detection}.}
   }
\label{fig:predictive_ensemble_sod}
\end{figure}

\begin{figure}[tp]
   \begin{center}
   \begin{tabular}{c@{ }c@{ }c@{ }c@{ }c@{ }c@{ }c@{ }c@{ }}
   {\includegraphics[width=0.11\linewidth]{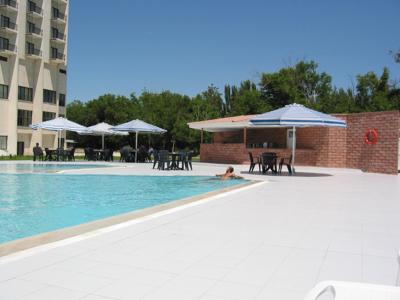}} &
   {\includegraphics[width=0.11\linewidth]{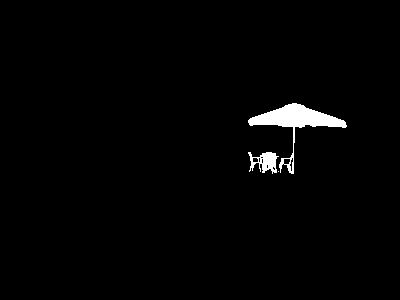}} &
   {\includegraphics[width=0.11\linewidth]{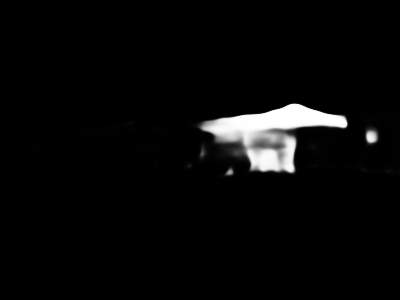}} &
   {\includegraphics[width=0.11\linewidth]{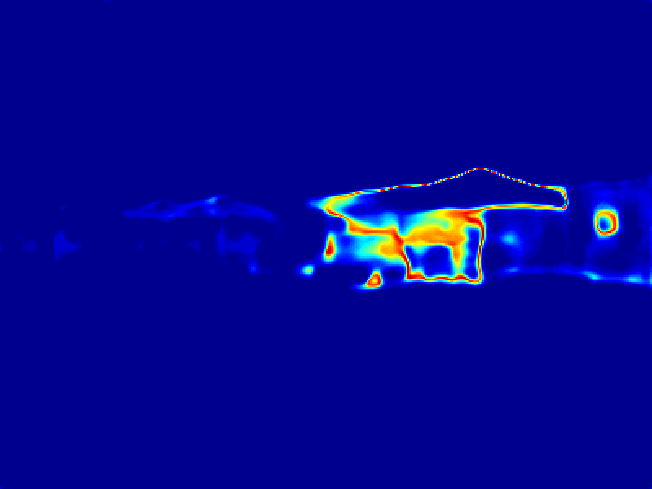}} &
   {\includegraphics[width=0.11\linewidth]{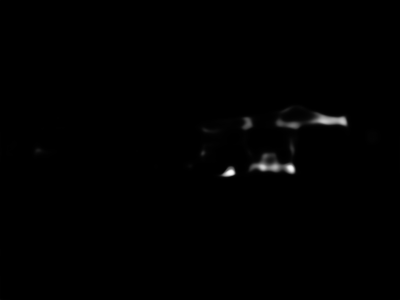}}&
   {\includegraphics[width=0.11\linewidth]{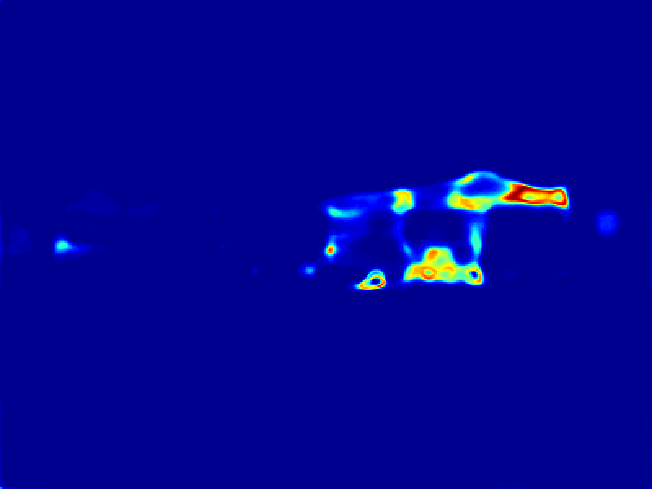}}&
   {\includegraphics[width=0.11\linewidth]{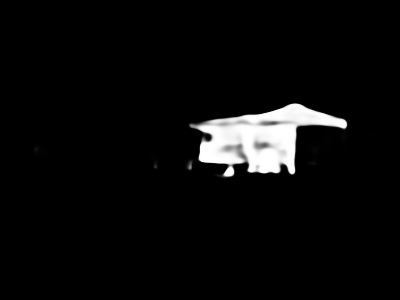}}&
   {\includegraphics[width=0.11\linewidth]{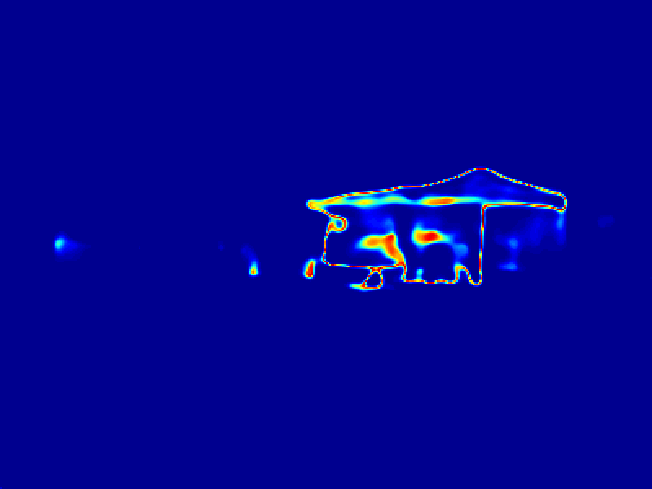}} \\
      {\includegraphics[width=0.11\linewidth]{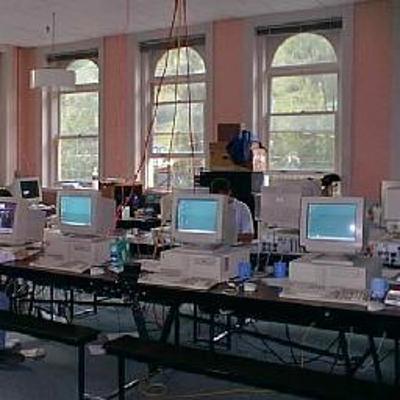}} &
   {\includegraphics[width=0.11\linewidth]{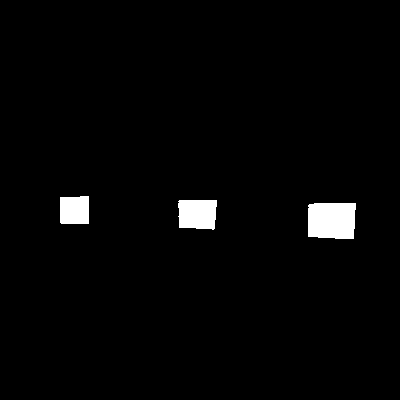}} &
   {\includegraphics[width=0.11\linewidth]{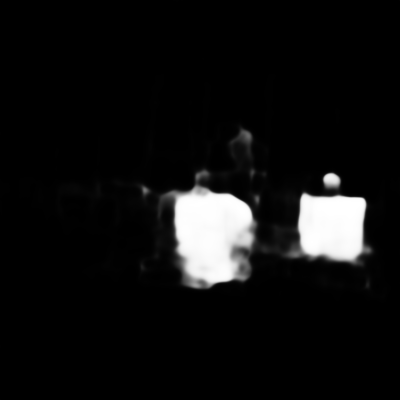}} &
   {\includegraphics[width=0.11\linewidth]{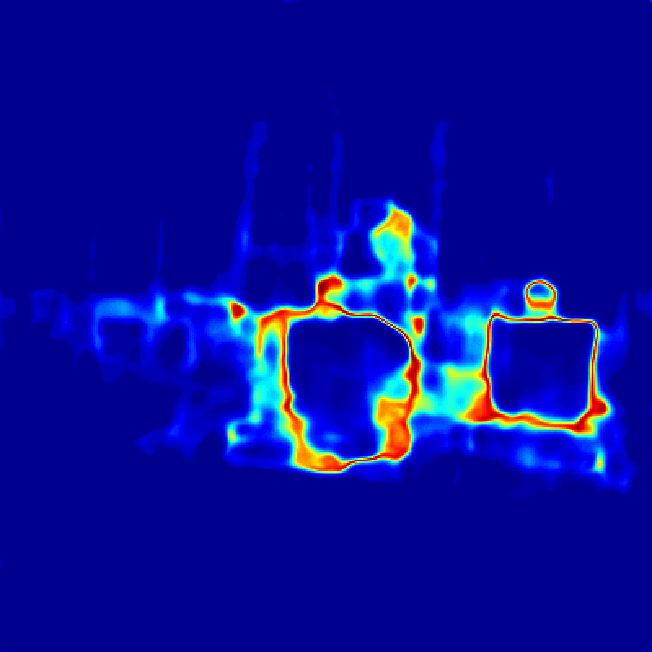}} &
   {\includegraphics[width=0.11\linewidth]{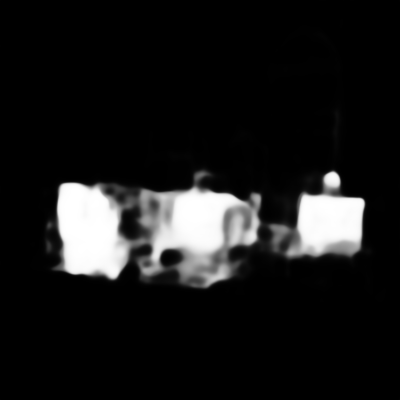}}&
   {\includegraphics[width=0.11\linewidth]{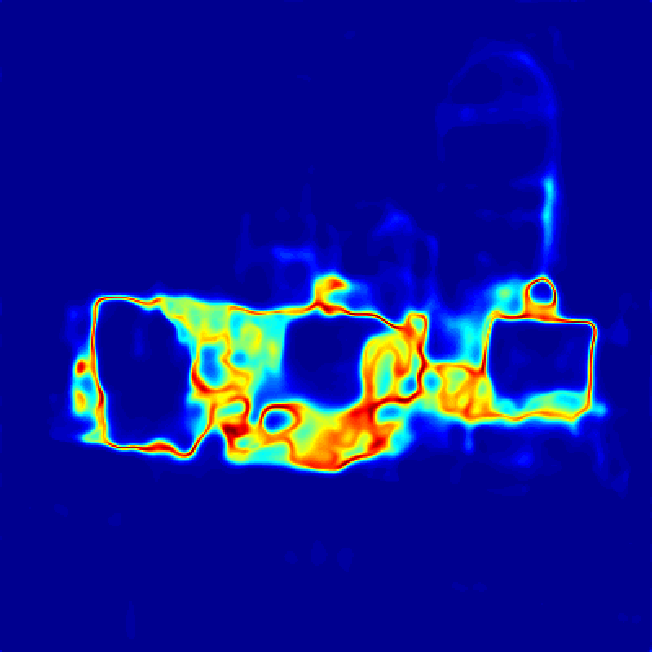}}&
   {\includegraphics[width=0.11\linewidth]{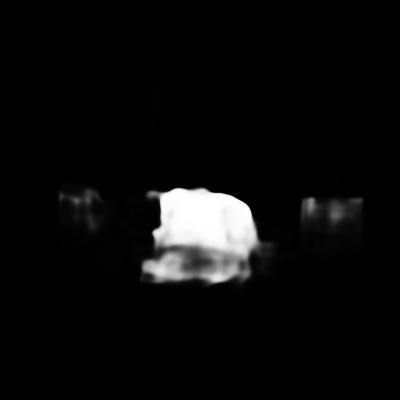}}&
   {\includegraphics[width=0.11\linewidth]{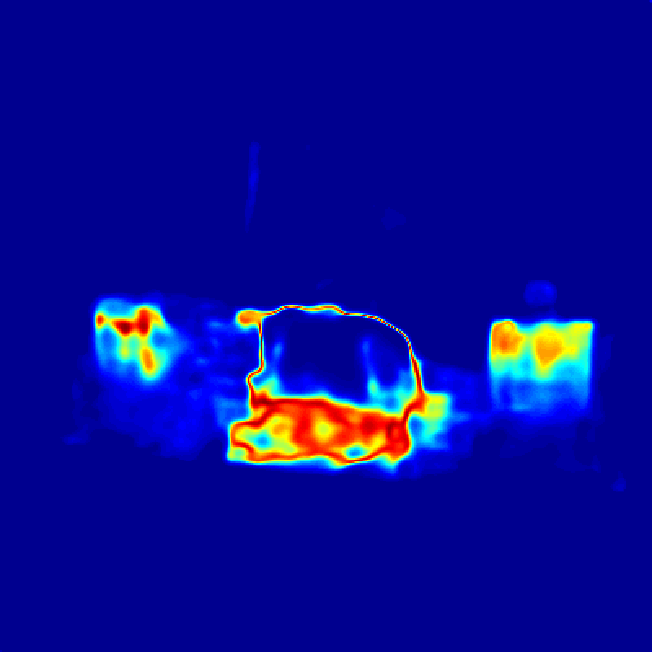}} \\
      {\includegraphics[width=0.11\linewidth]{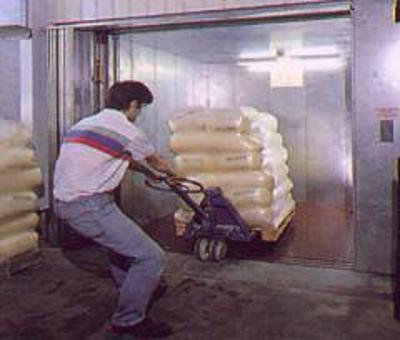}} &
   {\includegraphics[width=0.11\linewidth]{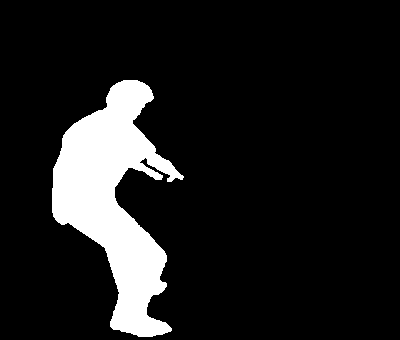}} &
   {\includegraphics[width=0.11\linewidth]{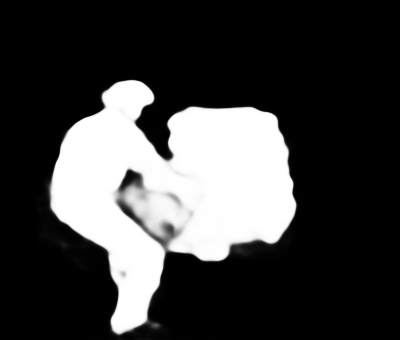}} &
   {\includegraphics[width=0.11\linewidth]{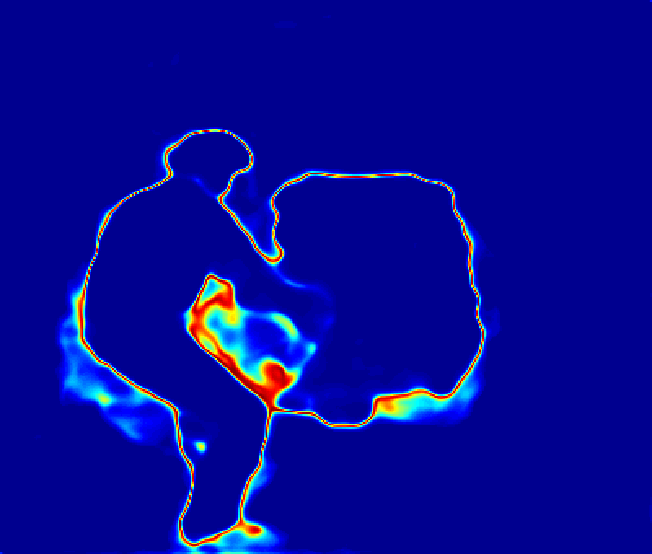}} &
   {\includegraphics[width=0.11\linewidth]{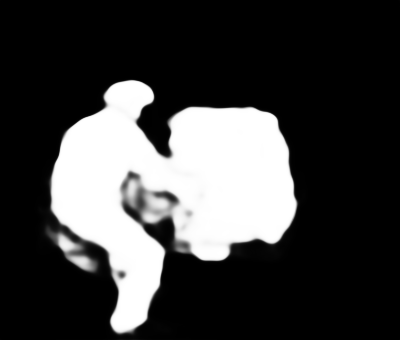}}&
   {\includegraphics[width=0.11\linewidth]{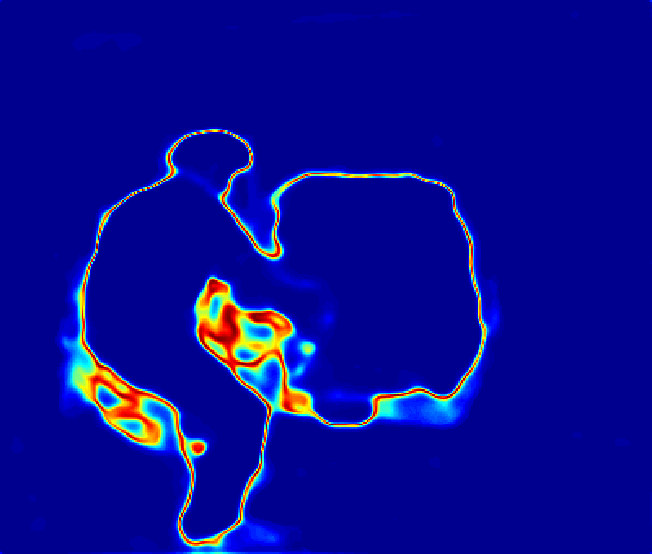}}&
   {\includegraphics[width=0.11\linewidth]{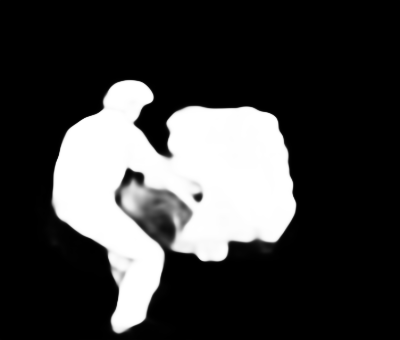}}&
   {\includegraphics[width=0.11\linewidth]{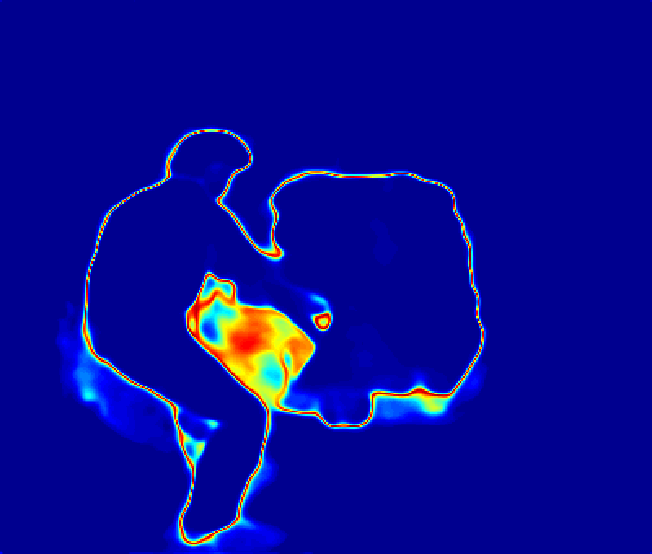}} \\
      {\includegraphics[width=0.11\linewidth]{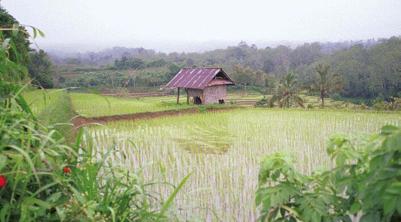}} &
   {\includegraphics[width=0.11\linewidth]{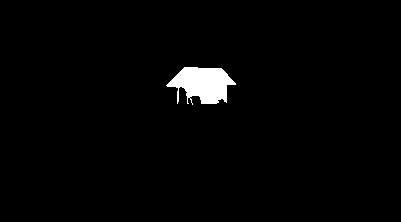}} &
   {\includegraphics[width=0.11\linewidth]{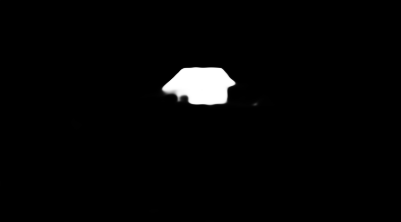}} &
   {\includegraphics[width=0.11\linewidth]{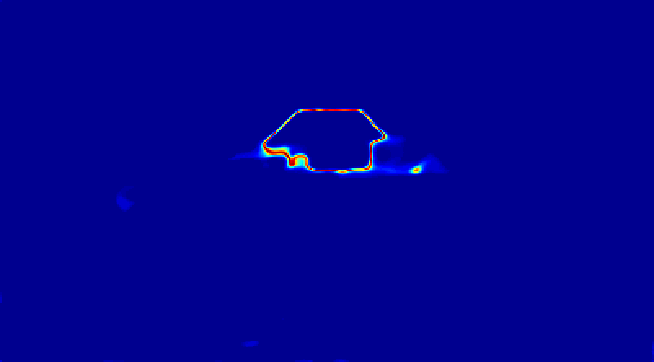}} &
   {\includegraphics[width=0.11\linewidth]{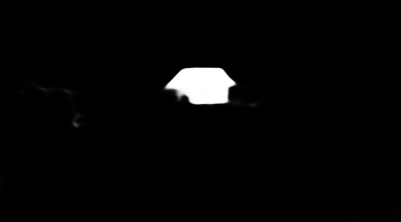}}&
   {\includegraphics[width=0.11\linewidth]{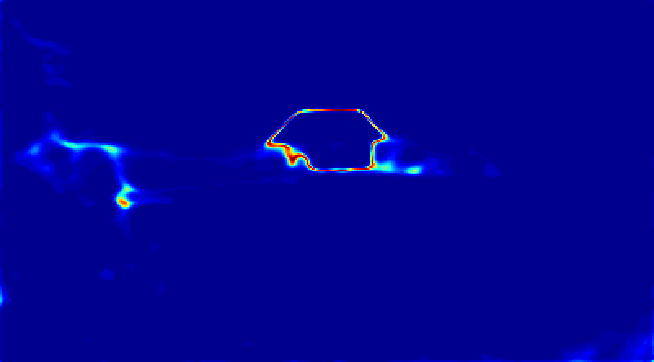}}&
   {\includegraphics[width=0.11\linewidth]{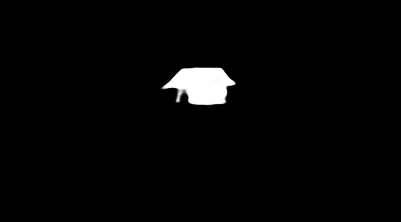}}&
   {\includegraphics[width=0.11\linewidth]{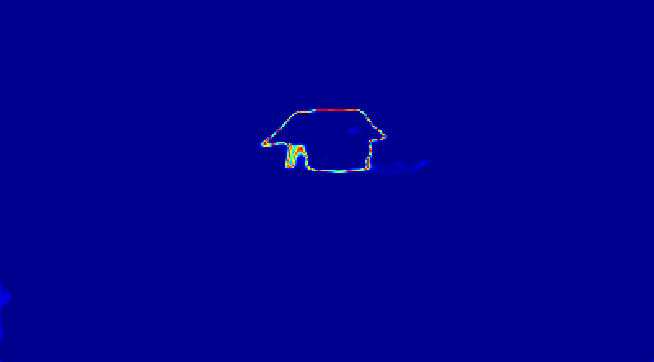}} \\
      {\includegraphics[width=0.11\linewidth]{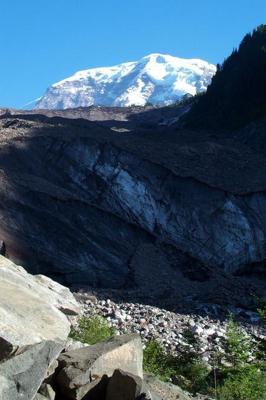}} &
   {\includegraphics[width=0.11\linewidth]{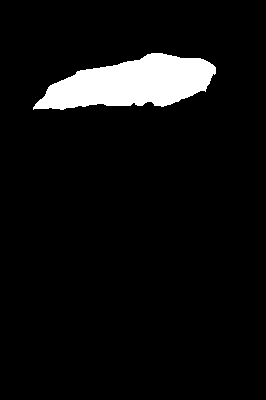}} &
   {\includegraphics[width=0.11\linewidth]{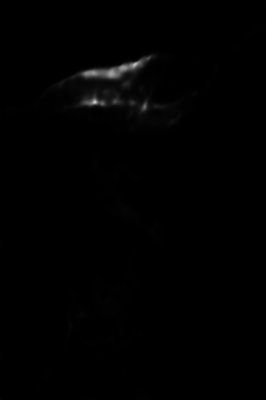}} &
   {\includegraphics[width=0.11\linewidth]{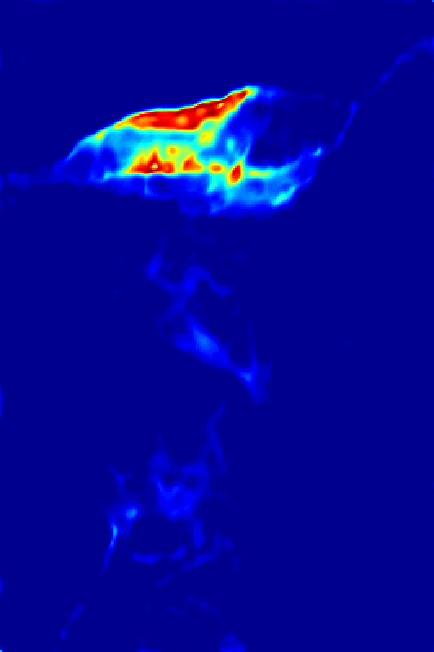}} &
   {\includegraphics[width=0.11\linewidth]{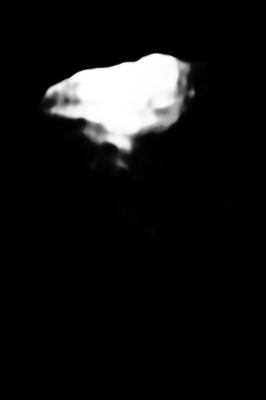}}&
   {\includegraphics[width=0.11\linewidth]{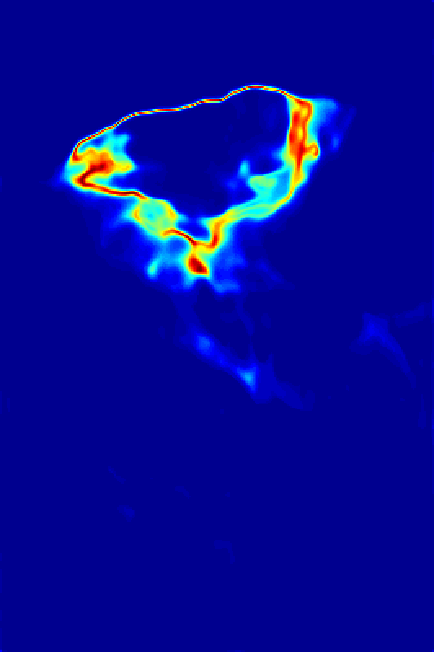}}&
   {\includegraphics[width=0.11\linewidth]{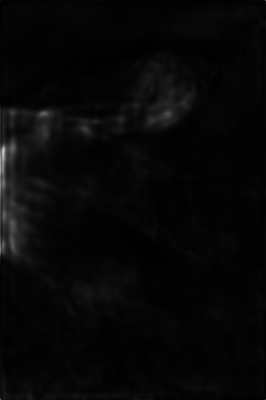}}&
   {\includegraphics[width=0.11\linewidth]{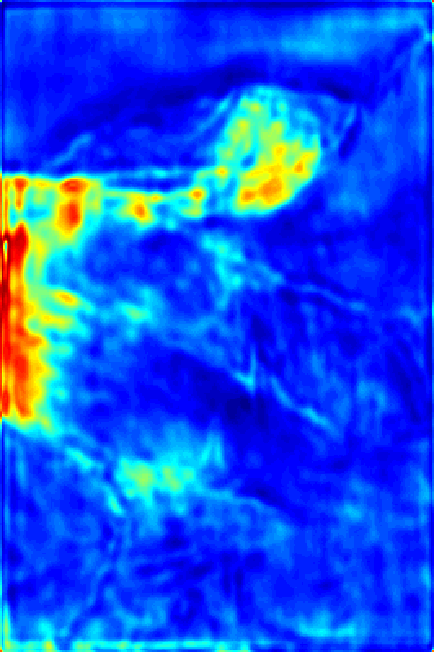}} \\
   \footnotesize{Image}&\footnotesize{GT}&\footnotesize{MD}&\footnotesize{$U_a$}&\footnotesize{DE}&\footnotesize{$U_a$}&\footnotesize{SE}&\footnotesize{$U_a$}\\
   \end{tabular}
   \end{center}
   \caption{\footnotesize{Aleatoric uncertainty of ensemble based solutions for \textbf{salient object detection}.}
   }
\label{fig:aleatoric_ensemble_sod}
\end{figure}

\begin{figure}[tp]
   \begin{center}
   \begin{tabular}{c@{ }c@{ }c@{ }c@{ }c@{ }c@{ }c@{ }c@{ }}
   {\includegraphics[width=0.11\linewidth]{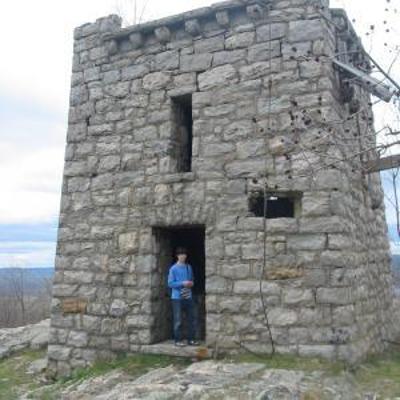}} &
   {\includegraphics[width=0.11\linewidth]{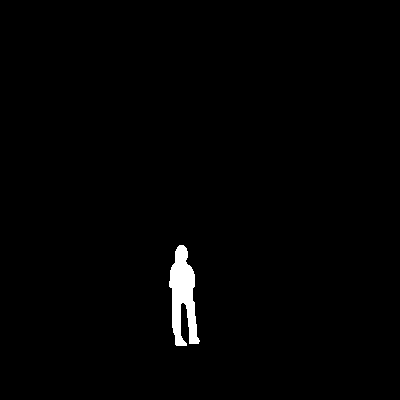}} &
   {\includegraphics[width=0.11\linewidth]{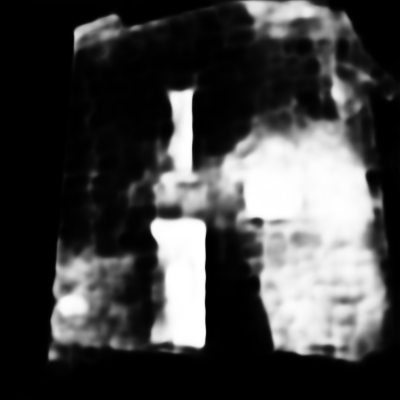}} &
   {\includegraphics[width=0.11\linewidth]{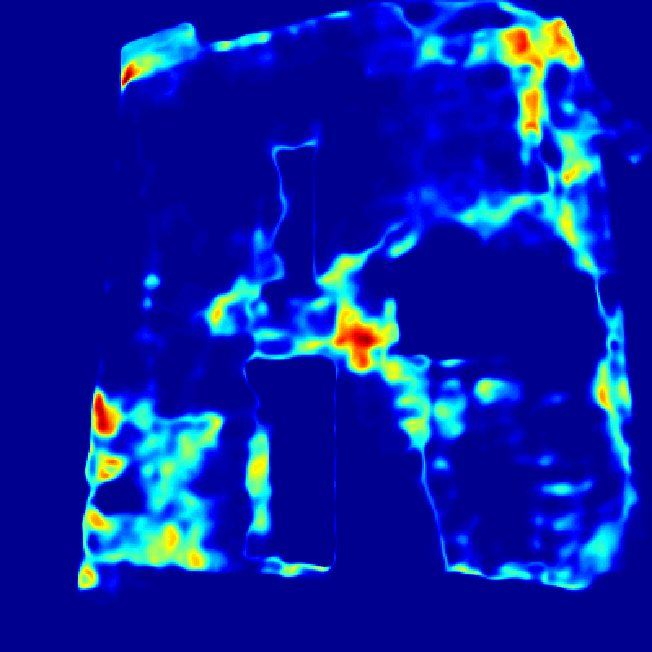}} &
   {\includegraphics[width=0.11\linewidth]{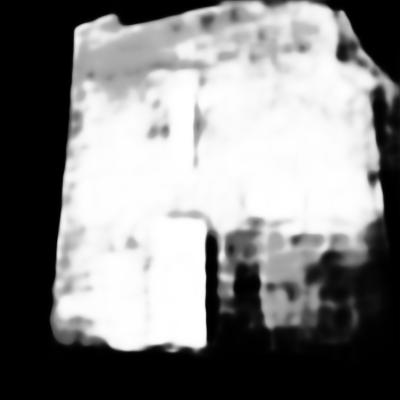}}&
   {\includegraphics[width=0.11\linewidth]{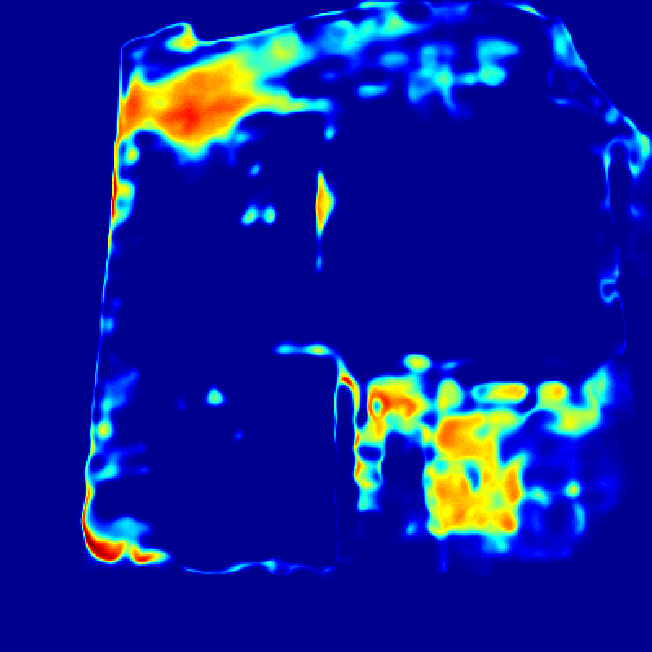}}&
   {\includegraphics[width=0.11\linewidth]{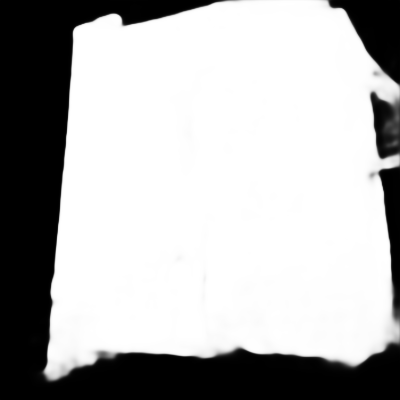}}&
   {\includegraphics[width=0.11\linewidth]{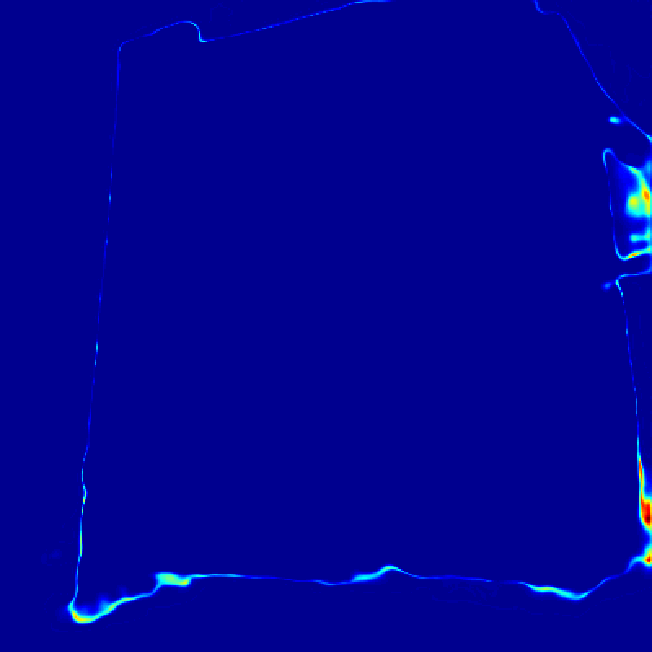}} \\
      {\includegraphics[width=0.11\linewidth]{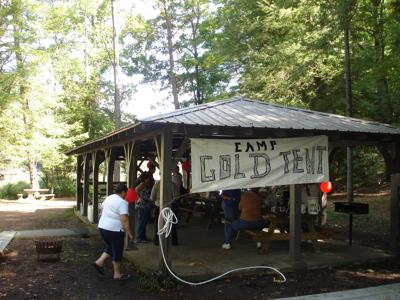}} &
   {\includegraphics[width=0.11\linewidth]{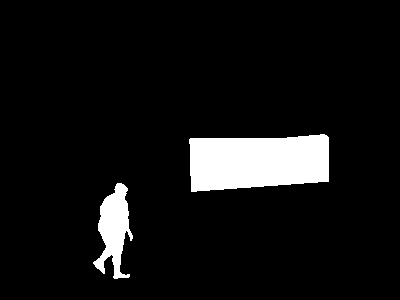}} &
   {\includegraphics[width=0.11\linewidth]{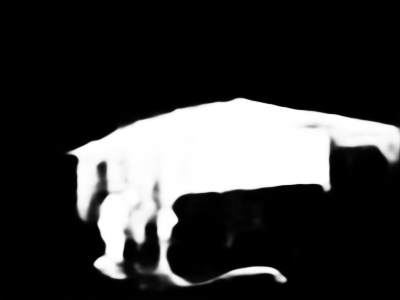}} &
   {\includegraphics[width=0.11\linewidth]{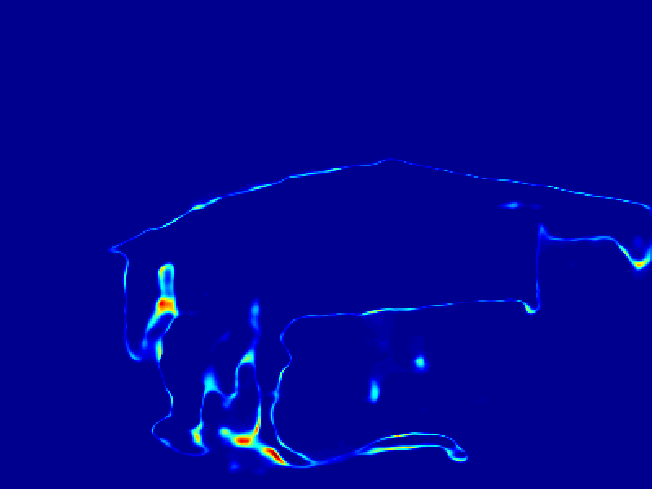}} &
   {\includegraphics[width=0.11\linewidth]{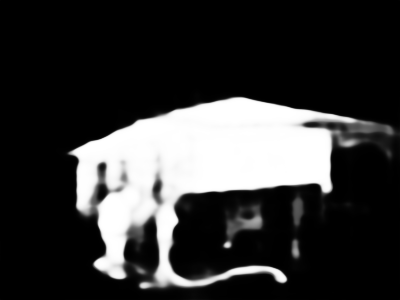}}&
   {\includegraphics[width=0.11\linewidth]{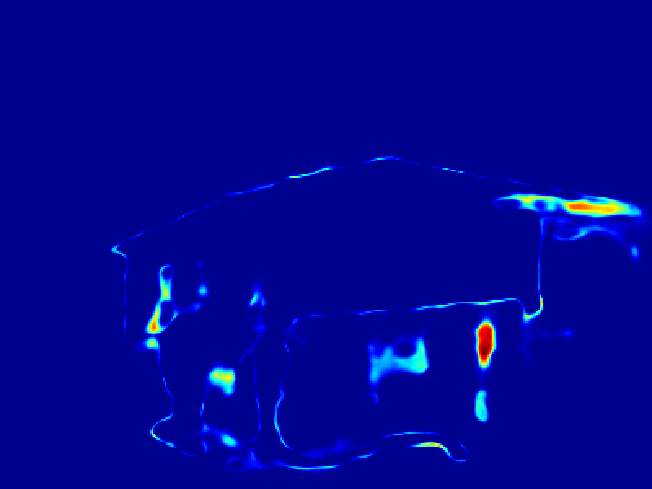}}&
   {\includegraphics[width=0.11\linewidth]{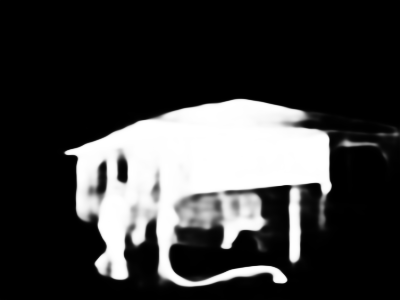}}&
   {\includegraphics[width=0.11\linewidth]{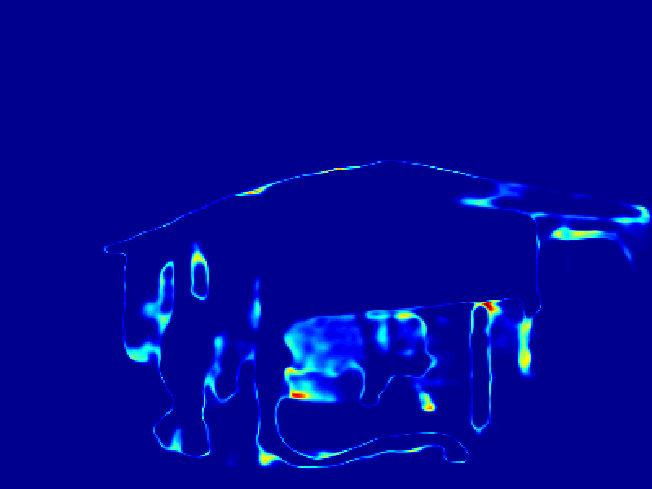}} \\
      {\includegraphics[width=0.11\linewidth]{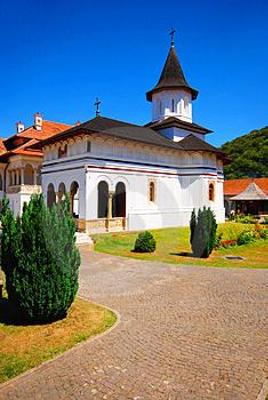}} &
   {\includegraphics[width=0.11\linewidth]{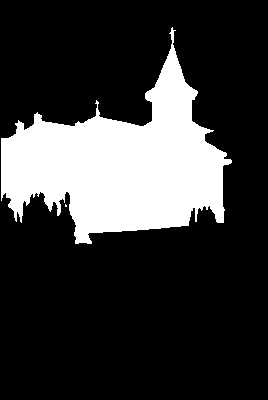}} &
   {\includegraphics[width=0.11\linewidth]{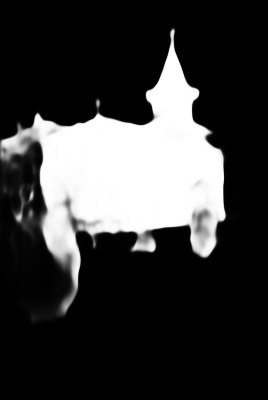}} &
   {\includegraphics[width=0.11\linewidth]{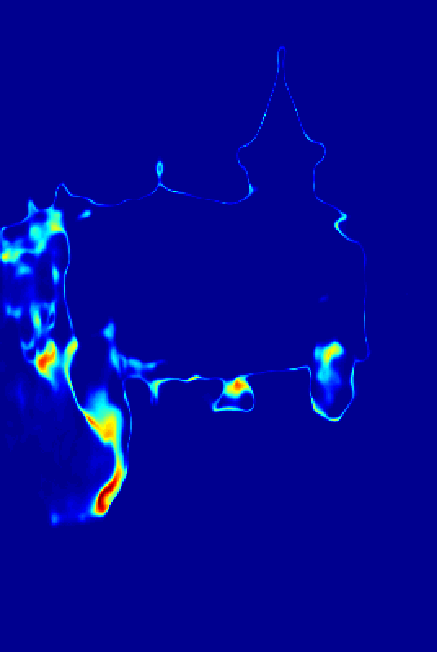}} &
   {\includegraphics[width=0.11\linewidth]{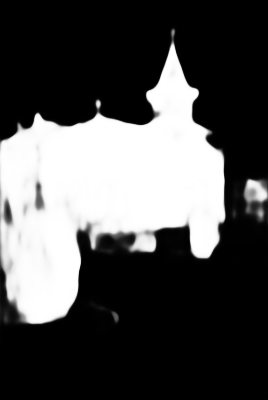}}&
   {\includegraphics[width=0.11\linewidth]{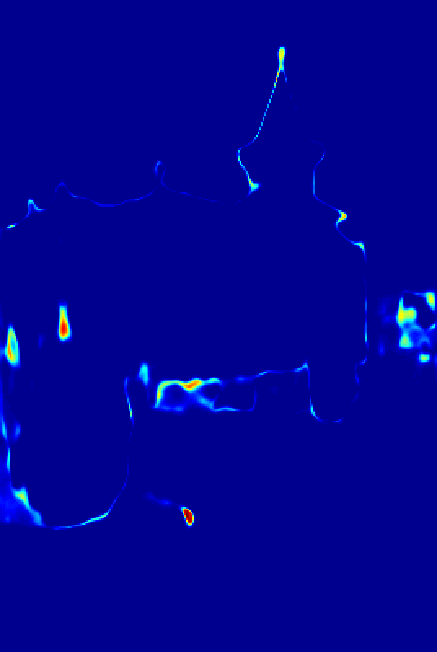}}&
   {\includegraphics[width=0.11\linewidth]{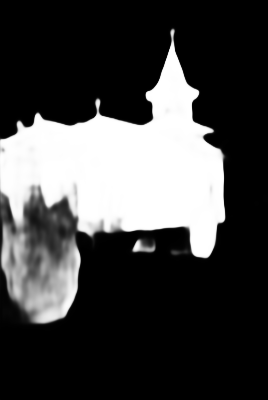}}&
   {\includegraphics[width=0.11\linewidth]{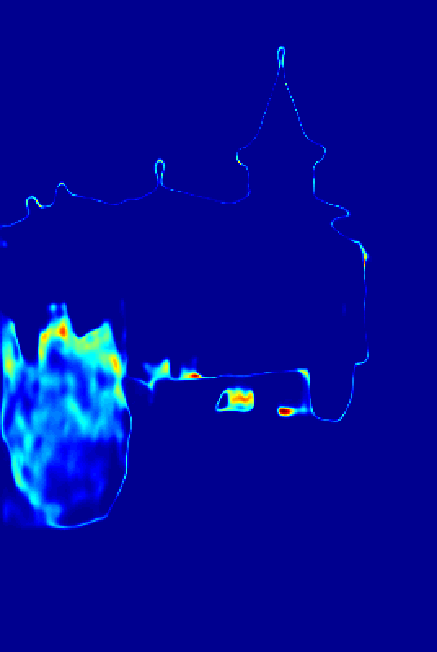}} \\
      {\includegraphics[width=0.11\linewidth]{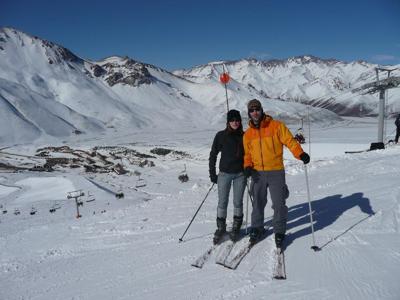}} &
   {\includegraphics[width=0.11\linewidth]{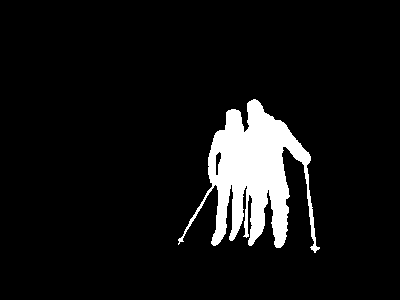}} &
   {\includegraphics[width=0.11\linewidth]{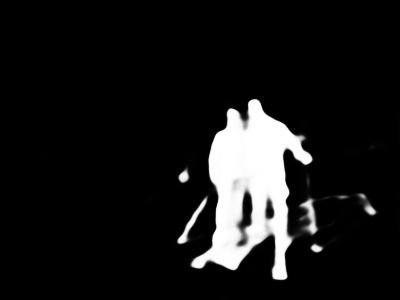}} &
   {\includegraphics[width=0.11\linewidth]{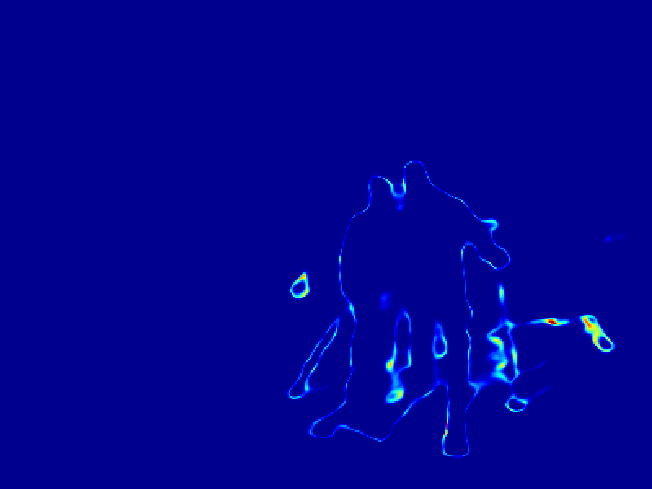}} &
   {\includegraphics[width=0.11\linewidth]{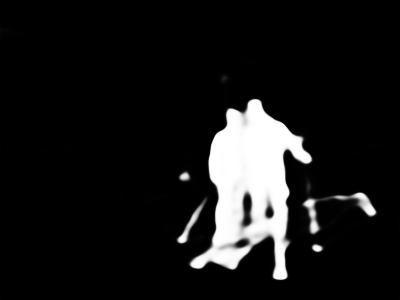}}&
   {\includegraphics[width=0.11\linewidth]{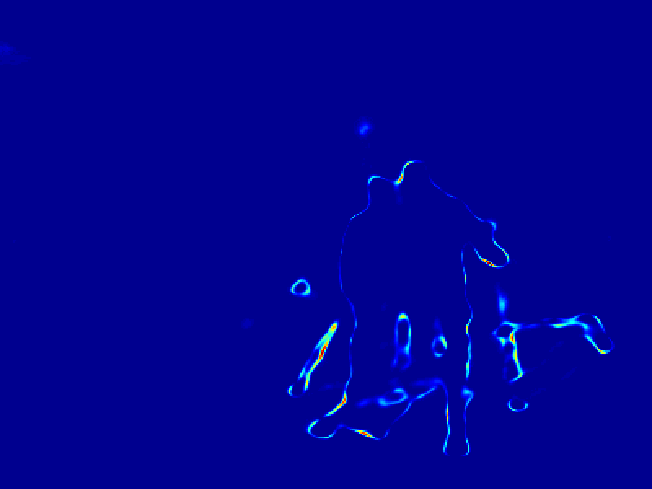}}&
   {\includegraphics[width=0.11\linewidth]{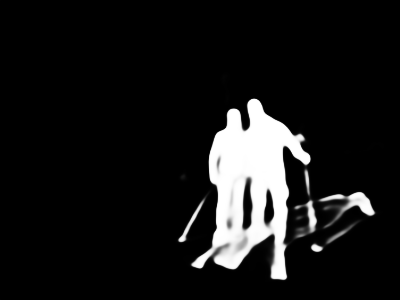}}&
   {\includegraphics[width=0.11\linewidth]{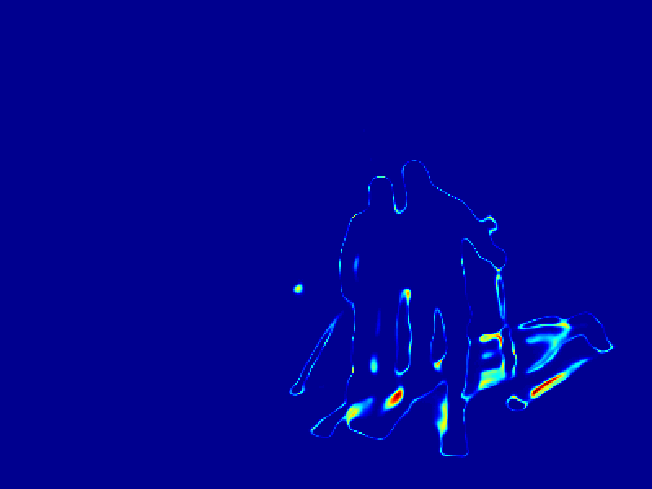}} \\
      {\includegraphics[width=0.11\linewidth]{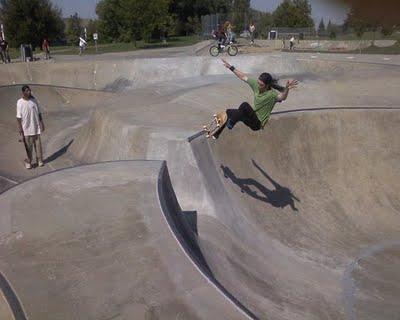}} &
   {\includegraphics[width=0.11\linewidth]{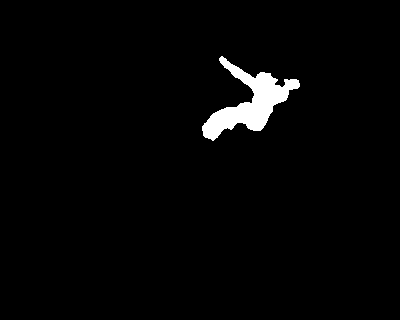}} &
   {\includegraphics[width=0.11\linewidth]{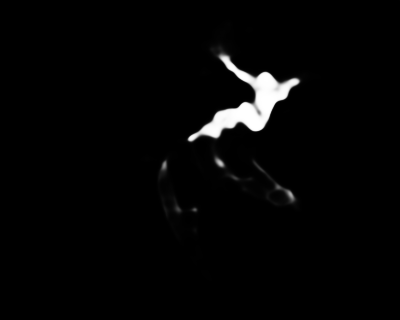}} &
   {\includegraphics[width=0.11\linewidth]{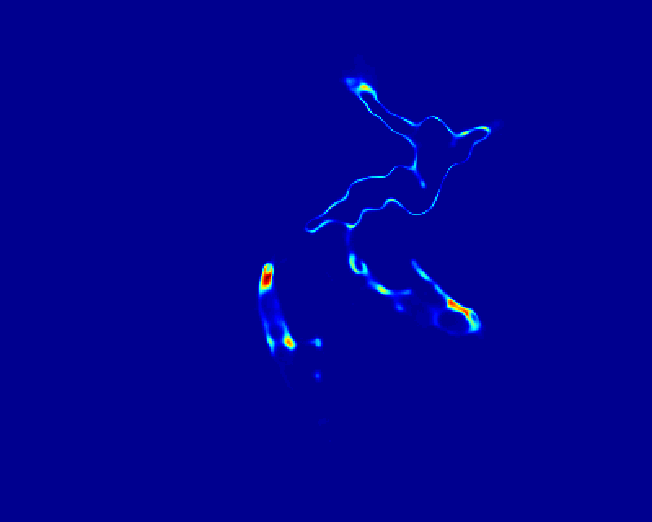}} &
   {\includegraphics[width=0.11\linewidth]{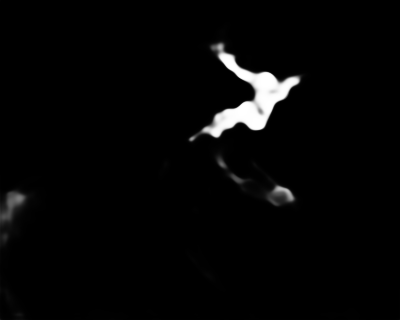}}&
   {\includegraphics[width=0.11\linewidth]{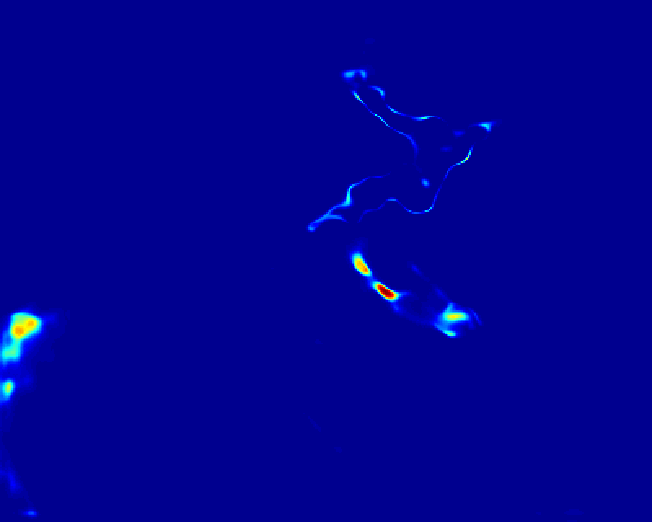}}&
   {\includegraphics[width=0.11\linewidth]{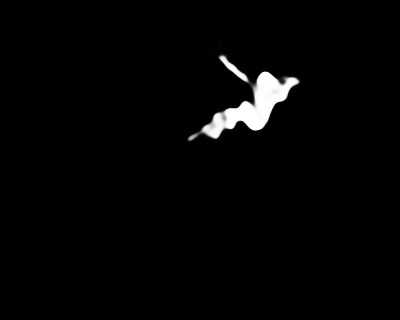}}&
   {\includegraphics[width=0.11\linewidth]{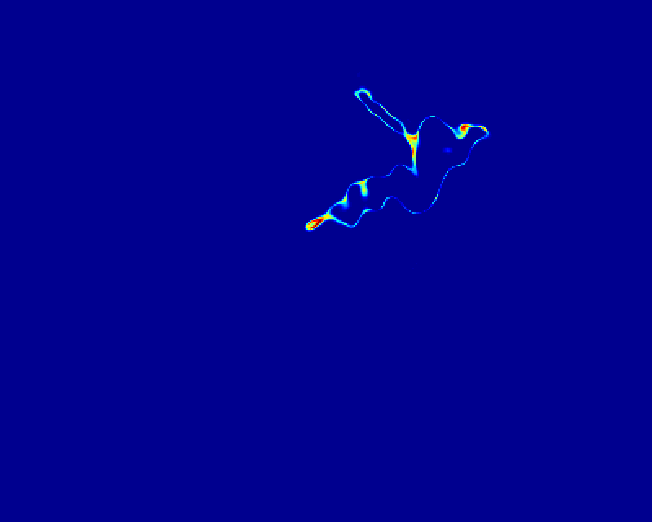}} \\
   \footnotesize{Image}&\footnotesize{GT}&\footnotesize{MD}&\footnotesize{$U_e$}&\footnotesize{DE}&\footnotesize{$U_e$}&\footnotesize{SE}&\footnotesize{$U_e$}\\
   \end{tabular}
   \end{center}
   \caption{\footnotesize{Epistemic uncertainty of ensemble based solutions for \textbf{salient object detection}.}
   }
\label{fig:epistemic_ensemble_sod}
\end{figure}

\begin{figure*}[tp]
   \begin{center}
   \begin{tabular}{c@{ }c@{ }c@{ }c@{ }c@{ }c@{ }c@{ }c@{ }c@{ }c@{ }}
   {\includegraphics[width=0.095\linewidth]{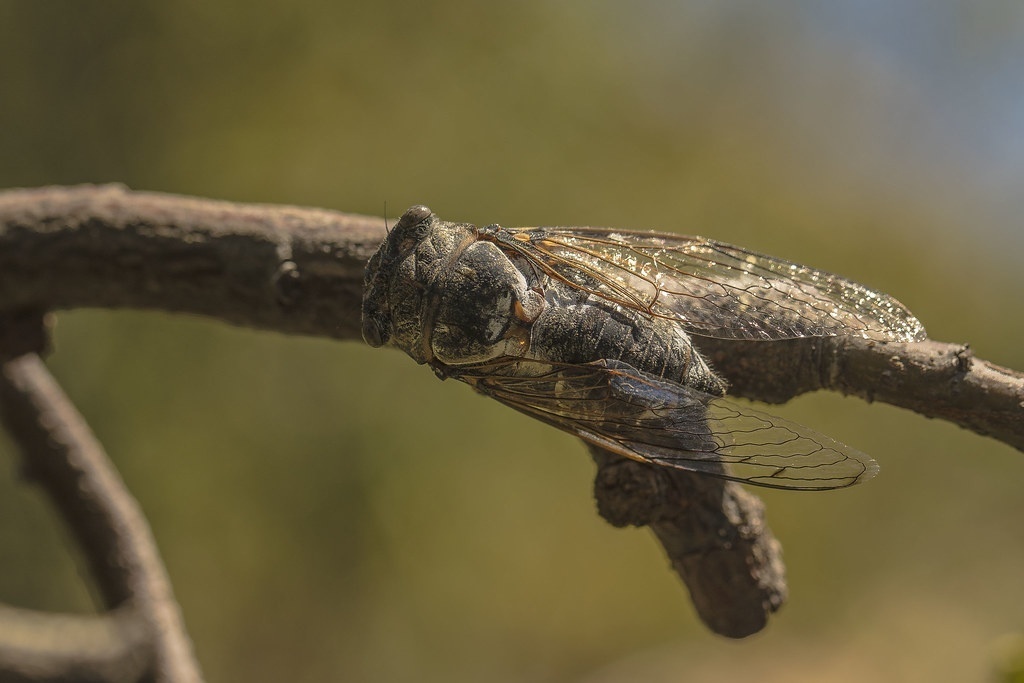}} &
   {\includegraphics[width=0.095\linewidth]{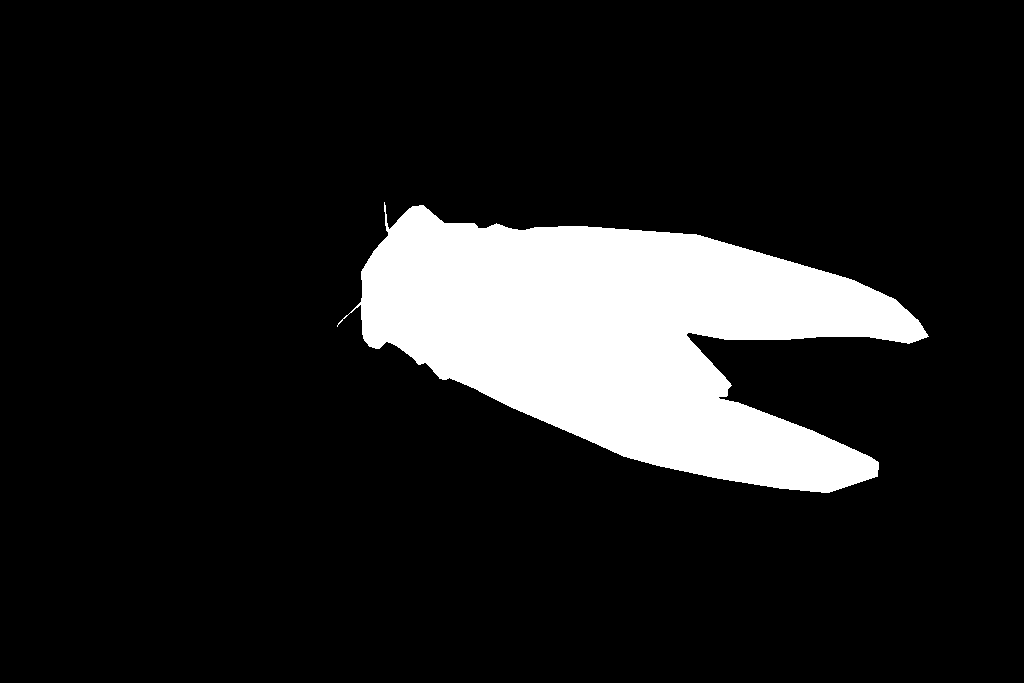}} &
   {\includegraphics[width=0.095\linewidth]{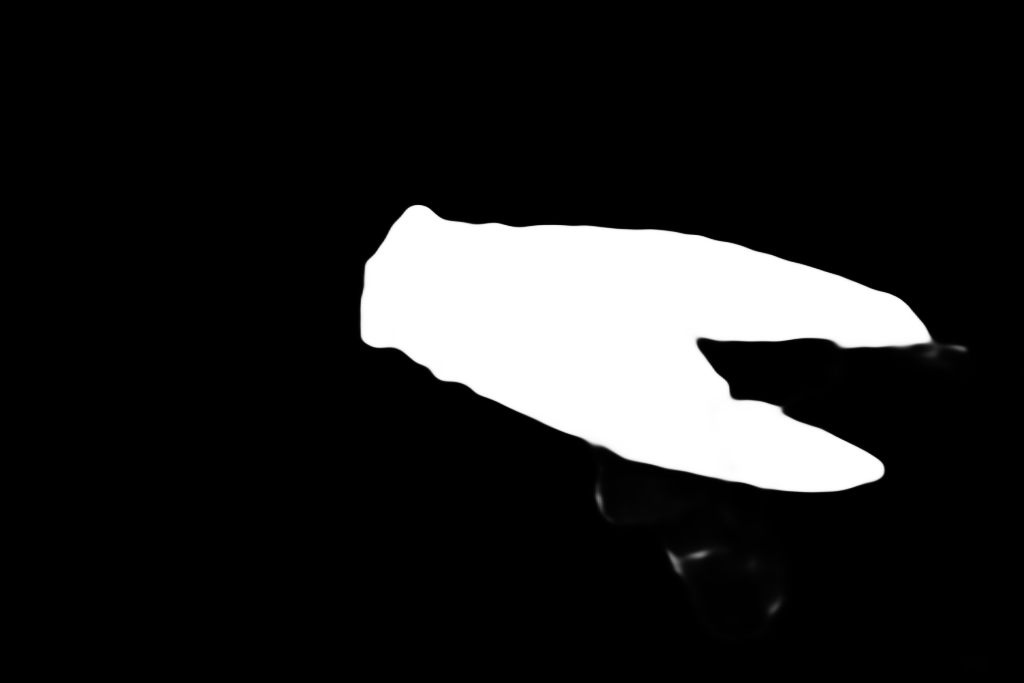}} &
   {\includegraphics[width=0.095\linewidth]{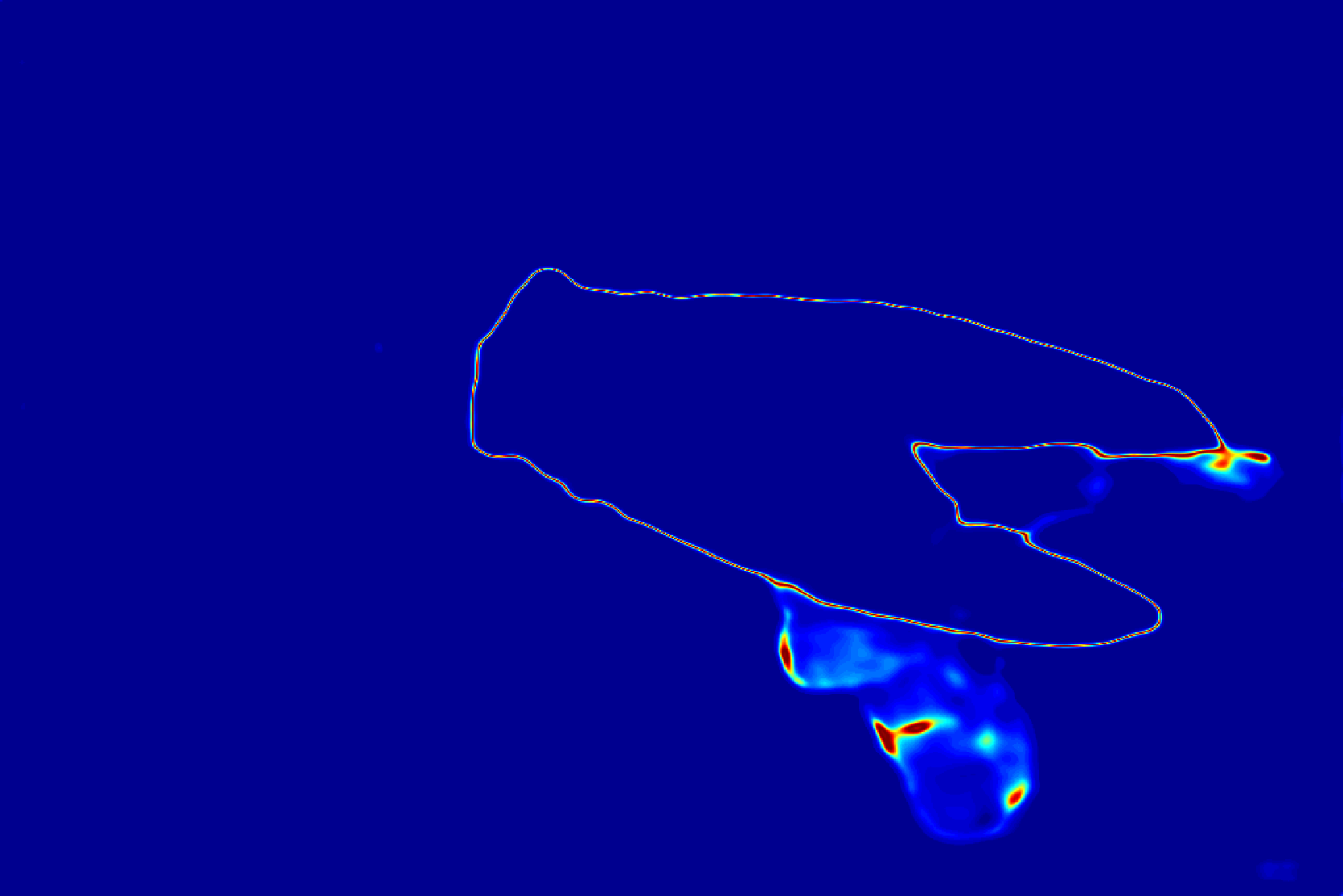}} &
   {\includegraphics[width=0.095\linewidth]{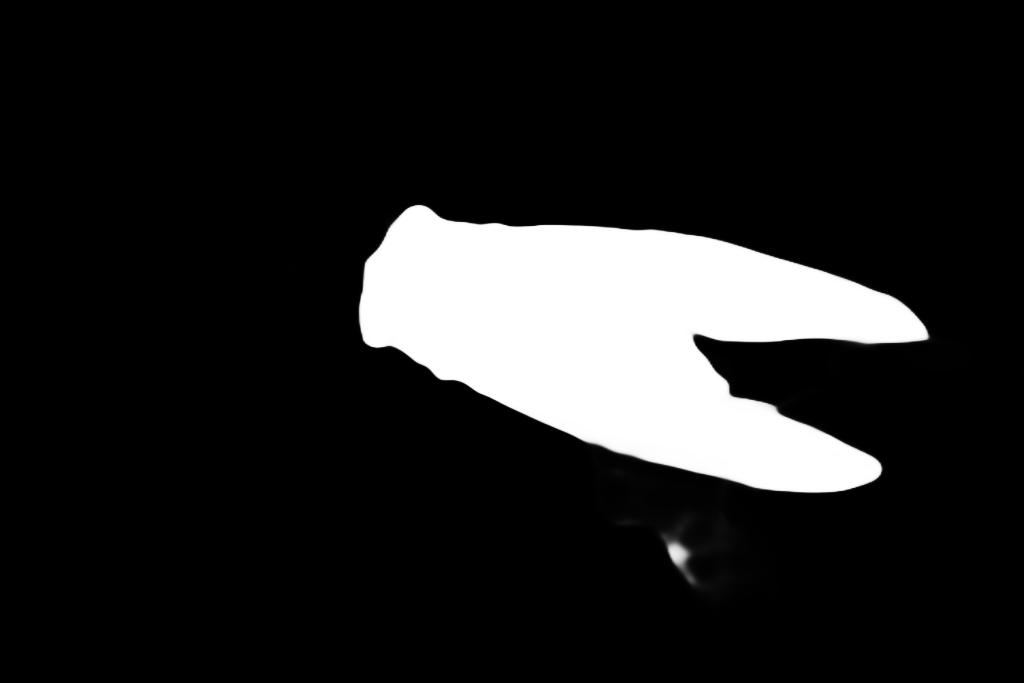}}&
   {\includegraphics[width=0.095\linewidth]{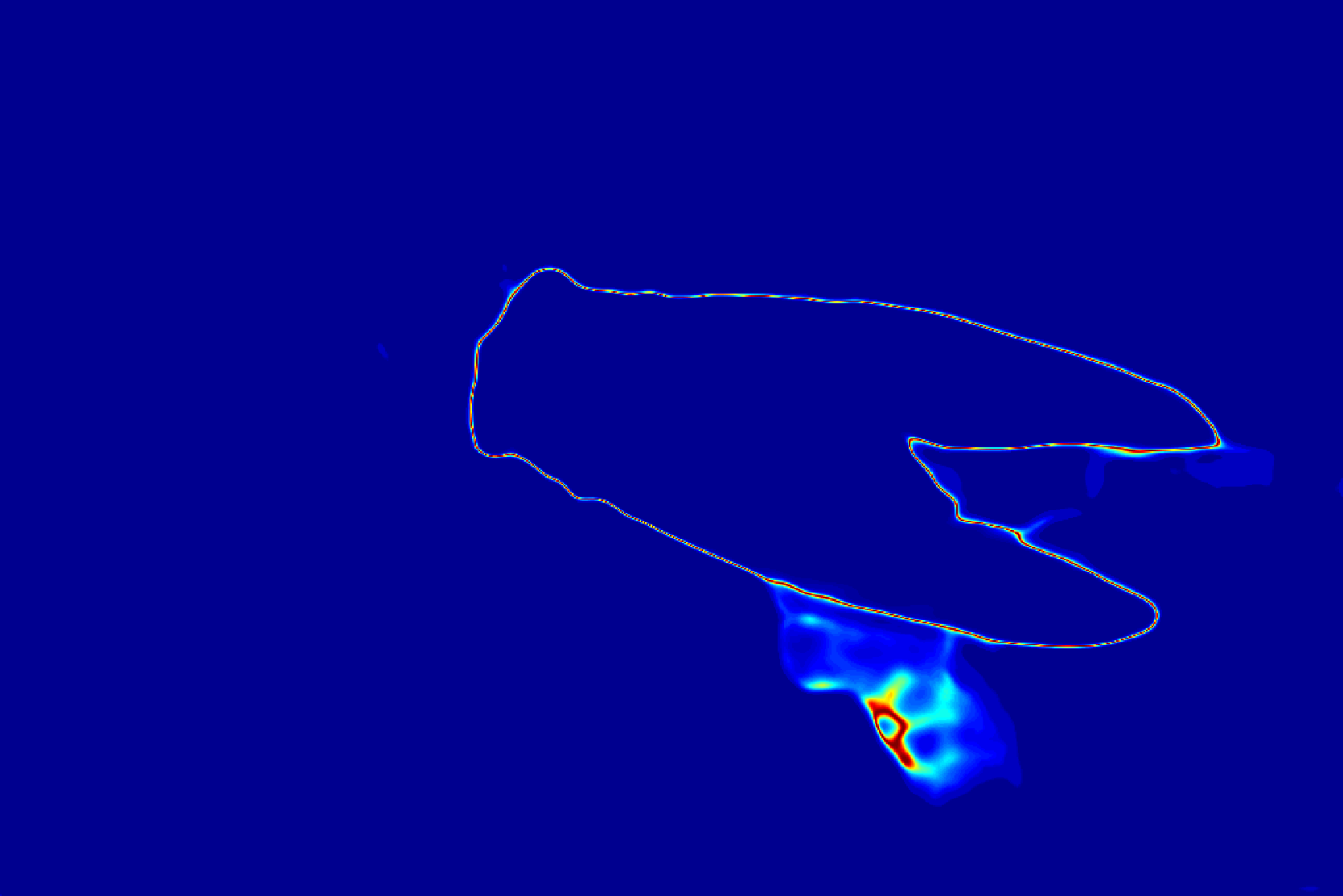}}&
   {\includegraphics[width=0.095\linewidth]{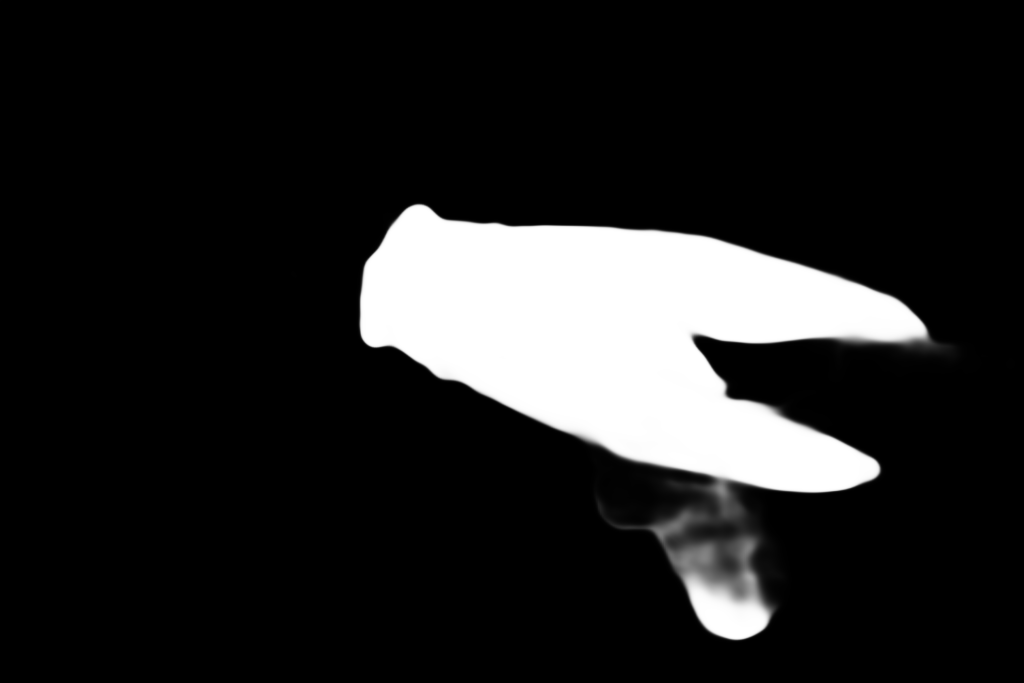}}&
   {\includegraphics[width=0.095\linewidth]{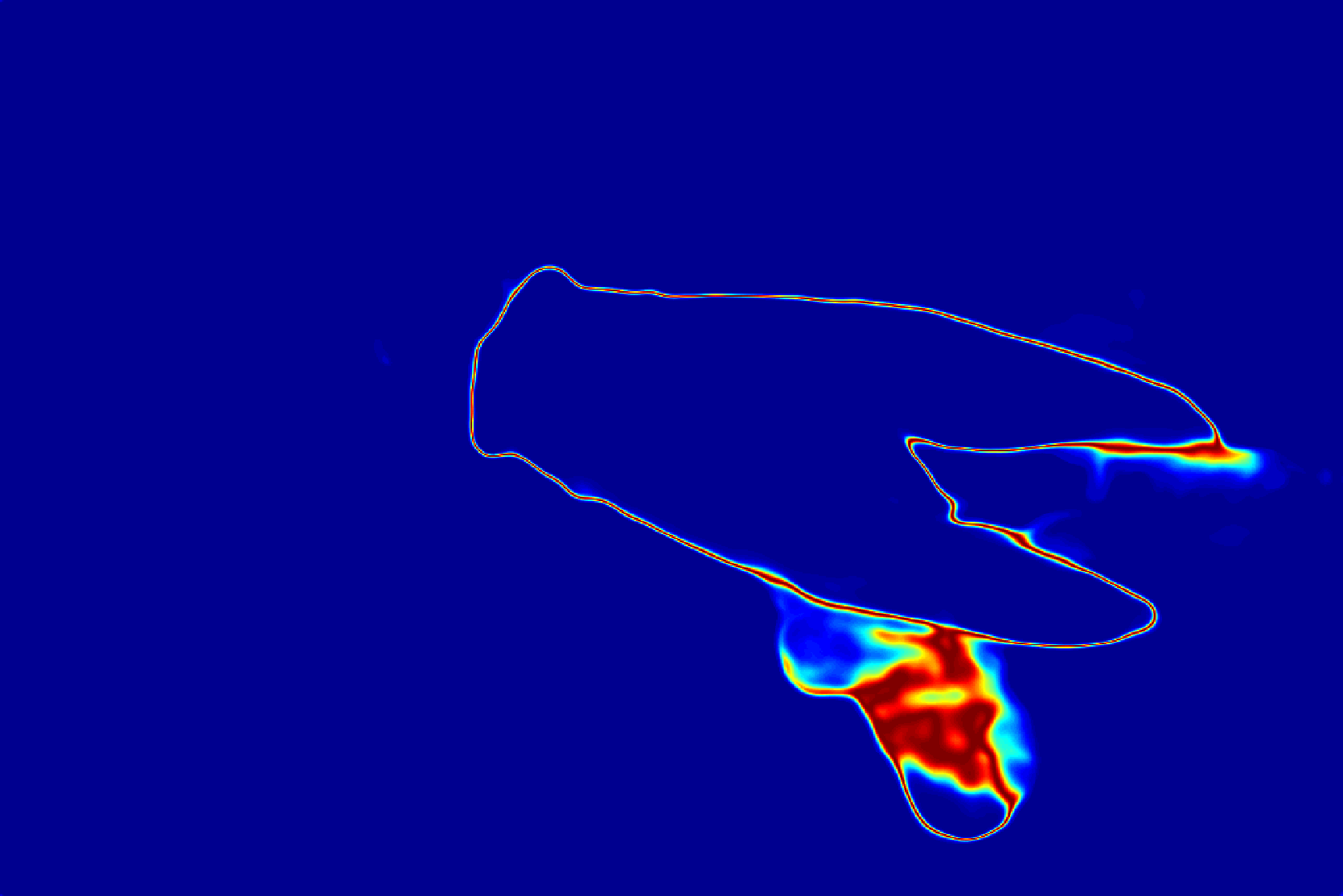}}&
   {\includegraphics[width=0.095\linewidth]{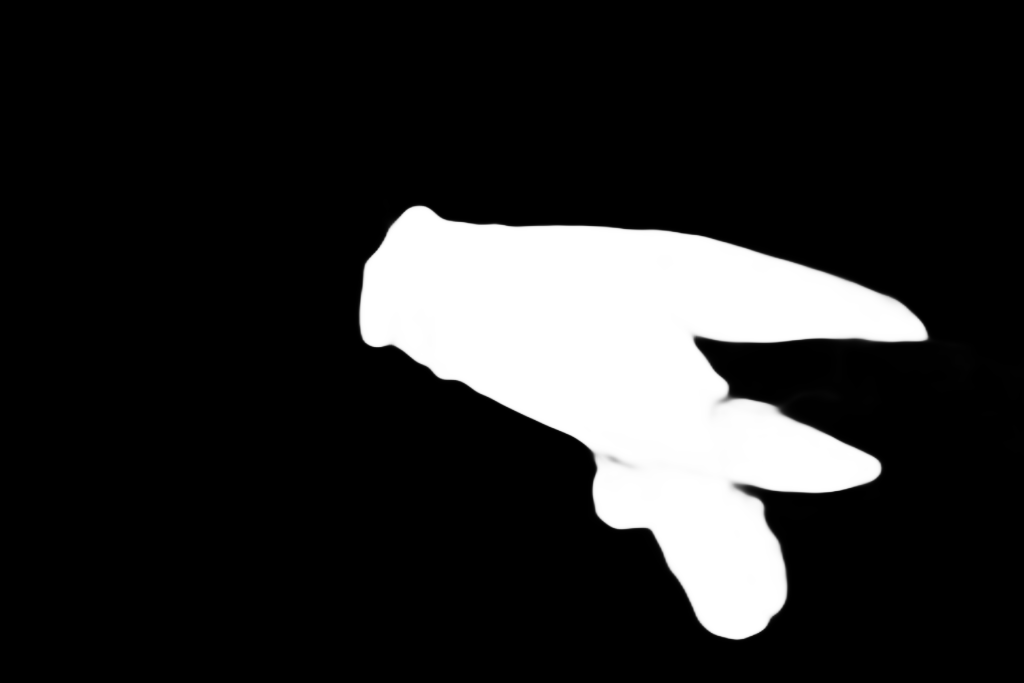}}&
   {\includegraphics[width=0.095\linewidth]{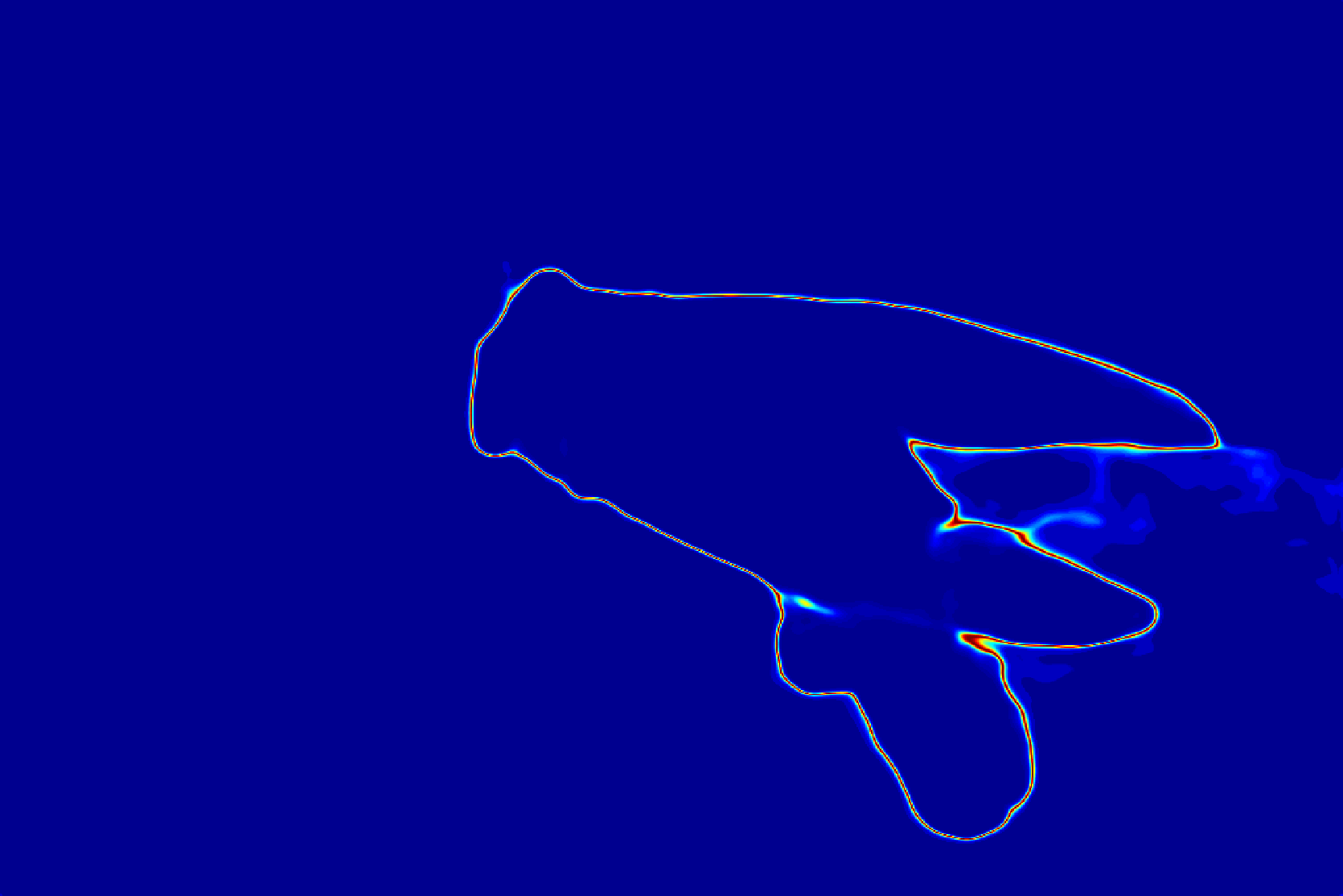}} \\
   {\includegraphics[width=0.095\linewidth]{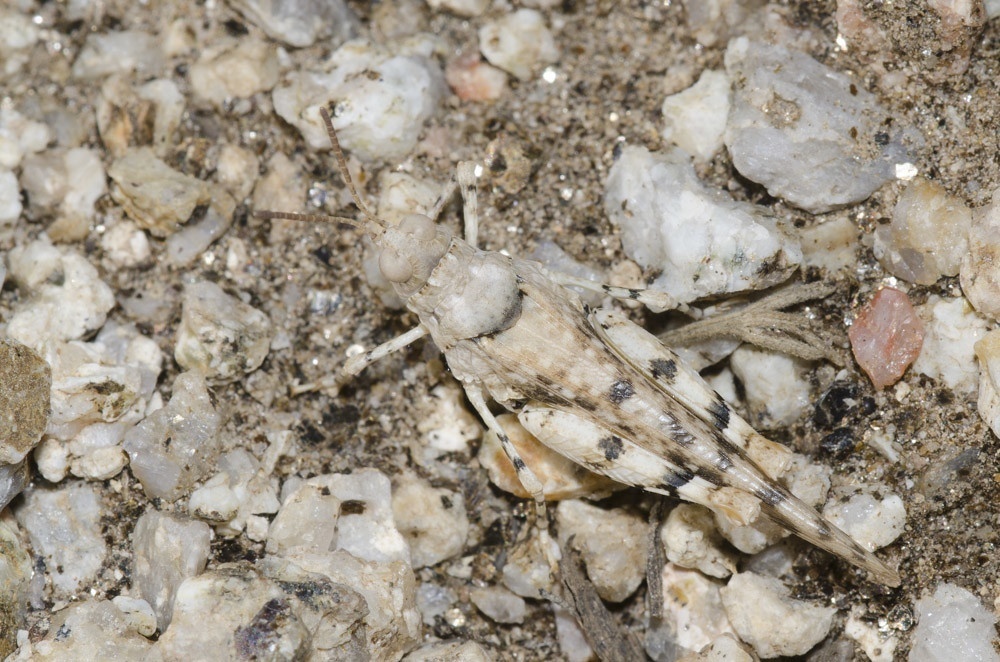}} &
   {\includegraphics[width=0.095\linewidth]{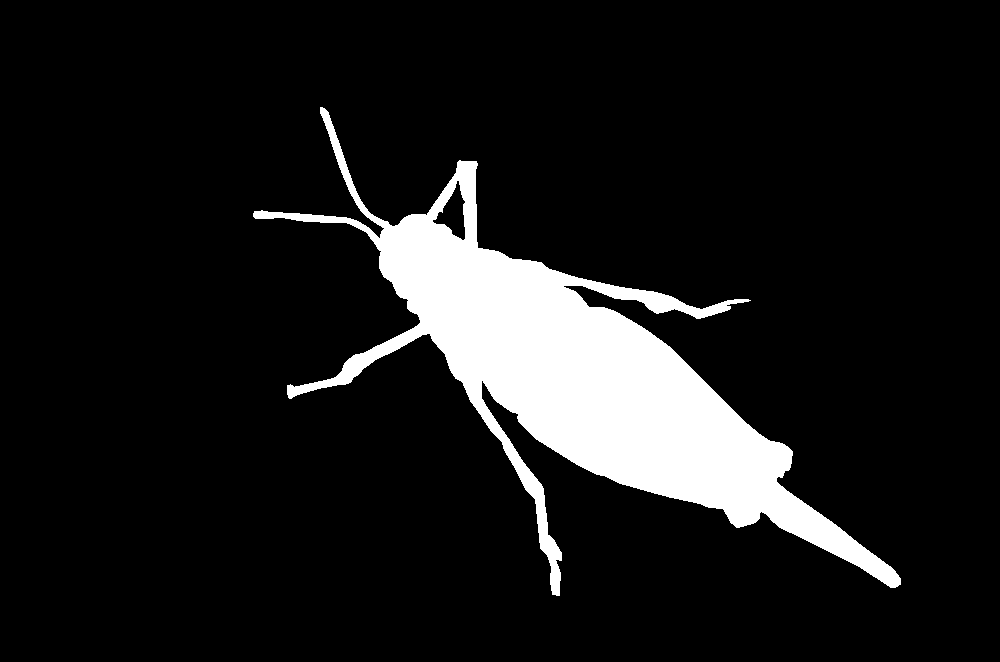}} &
   {\includegraphics[width=0.095\linewidth]{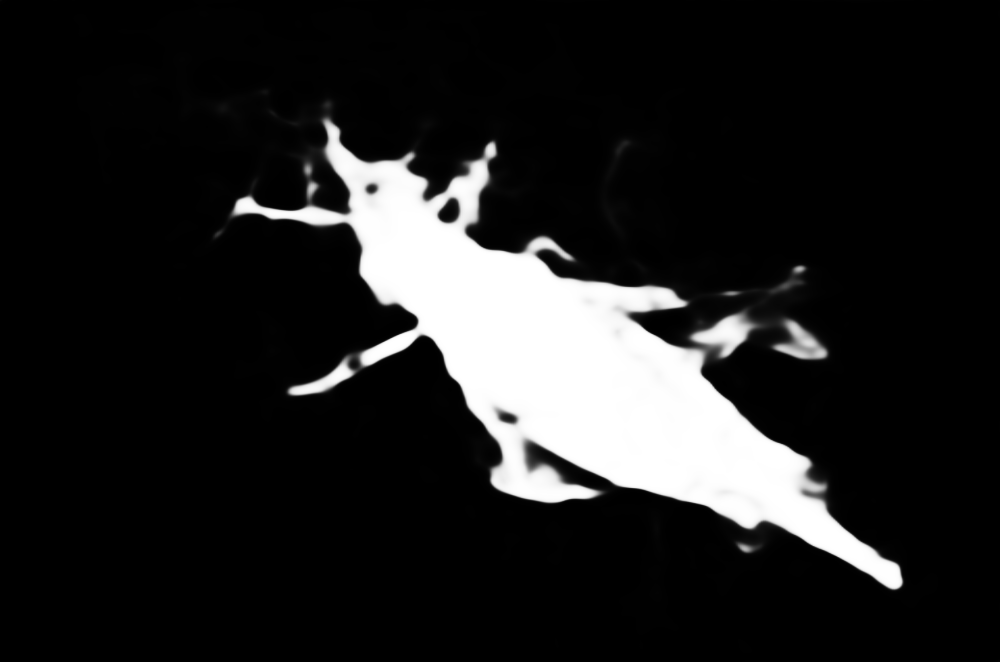}} &
   {\includegraphics[width=0.095\linewidth]{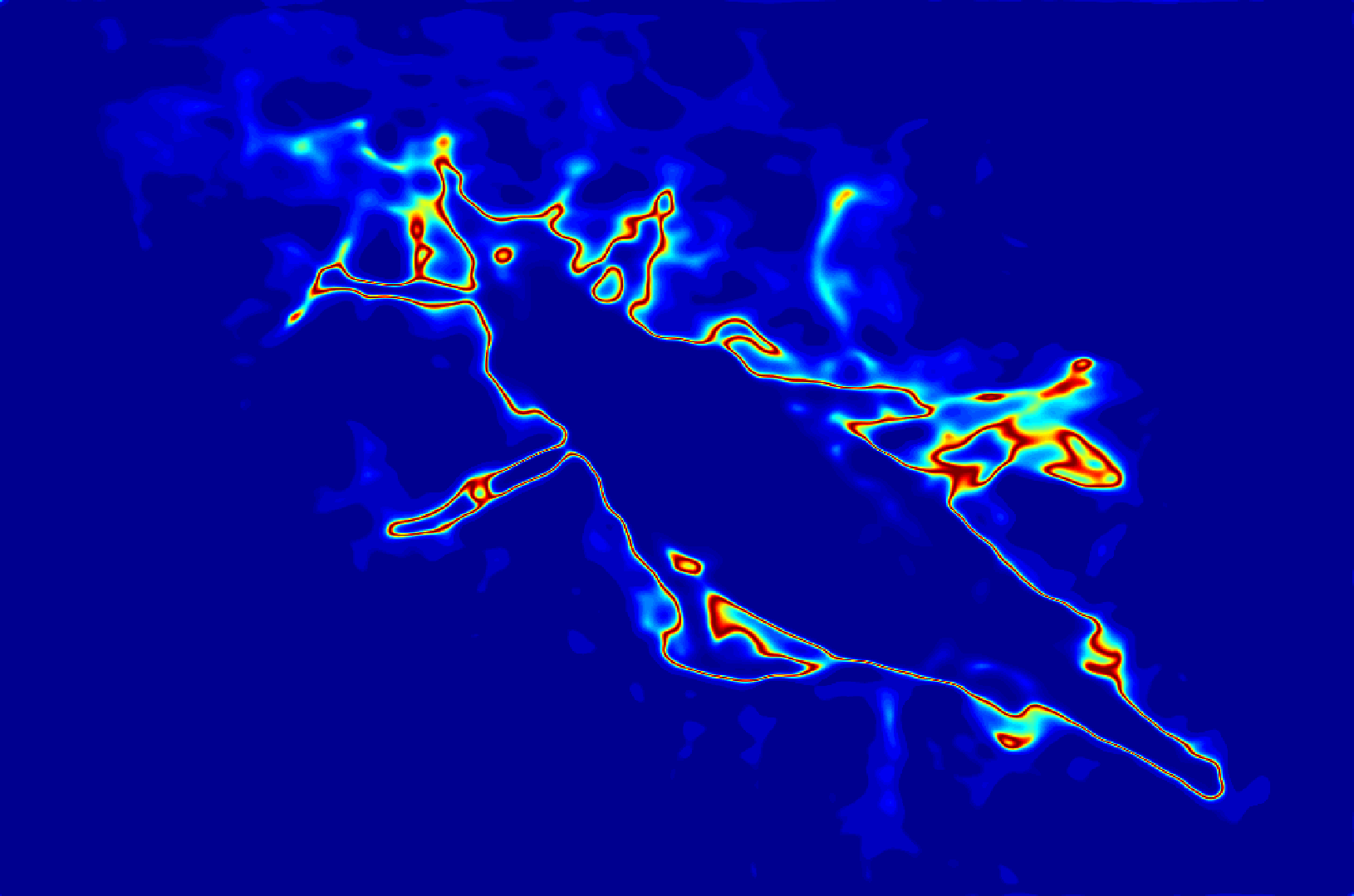}} &
   {\includegraphics[width=0.095\linewidth]{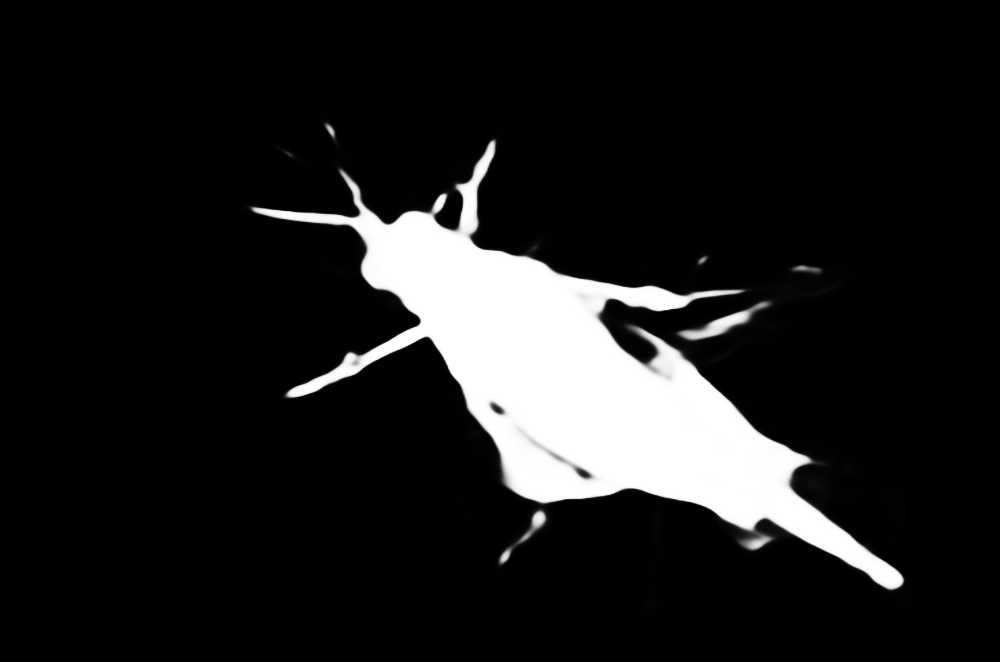}}&
   {\includegraphics[width=0.095\linewidth]{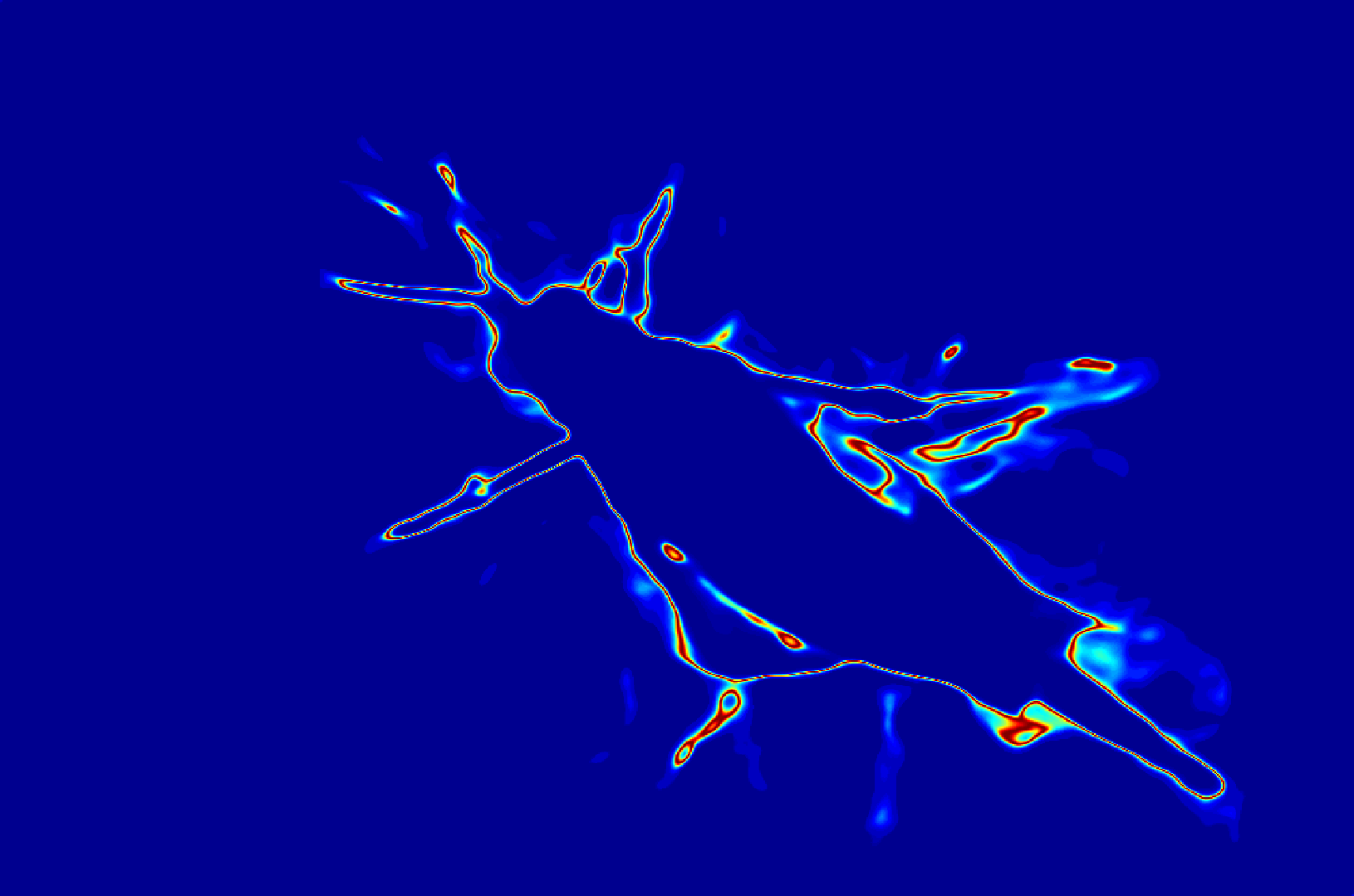}}&
   {\includegraphics[width=0.095\linewidth]{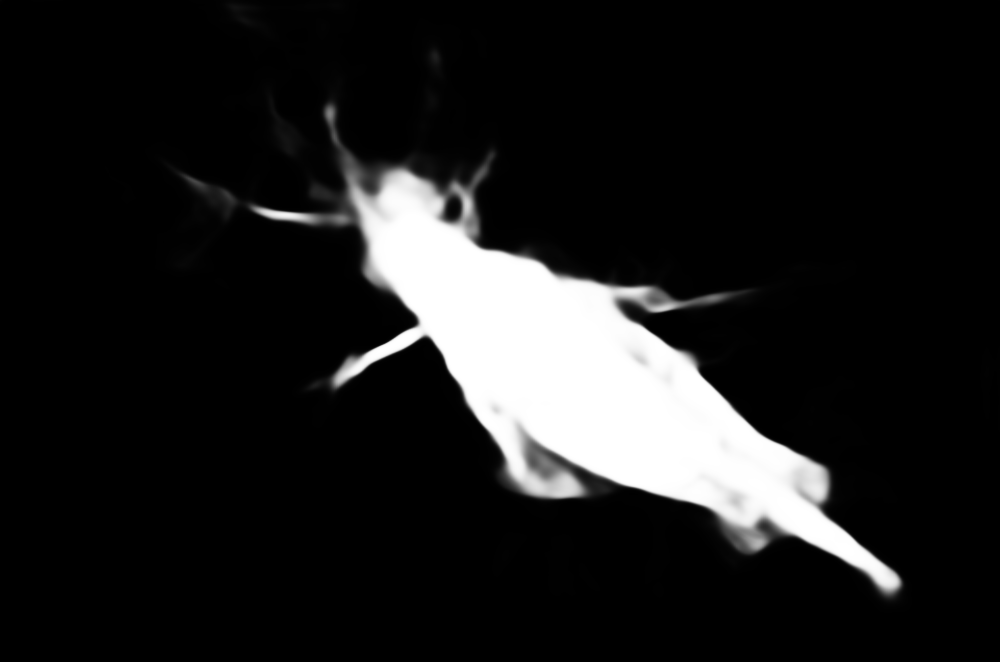}}&
   {\includegraphics[width=0.095\linewidth]{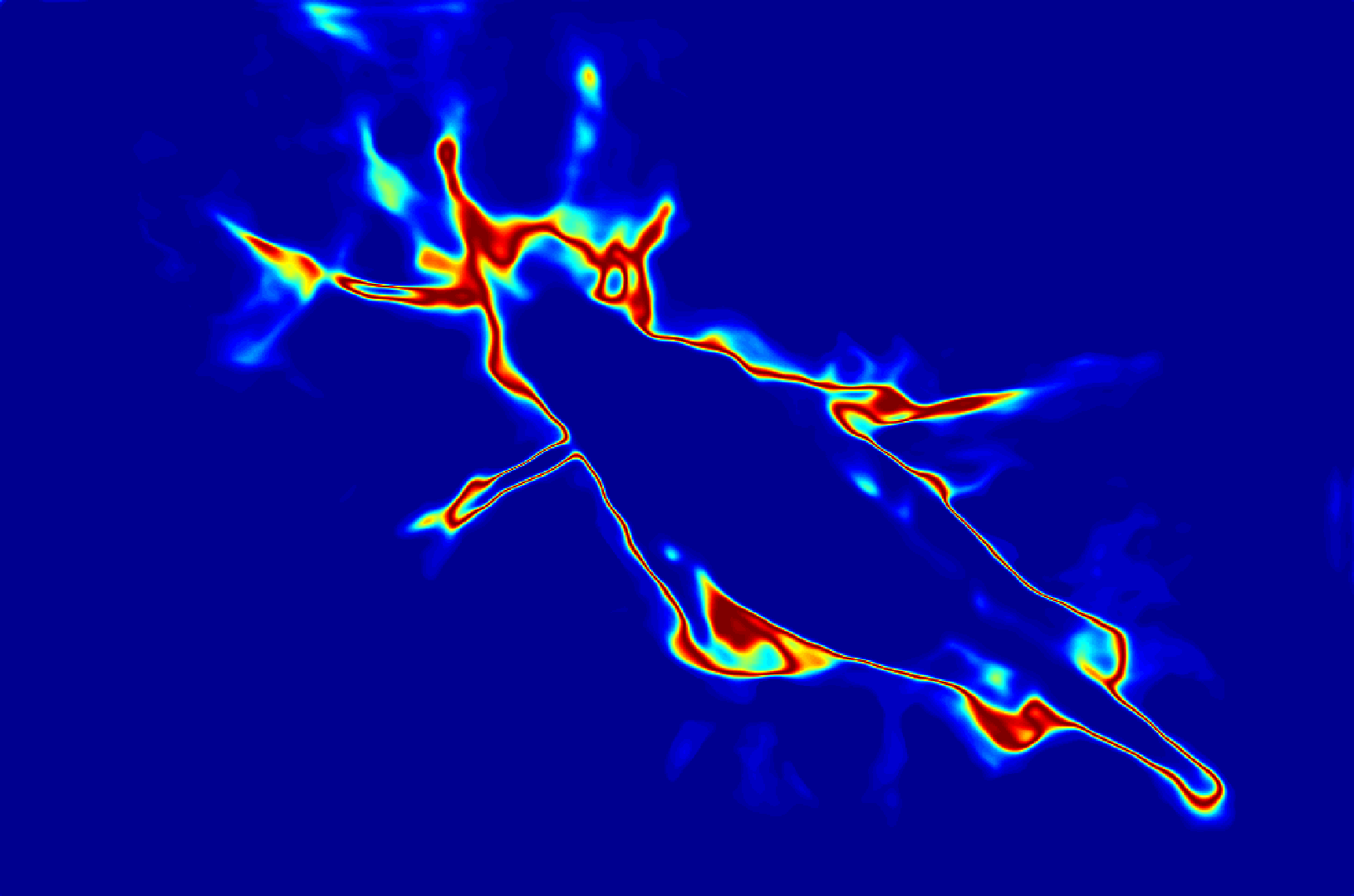}}&
   {\includegraphics[width=0.095\linewidth]{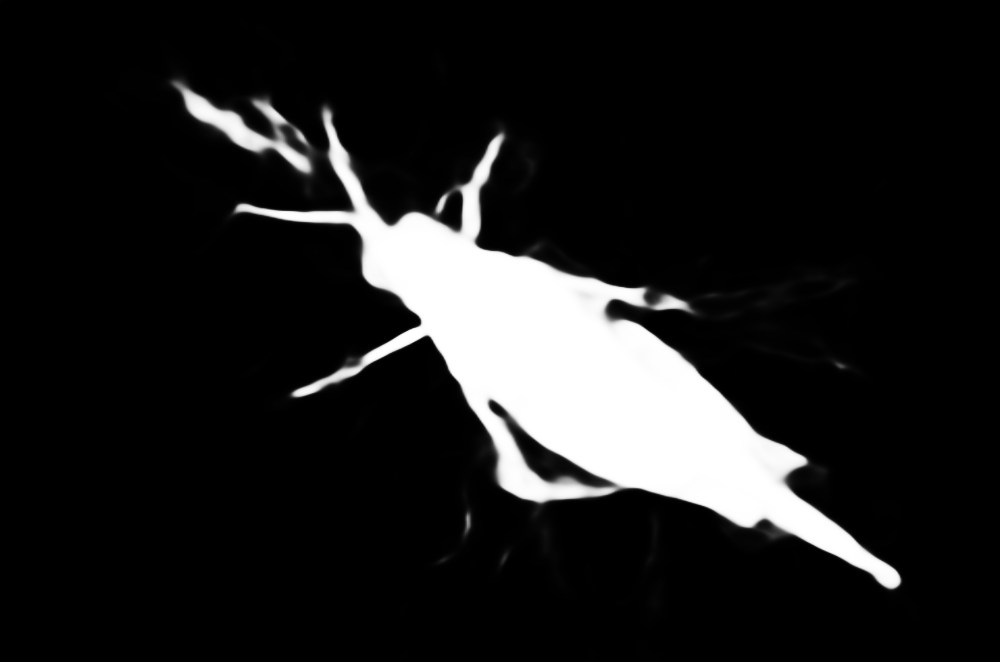}}&
   {\includegraphics[width=0.095\linewidth]{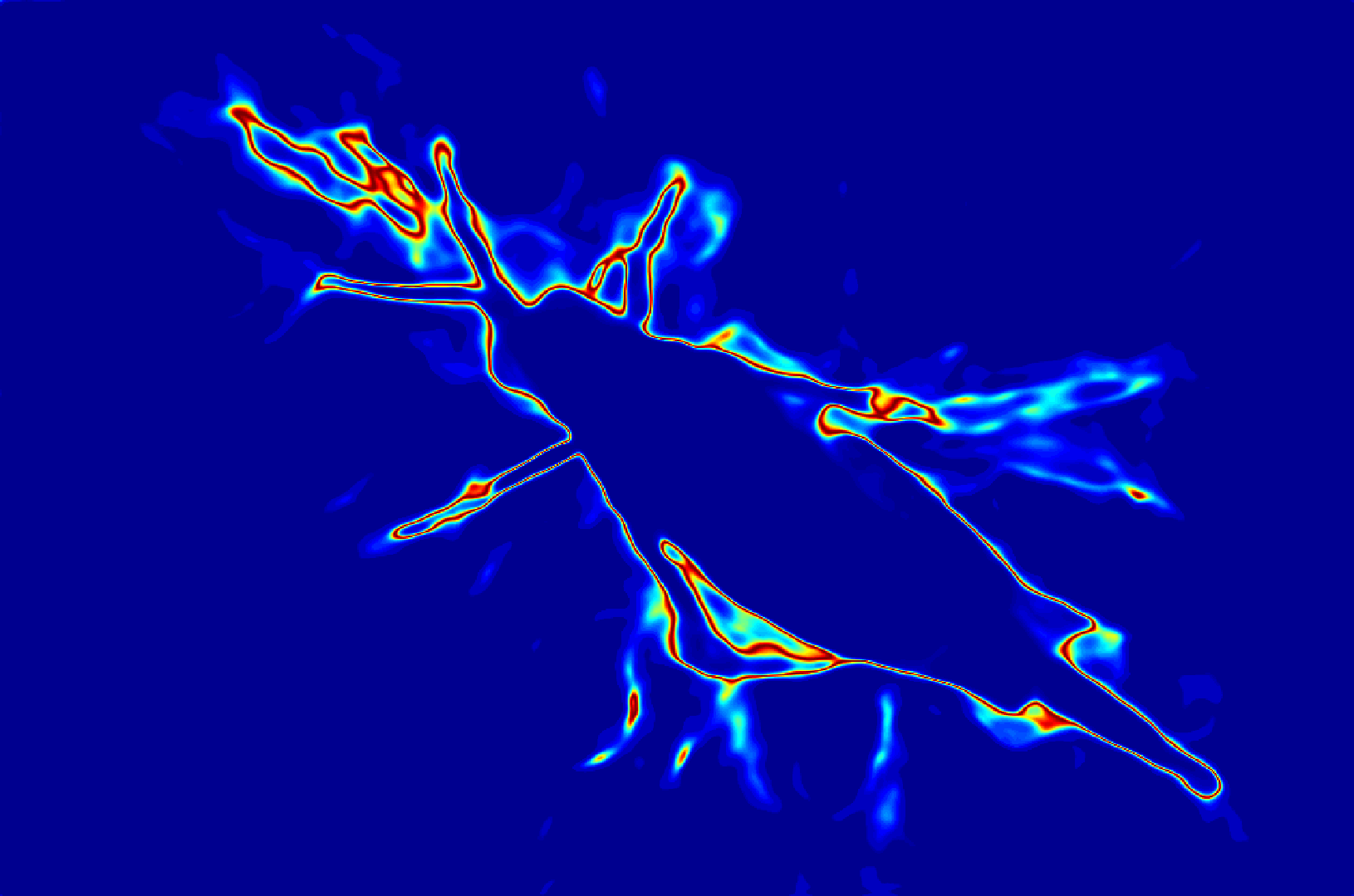}} \\
   {\includegraphics[width=0.095\linewidth]{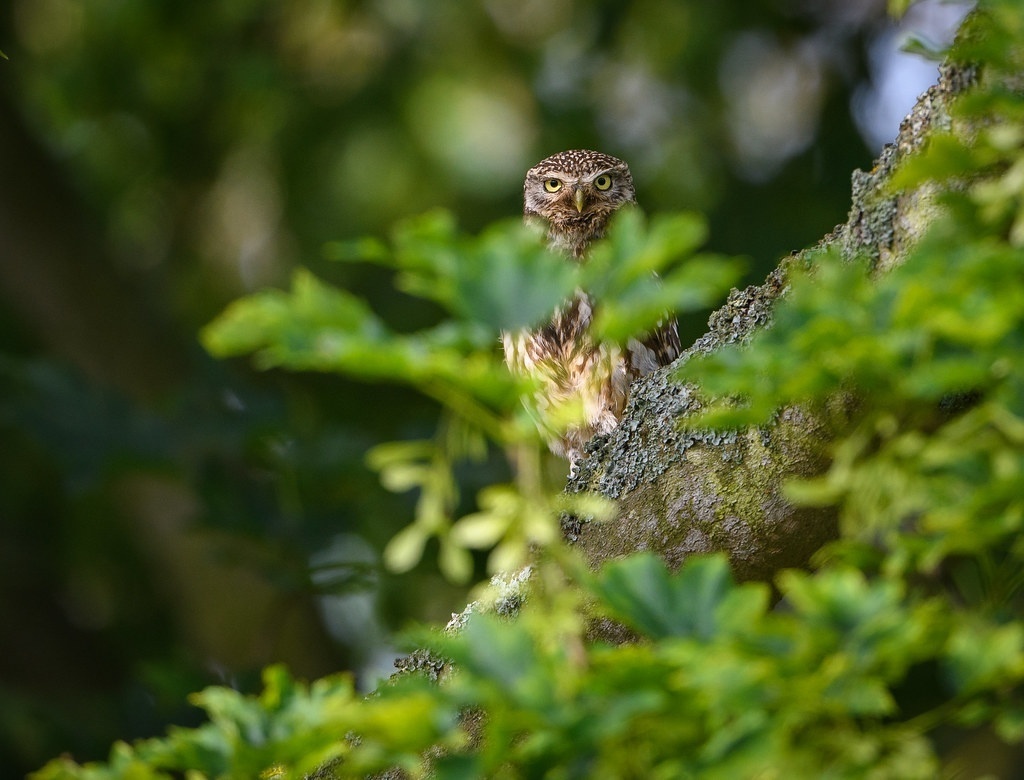}} &
   {\includegraphics[width=0.095\linewidth]{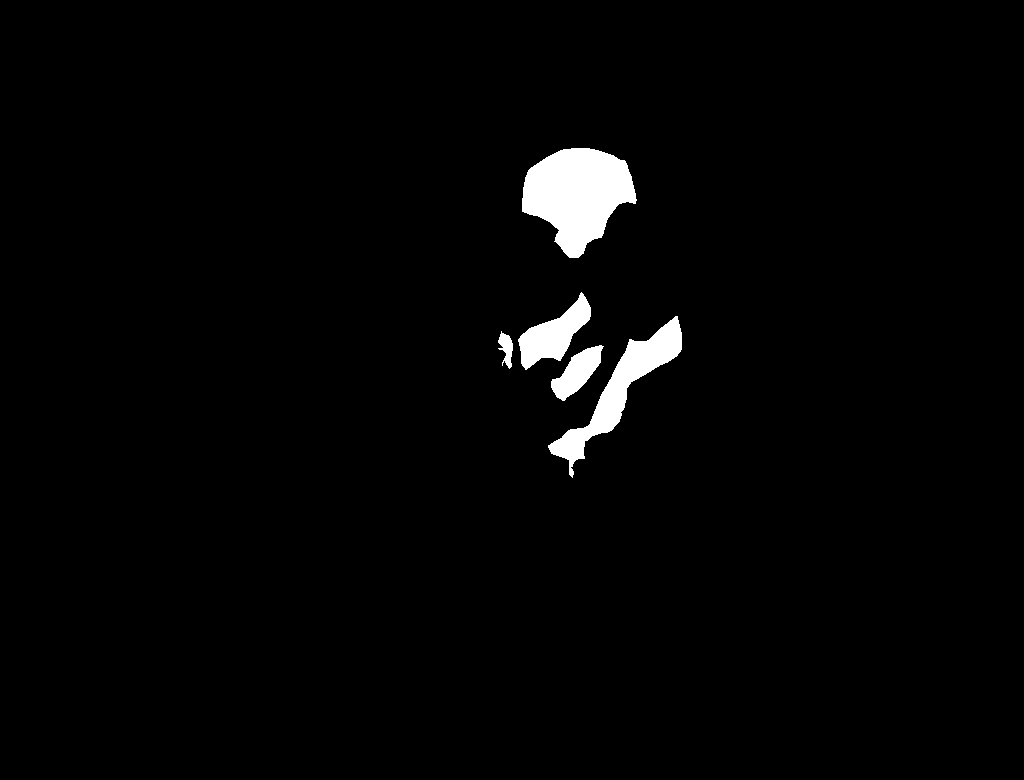}} &
   {\includegraphics[width=0.095\linewidth]{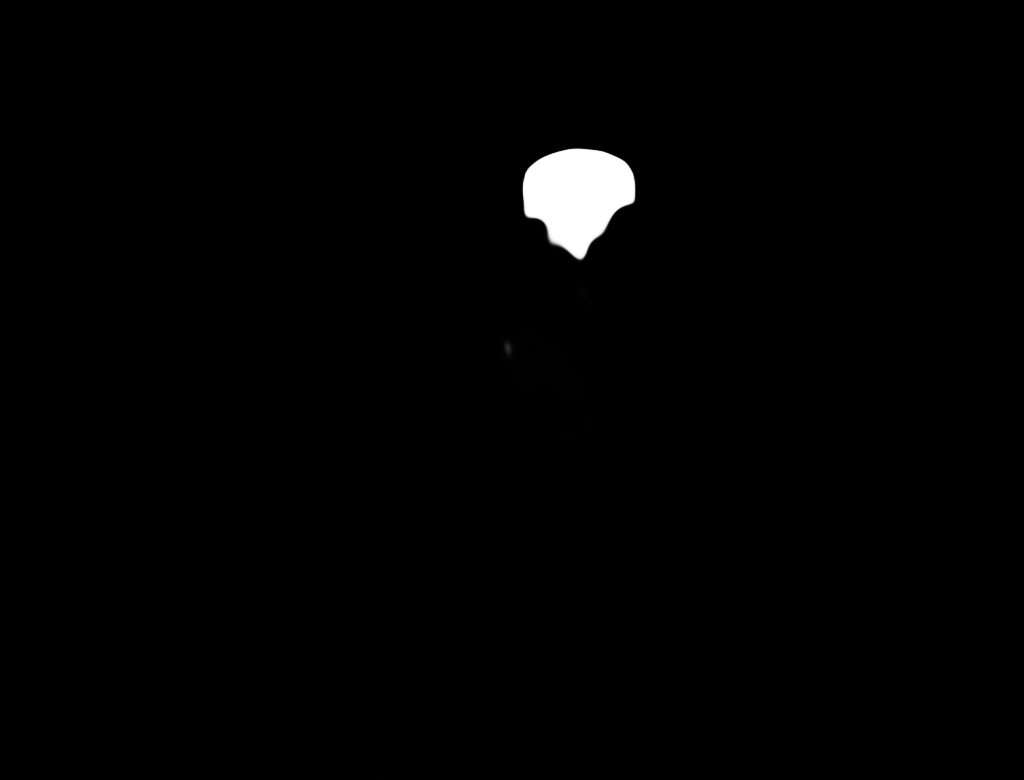}} &
   {\includegraphics[width=0.095\linewidth]{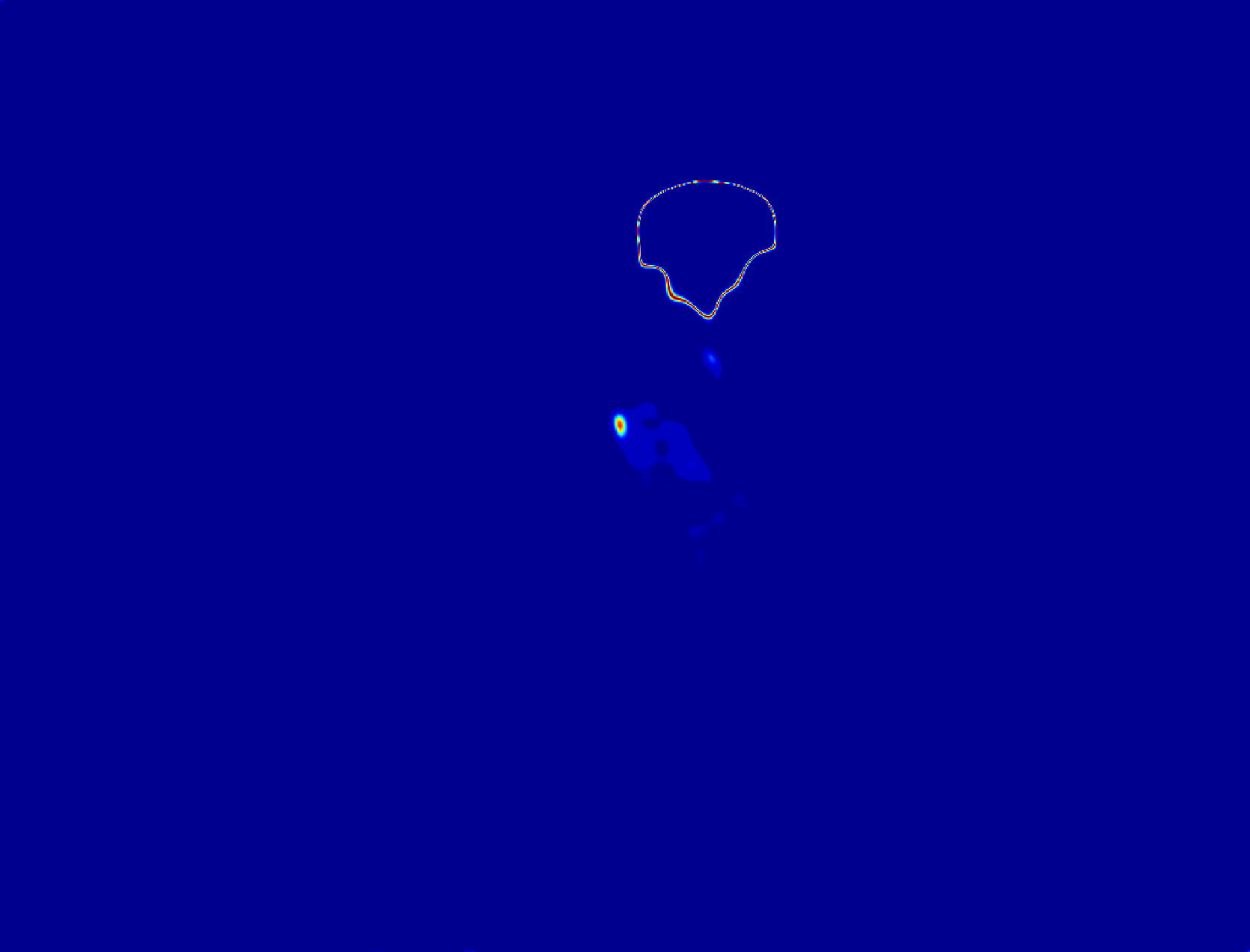}} &
   {\includegraphics[width=0.095\linewidth]{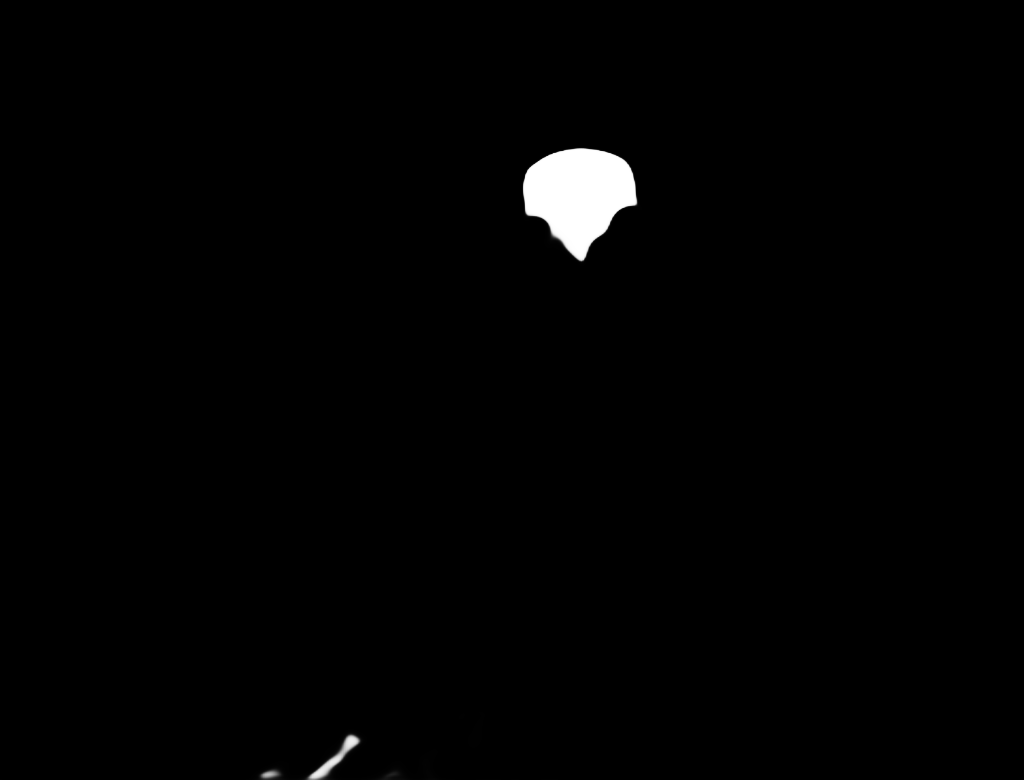}}&
   {\includegraphics[width=0.095\linewidth]{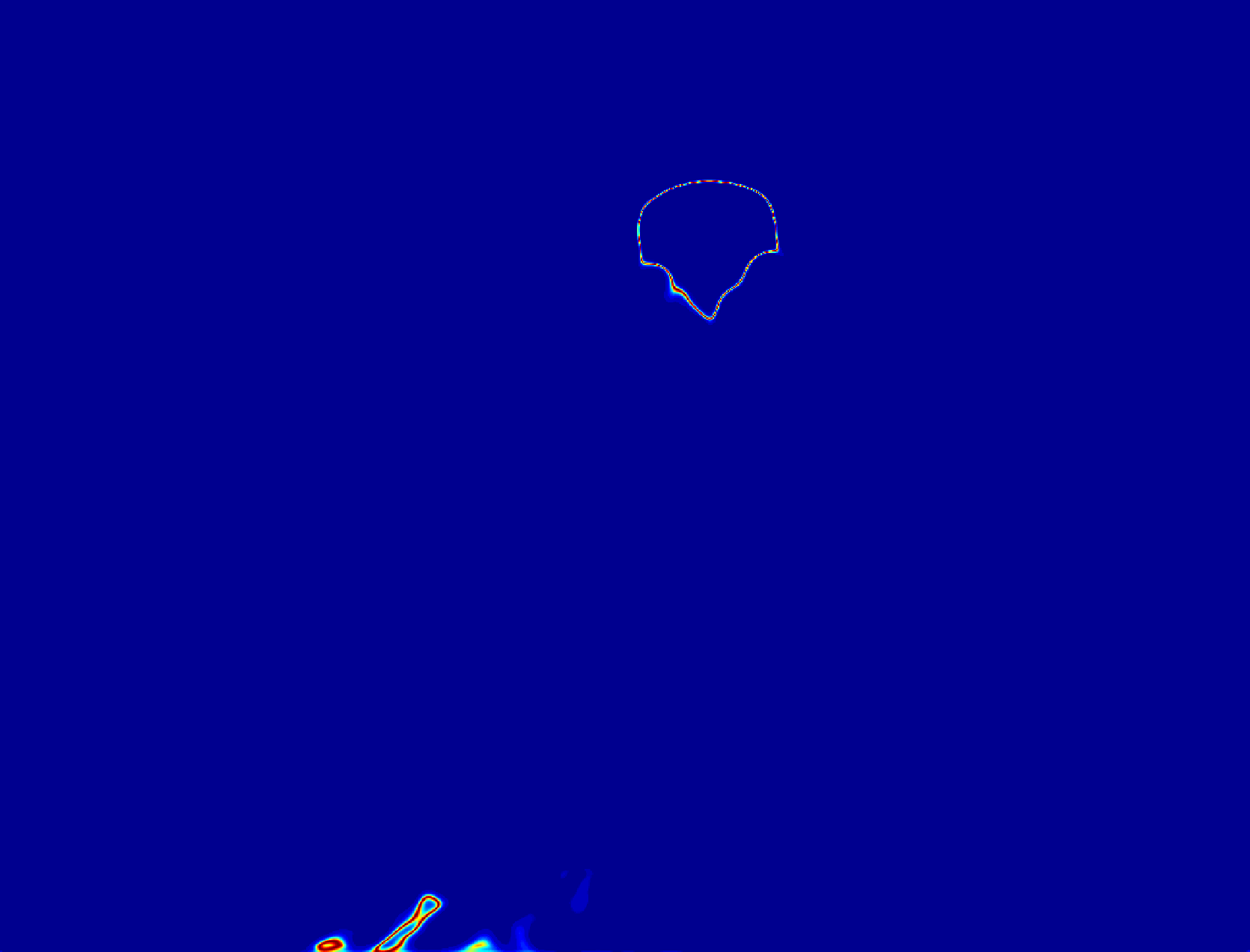}}&
   {\includegraphics[width=0.095\linewidth]{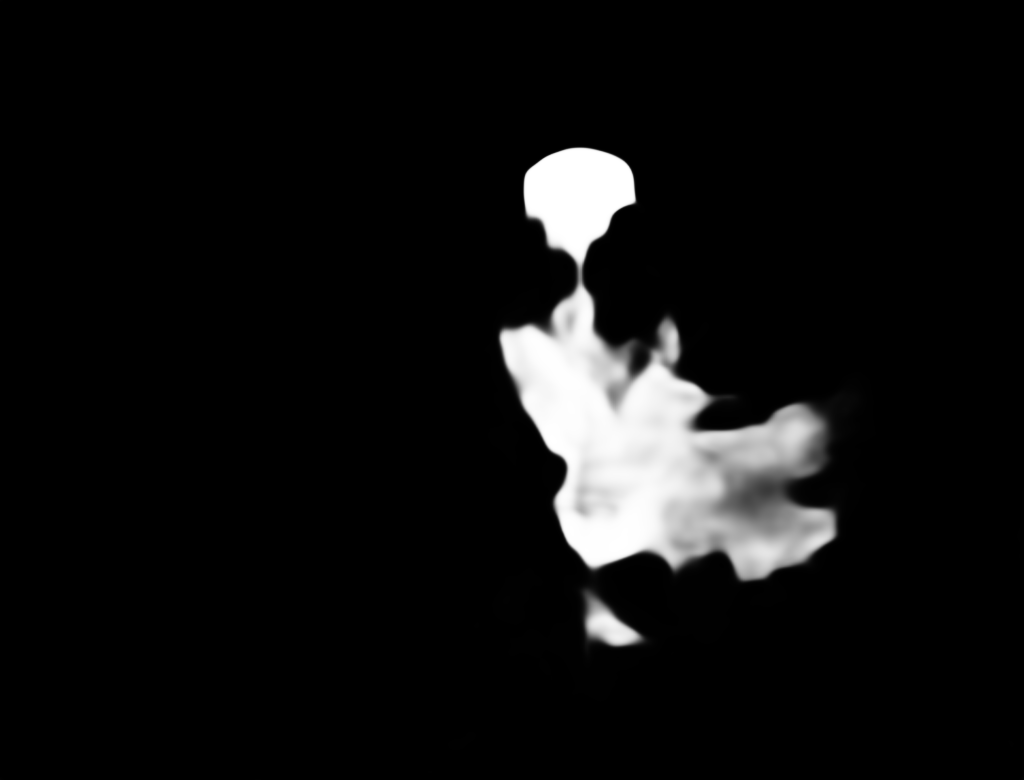}}&
   {\includegraphics[width=0.095\linewidth]{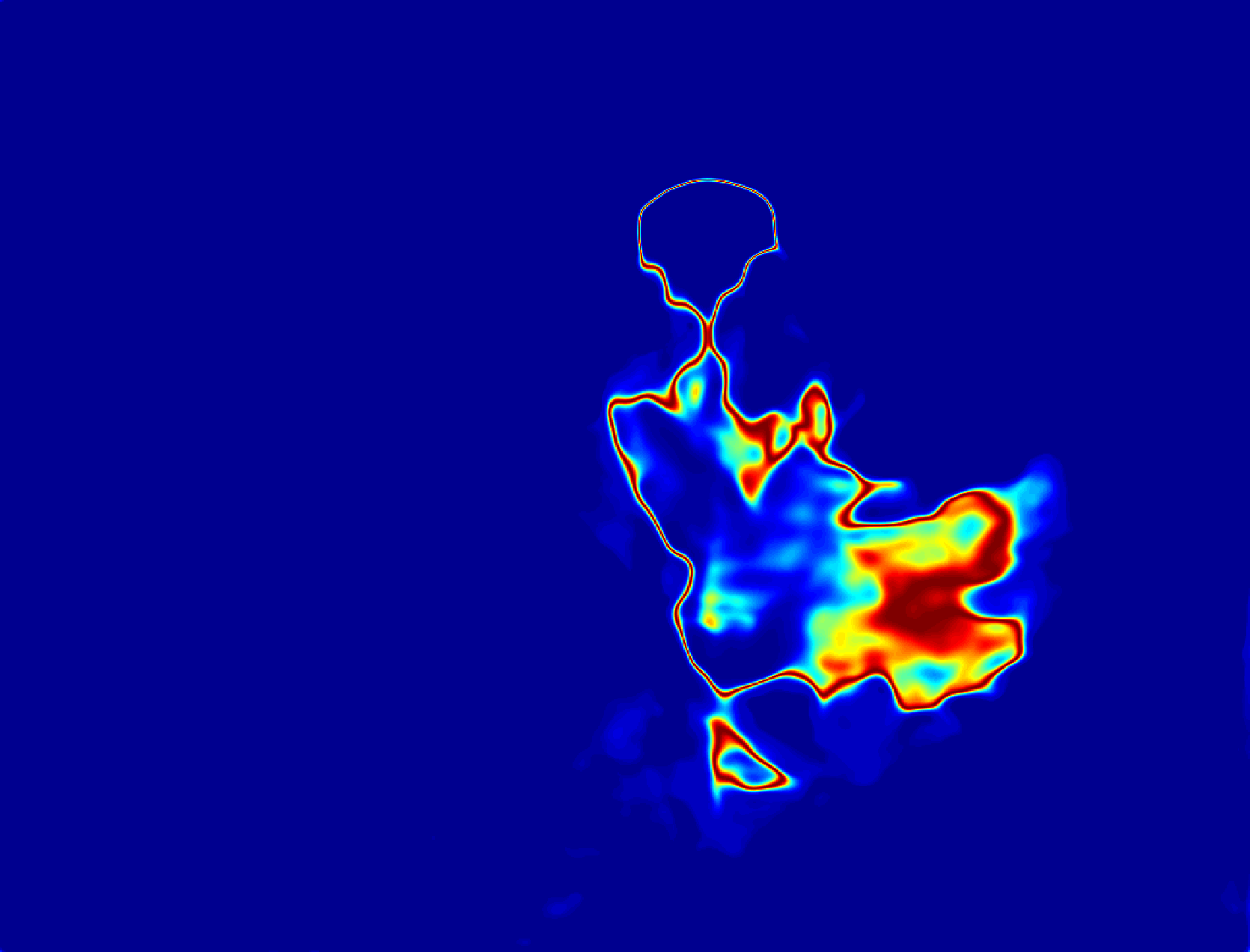}}&
   {\includegraphics[width=0.095\linewidth]{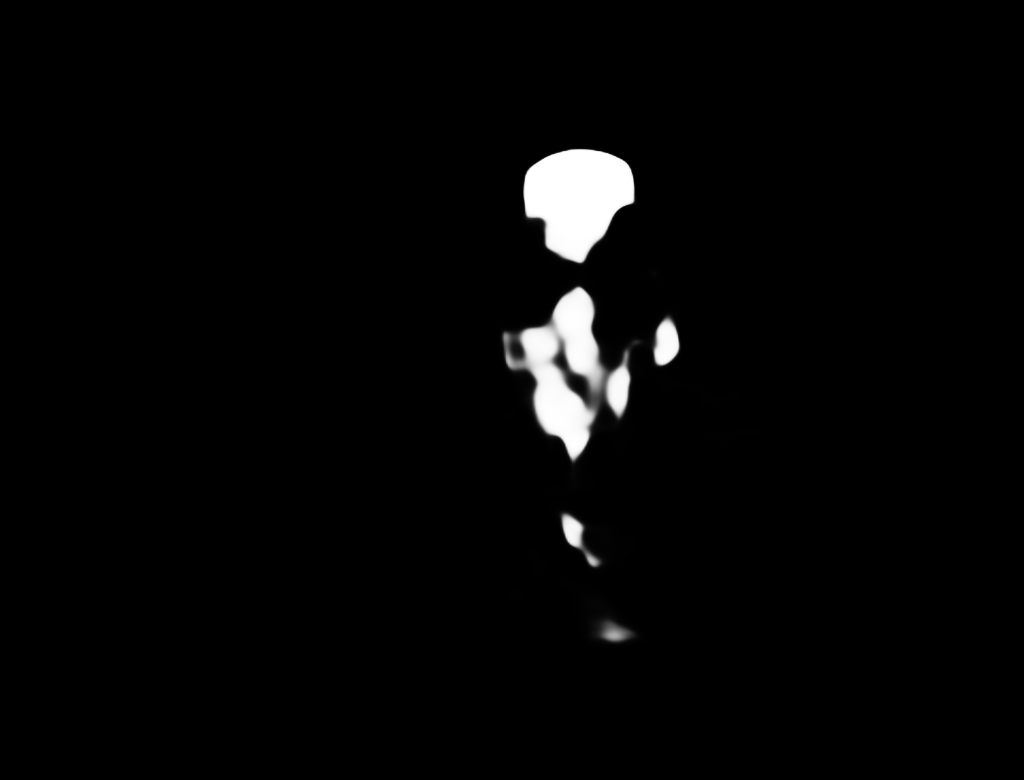}}&
   {\includegraphics[width=0.095\linewidth]{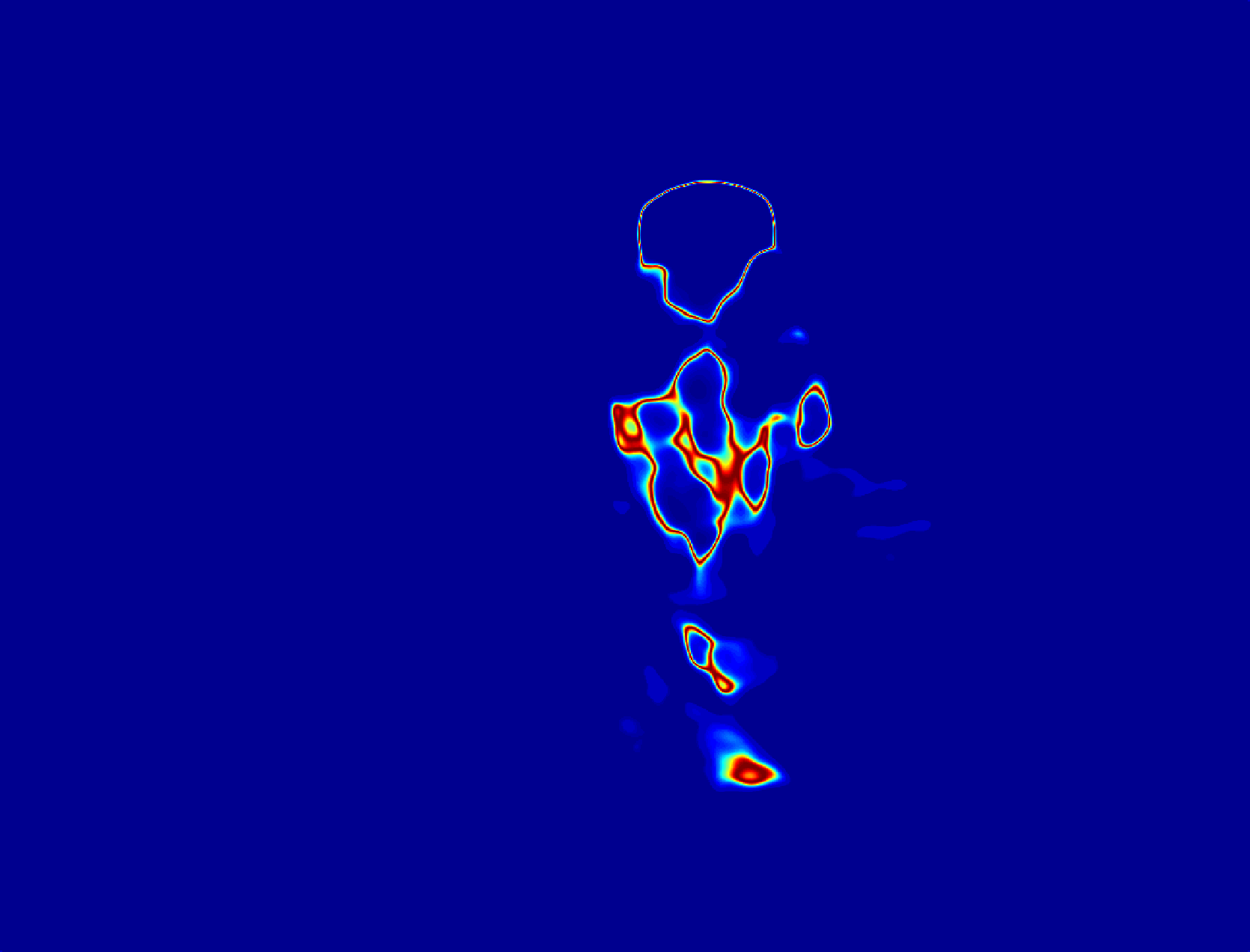}} \\
   \footnotesize{Image}&\footnotesize{GT}&\footnotesize{CVAE}&\footnotesize{$U_p$}&\footnotesize{CGAN}&\footnotesize{$U_p$}&\footnotesize{ABP}&\footnotesize{$U_p$}&\footnotesize{EBM}&\footnotesize{$U_p$}\\
   \end{tabular}
   \end{center}
   \caption{\footnotesize{Predictive uncertainty of generative model based solutions for \textbf{camouflaged object detection}.}
   }
\label{fig:predictive_generative_cod}
\end{figure*}

\begin{figure*}[tp]
   \begin{center}
   \begin{tabular}{c@{ }c@{ }c@{ }c@{ }c@{ }c@{ }c@{ }c@{ }c@{ }c@{ }}
   {\includegraphics[width=0.095\linewidth]{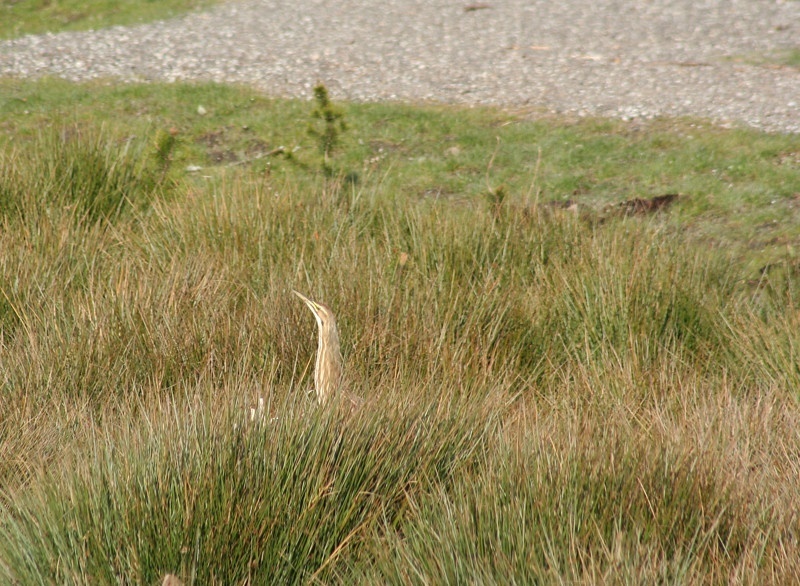}} &
   {\includegraphics[width=0.095\linewidth]{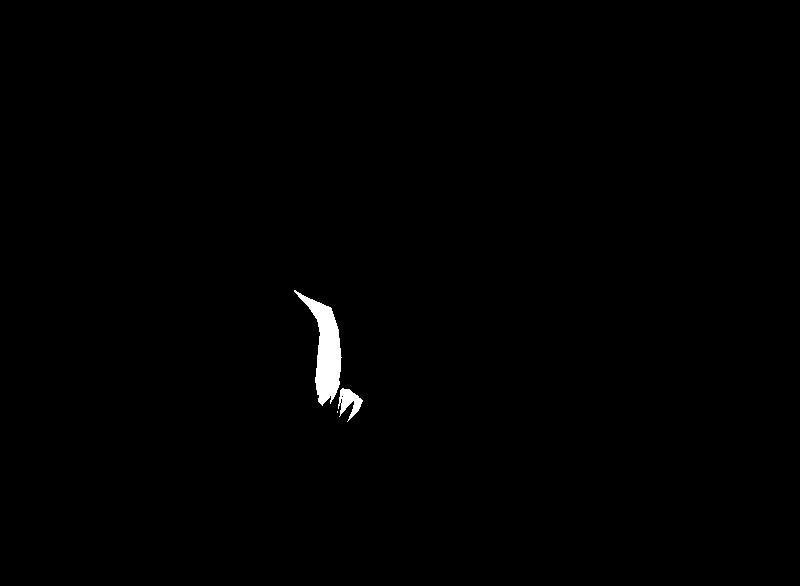}} &
   {\includegraphics[width=0.095\linewidth]{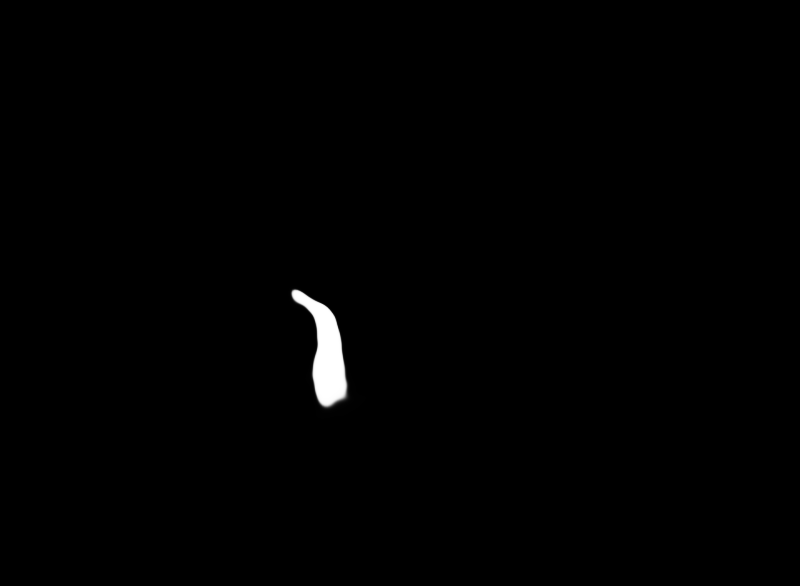}} &
   {\includegraphics[width=0.095\linewidth]{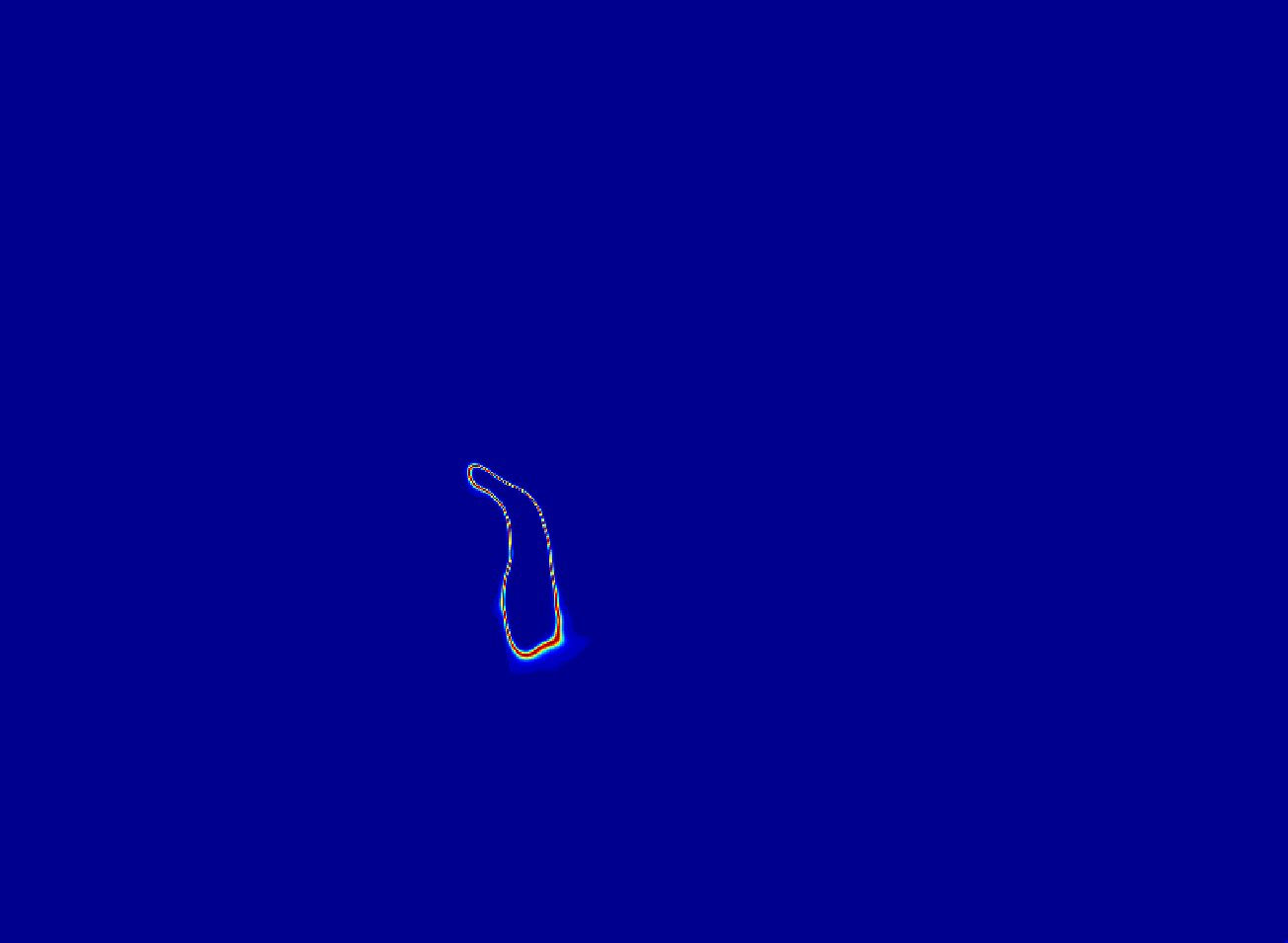}} &
   {\includegraphics[width=0.095\linewidth]{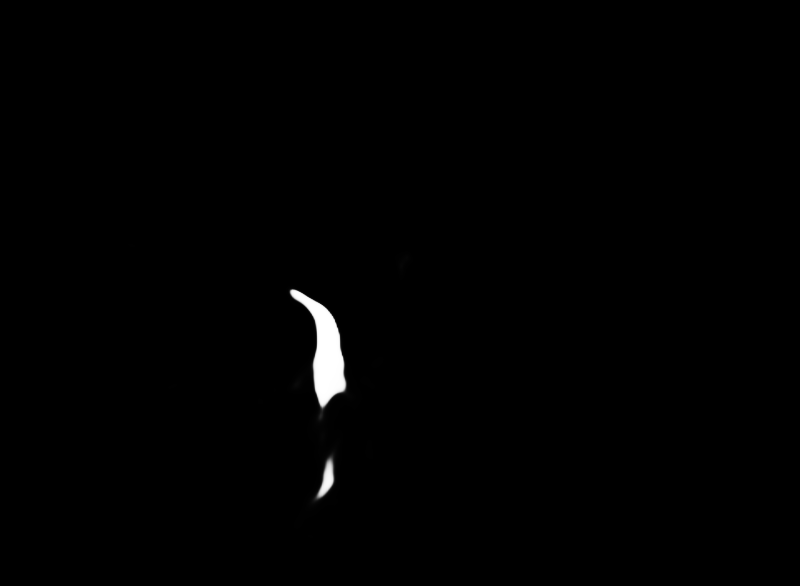}}&
   {\includegraphics[width=0.095\linewidth]{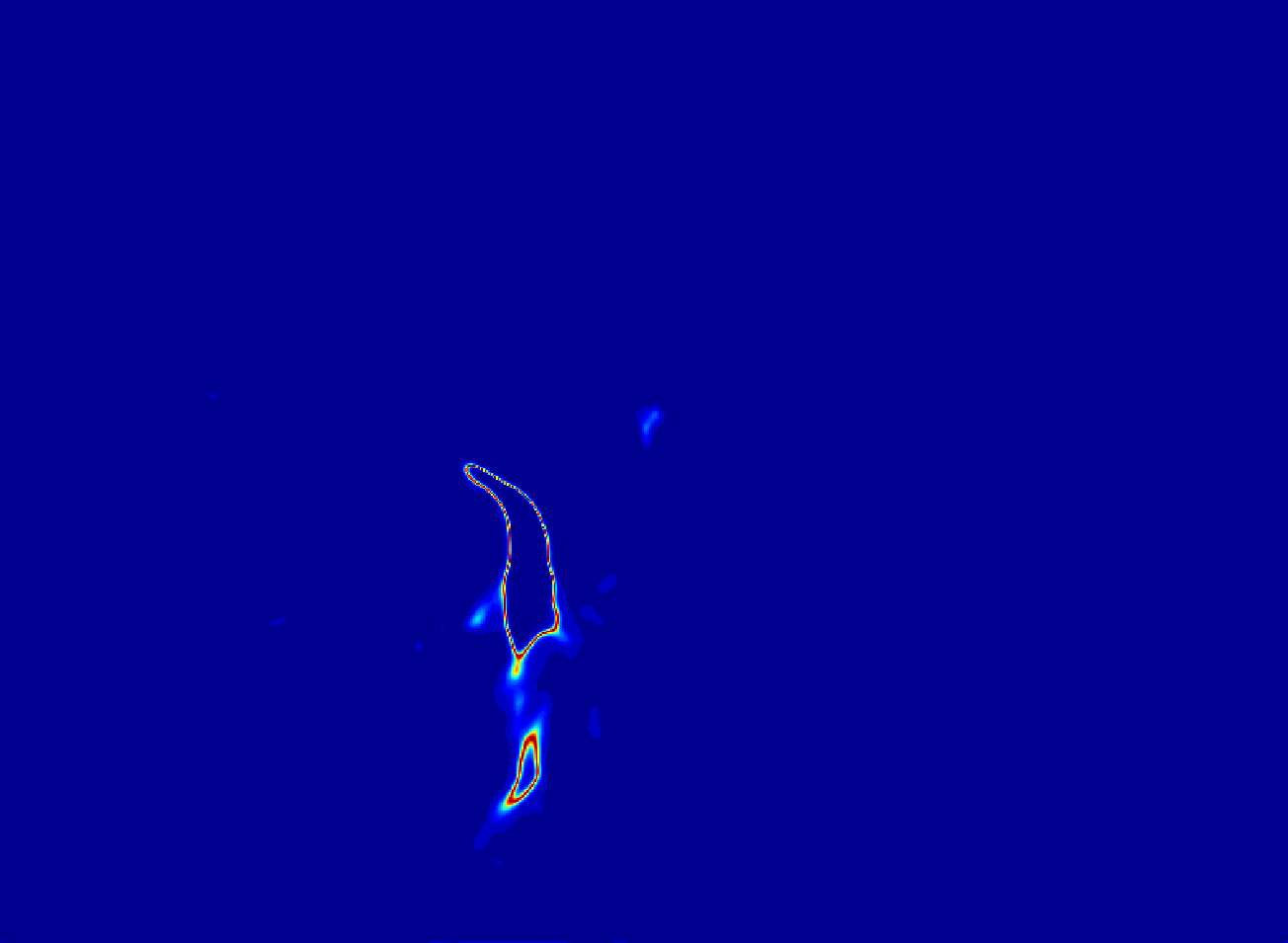}}&
   {\includegraphics[width=0.095\linewidth]{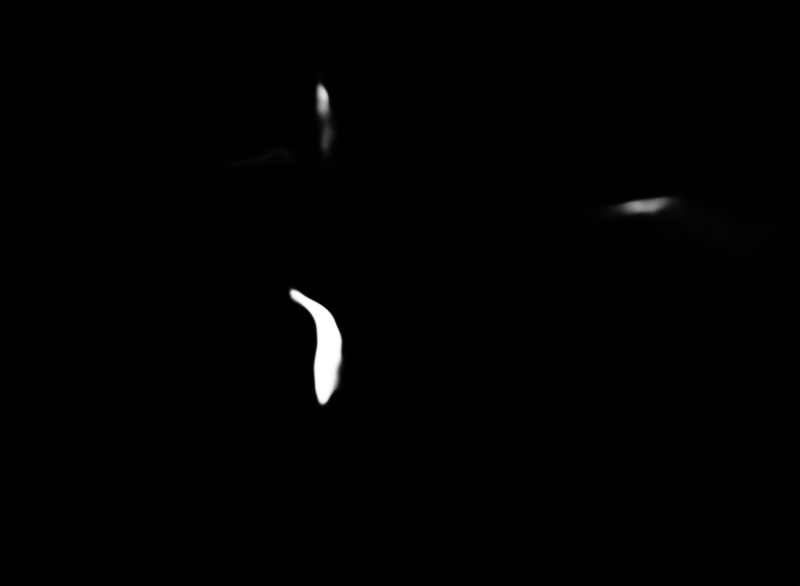}}&
   {\includegraphics[width=0.095\linewidth]{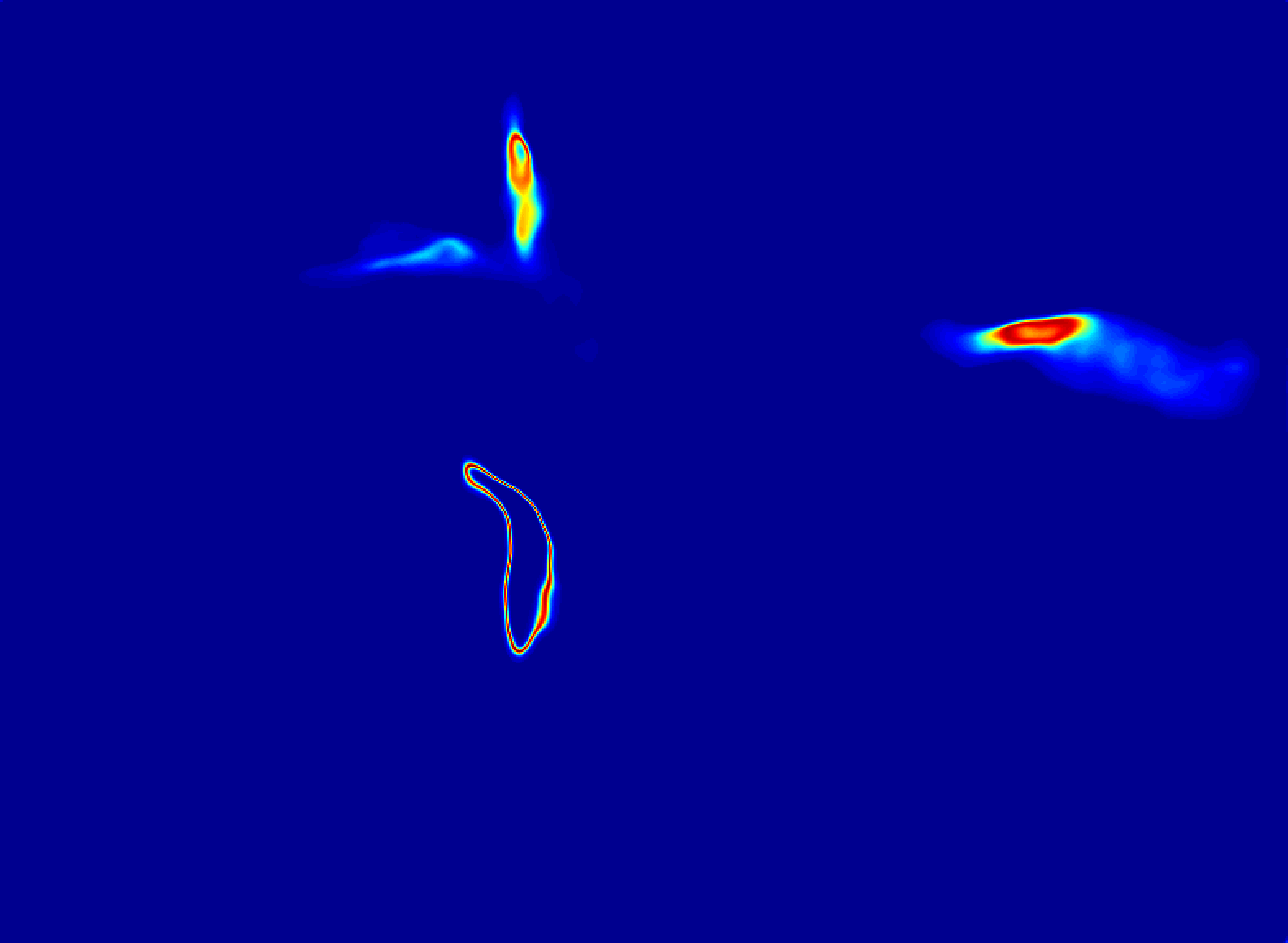}}&
   {\includegraphics[width=0.095\linewidth]{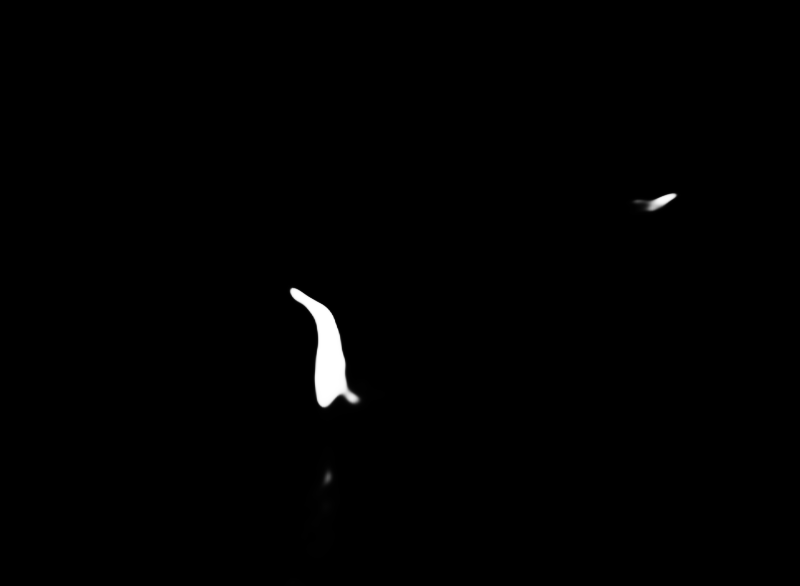}}&
   {\includegraphics[width=0.095\linewidth]{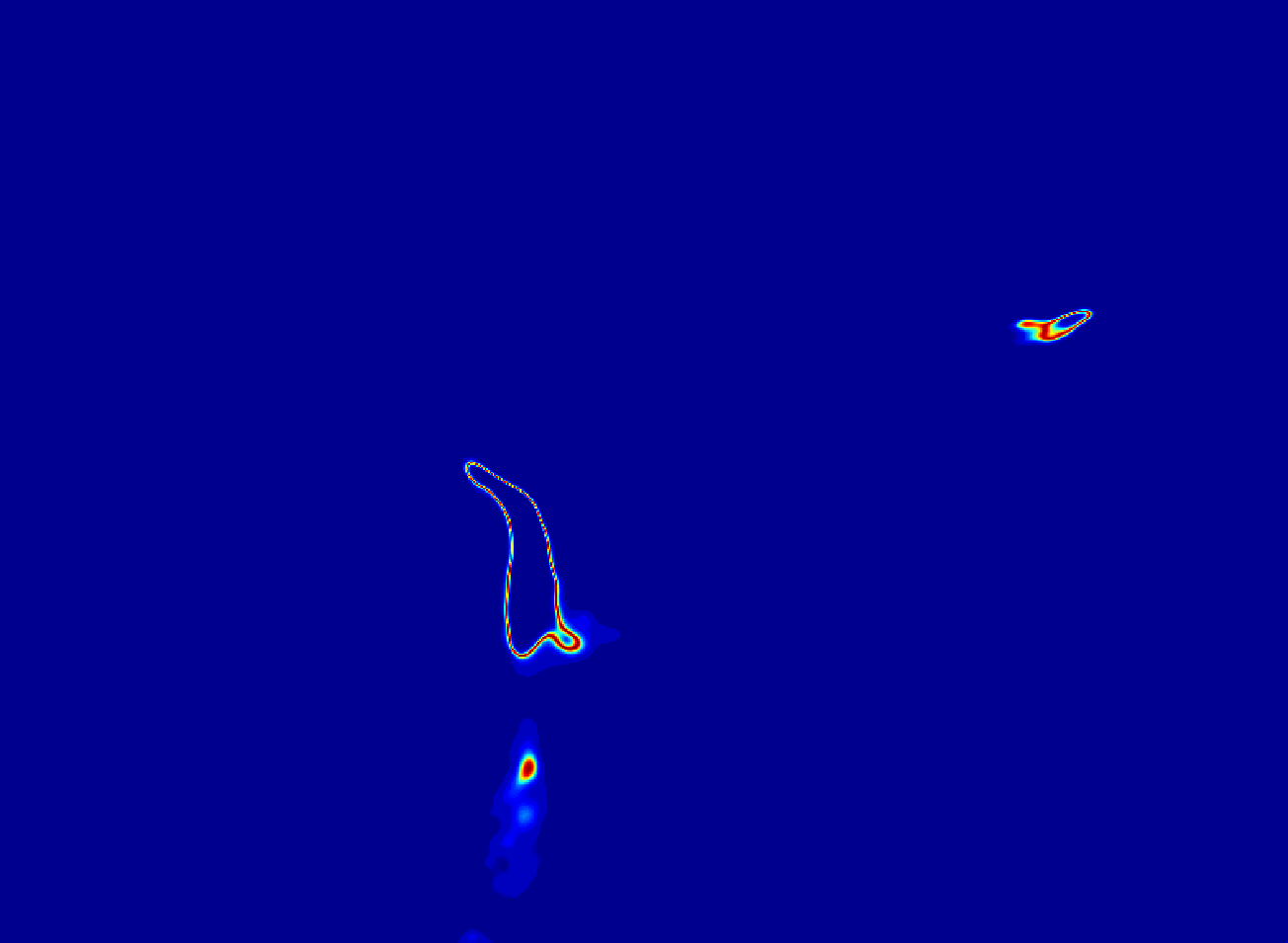}} \\
   {\includegraphics[width=0.095\linewidth]{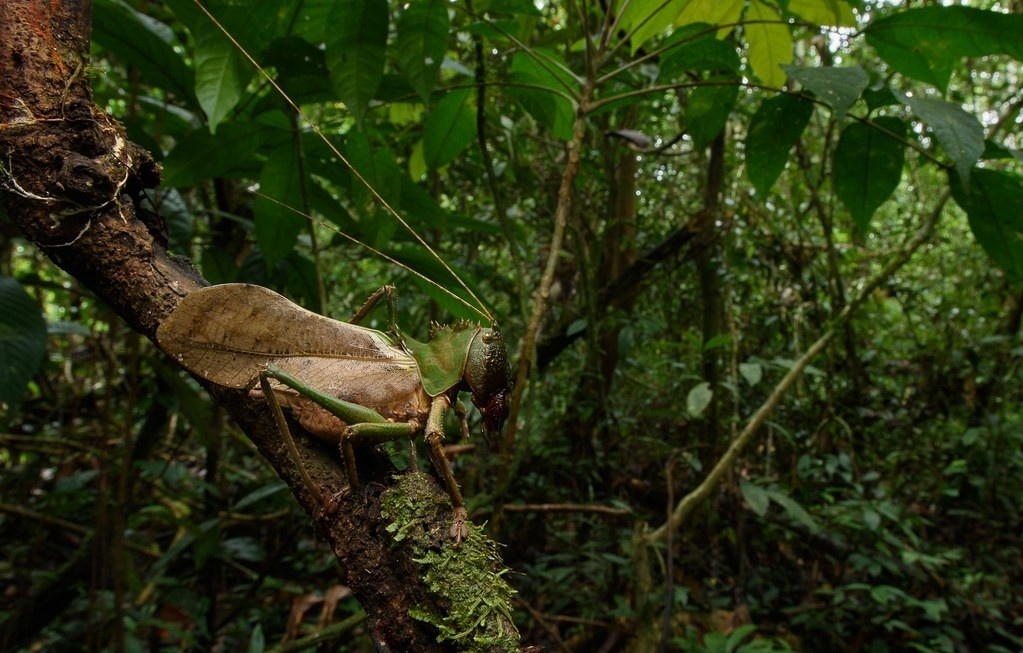}} &
   {\includegraphics[width=0.095\linewidth]{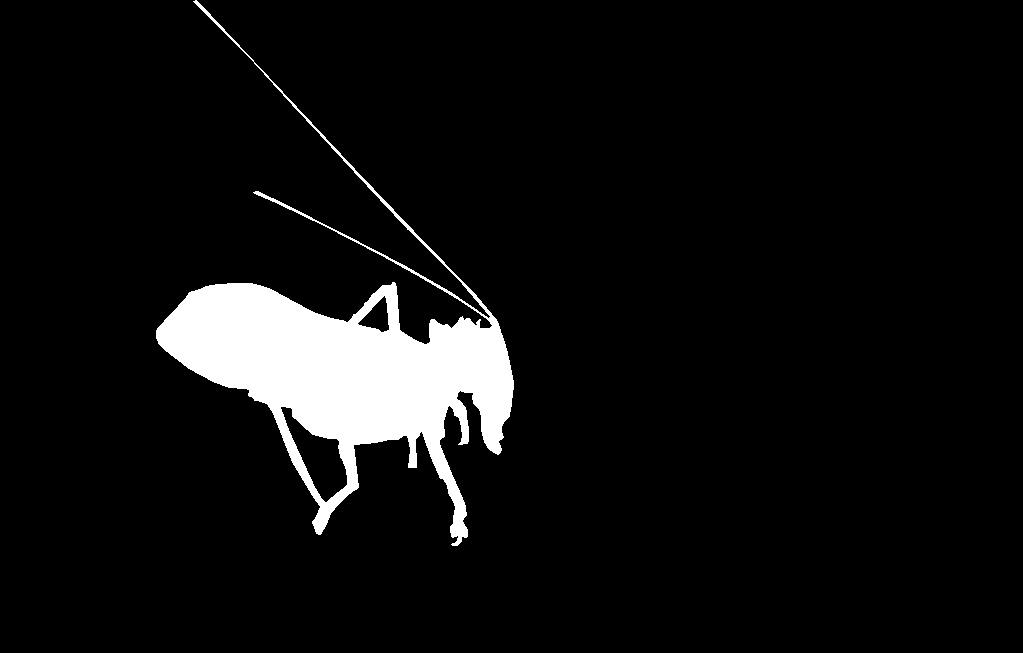}} &
   {\includegraphics[width=0.095\linewidth]{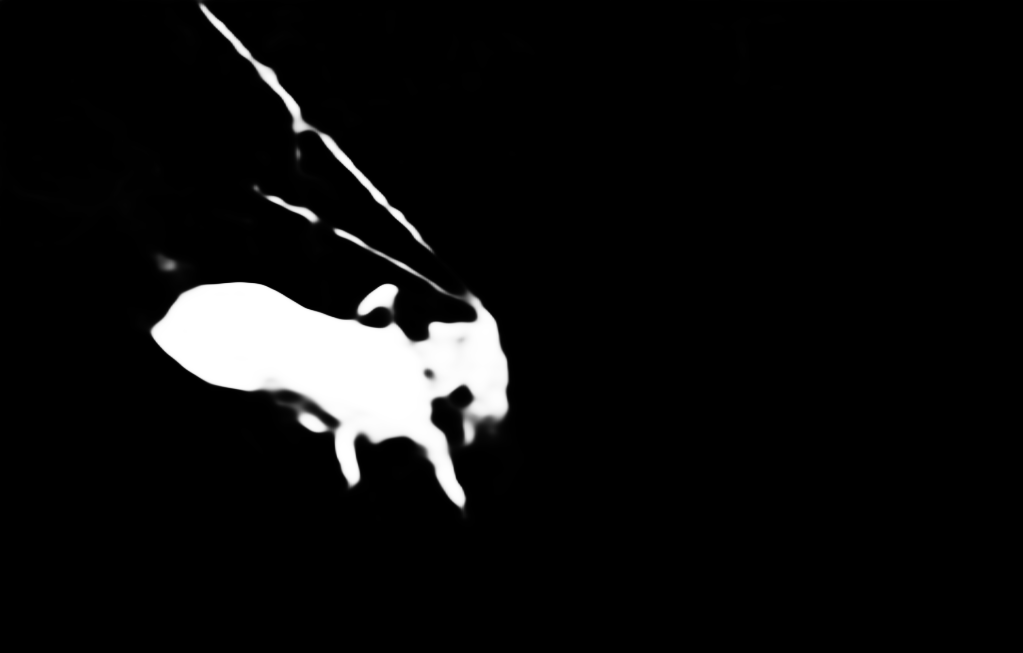}} &
   {\includegraphics[width=0.095\linewidth]{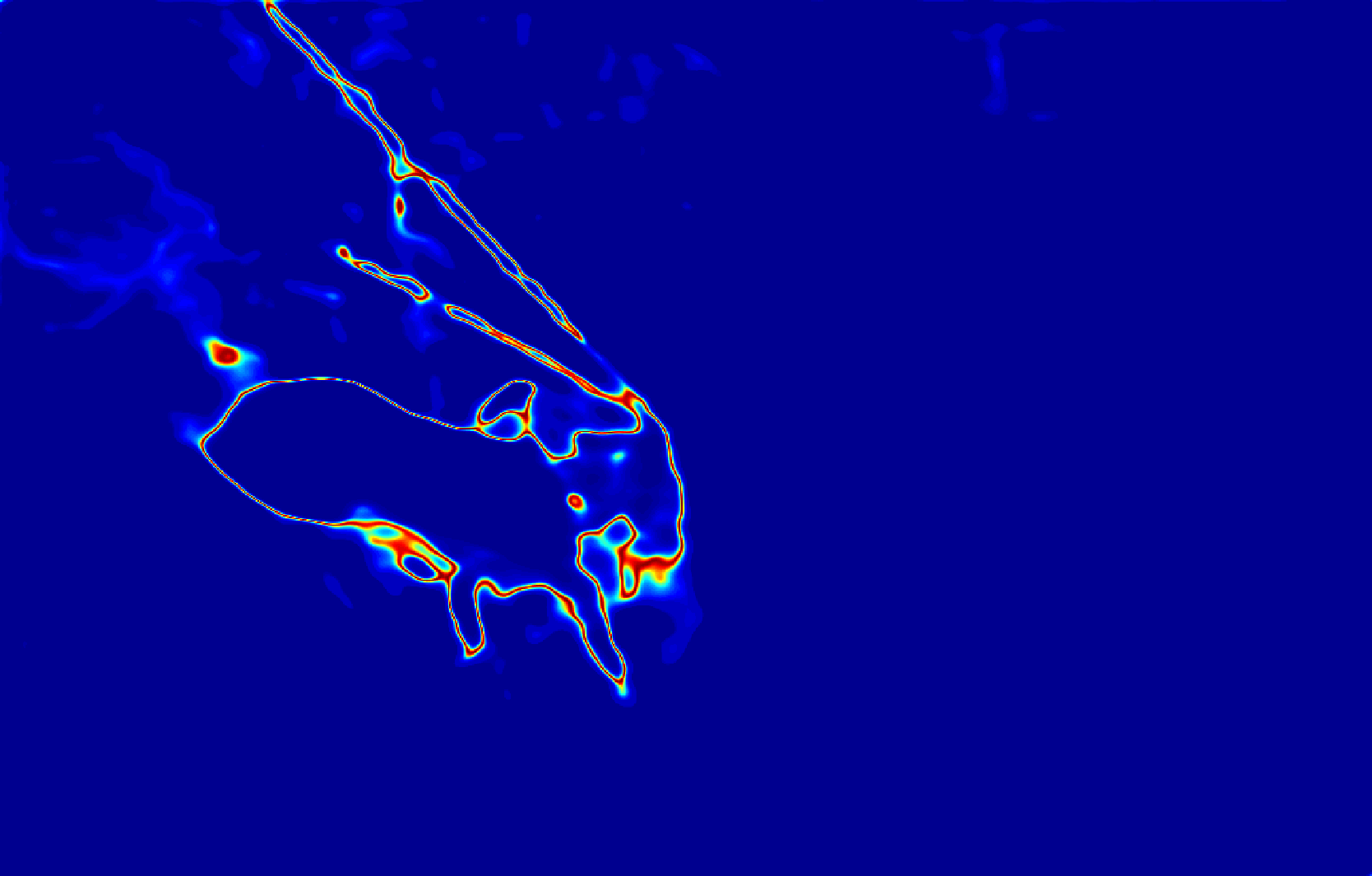}} &
   {\includegraphics[width=0.095\linewidth]{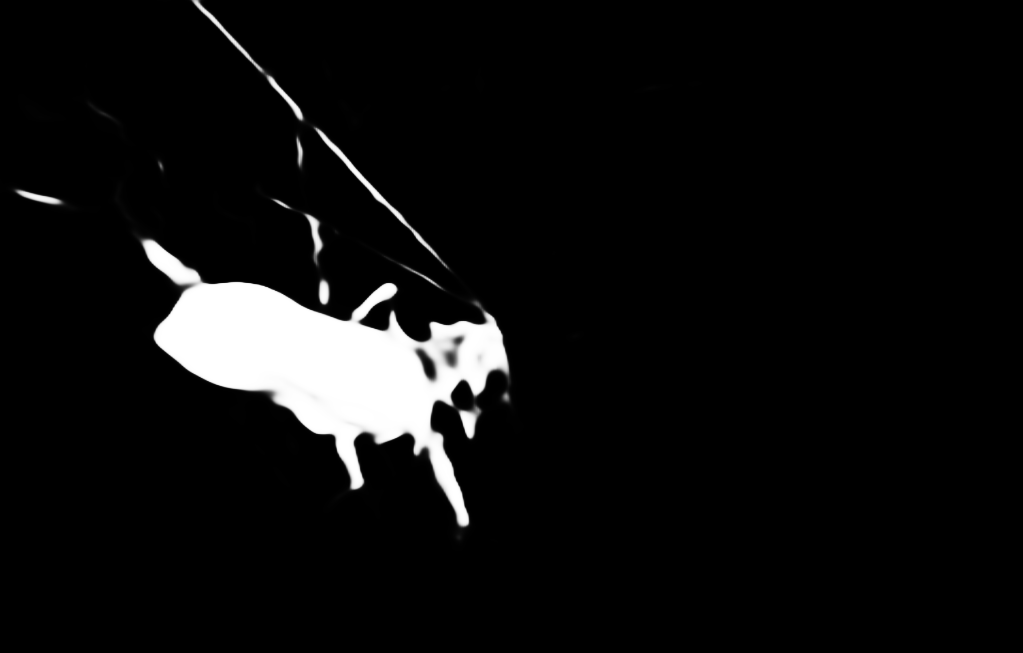}}&
   {\includegraphics[width=0.095\linewidth]{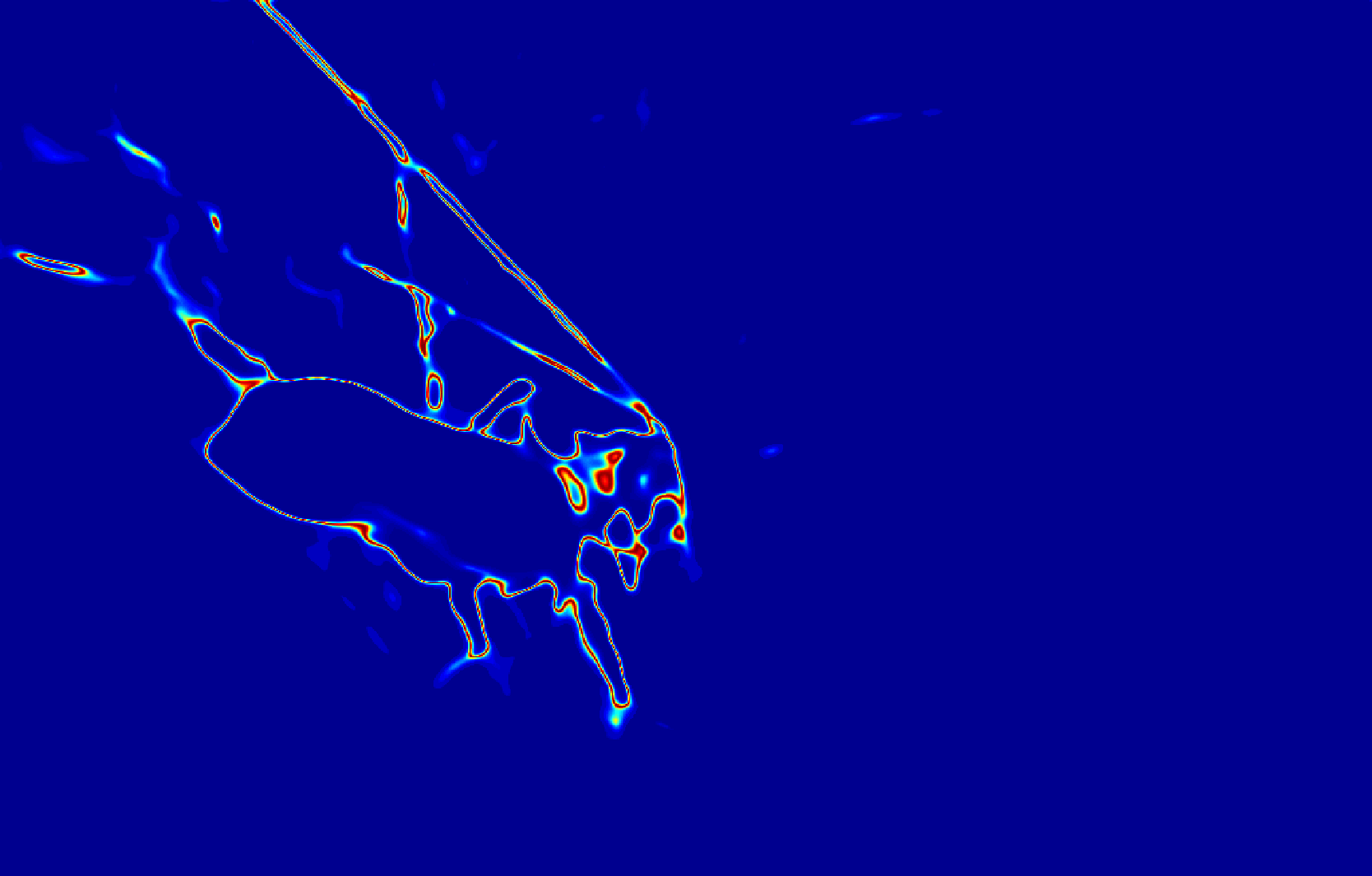}}&
   {\includegraphics[width=0.095\linewidth]{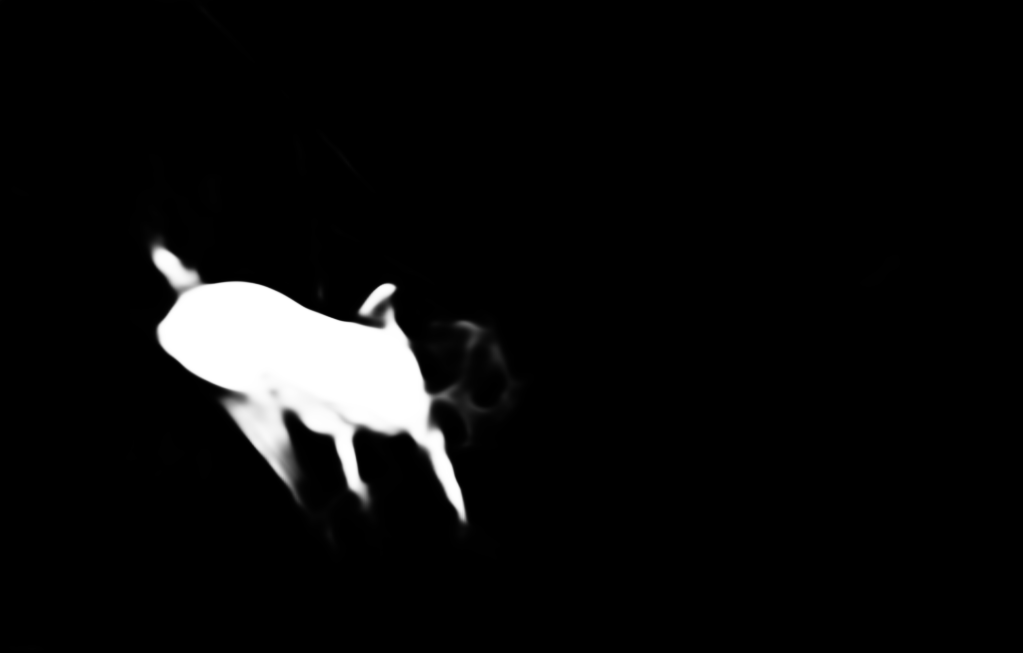}}&
   {\includegraphics[width=0.095\linewidth]{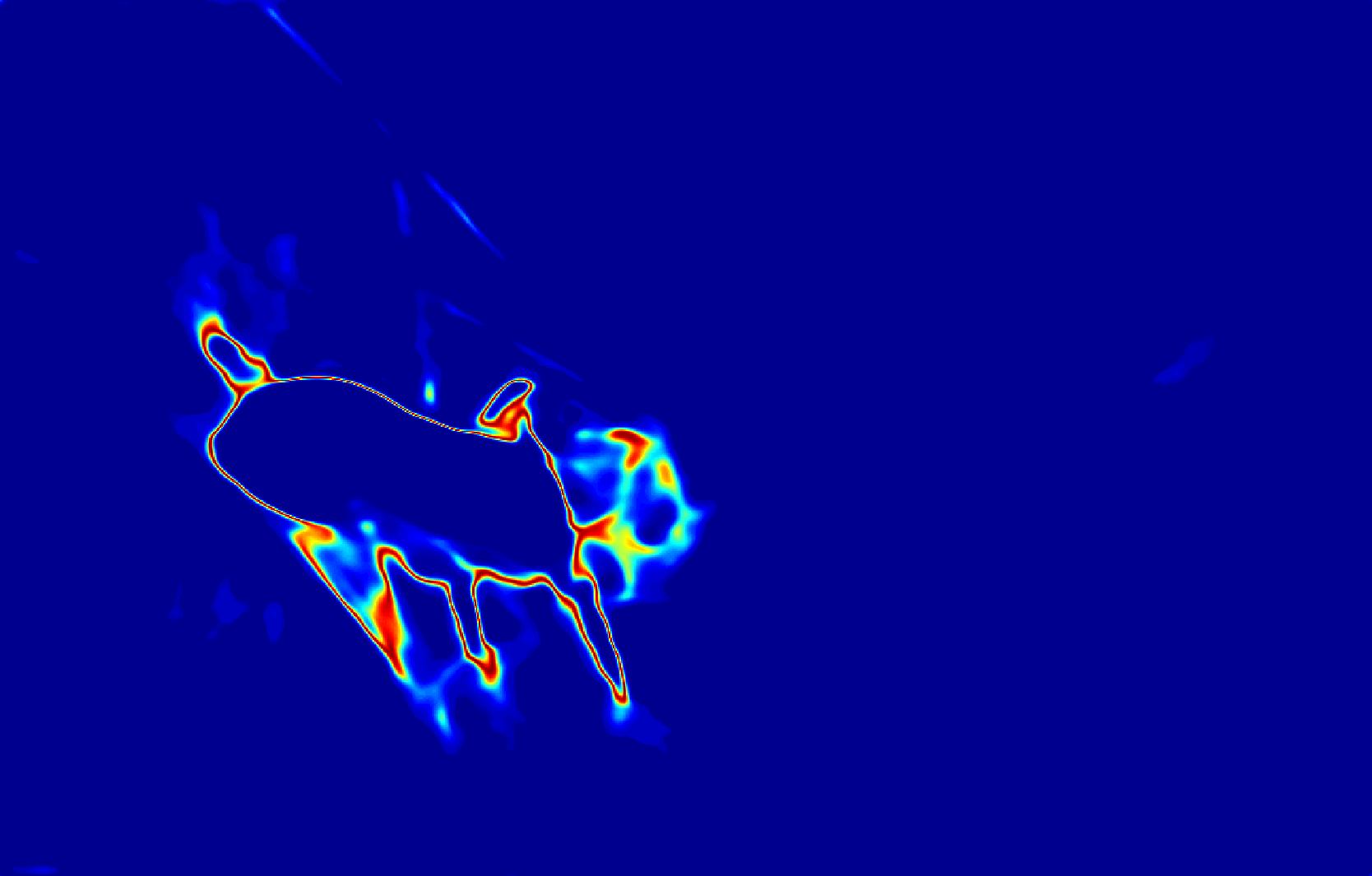}}&
   {\includegraphics[width=0.095\linewidth]{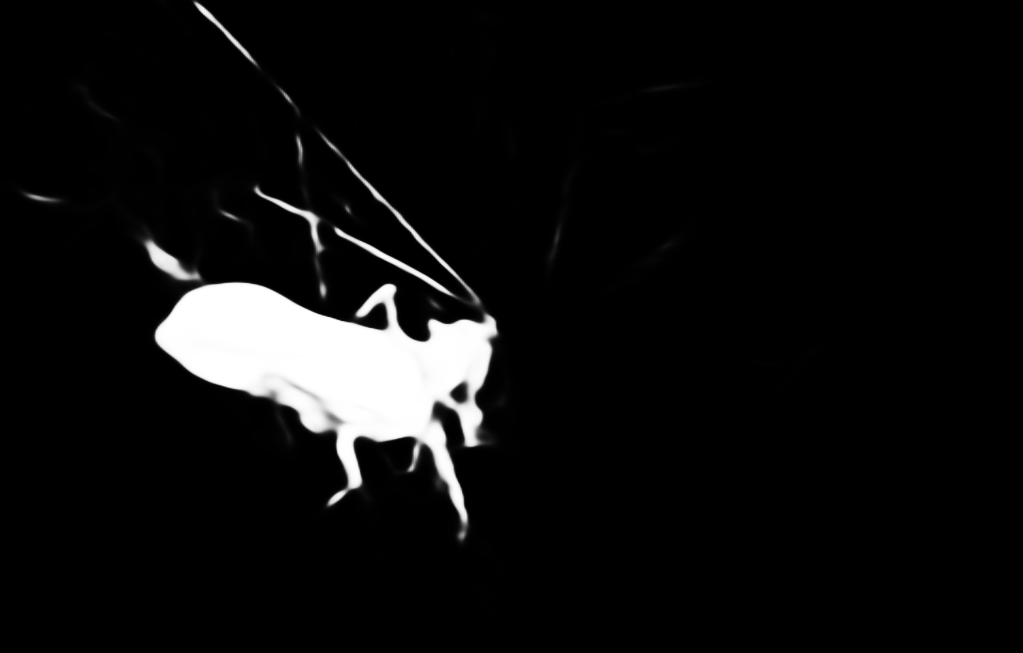}}&
   {\includegraphics[width=0.095\linewidth]{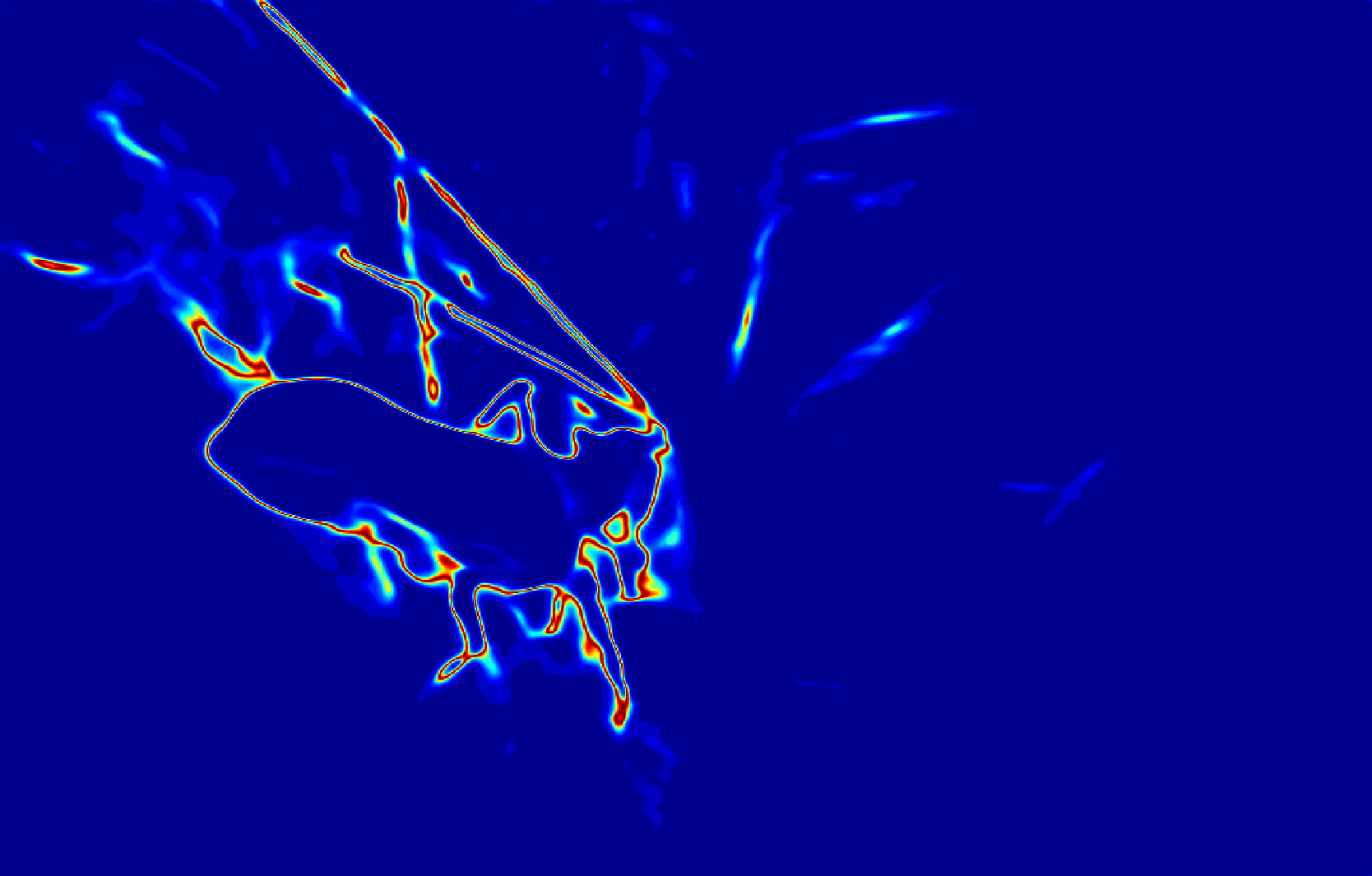}} \\
   {\includegraphics[width=0.095\linewidth]{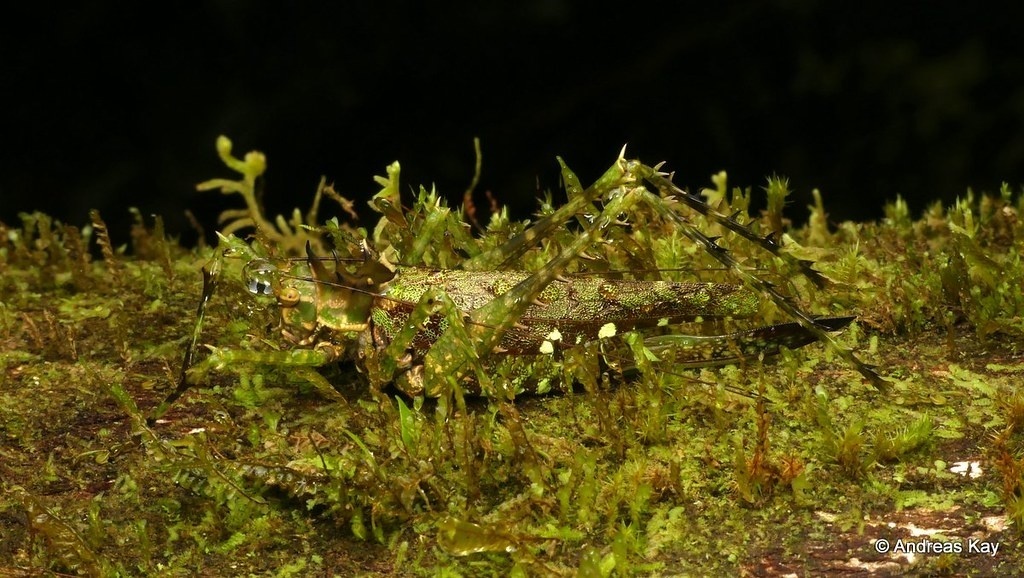}} &
   {\includegraphics[width=0.095\linewidth]{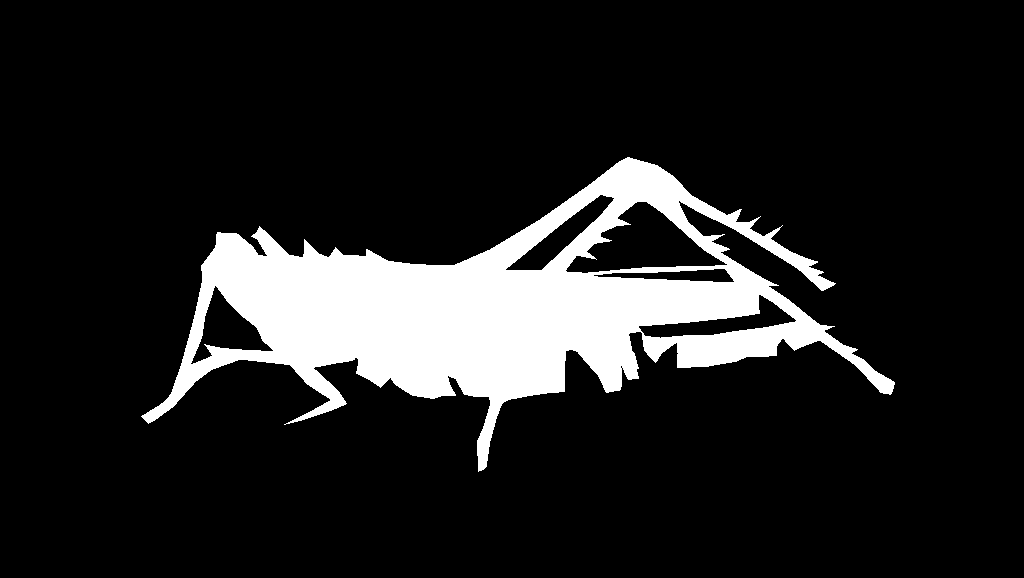}} &
   {\includegraphics[width=0.095\linewidth]{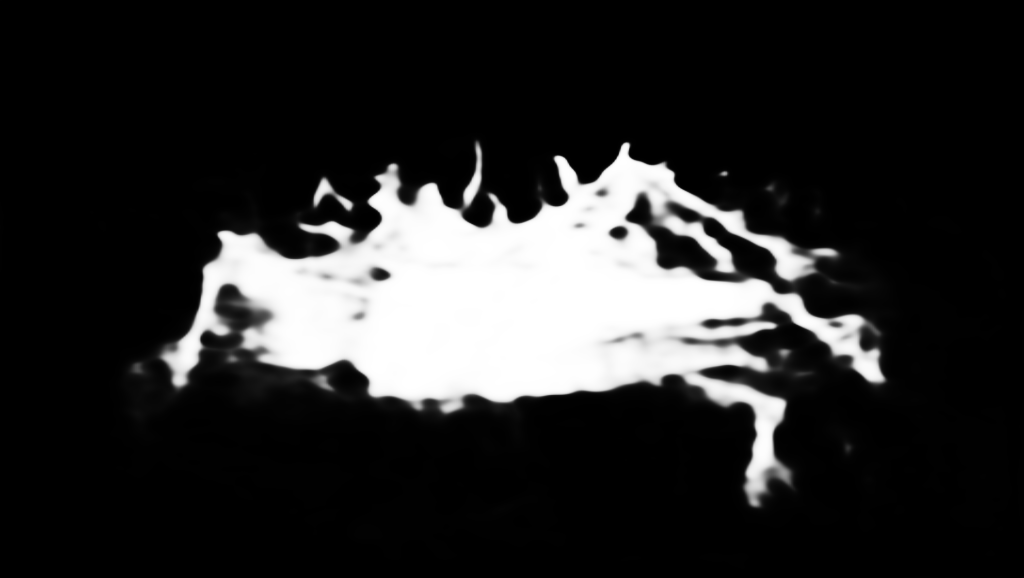}} &
   {\includegraphics[width=0.095\linewidth]{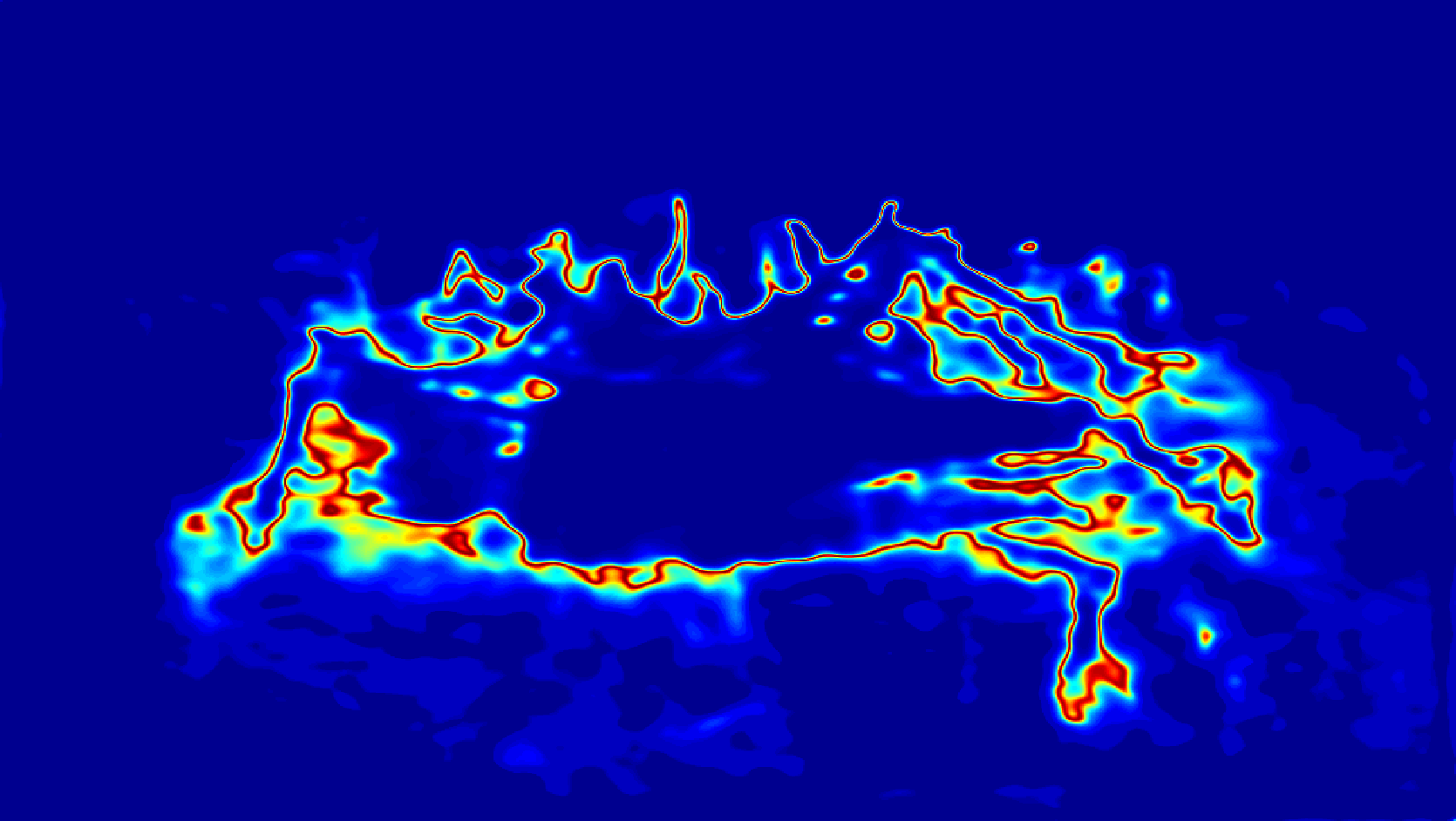}} &
   {\includegraphics[width=0.095\linewidth]{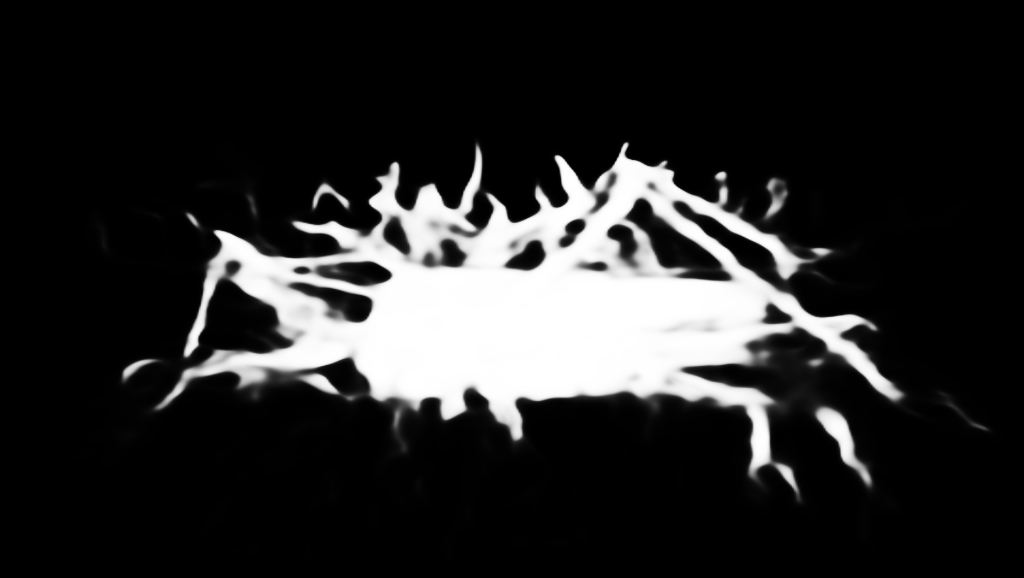}}&
   {\includegraphics[width=0.095\linewidth]{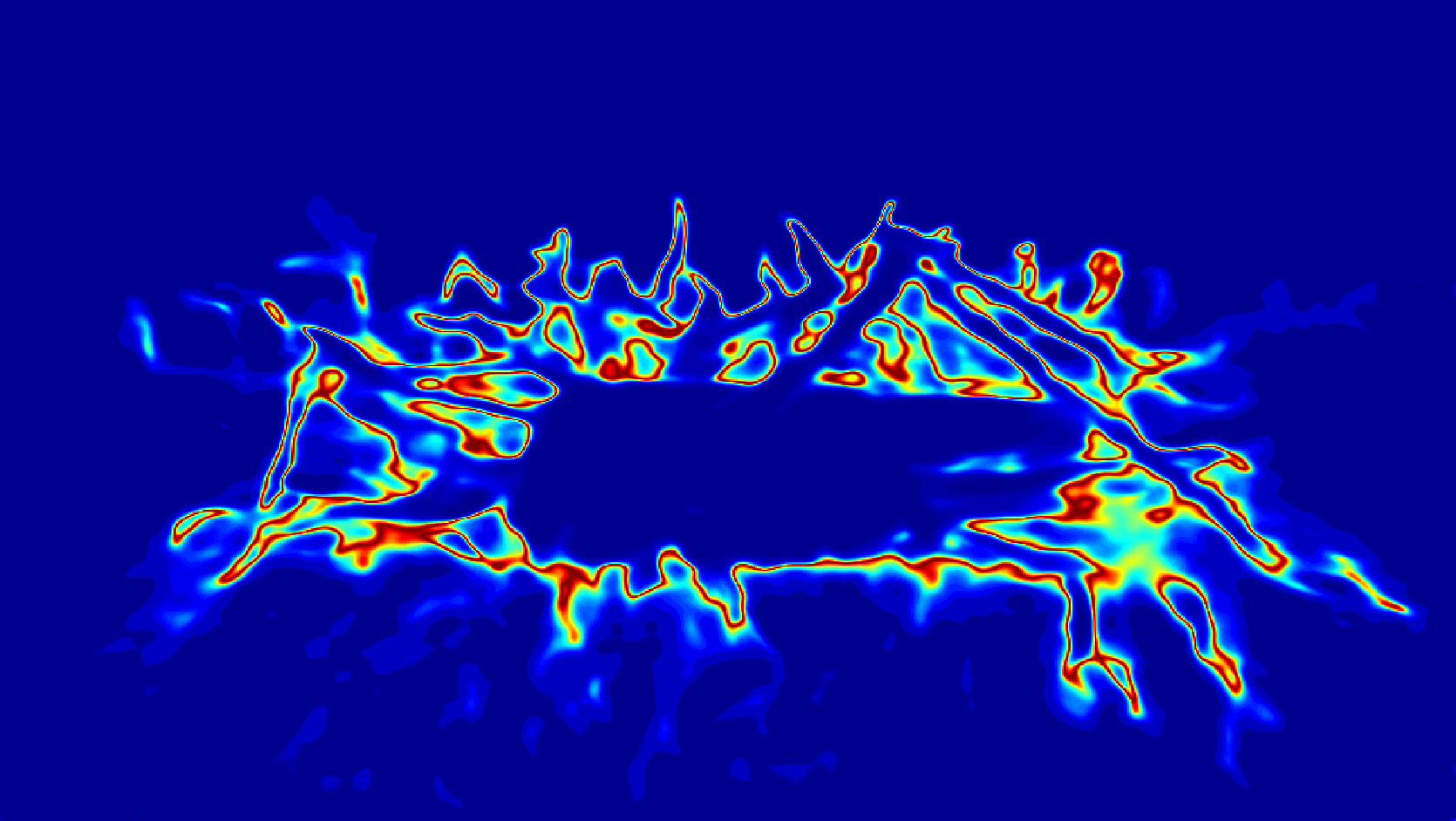}}&
   {\includegraphics[width=0.095\linewidth]{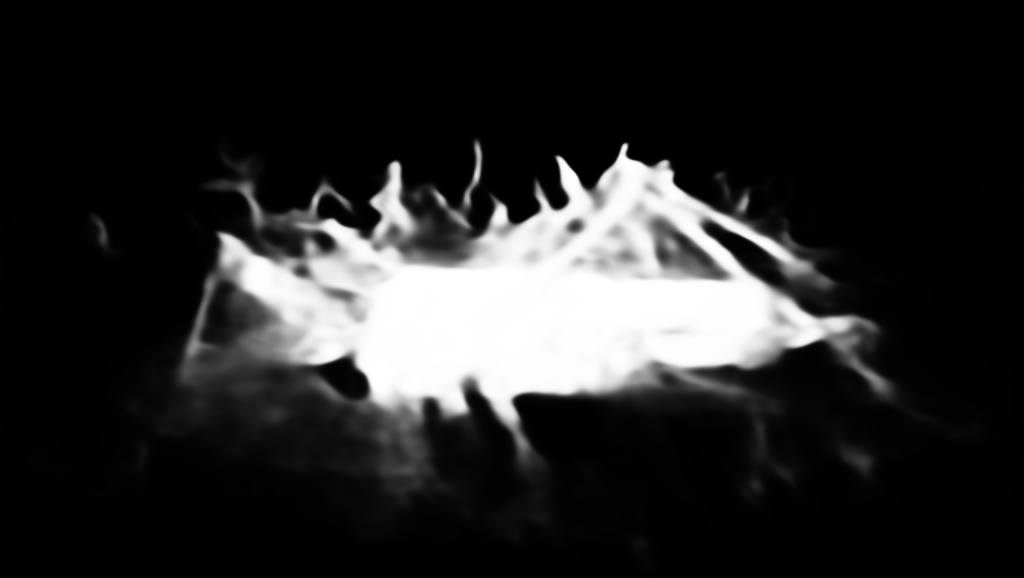}}&
   {\includegraphics[width=0.095\linewidth]{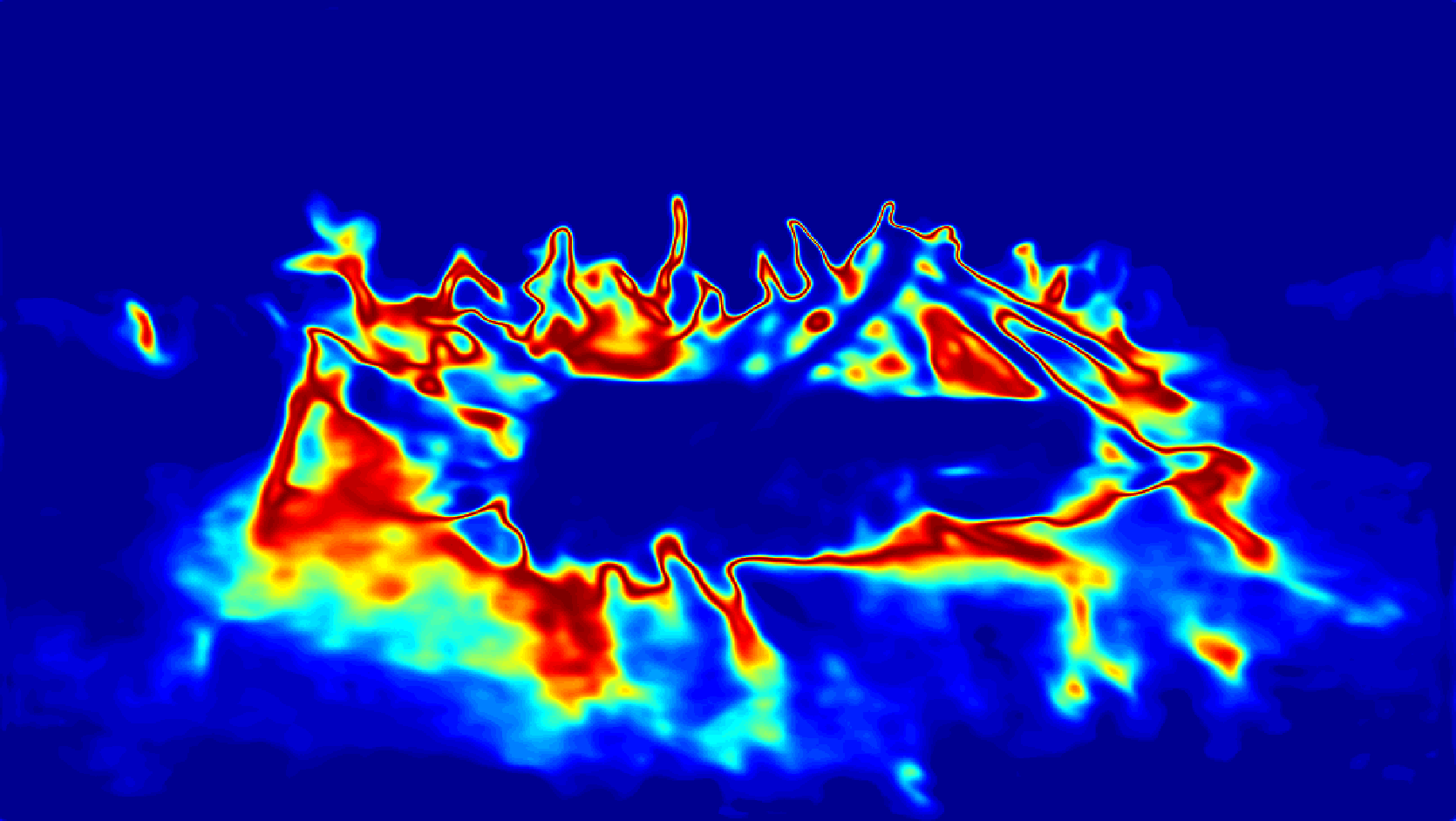}}&
   {\includegraphics[width=0.095\linewidth]{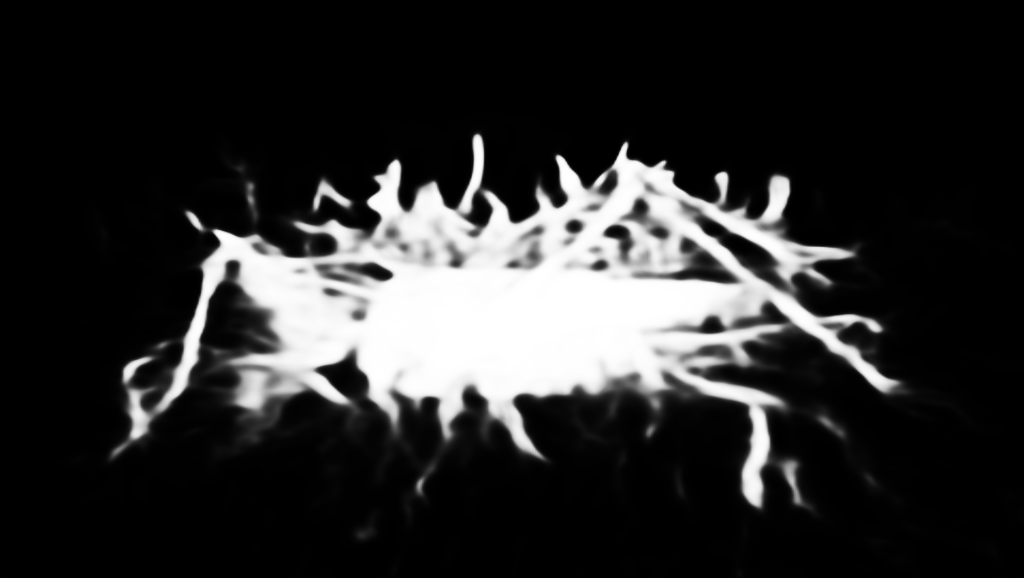}}&
   {\includegraphics[width=0.095\linewidth]{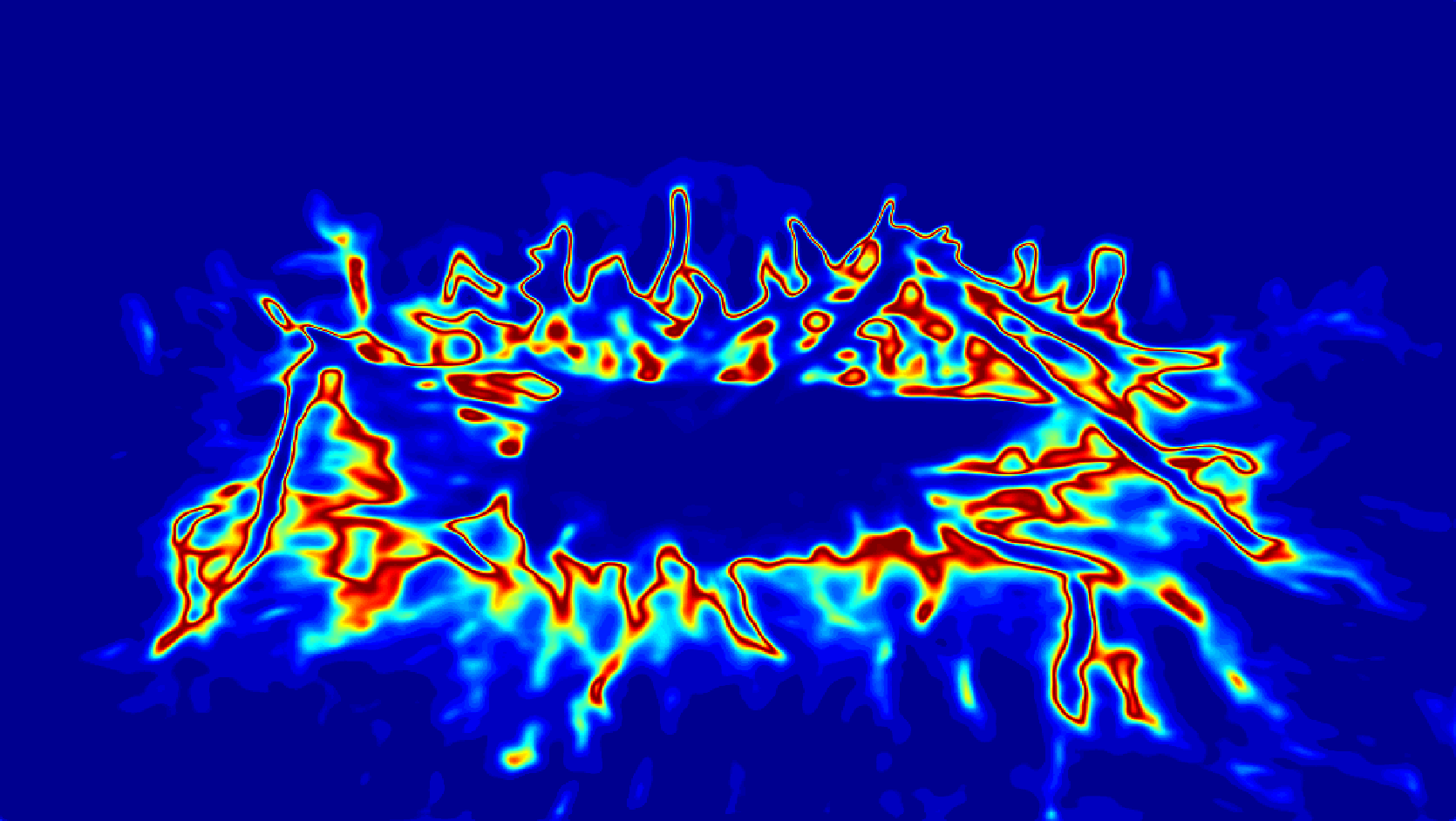}} \\
   \footnotesize{Image}&\footnotesize{GT}&\footnotesize{CVAE}&\footnotesize{$U_a$}&\footnotesize{CGAN}&\footnotesize{$U_a$}&\footnotesize{ABP}&\footnotesize{$U_a$}&\footnotesize{EBM}&\footnotesize{$U_a$}\\
   \end{tabular}
   \end{center}
   \caption{\footnotesize{Aleatoric uncertainty of generative model based solutions for \textbf{camouflaged object detection}.}
   }
\label{fig:aleatoric_generative_cod}
\end{figure*}

\begin{figure*}[tp]
   \begin{center}
   \begin{tabular}{c@{ }c@{ }c@{ }c@{ }c@{ }c@{ }c@{ }c@{ }c@{ }c@{ }}
   {\includegraphics[width=0.095\linewidth]{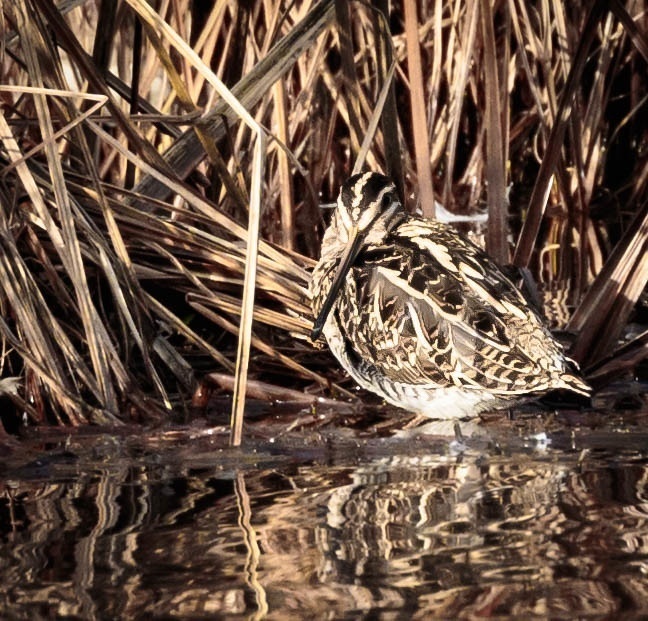}} &
   {\includegraphics[width=0.095\linewidth]{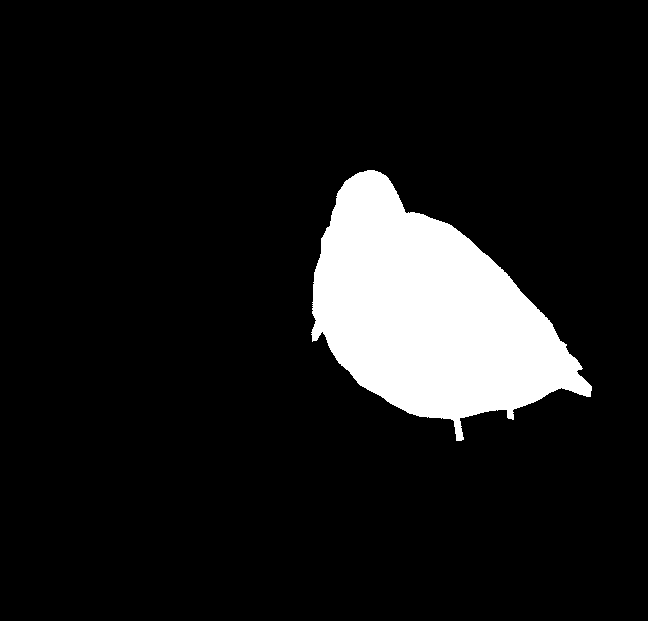}} &
   {\includegraphics[width=0.095\linewidth]{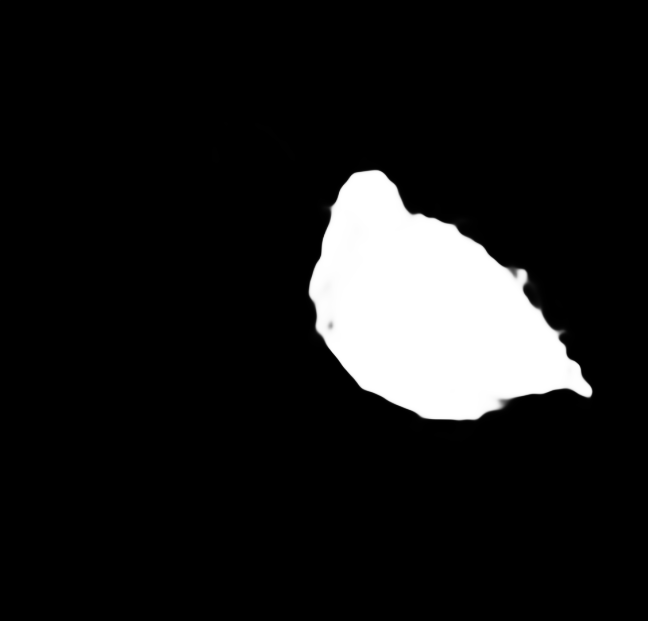}} &
   {\includegraphics[width=0.095\linewidth]{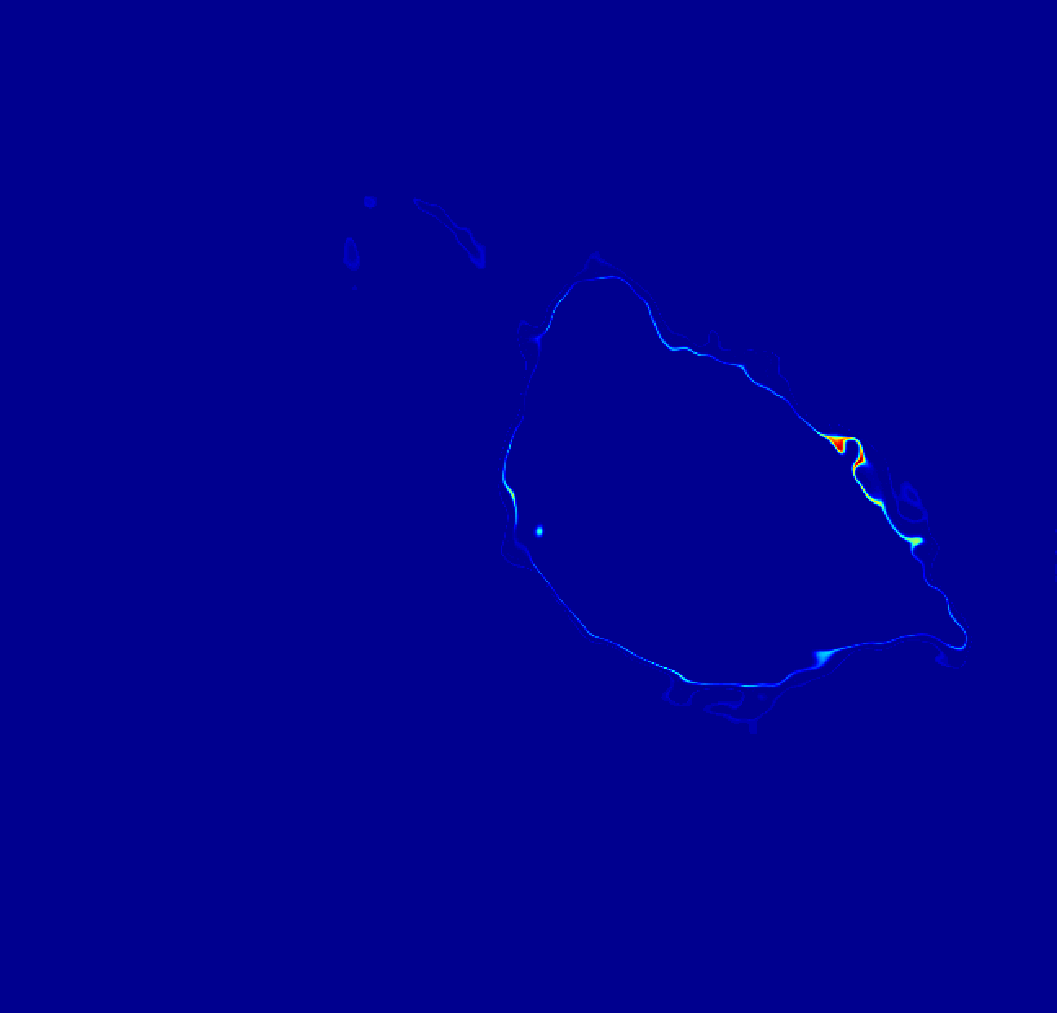}} &
   {\includegraphics[width=0.095\linewidth]{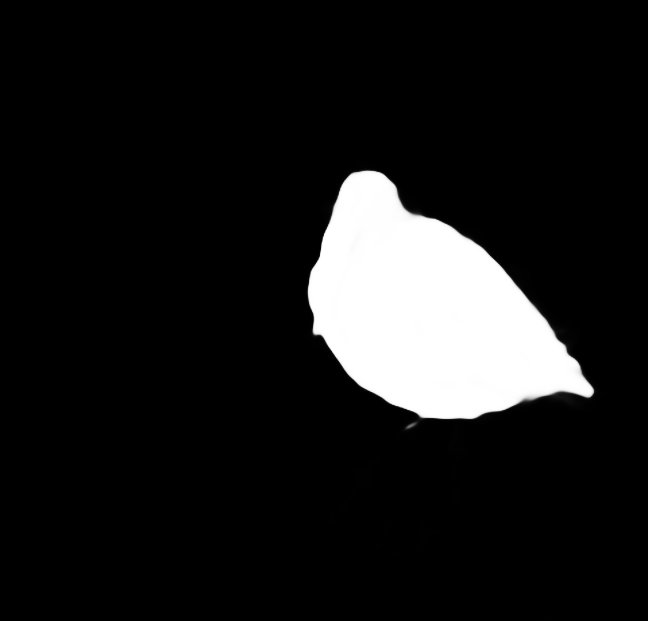}}&
   {\includegraphics[width=0.095\linewidth]{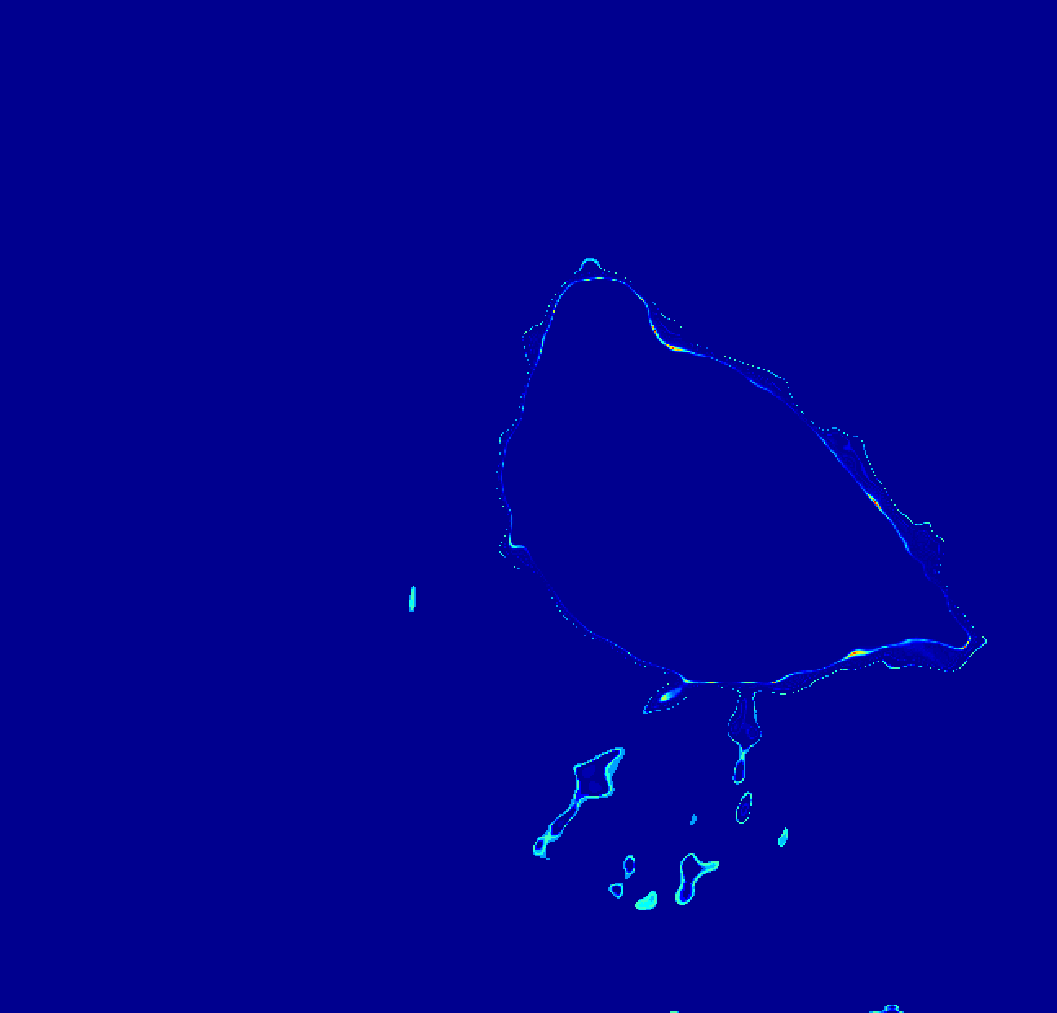}}&
   {\includegraphics[width=0.095\linewidth]{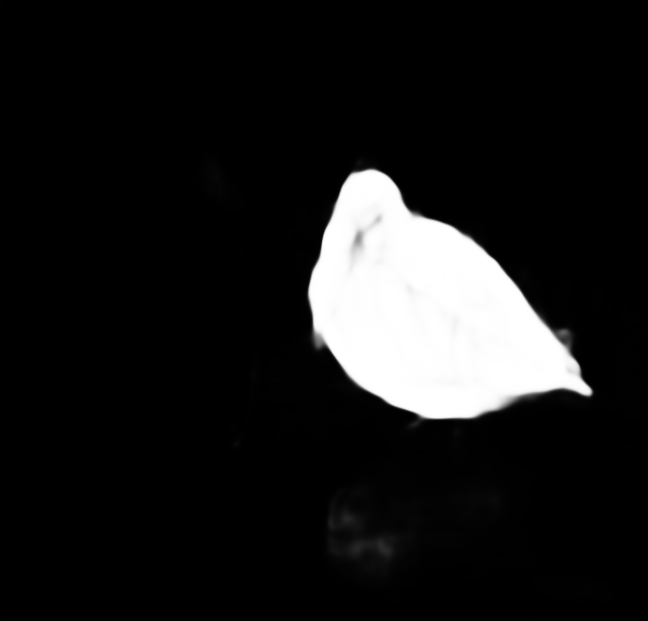}}&
   {\includegraphics[width=0.095\linewidth]{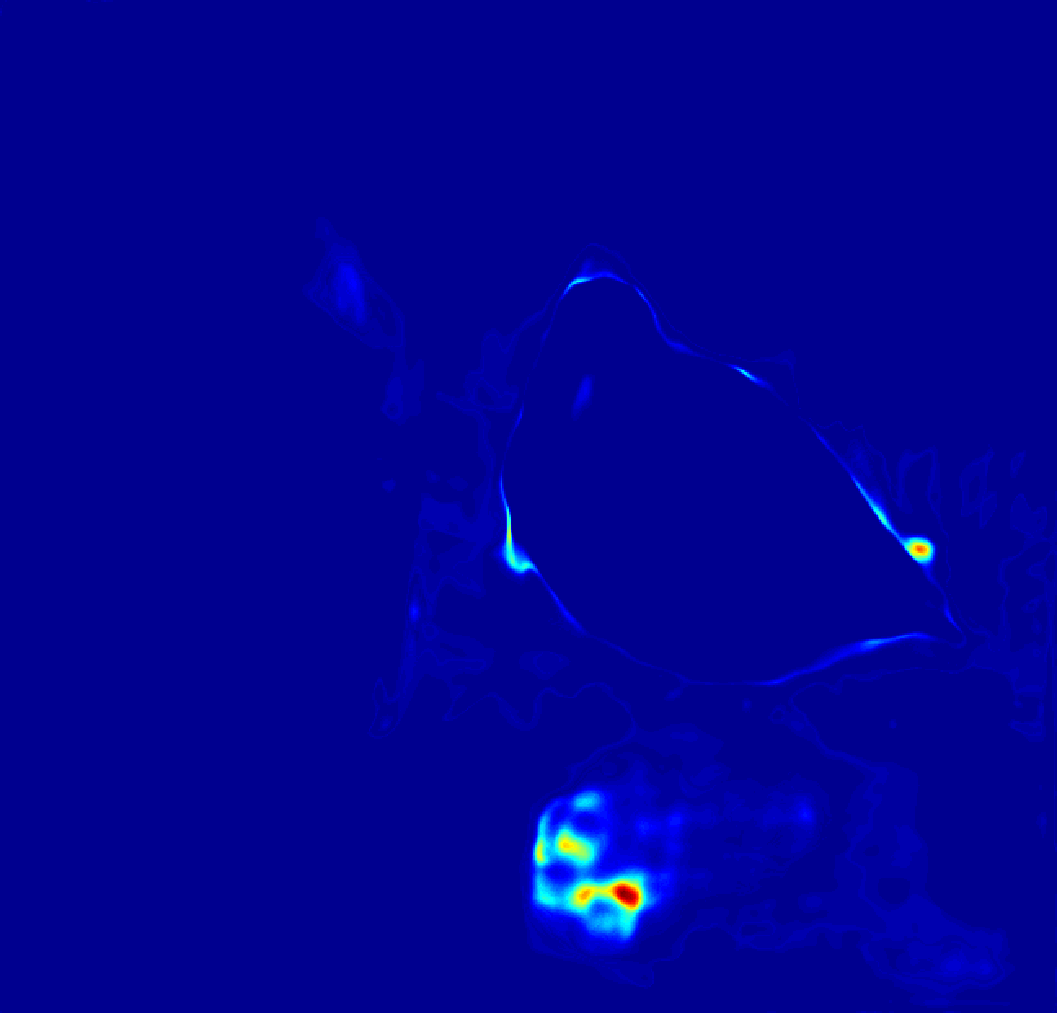}}&
   {\includegraphics[width=0.095\linewidth]{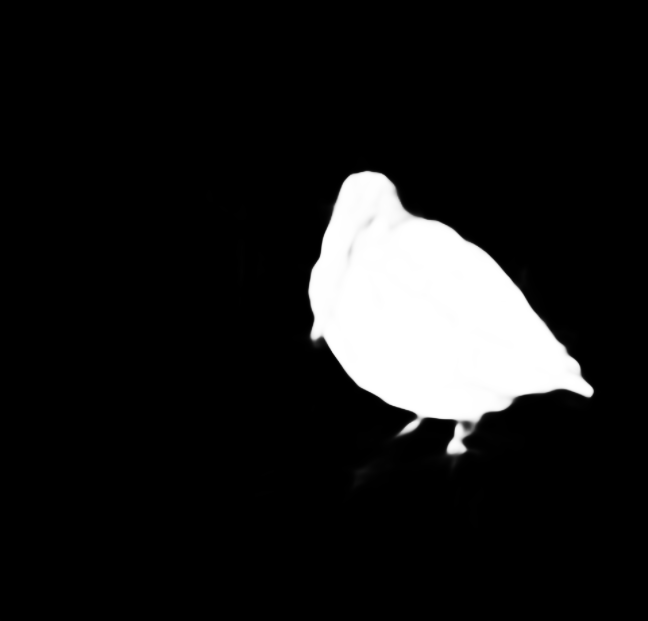}}&
   {\includegraphics[width=0.095\linewidth]{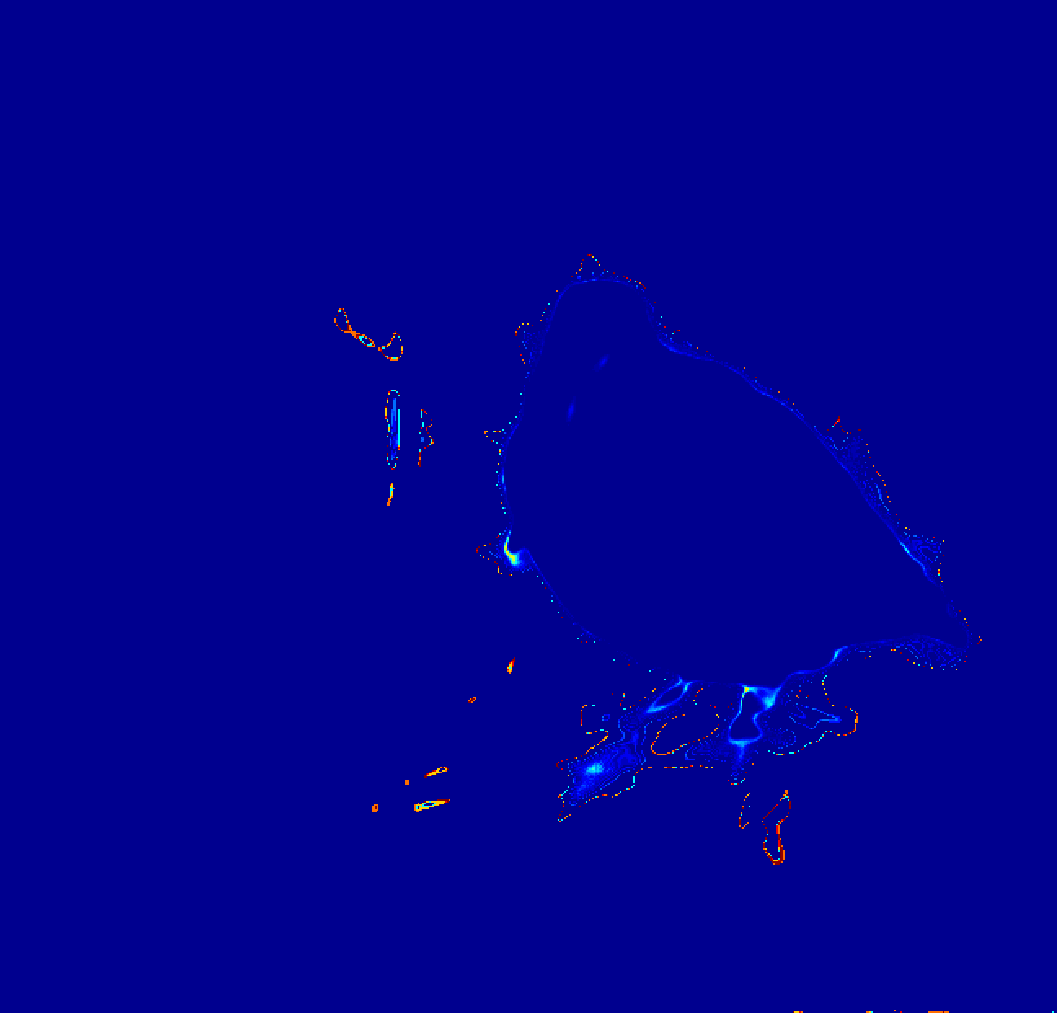}} \\
   {\includegraphics[width=0.095\linewidth]{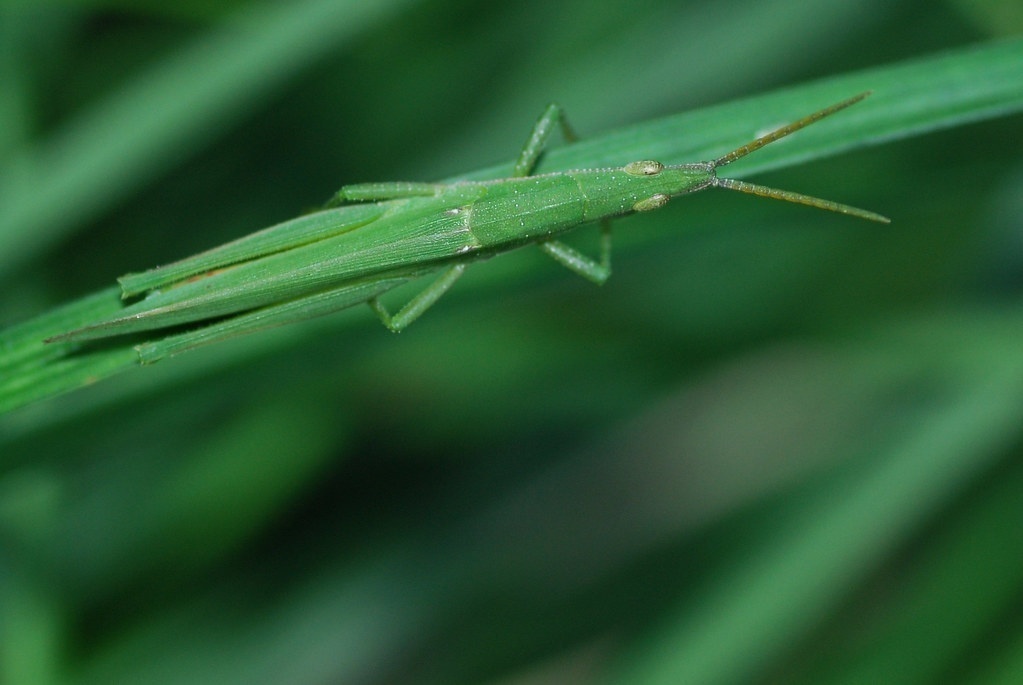}} &
   {\includegraphics[width=0.095\linewidth]{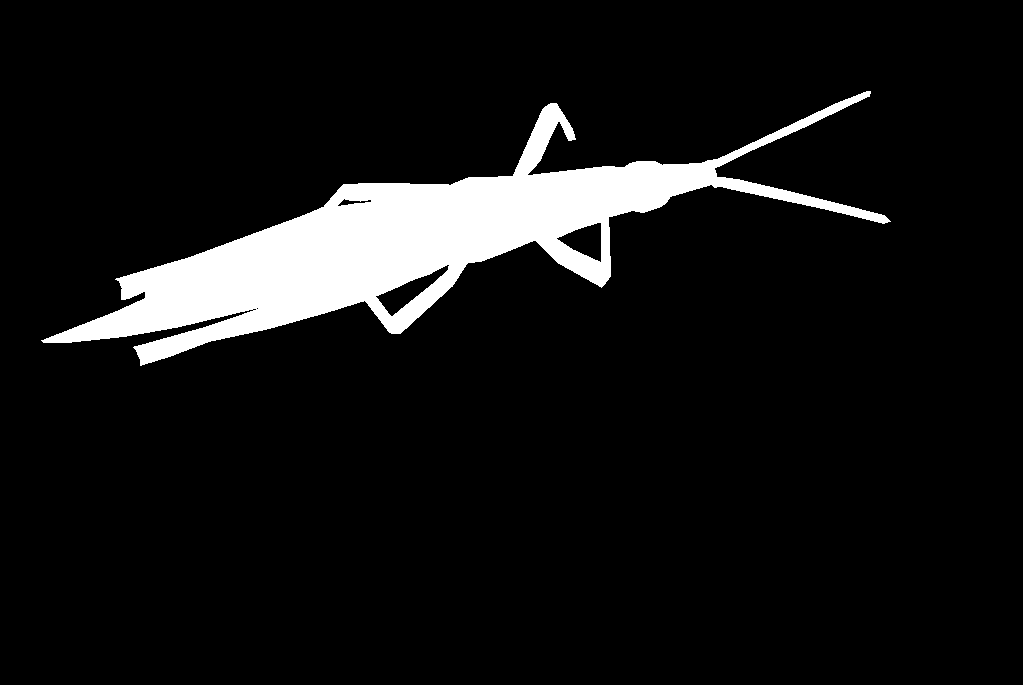}} &
   {\includegraphics[width=0.095\linewidth]{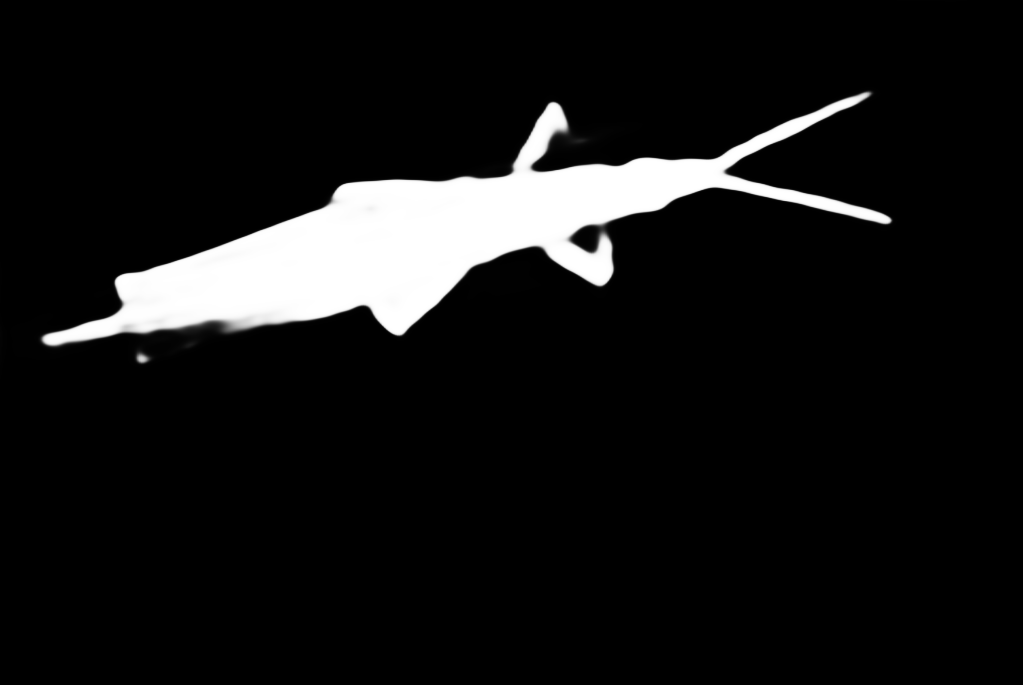}} &
   {\includegraphics[width=0.095\linewidth]{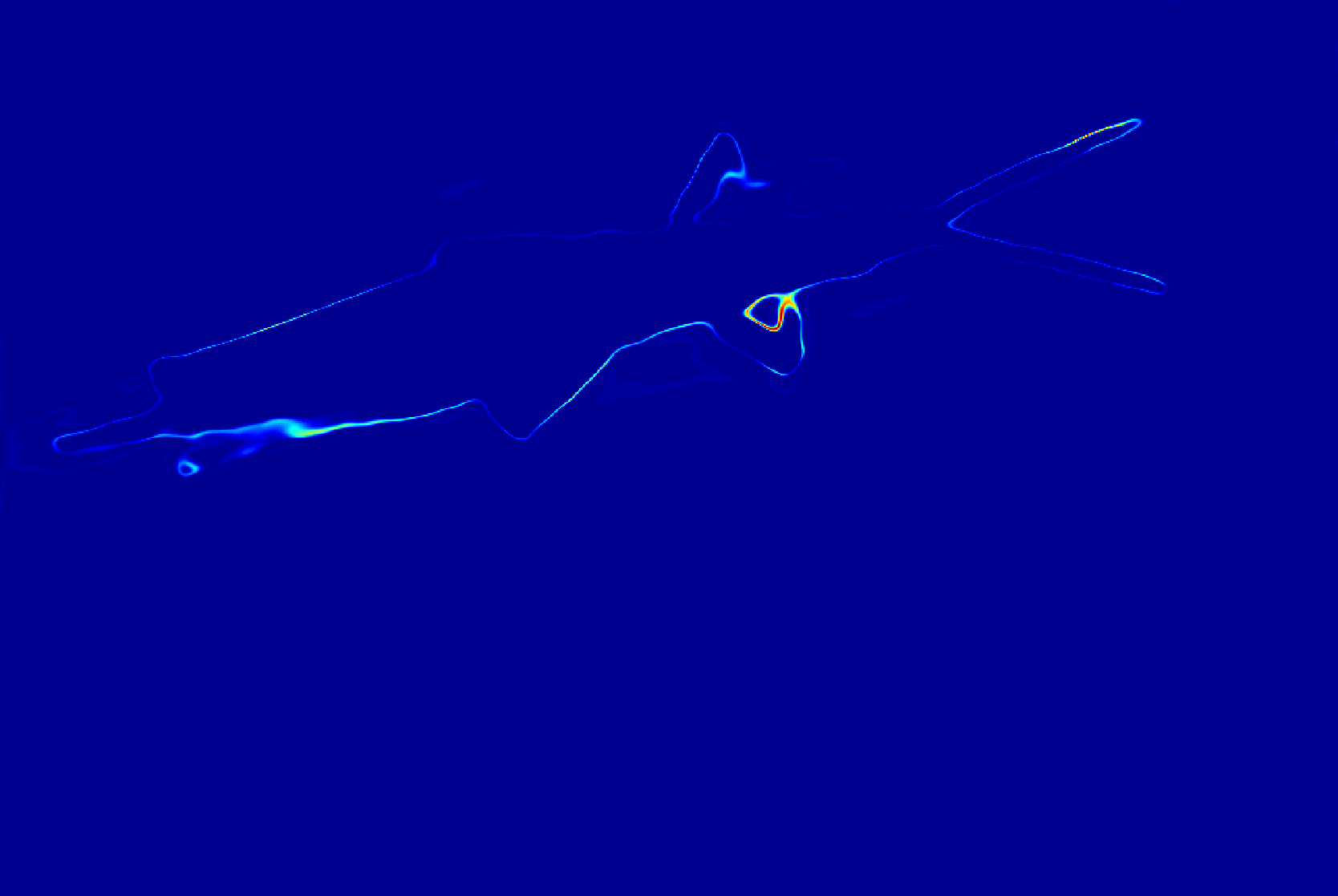}} &
   {\includegraphics[width=0.095\linewidth]{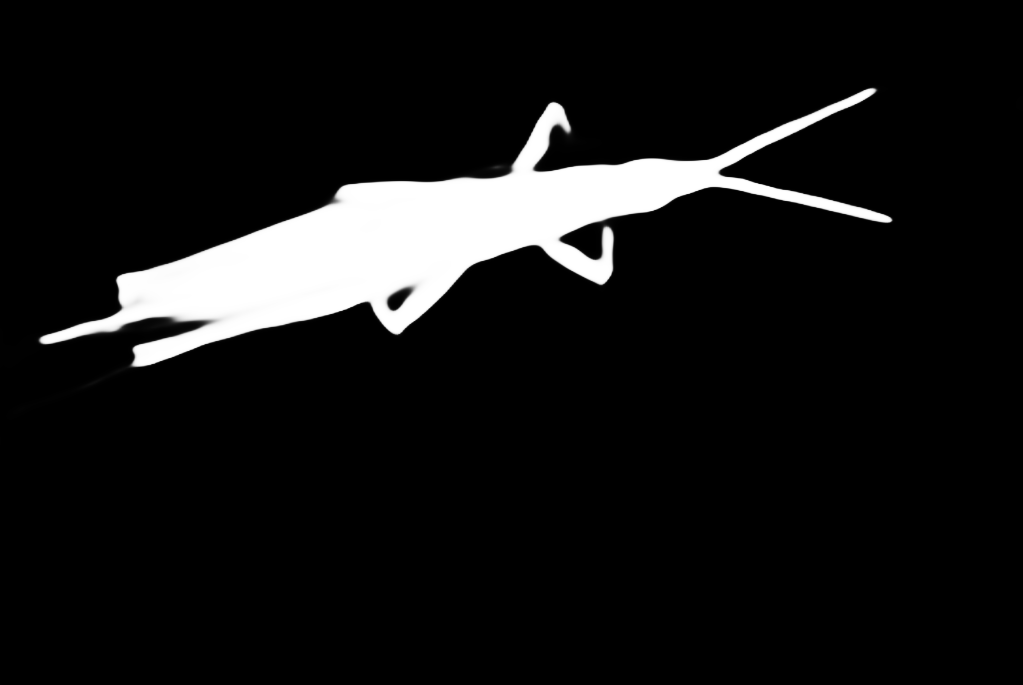}}&
   {\includegraphics[width=0.095\linewidth]{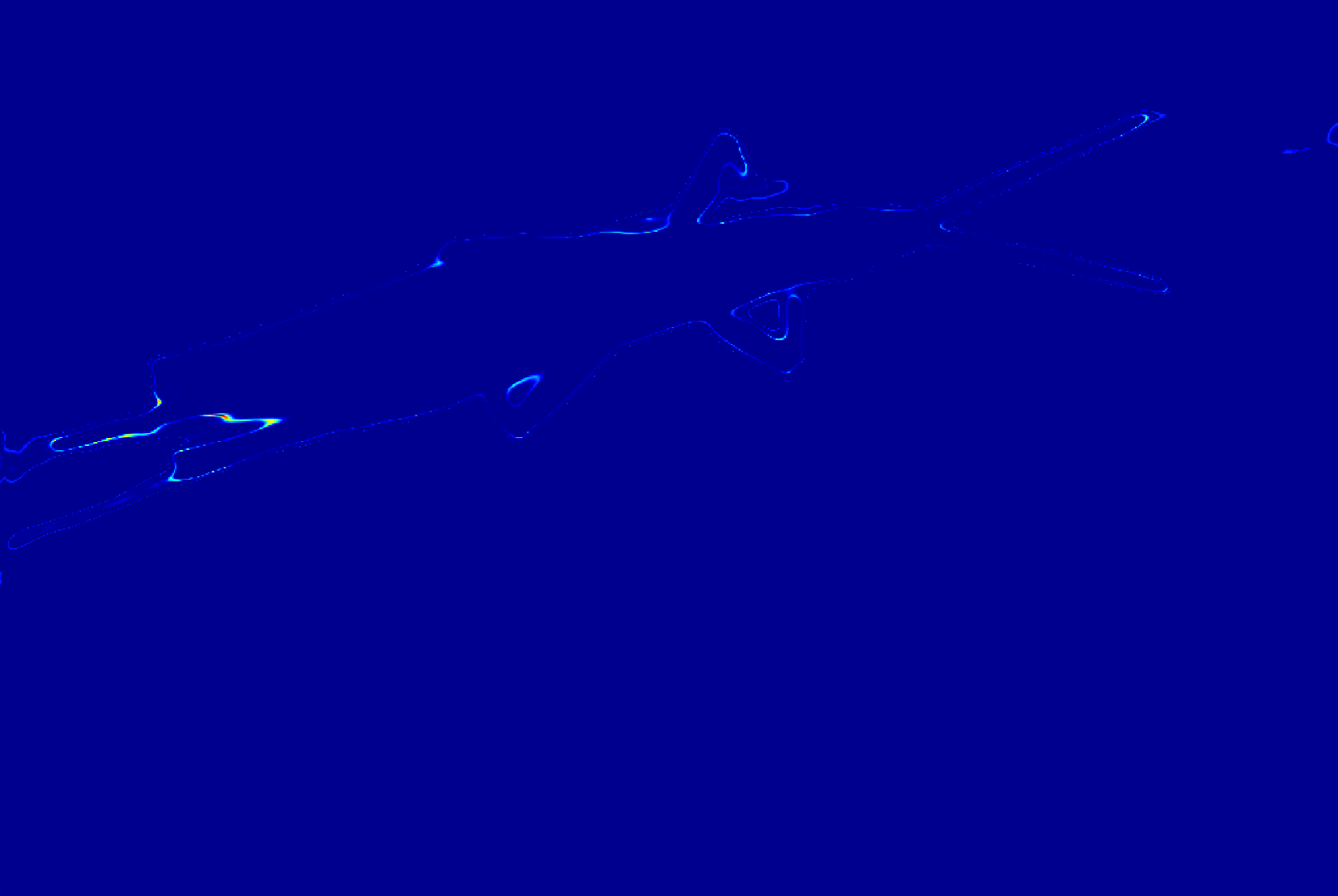}}&
   {\includegraphics[width=0.095\linewidth]{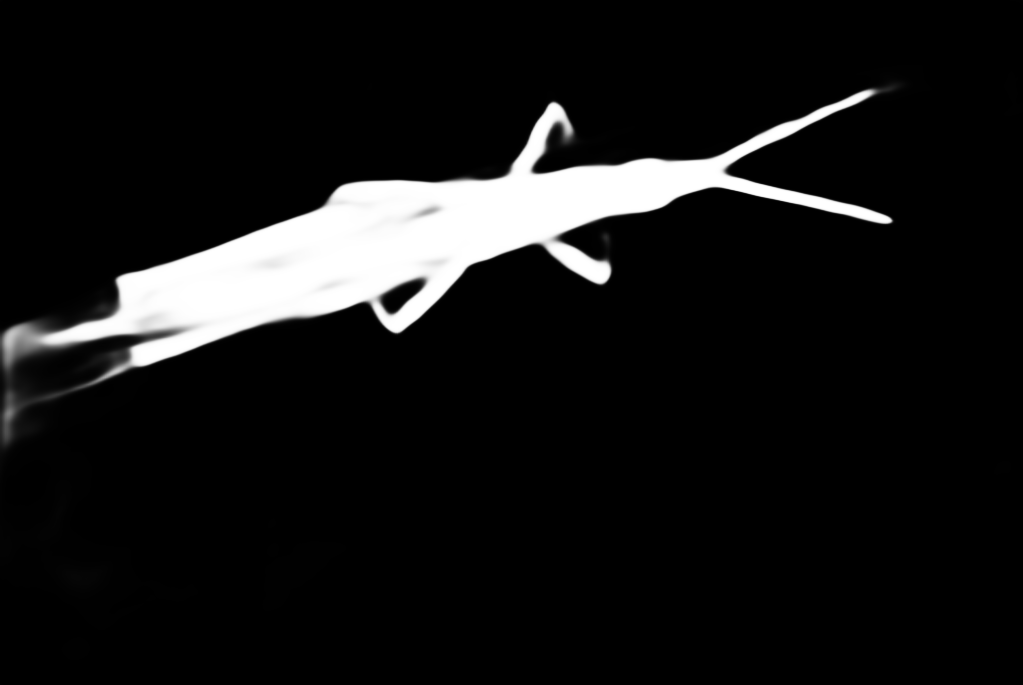}}&
   {\includegraphics[width=0.095\linewidth]{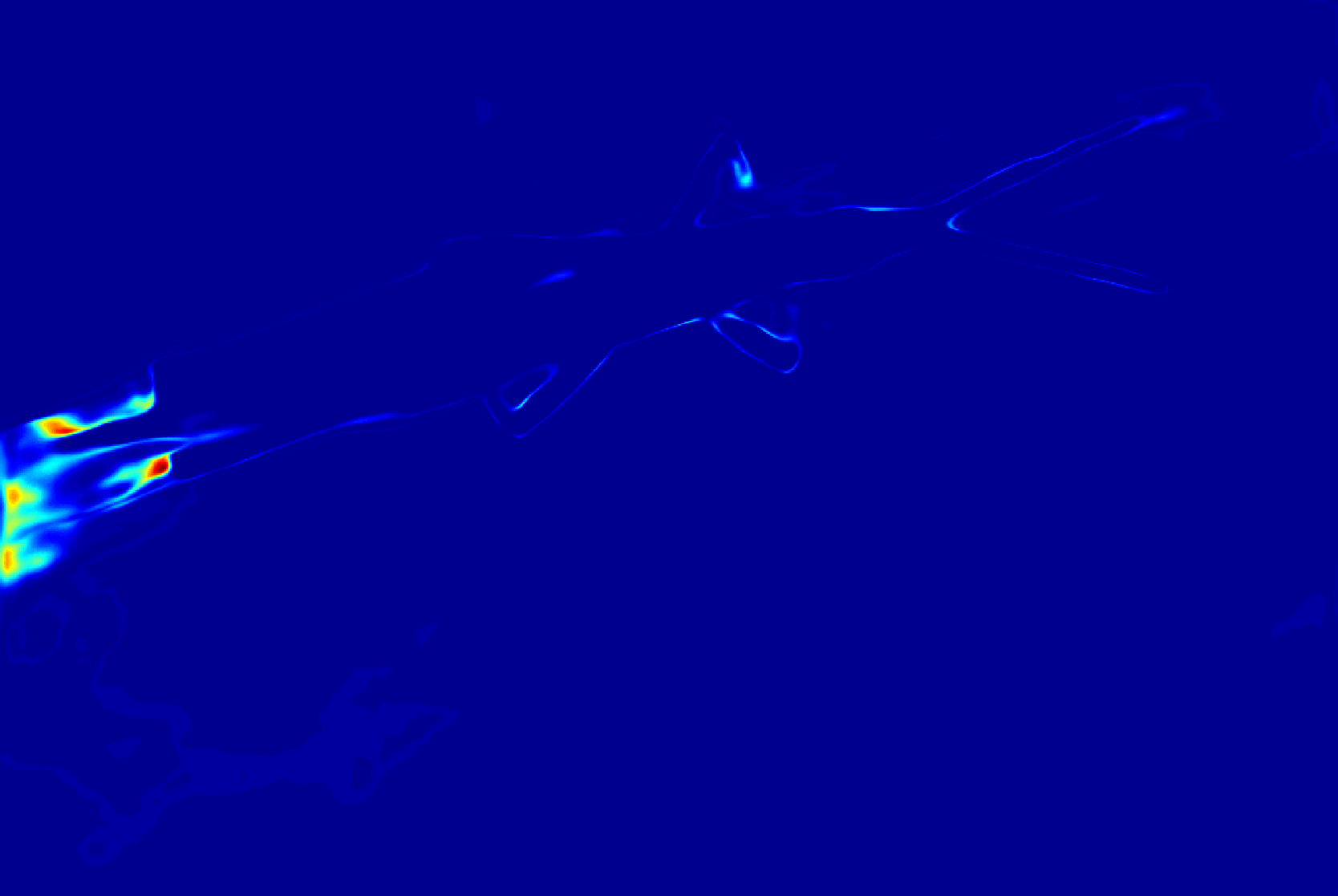}}&
   {\includegraphics[width=0.095\linewidth]{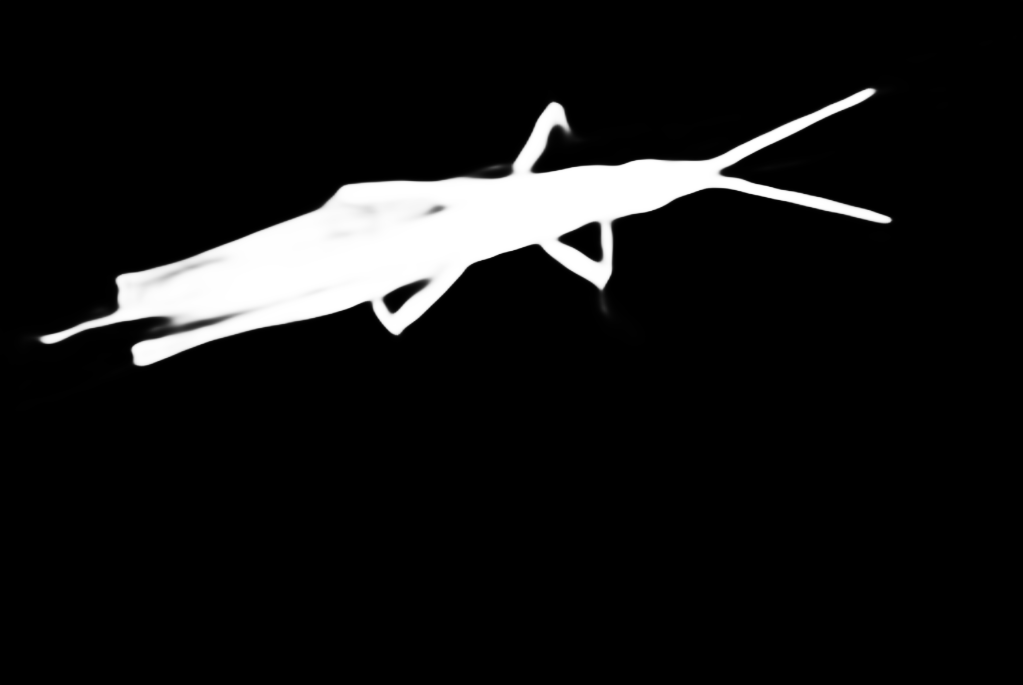}}&
   {\includegraphics[width=0.095\linewidth]{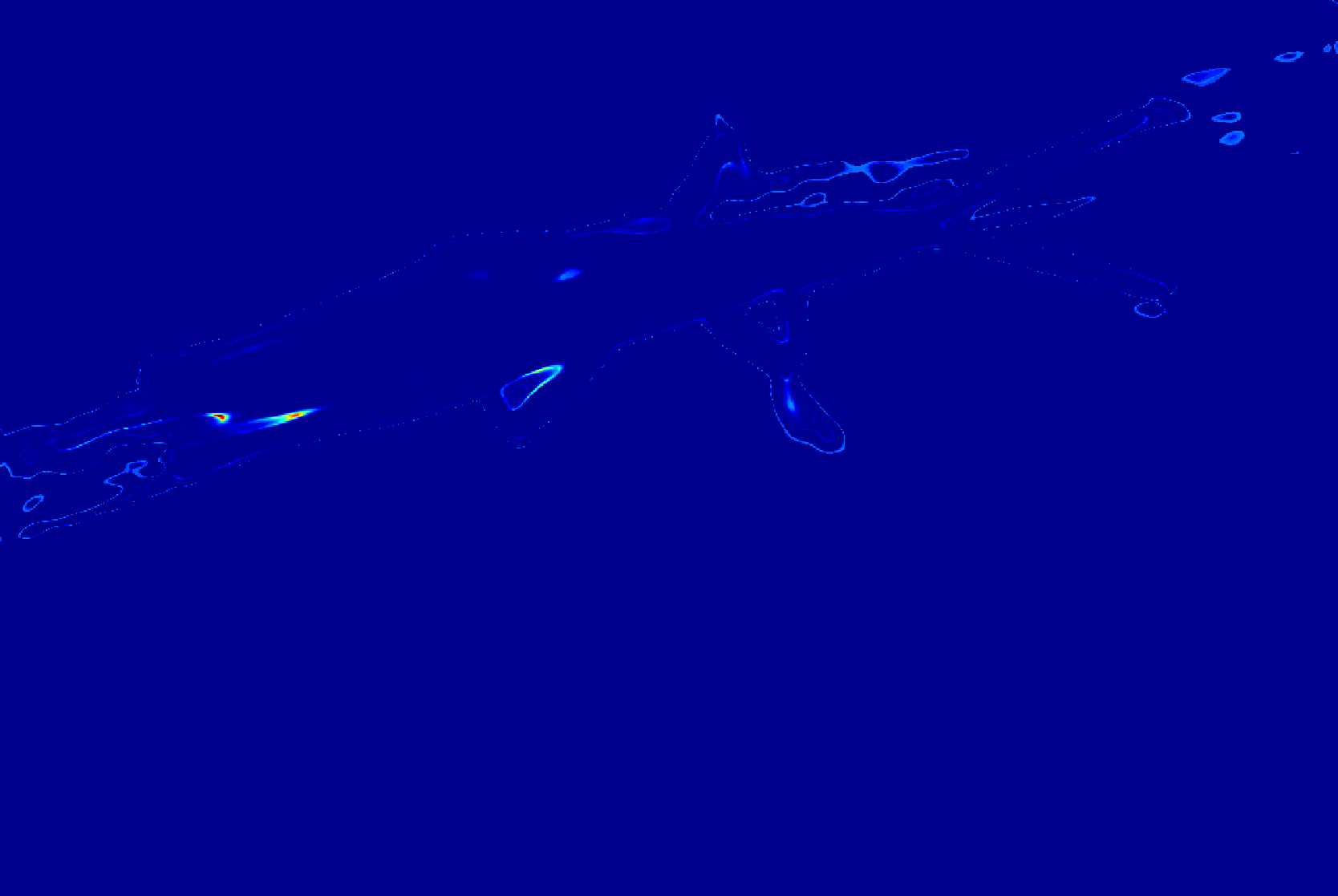}} \\
   {\includegraphics[width=0.095\linewidth]{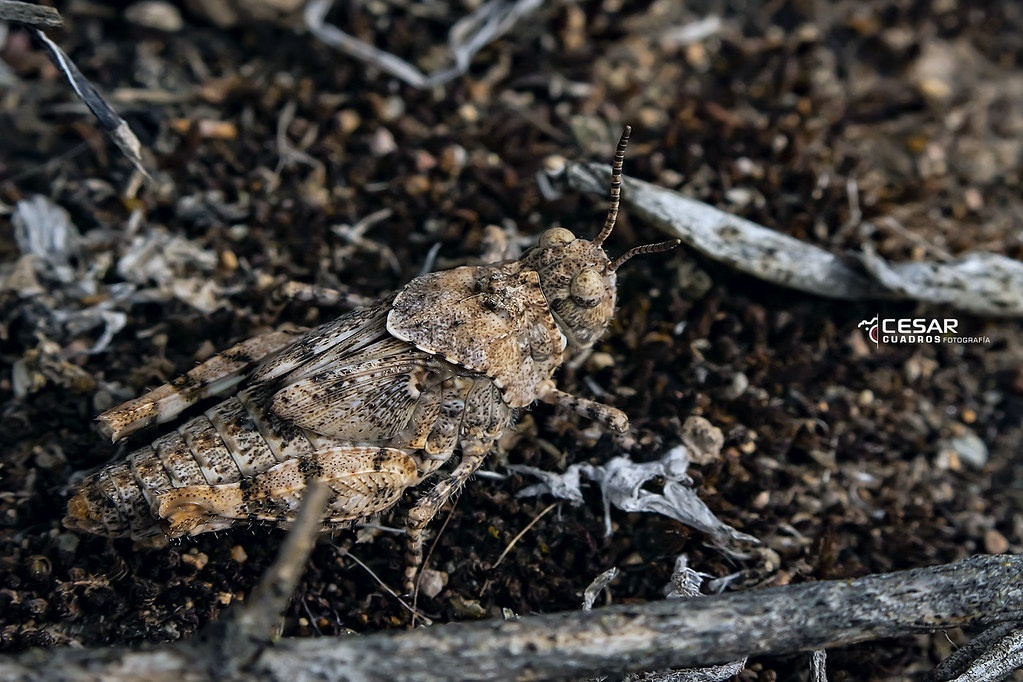}} &
   {\includegraphics[width=0.095\linewidth]{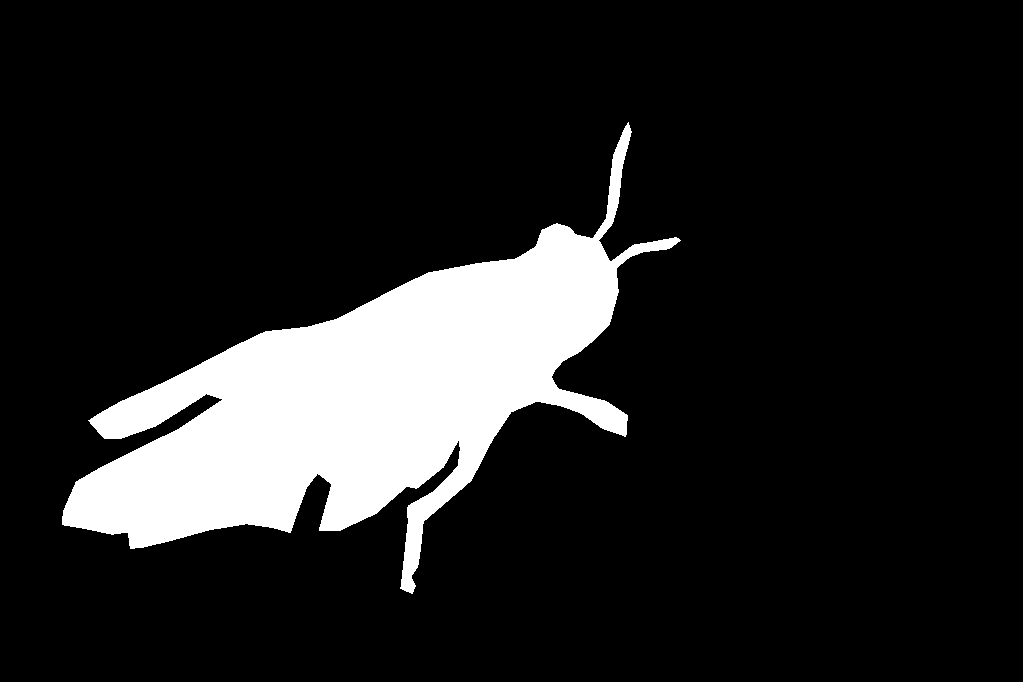}} &
   {\includegraphics[width=0.095\linewidth]{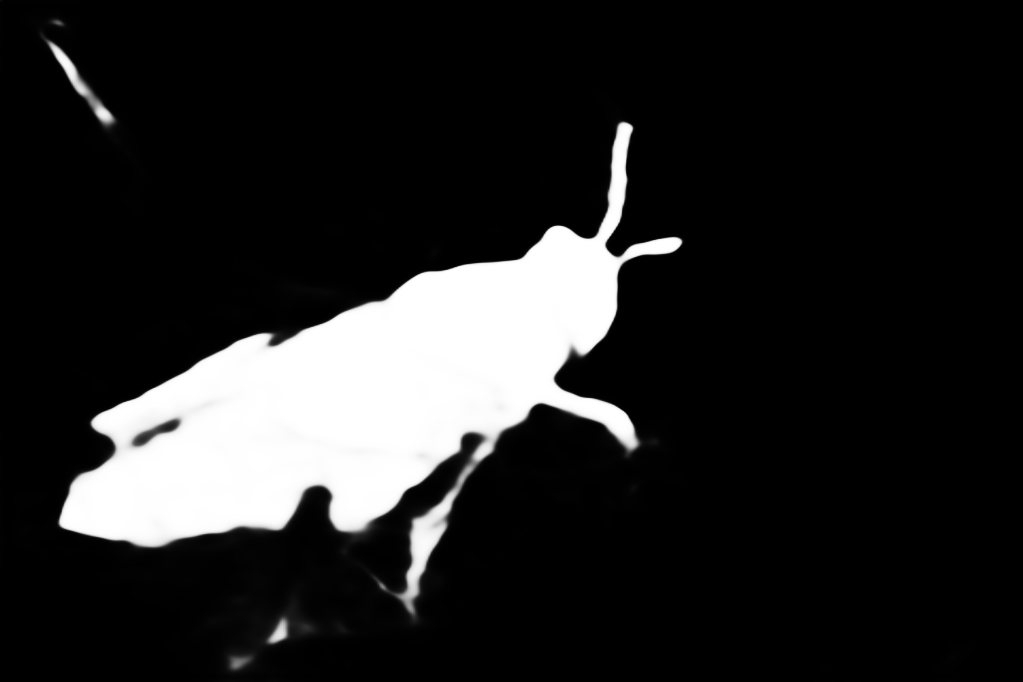}} &
   {\includegraphics[width=0.095\linewidth]{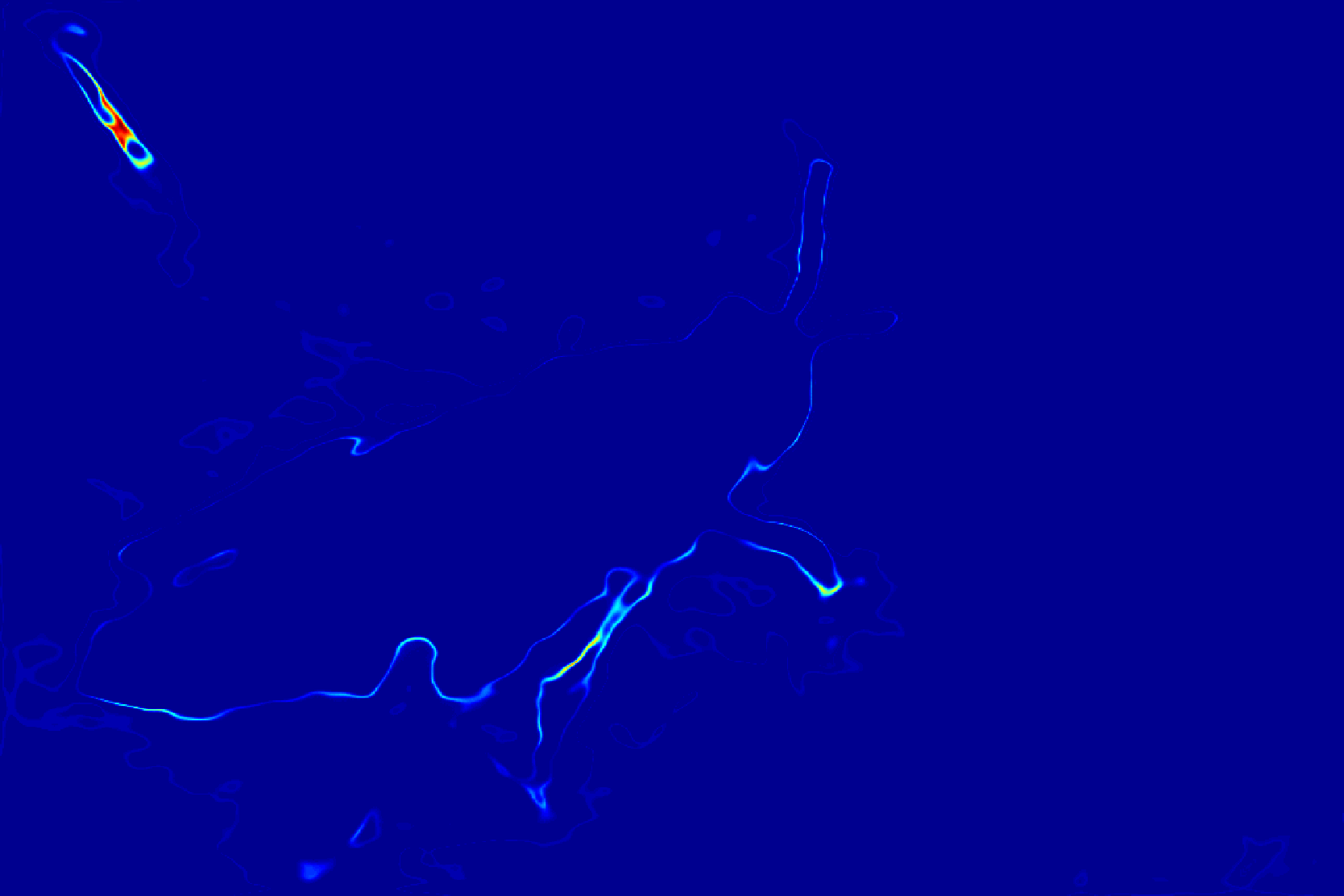}} &
   {\includegraphics[width=0.095\linewidth]{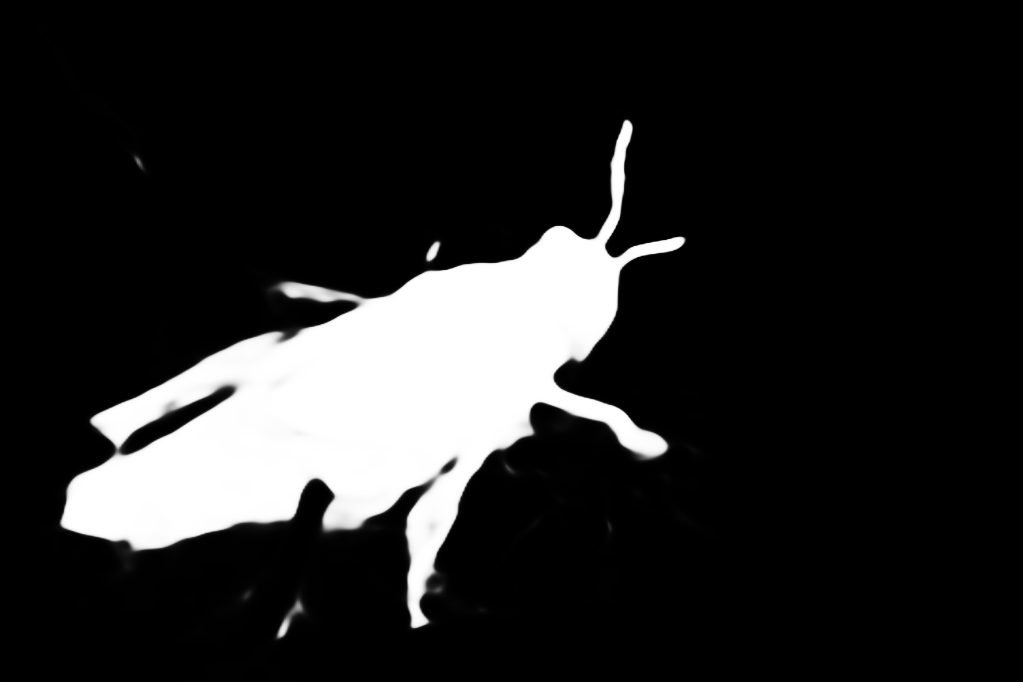}}&
   {\includegraphics[width=0.095\linewidth]{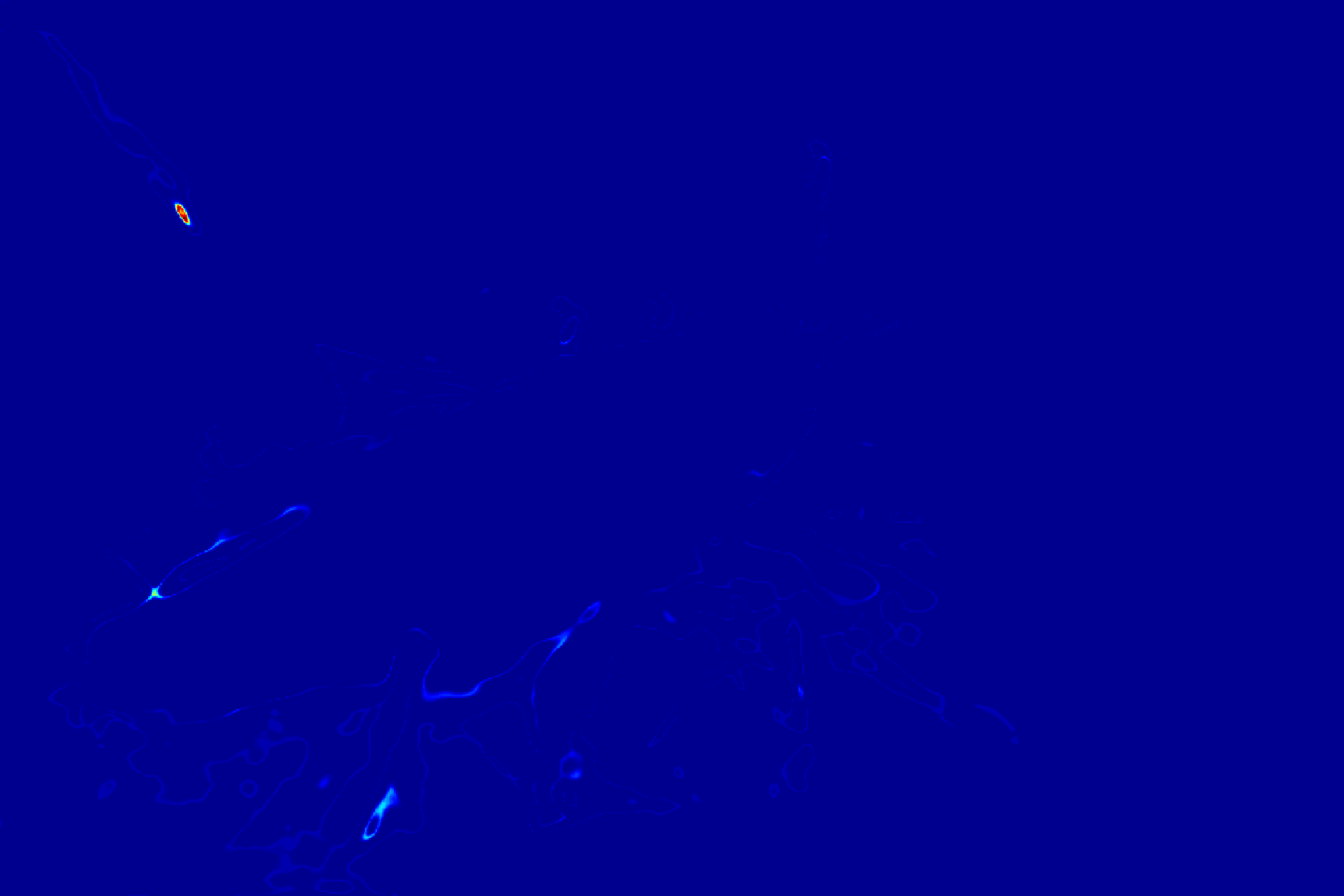}}&
   {\includegraphics[width=0.095\linewidth]{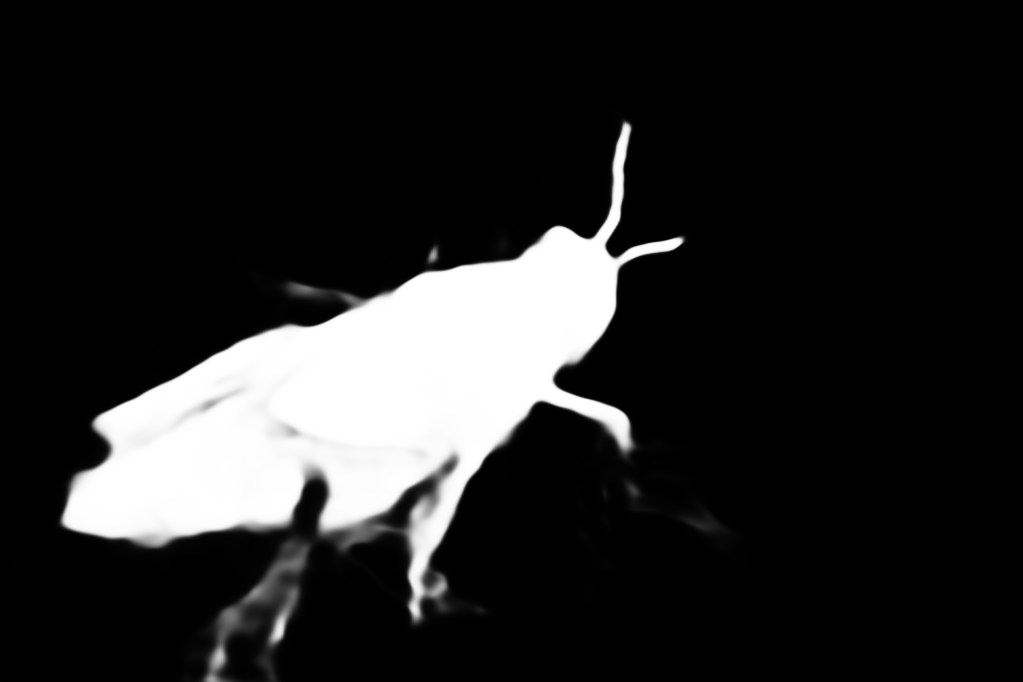}}&
   {\includegraphics[width=0.095\linewidth]{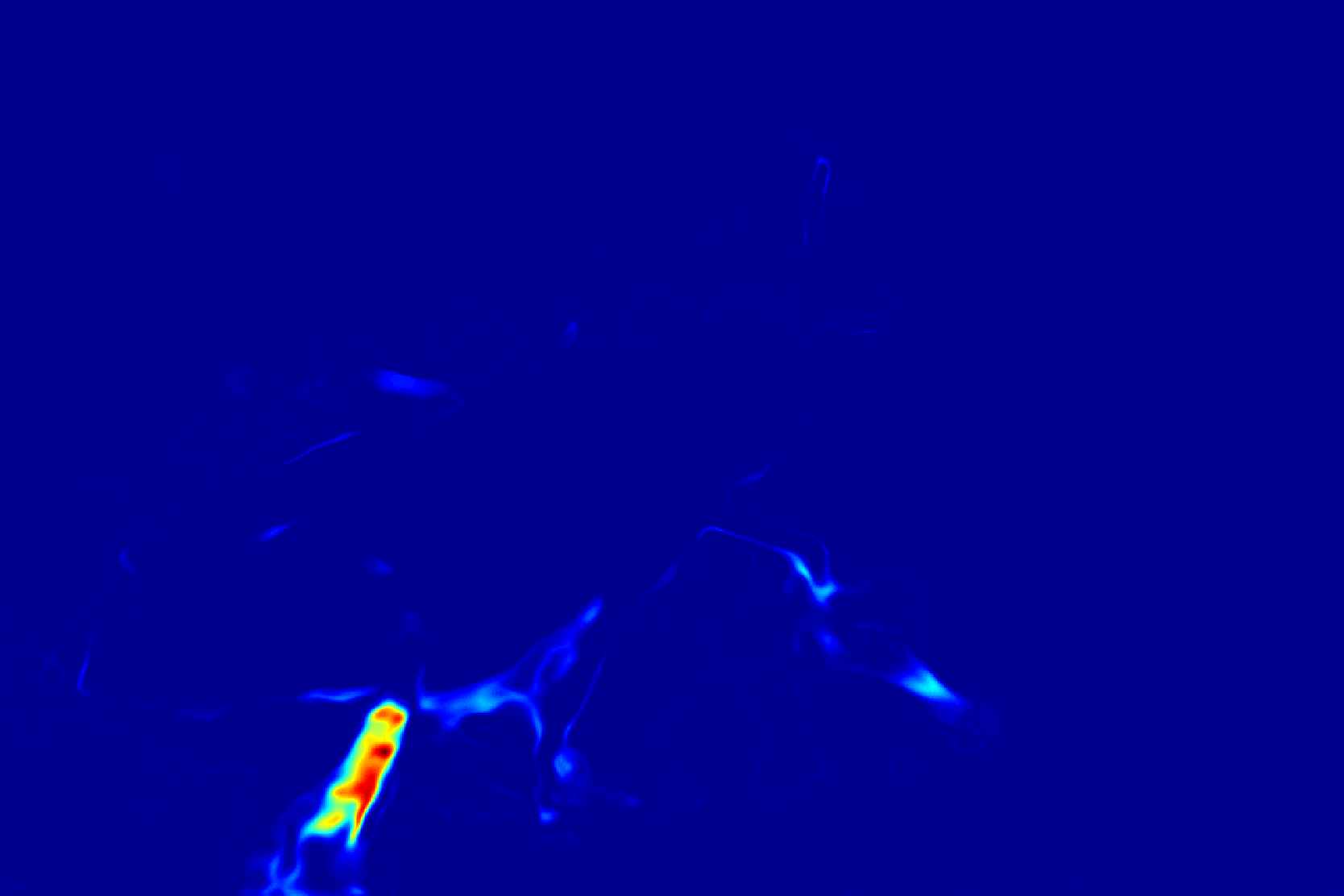}}&
   {\includegraphics[width=0.095\linewidth]{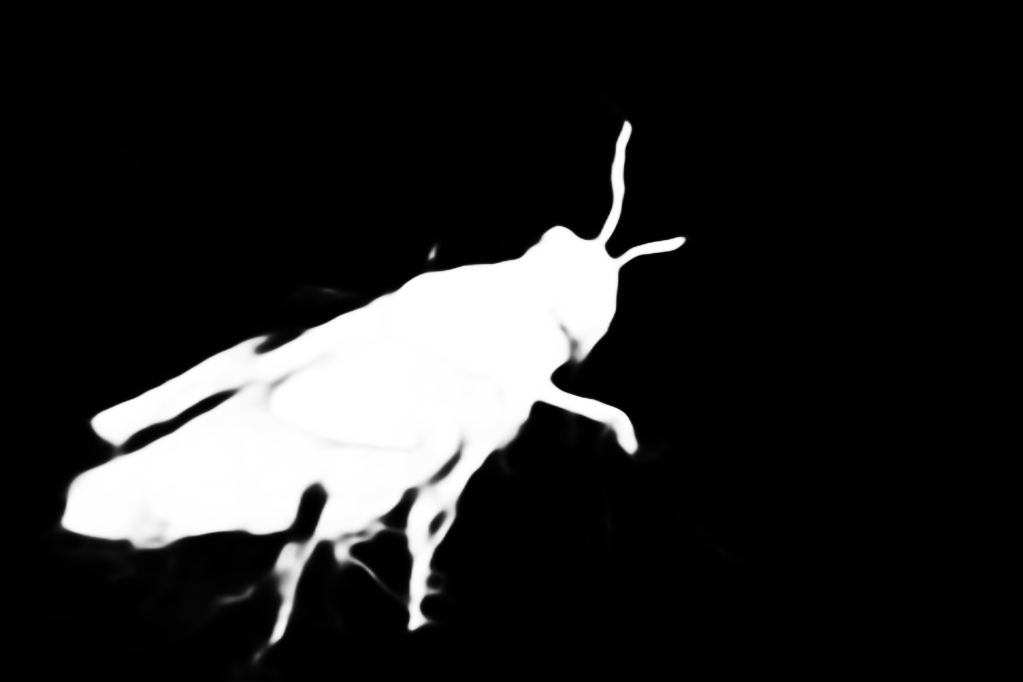}}&
   {\includegraphics[width=0.095\linewidth]{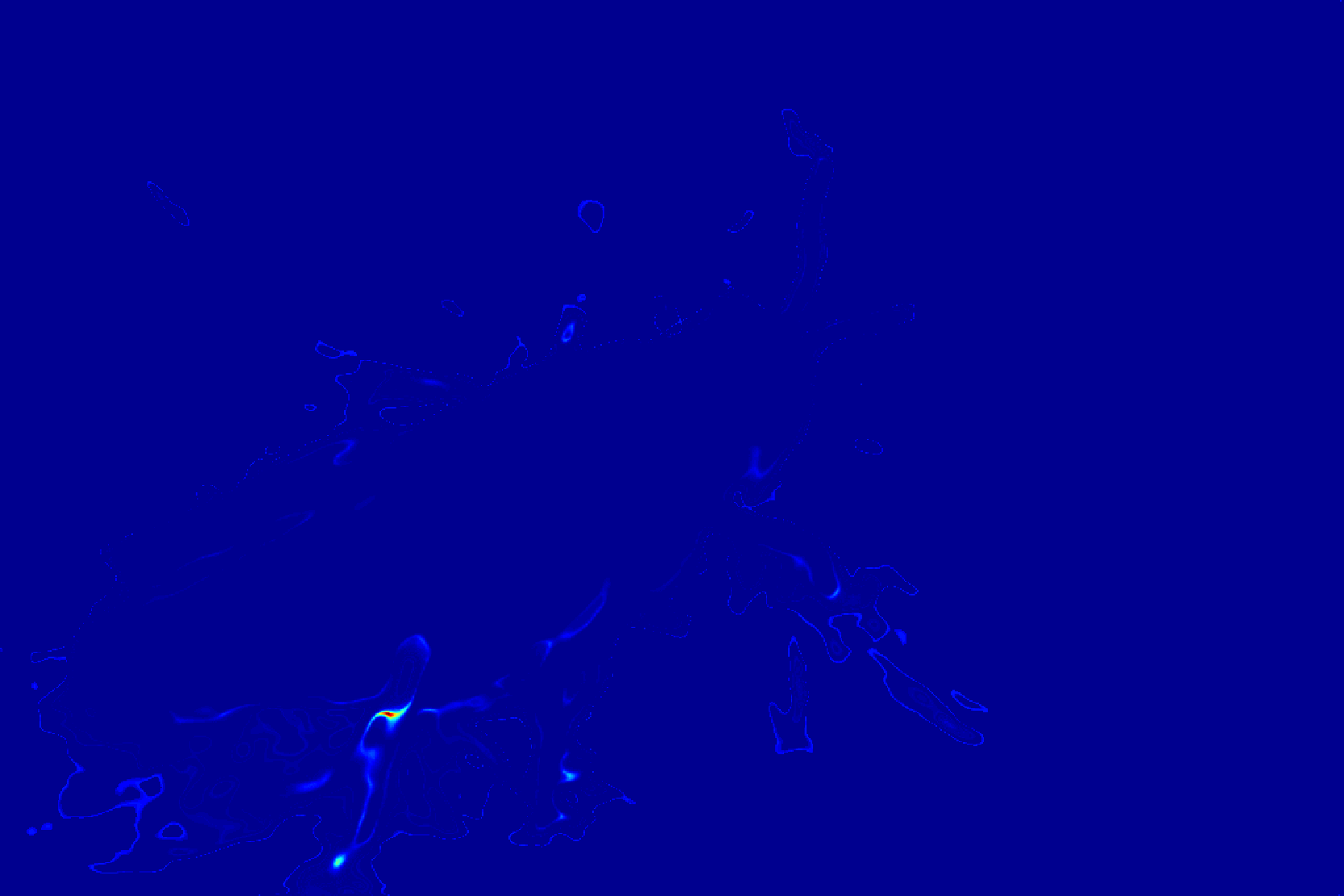}} \\
   \footnotesize{Image}&\footnotesize{GT}&\footnotesize{CVAE}&\footnotesize{$U_e$}&\footnotesize{CGAN}&\footnotesize{$U_e$}&\footnotesize{ABP}&\footnotesize{$U_e$}&\footnotesize{EBM}&\footnotesize{$U_e$}\\
   \end{tabular}
   \end{center}
   \caption{\footnotesize{Epistemic uncertainty of generative model based solutions for \textbf{camouflaged object detection}.}
   }
\label{fig:epistemic_generative_cod}
\end{figure*}

\noindent\textbf{Uncertainty for Monocular Depth Estimation}: Monocular depth estimation aims to estimate depth of a scene from a single RGB image \cite{Make3D_single3D,midas_tpami}. As a distance perception strategy, the accuracy and model's confidence in its prediction is important for real-life applications, \ie~it is essential for an autonomous driving system to understand when it makes a mistake \cite{Kendall_2018_CVPR,Poggi_2020_CVPR,bayesian_depth_completion,bayesian_single_view_depth,depth_predictive_uncertainty_icra21,adaptive_confidence_semi_depth}. In this way, uncertainty estimation techniques that can produce both accurate prediction and reliable uncertainty map are necessary for monocular depth estimation to be used in real-life scenarios.

\noindent\textbf{Uncertainty for Image Deblurring:} Image deblurring \cite{Tran_2021_CVPR} aims to recover a sharp image from its blurred version. Due to the difficulty in recovering the detail information of objects of different categories, there exists inherent uncertainty for image deblurring \cite{ryasarla_UMSN,confidence_guide_turbulence_mitigation,restore_image_turbulence}. Further, as there exists ambiguity in estimating the true blur kernel space that leads to the degraded blurring image, it is necessary to estimate the uncertainty within the image deblurring task.

\subsection{Uncertainty Estimation}
We estimate uncertainty for each task, and show experimental results for each task. Specifically, we analyse the ensemble based uncertainty estimation techniques, the generative model based techniques and the Bayesian latent variable model based solutions.

\begin{figure*}[tp]
   \begin{center}
   \begin{tabular}{c@{ }c@{ }c@{ }c@{ }c@{ }c@{ }c@{ }c@{ }c@{ }c@{ }}
   {\includegraphics[width=0.093\linewidth]{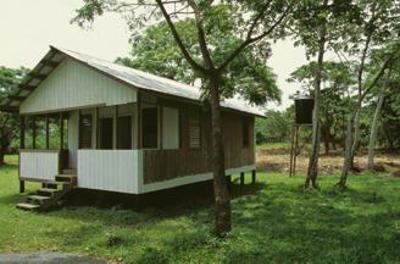}} &
   {\includegraphics[width=0.093\linewidth]{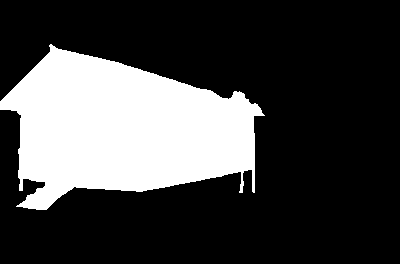}} &
   {\includegraphics[width=0.093\linewidth]{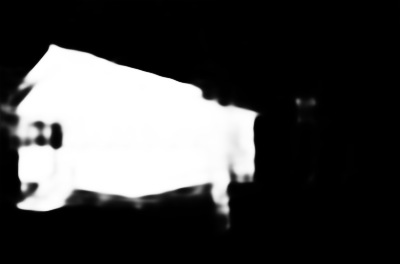}} &
   {\includegraphics[width=0.093\linewidth]{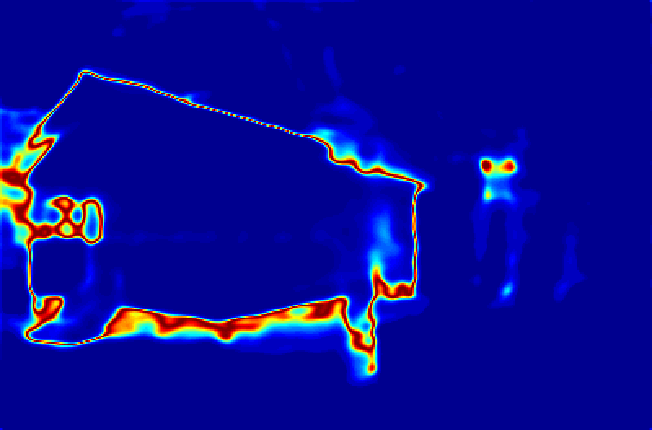}} &
   {\includegraphics[width=0.093\linewidth]{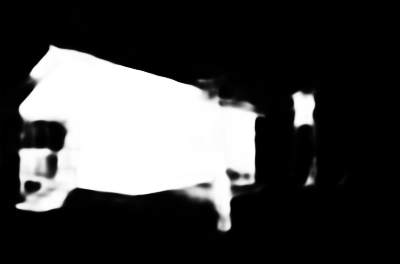}}&
   {\includegraphics[width=0.093\linewidth]{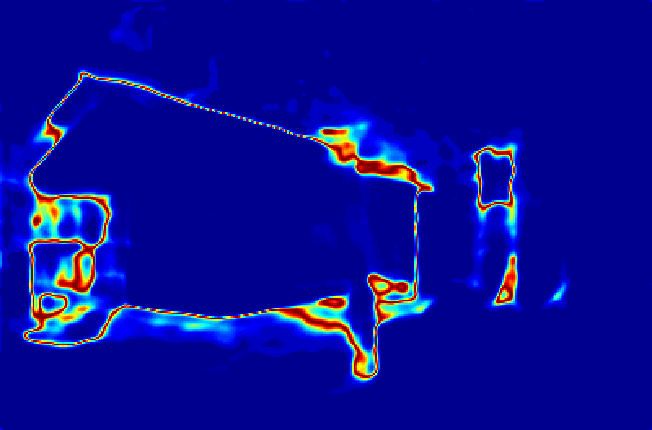}}&
   {\includegraphics[width=0.093\linewidth]{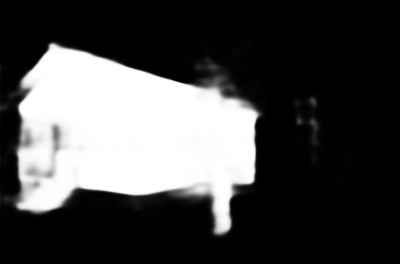}}&
   {\includegraphics[width=0.093\linewidth]{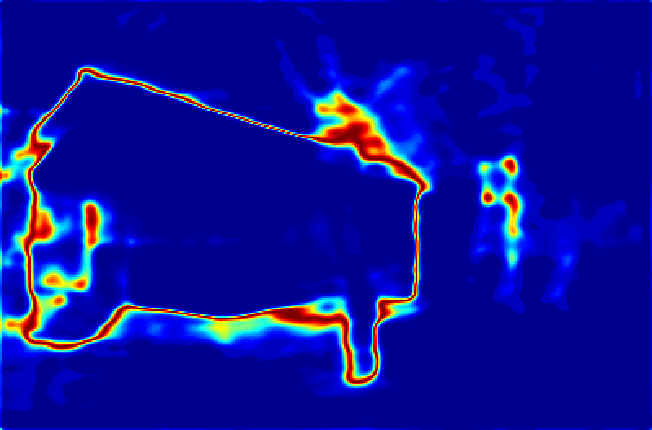}}&
   {\includegraphics[width=0.093\linewidth]{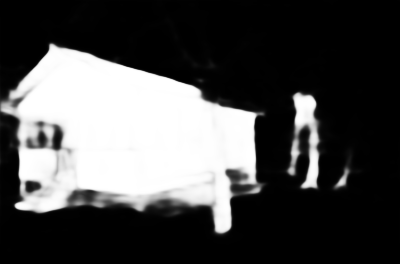}}&
   {\includegraphics[width=0.093\linewidth]{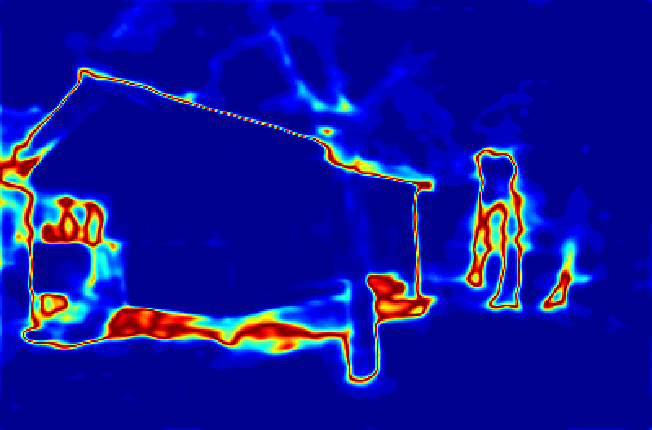}} \\
   {\includegraphics[width=0.093\linewidth]{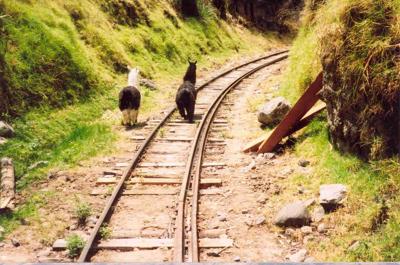}} &
   {\includegraphics[width=0.093\linewidth]{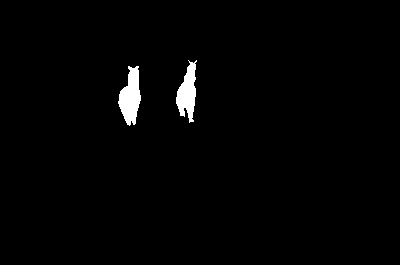}} &
   {\includegraphics[width=0.093\linewidth]{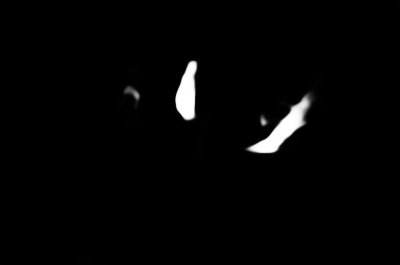}} &
   {\includegraphics[width=0.093\linewidth]{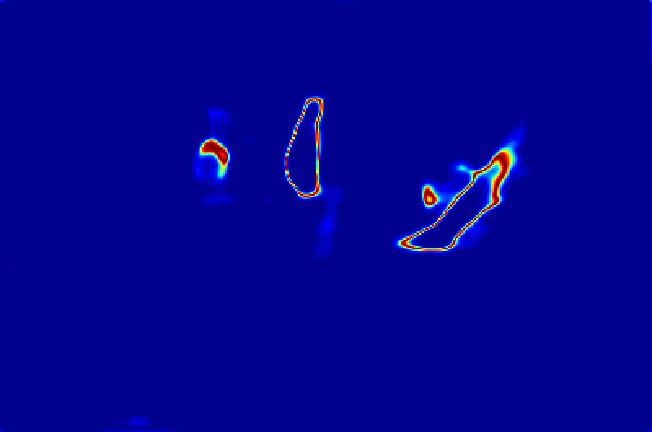}} &
   {\includegraphics[width=0.093\linewidth]{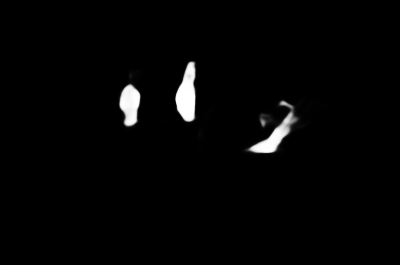}}&
   {\includegraphics[width=0.093\linewidth]{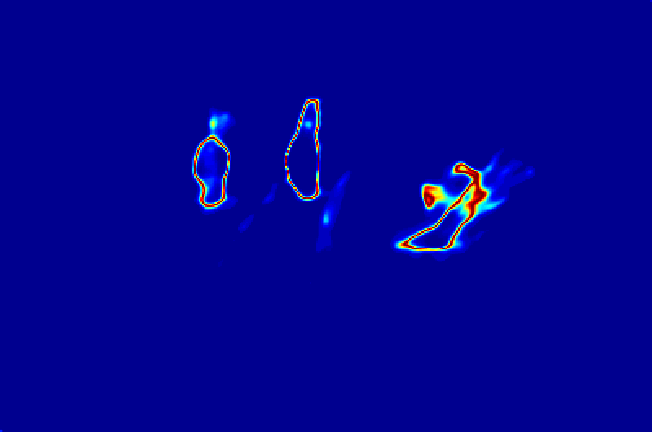}}&
   {\includegraphics[width=0.093\linewidth]{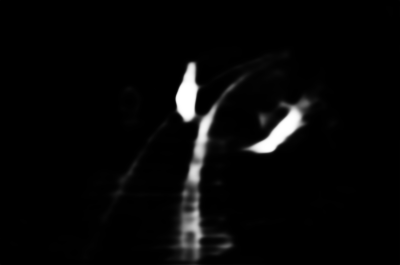}}&
   {\includegraphics[width=0.093\linewidth]{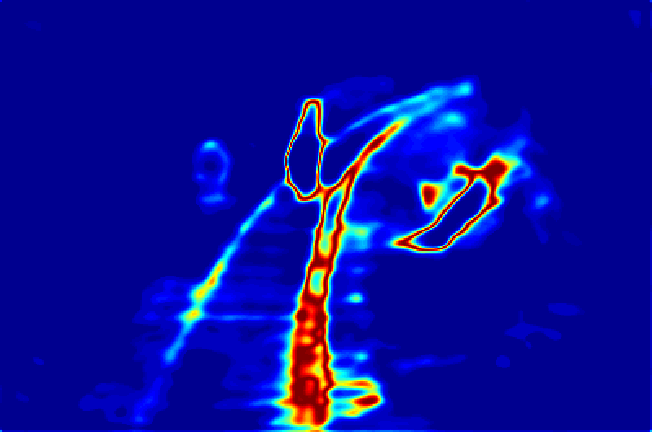}}&
   {\includegraphics[width=0.093\linewidth]{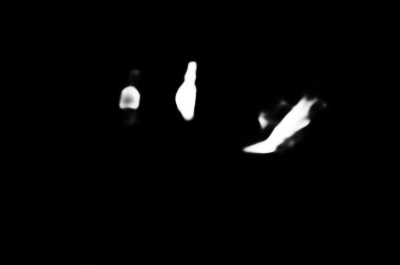}}&
   {\includegraphics[width=0.093\linewidth]{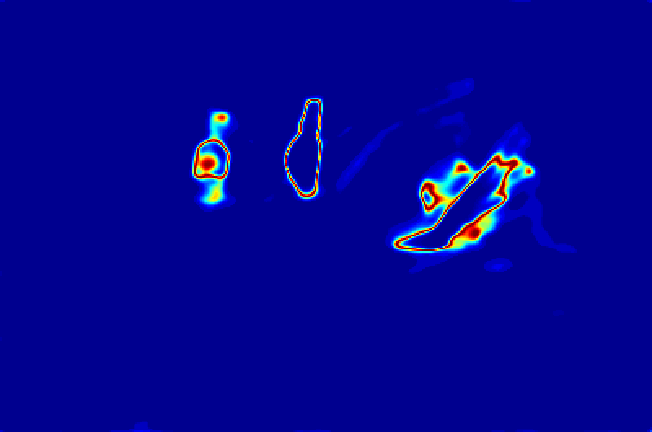}} \\
   {\includegraphics[width=0.093\linewidth]{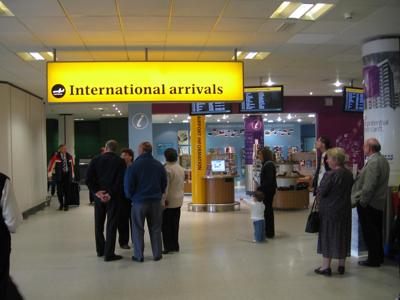}} &
   {\includegraphics[width=0.093\linewidth]{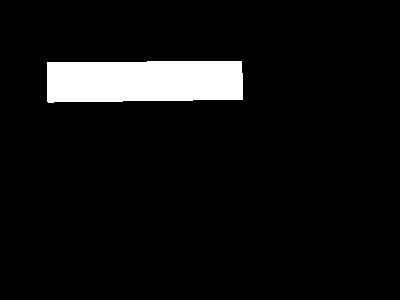}} &
   {\includegraphics[width=0.093\linewidth]{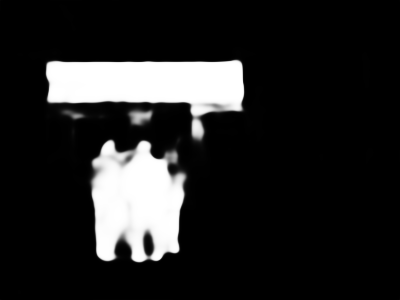}} &
   {\includegraphics[width=0.093\linewidth]{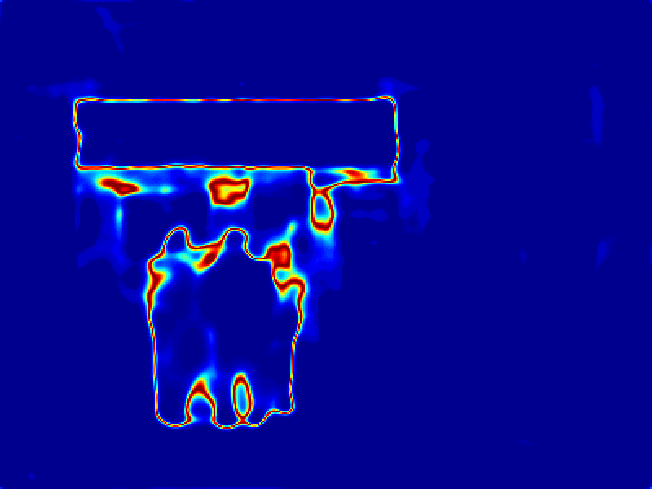}} &
   {\includegraphics[width=0.093\linewidth]{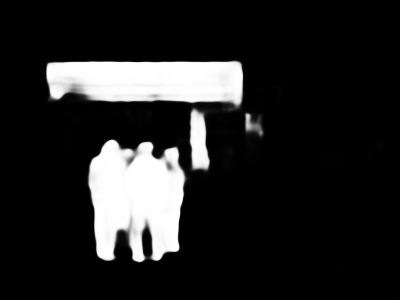}}&
   {\includegraphics[width=0.093\linewidth]{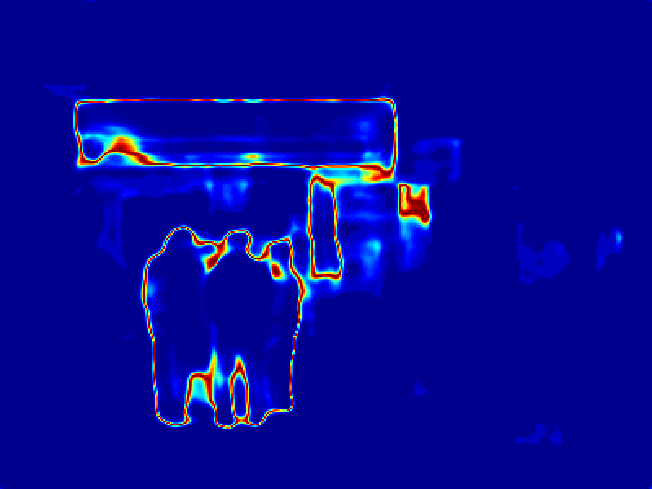}}&
   {\includegraphics[width=0.093\linewidth]{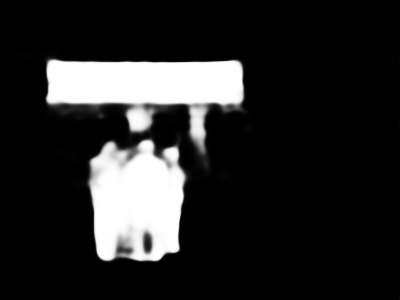}}&
   {\includegraphics[width=0.093\linewidth]{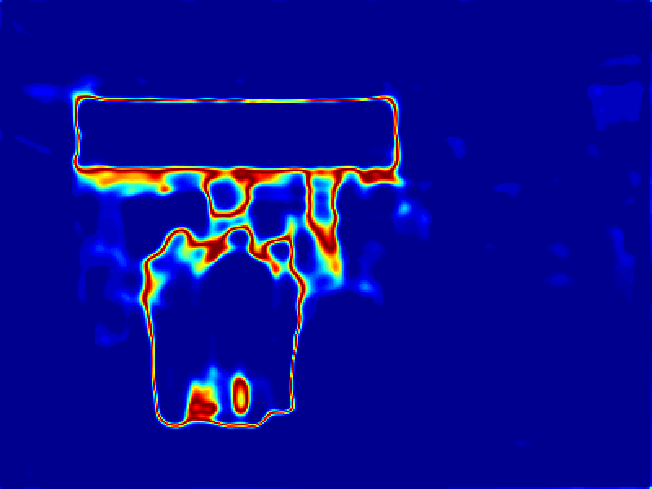}}&
   {\includegraphics[width=0.093\linewidth]{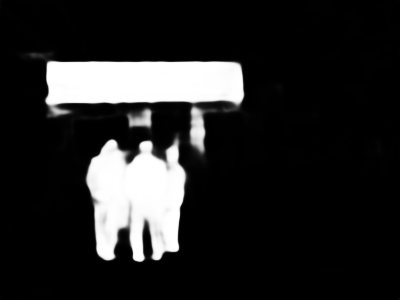}}&
   {\includegraphics[width=0.093\linewidth]{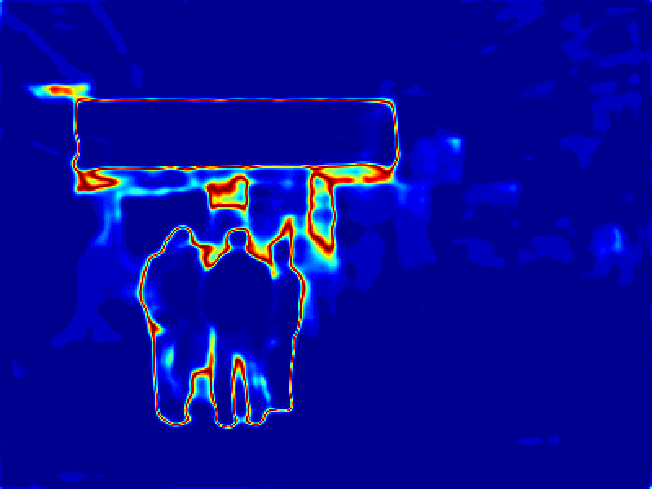}} \\
   \footnotesize{Image}&\footnotesize{GT}&\footnotesize{CVAE}&\footnotesize{$U_p$}&\footnotesize{CGAN}&\footnotesize{$U_p$}&\footnotesize{ABP}&\footnotesize{$U_p$}&\footnotesize{EBM}&\footnotesize{$U_p$}\\
   \end{tabular}
   \end{center}
   \caption{\footnotesize{Predictive uncertainty of generative model based solutions for \textbf{salient object detection}.}
   }
\label{fig:predictive_generative_sod}
\end{figure*}

\begin{figure*}[tp]
   \begin{center}
   \begin{tabular}{c@{ }c@{ }c@{ }c@{ }c@{ }c@{ }c@{ }c@{ }c@{ }c@{ }}
   {\includegraphics[width=0.095\linewidth]{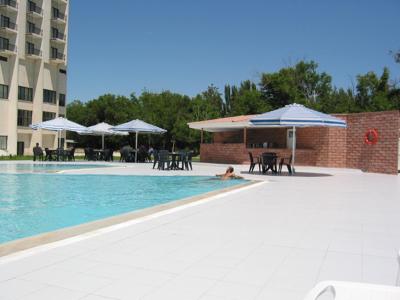}} &
   {\includegraphics[width=0.095\linewidth]{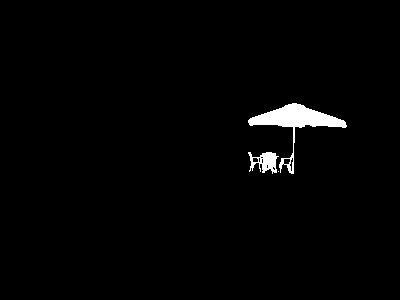}} &
   {\includegraphics[width=0.095\linewidth]{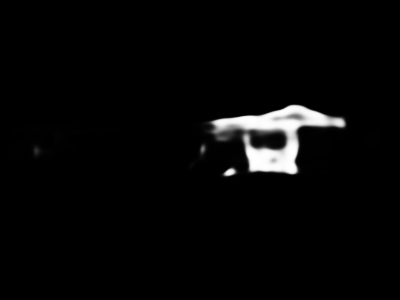}} &
   {\includegraphics[width=0.095\linewidth]{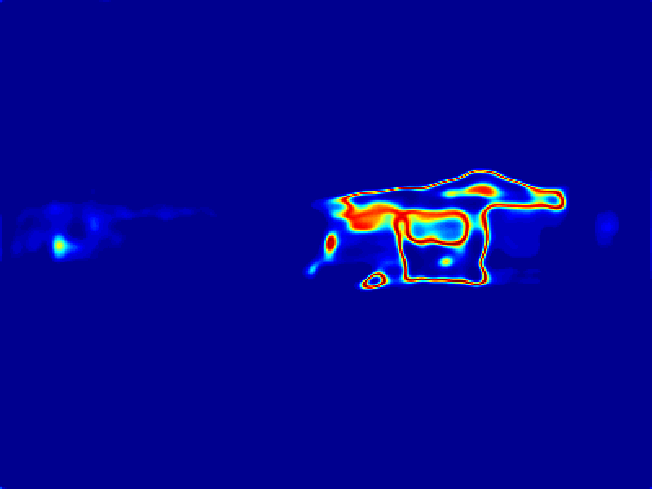}} &
   {\includegraphics[width=0.095\linewidth]{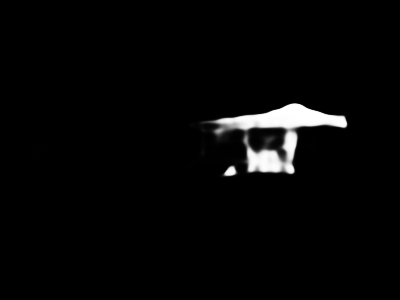}}&
   {\includegraphics[width=0.095\linewidth]{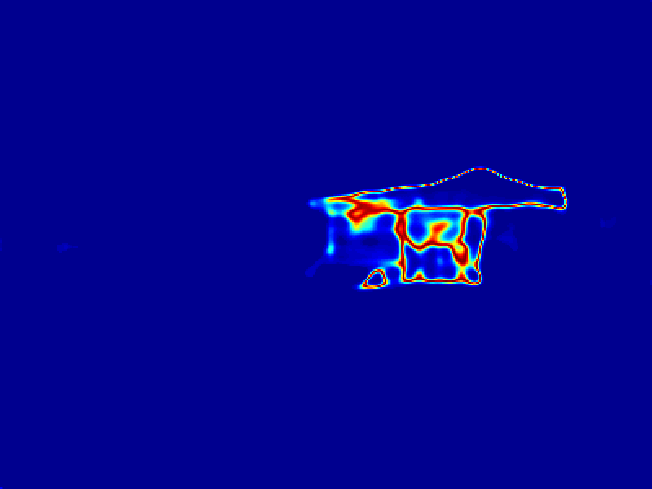}}&
   {\includegraphics[width=0.095\linewidth]{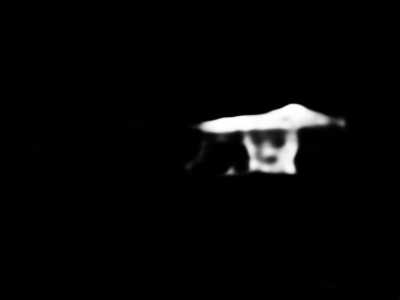}}&
   {\includegraphics[width=0.095\linewidth]{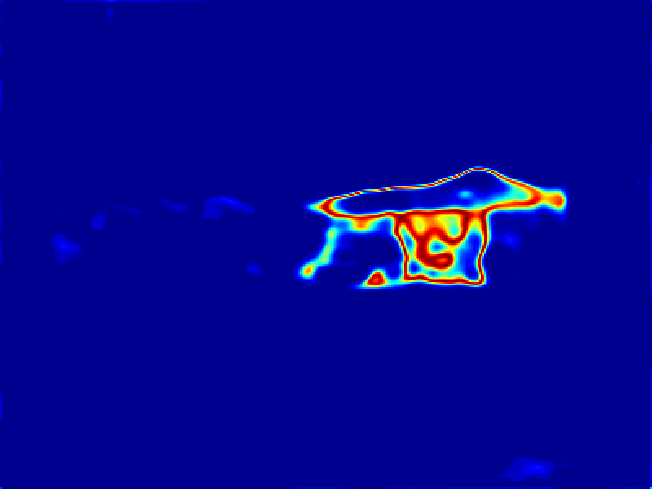}}&
   {\includegraphics[width=0.095\linewidth]{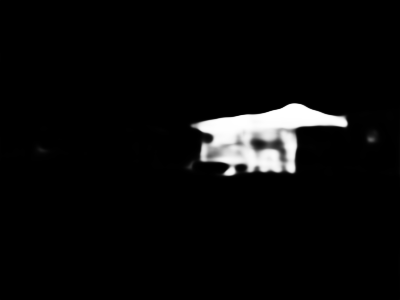}}&
   {\includegraphics[width=0.095\linewidth]{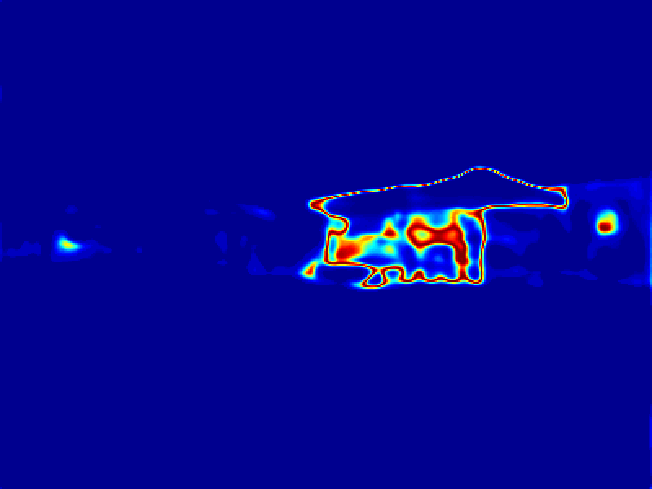}} \\
   {\includegraphics[width=0.095\linewidth]{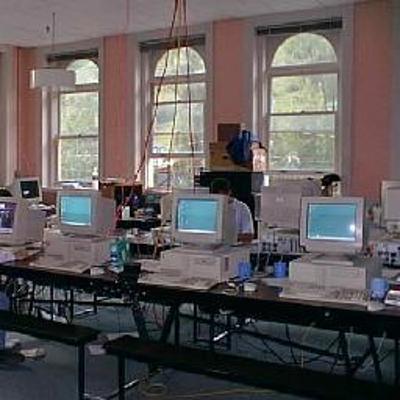}} &
   {\includegraphics[width=0.095\linewidth]{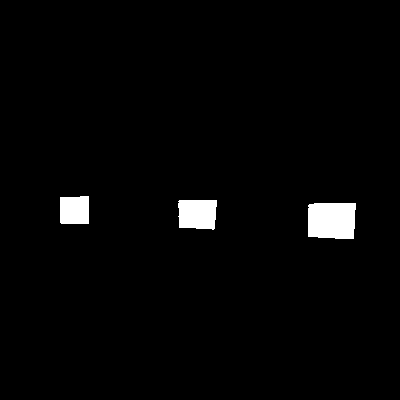}} &
   {\includegraphics[width=0.095\linewidth]{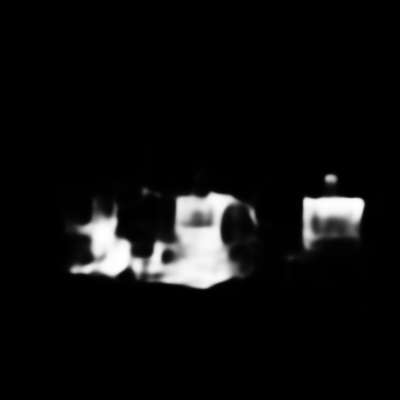}} &
   {\includegraphics[width=0.095\linewidth]{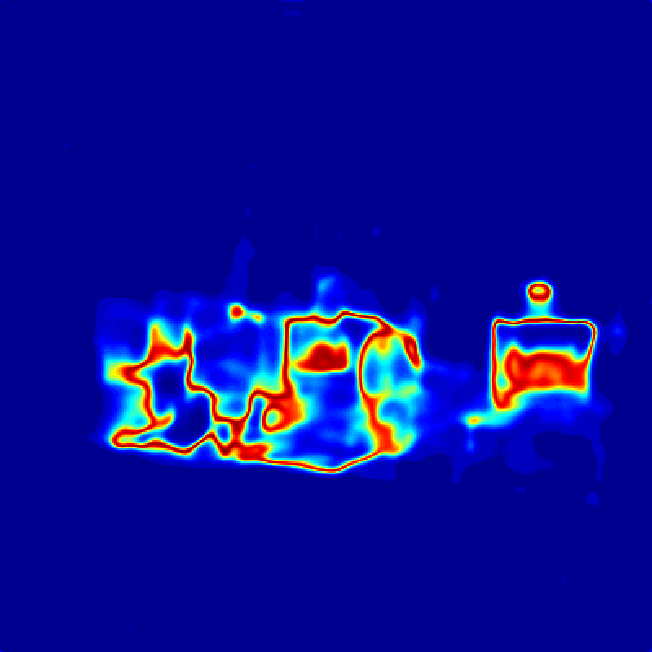}} &
   {\includegraphics[width=0.095\linewidth]{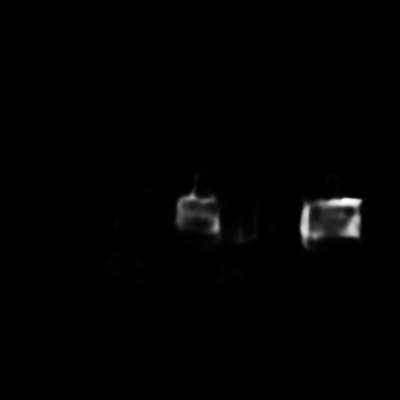}}&
   {\includegraphics[width=0.095\linewidth]{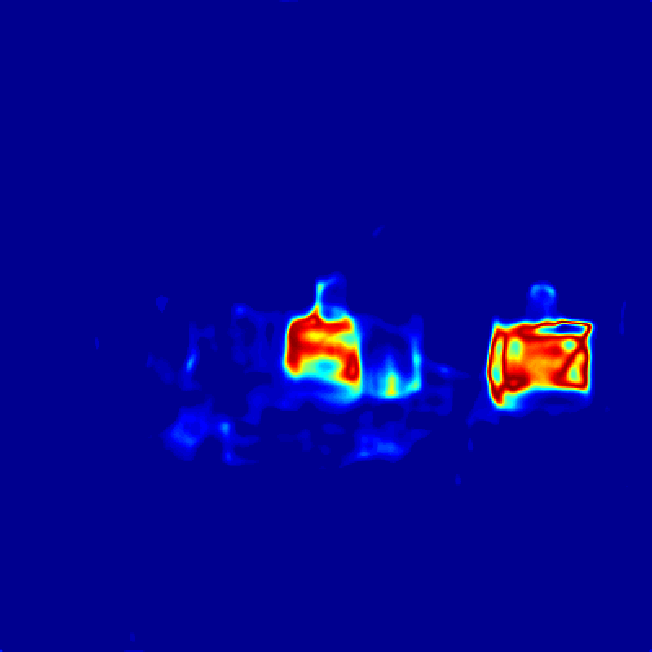}}&
   {\includegraphics[width=0.095\linewidth]{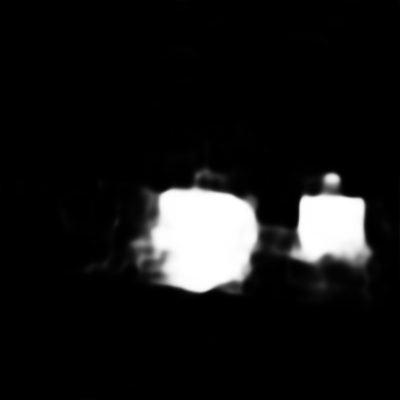}}&
   {\includegraphics[width=0.095\linewidth]{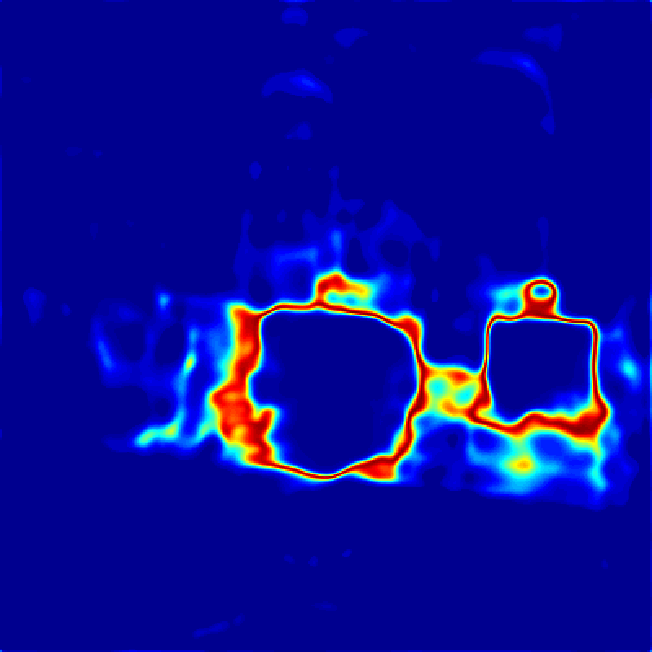}}&
   {\includegraphics[width=0.095\linewidth]{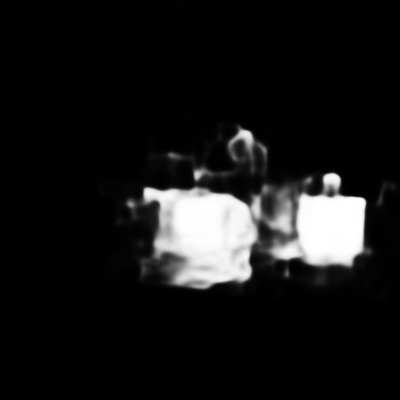}}&
   {\includegraphics[width=0.095\linewidth]{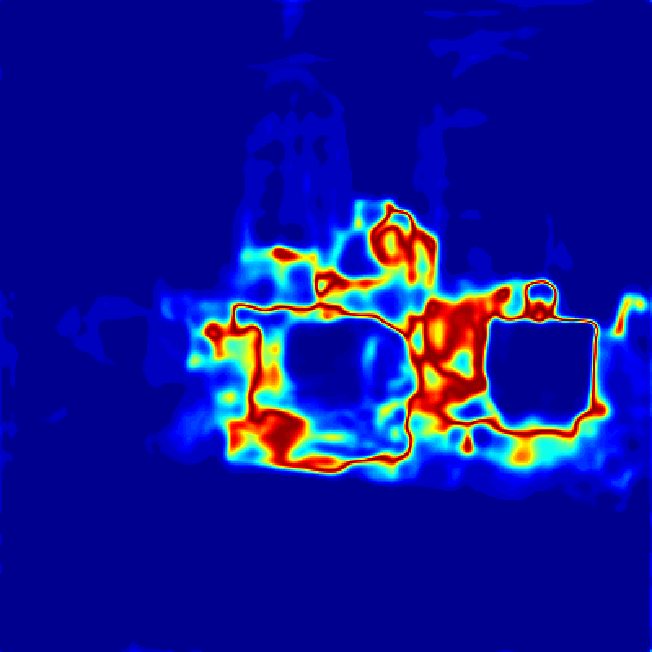}} \\
   {\includegraphics[width=0.095\linewidth]{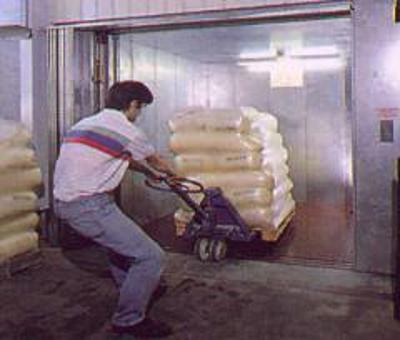}} &
   {\includegraphics[width=0.095\linewidth]{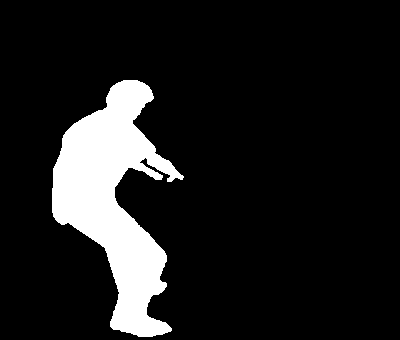}} &
   {\includegraphics[width=0.095\linewidth]{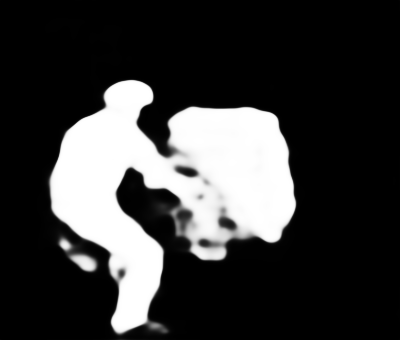}} &
   {\includegraphics[width=0.095\linewidth]{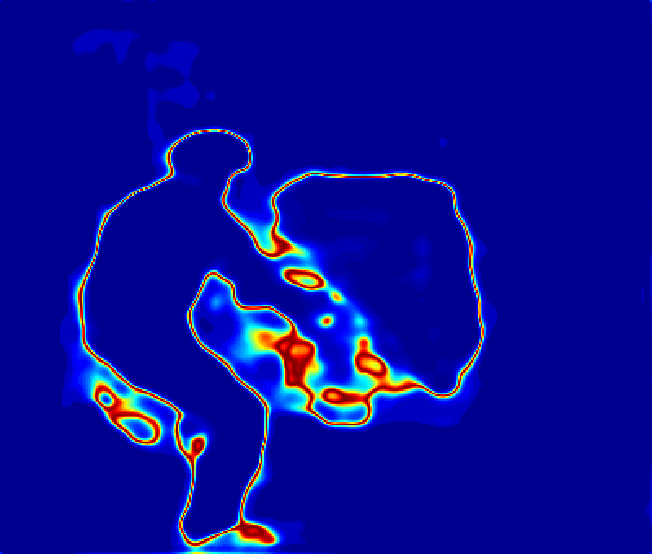}} &
   {\includegraphics[width=0.095\linewidth]{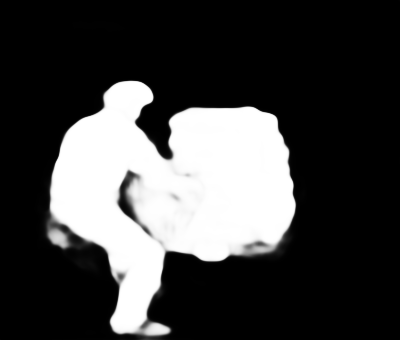}}&
   {\includegraphics[width=0.095\linewidth]{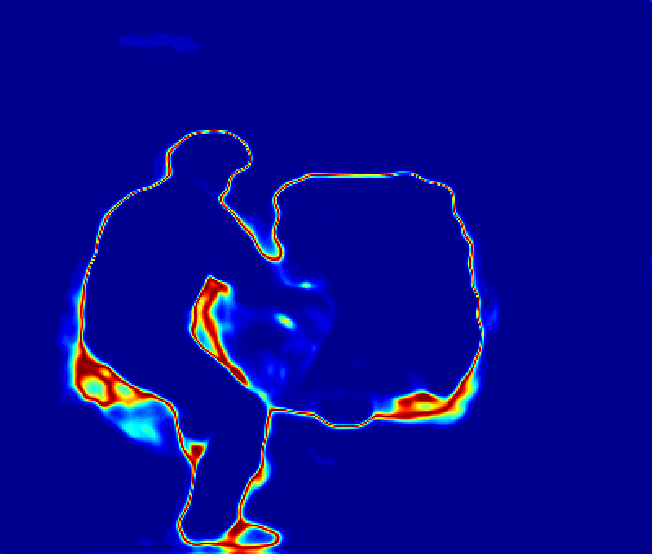}}&
   {\includegraphics[width=0.095\linewidth]{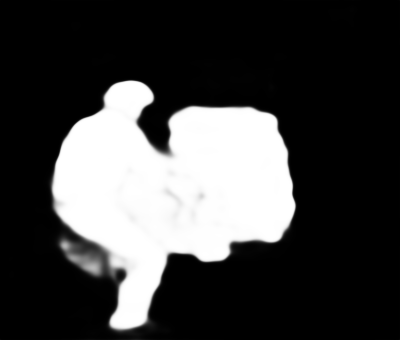}}&
   {\includegraphics[width=0.095\linewidth]{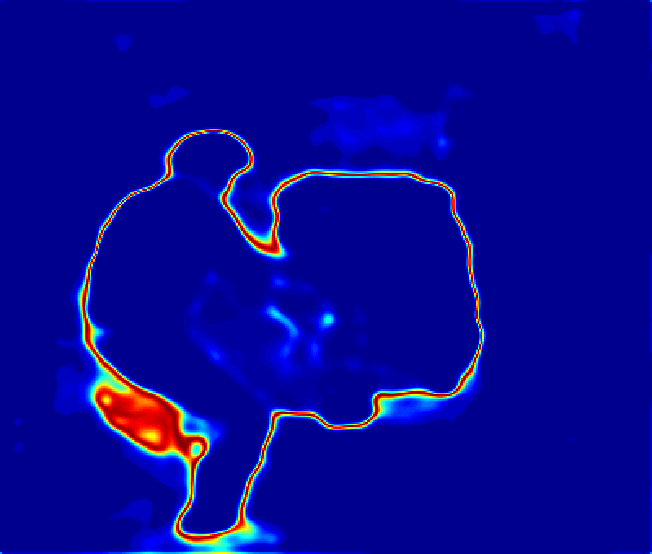}}&
   {\includegraphics[width=0.095\linewidth]{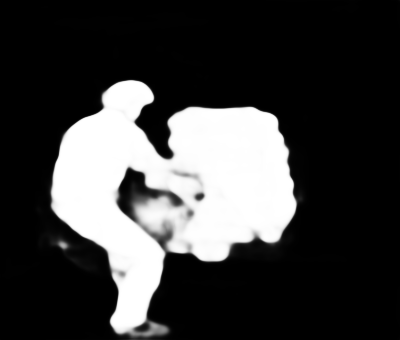}}&
   {\includegraphics[width=0.095\linewidth]{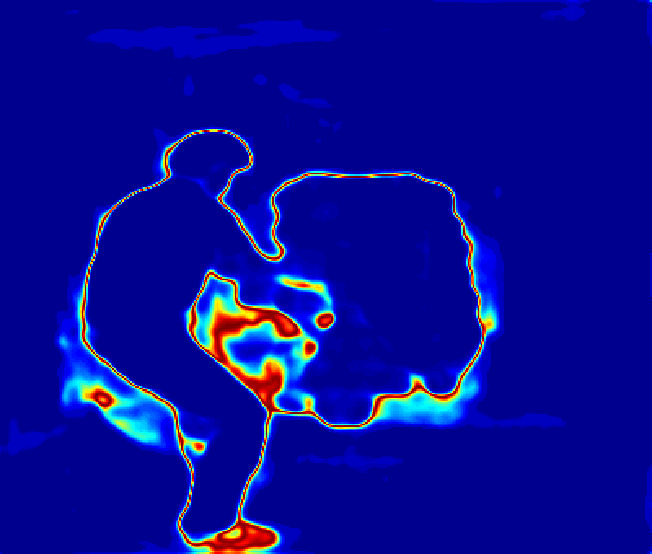}} \\
   \footnotesize{Image}&\footnotesize{GT}&\footnotesize{CVAE}&\footnotesize{$U_a$}&\footnotesize{CGAN}&\footnotesize{$U_a$}&\footnotesize{ABP}&\footnotesize{$U_a$}&\footnotesize{EBM}&\footnotesize{$U_a$}\\
   \end{tabular}
   \end{center}
   \caption{\footnotesize{Aleatoric uncertainty of generative model based solutions for \textbf{salient object detection}.}
   }
\label{fig:aleatoric_generative_sod}
\end{figure*}

\begin{figure*}[tp]
   \begin{center}
   \begin{tabular}{c@{ }c@{ }c@{ }c@{ }c@{ }c@{ }c@{ }c@{ }c@{ }c@{ }}
   {\includegraphics[width=0.095\linewidth]{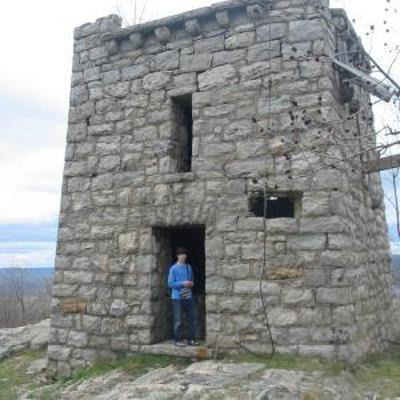}} &
   {\includegraphics[width=0.095\linewidth]{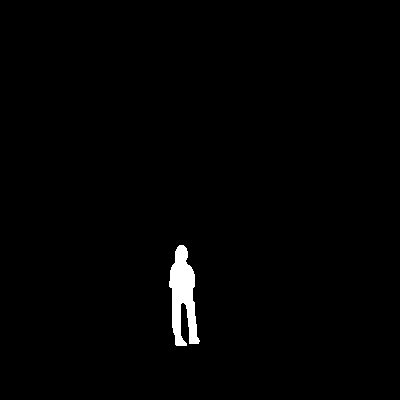}} &
   {\includegraphics[width=0.095\linewidth]{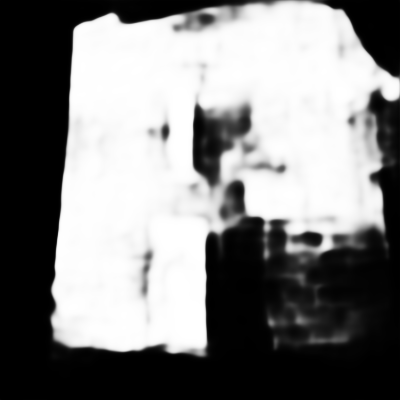}} &
   {\includegraphics[width=0.095\linewidth]{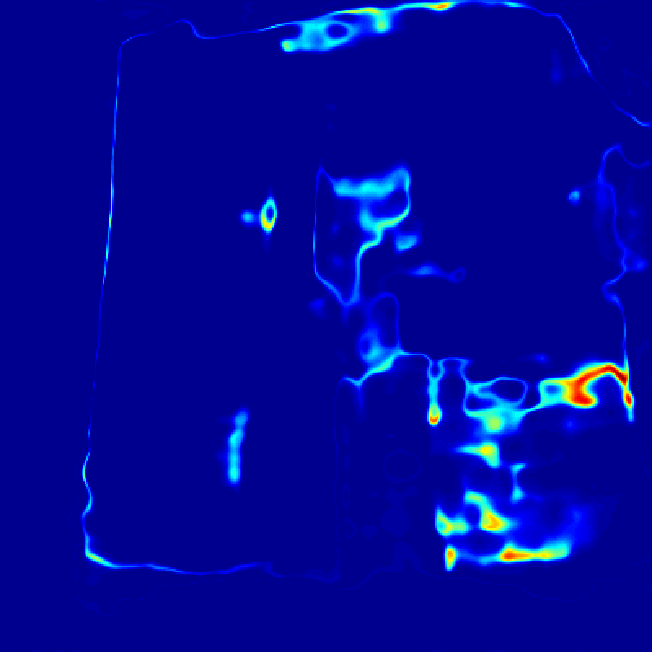}} &
   {\includegraphics[width=0.095\linewidth]{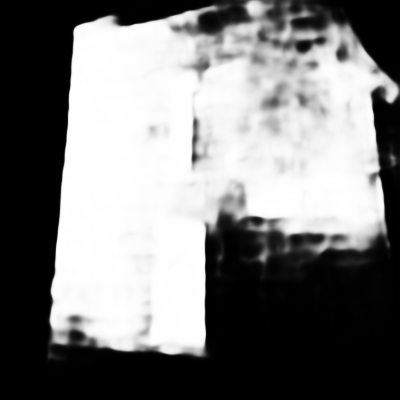}}&
   {\includegraphics[width=0.095\linewidth]{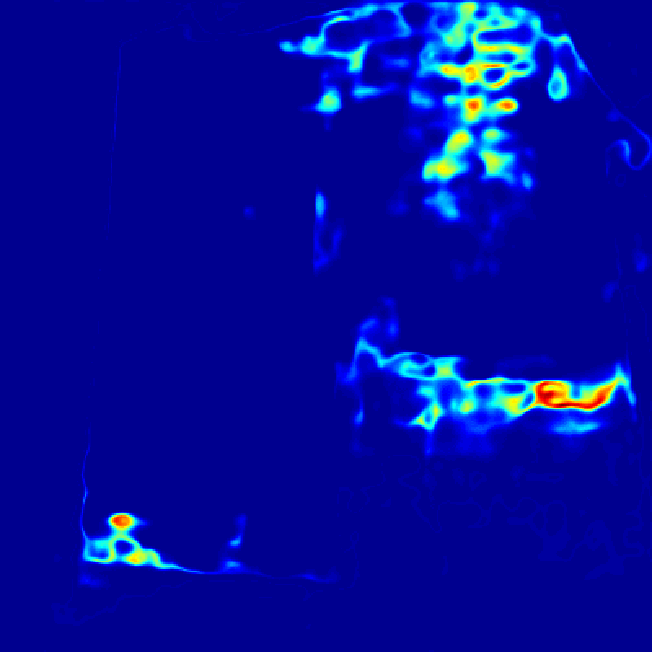}}&
   {\includegraphics[width=0.095\linewidth]{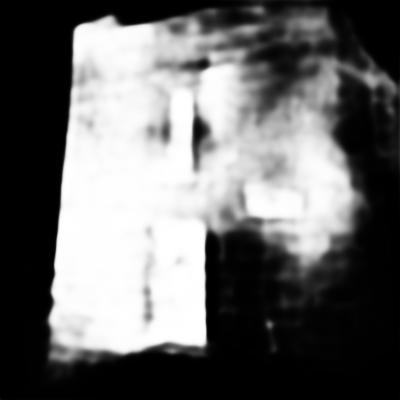}}&
   {\includegraphics[width=0.095\linewidth]{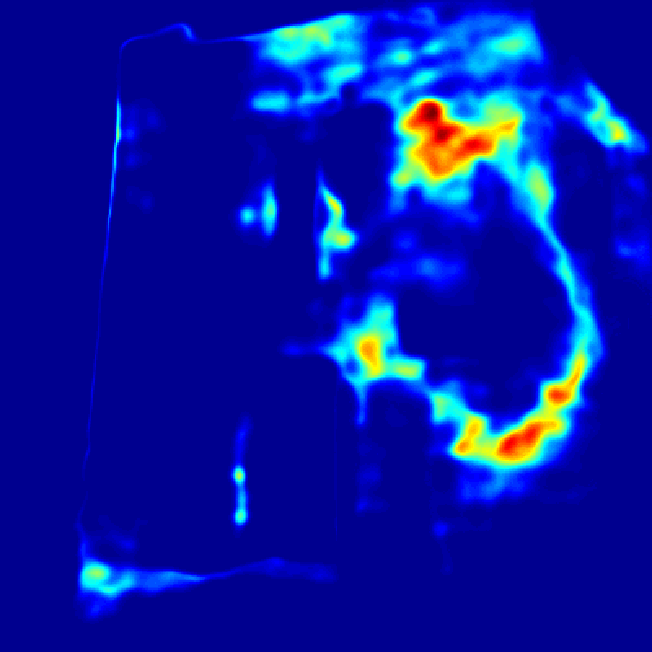}}&
   {\includegraphics[width=0.095\linewidth]{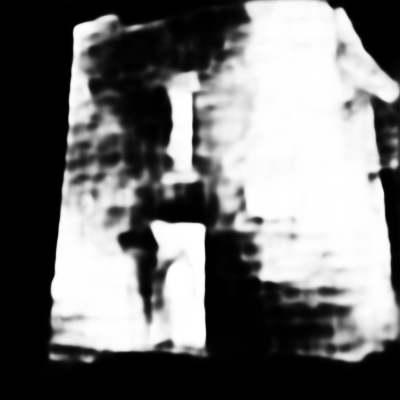}}&
   {\includegraphics[width=0.095\linewidth]{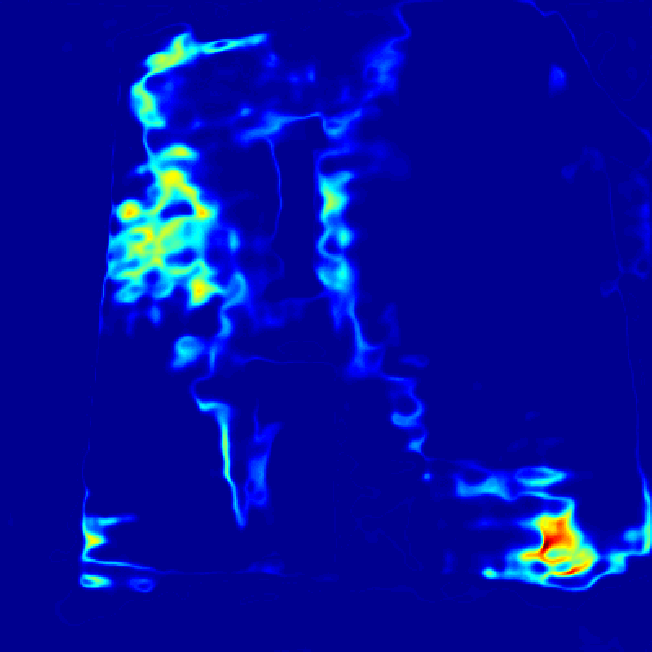}} \\
   {\includegraphics[width=0.095\linewidth]{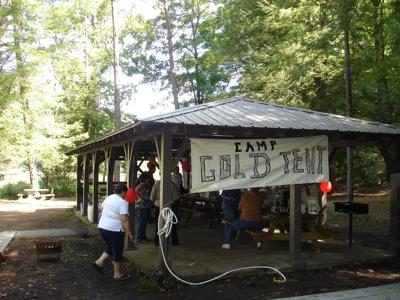}} &
   {\includegraphics[width=0.095\linewidth]{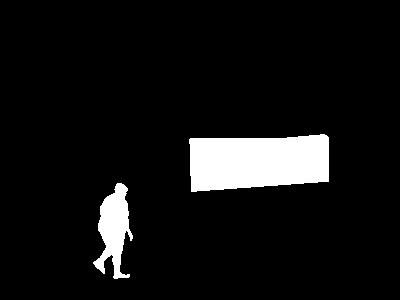}} &
   {\includegraphics[width=0.095\linewidth]{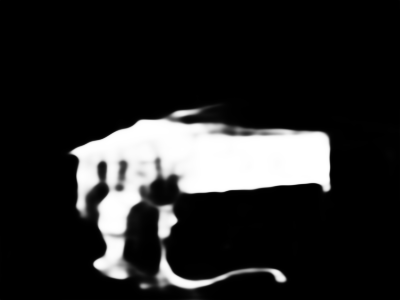}} &
   {\includegraphics[width=0.095\linewidth]{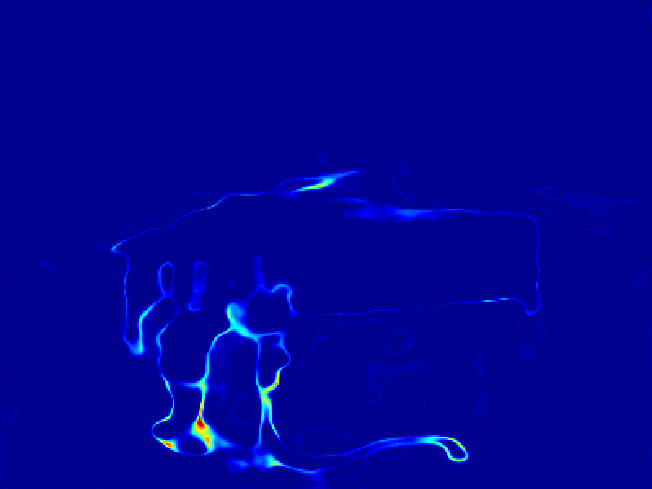}} &
   {\includegraphics[width=0.095\linewidth]{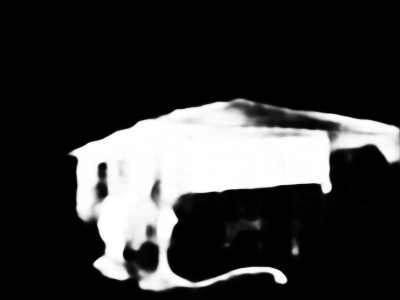}}&
   {\includegraphics[width=0.095\linewidth]{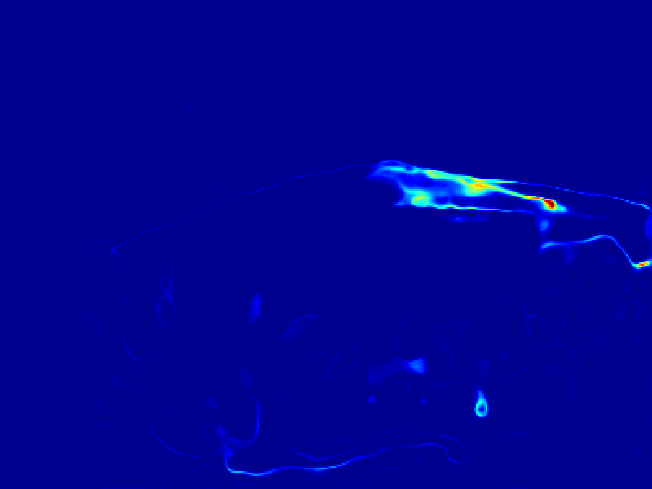}}&
   {\includegraphics[width=0.095\linewidth]{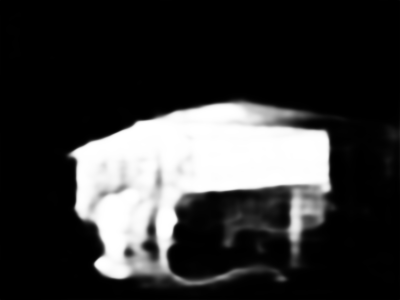}}&
   {\includegraphics[width=0.095\linewidth]{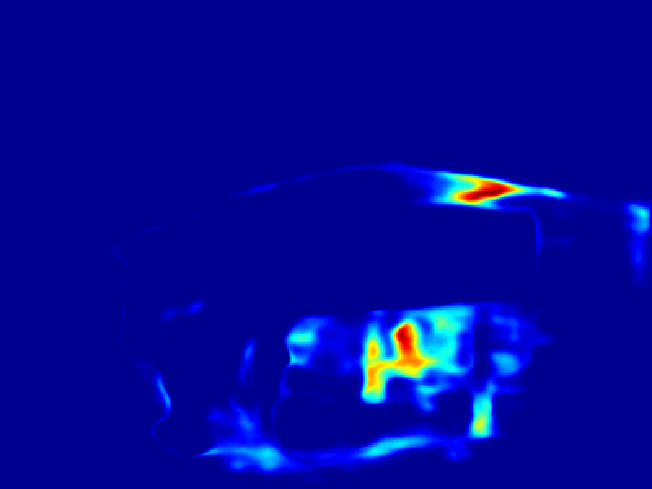}}&
   {\includegraphics[width=0.095\linewidth]{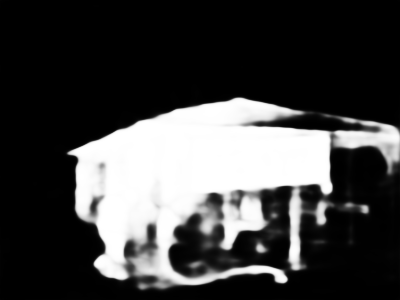}}&
   {\includegraphics[width=0.095\linewidth]{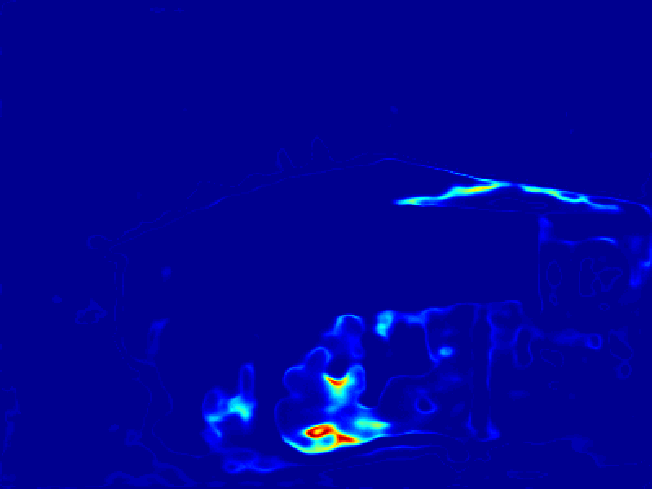}} \\
   {\includegraphics[width=0.095\linewidth]{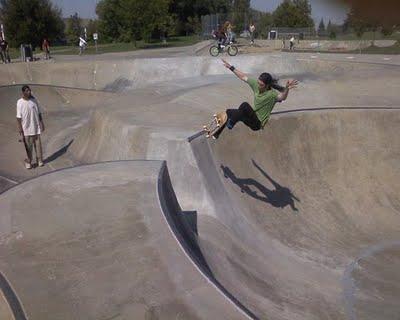}} &
   {\includegraphics[width=0.095\linewidth]{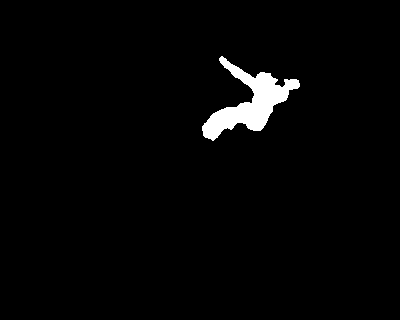}} &
   {\includegraphics[width=0.095\linewidth]{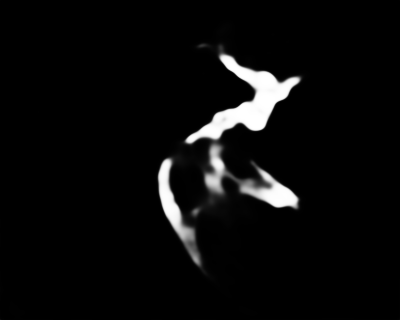}} &
   {\includegraphics[width=0.095\linewidth]{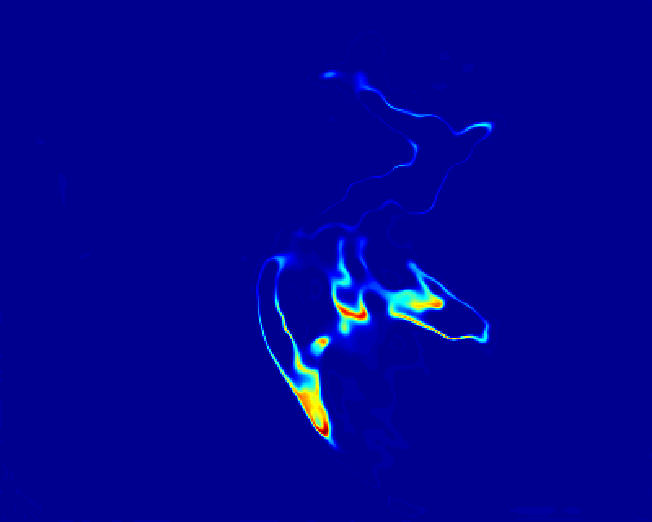}} &
   {\includegraphics[width=0.095\linewidth]{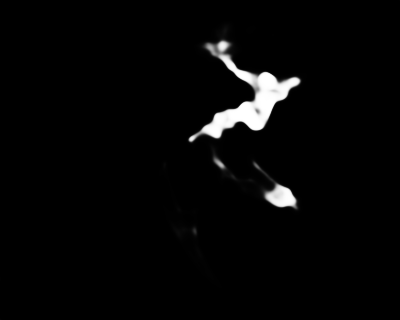}}&
   {\includegraphics[width=0.095\linewidth]{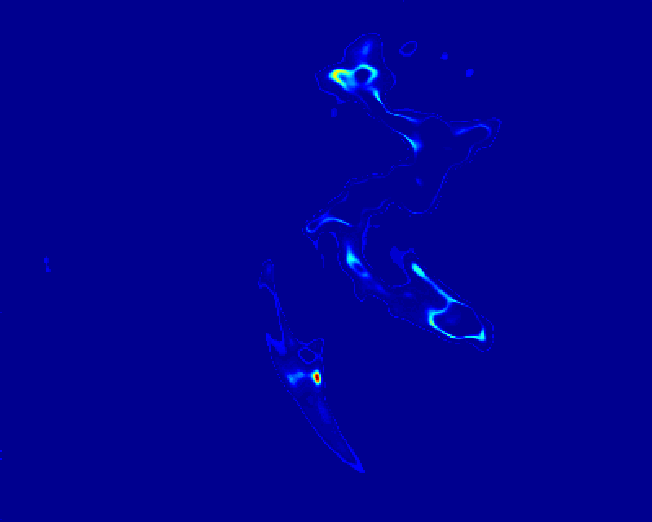}}&
   {\includegraphics[width=0.095\linewidth]{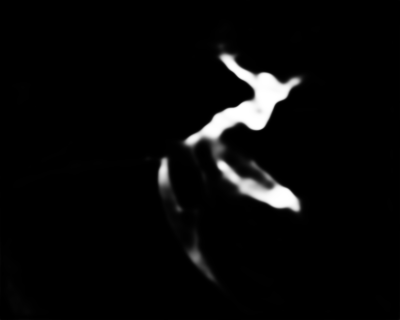}}&
   {\includegraphics[width=0.095\linewidth]{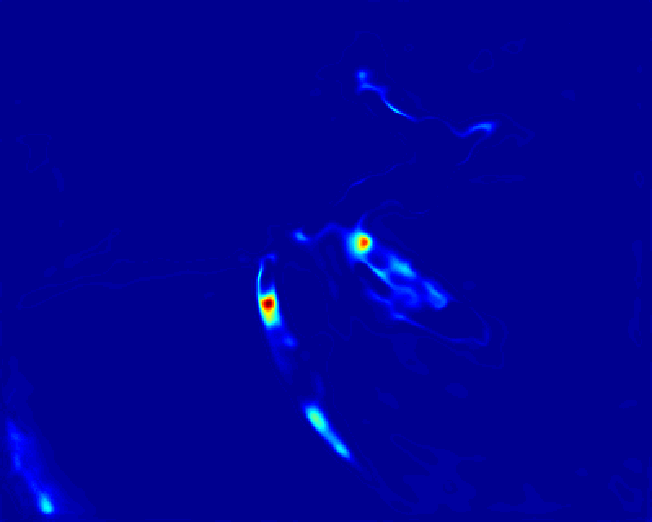}}&
   {\includegraphics[width=0.095\linewidth]{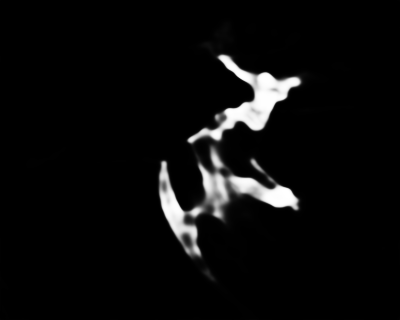}}&
   {\includegraphics[width=0.095\linewidth]{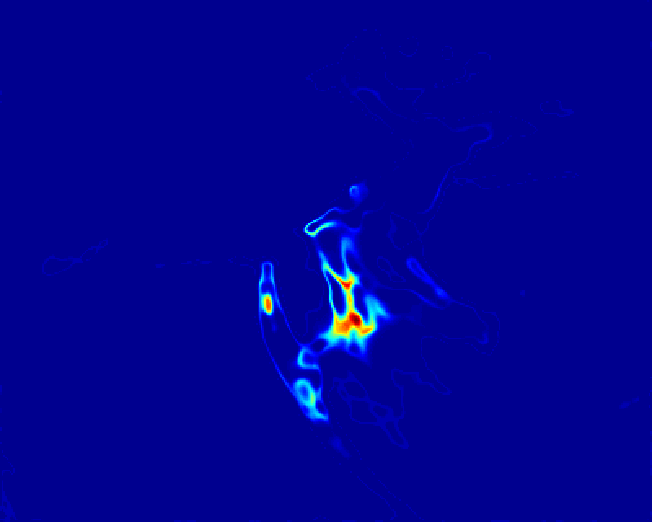}} \\
   \footnotesize{Image}&\footnotesize{GT}&\footnotesize{CVAE}&\footnotesize{$U_e$}&\footnotesize{CGAN}&\footnotesize{$U_e$}&\footnotesize{ABP}&\footnotesize{$U_e$}&\footnotesize{EBM}&\footnotesize{$U_e$}\\
   \end{tabular}
   \end{center}
   \caption{\footnotesize{Epistemic uncertainty of generative model based solutions for \textbf{salient object detection}.}
   }
\label{fig:epistemic_generative_sod}
\end{figure*}

\begin{table}[t!]
  \centering
  \scriptsize
  \renewcommand{\arraystretch}{1.2}
  \renewcommand{\tabcolsep}{1.55mm}
  \caption{Ensemble based solutions for \textbf{salient object detection}. $\uparrow$ indicates the higher the score the better, and vice versa for $\downarrow$.}
  \begin{tabular}{l|cc|cc|cc|cc}
  \toprule
  Method &\multicolumn{2}{c|}{DUTS~\cite{imagesaliency}}&\multicolumn{2}{c|}{DUT~\cite{Manifold-Ranking:CVPR-2013}}&\multicolumn{2}{c|}{HKU-IS~\cite{MDF:CVPR-2015}}&\multicolumn{2}{c}{PASCAL~\cite{pascal_s_dataset}} \\
    &$F_{\beta}\uparrow$&$\mathcal{M}\downarrow$&$F_{\beta}\uparrow$&$\mathcal{M}\downarrow$
    &$F_{\beta}\uparrow$&$\mathcal{M}\downarrow$
    &$F_{\beta}\uparrow$&$\mathcal{M}\downarrow$\\
  \hline
  Base & .842 & .037 & .760 & .055 & .904 & .030 & .828 & .064 \\
  MD & .854 & .036 & .763 & .056 & .911 & .028 & .840 & .061 \\
  DE & .828 & .040 & .738 & .061 & .897 & .031 & .825 & .065 \\
  \bottomrule
  \end{tabular}
  \label{tab:ablation_sod_ensemble}
\end{table}

\subsubsection{Ensemble based uncertainty estimation}
We show the base model performance as \enquote{Base} in Table \ref{tab:ablation_cod_ensemble}, which is a deterministic neural network with no ensemble strategies or generative model for uncertainty estimation. Then, we introduce MC-dropout \cite{Gal2016Dropout} and deep ensemble \cite{simple_scalable_uncertainty} to the base model respectively,
and show the model performance as \enquote{MD} and \enquote{DE}.
The relatively stable deterministic performance of \enquote{MD}
compared with the base model illustrate the MC-dropout
can keep the original deterministic prediction accuracy. However, we observe slightly worse performance of the deep ensemble solution. For the MC-dropout based model, although we perform random weights erasing, due to the huge capacity of the deep network, the final model may not deviate too much from the base model with the same initialization \cite{ResHe2015}. Within deep ensemble model, we attached five decoders \cite{midas_tpami} of the same structure but different initialization to generate five predictions. The different initialization of deep ensemble model compared with the base model explains their different predictions. As discussed in \cite{He_2019_ICCV_rethink_imagenet_pretrain}, model with more random initialization may need longer training time. As we fixed the maximum epoch for the ensemble based models, the deep ensemble based model may not converge to the best performance, which explians the slightly worse performance of the deep ensemble solution.

We further show the uncertainty maps of the ensemble based solutions for camouflaged object detection in Fig.~\ref{fig:predictive_ensemble_cod}, Fig.~\ref{fig:aleatoric_ensemble_cod} and Fig.~\ref{fig:epistemic_ensemble_cod}, representing the predictive uncertainty, aleatoric uncertainty and epistemic uncertainty respectively. The uncertainty maps indicate that aleatoric uncertainty usually focus on object boundary, where the labeling noise usually occurs. The epistemic uncertainty has less activation region than the aleatoric uncertainty, which usually highlight the hard regions with camouflage attributes that are not learned with the current training dataset.

Similarly, we train ensemble based salient object detection model and show model performance in Table \ref{tab:ablation_sod_ensemble}, and uncertainty maps in Fig.~\ref{fig:predictive_ensemble_sod}, Fig.~\ref{fig:aleatoric_ensemble_sod} and Fig.~\ref{fig:epistemic_ensemble_sod} respectively, representing the predictive uncertainty, aleatoric uncertainty and epistemic uncertainty. The conclusion is similar to the camouflaged object detection task.


\subsubsection{Generative model based uncertainty estimation}
We introduce four generative model based uncertainty estimation techniques, including three latent variable models, namely CVAE based framework \cite{cvae}, GAN based framework \cite{NIPS2014_5423_gan} ABP based framework \cite{ABP_aaai}, and one energy-based model \cite{LeCun06atutorial,coopnets} for prediction distribution estimation. We apply those generative model based solutions to camouflaged object detection and salient object detection, and show performance in Table \ref{tab:ablation_cod_generative} and Table \ref{tab:ablation_sod_generative} respectively. We observe similar performance of those generative model based methods compared with the base models. In general, the GAN \cite{NIPS2014_5423_gan} based framework achieves slightly better performance than the other generative models. The main reason is the adversarial loss term within the generator that serves as higher-order similarity measure.

\begin{figure}[tp]
   \begin{center}
   \begin{tabular}{c@{ }c@{ }c@{ }c@{ }c@{ }c@{ }}
   {\includegraphics[width=0.153\linewidth]{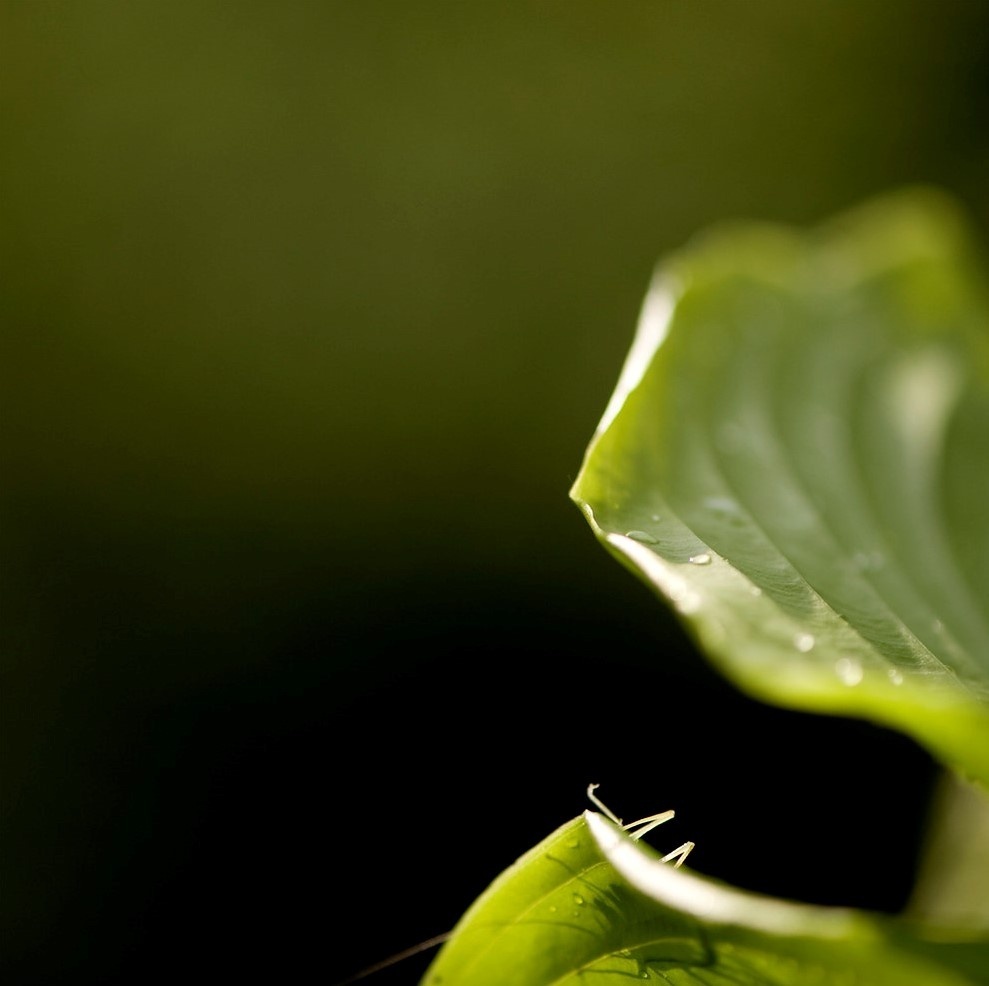}} &
   {\includegraphics[width=0.153\linewidth]{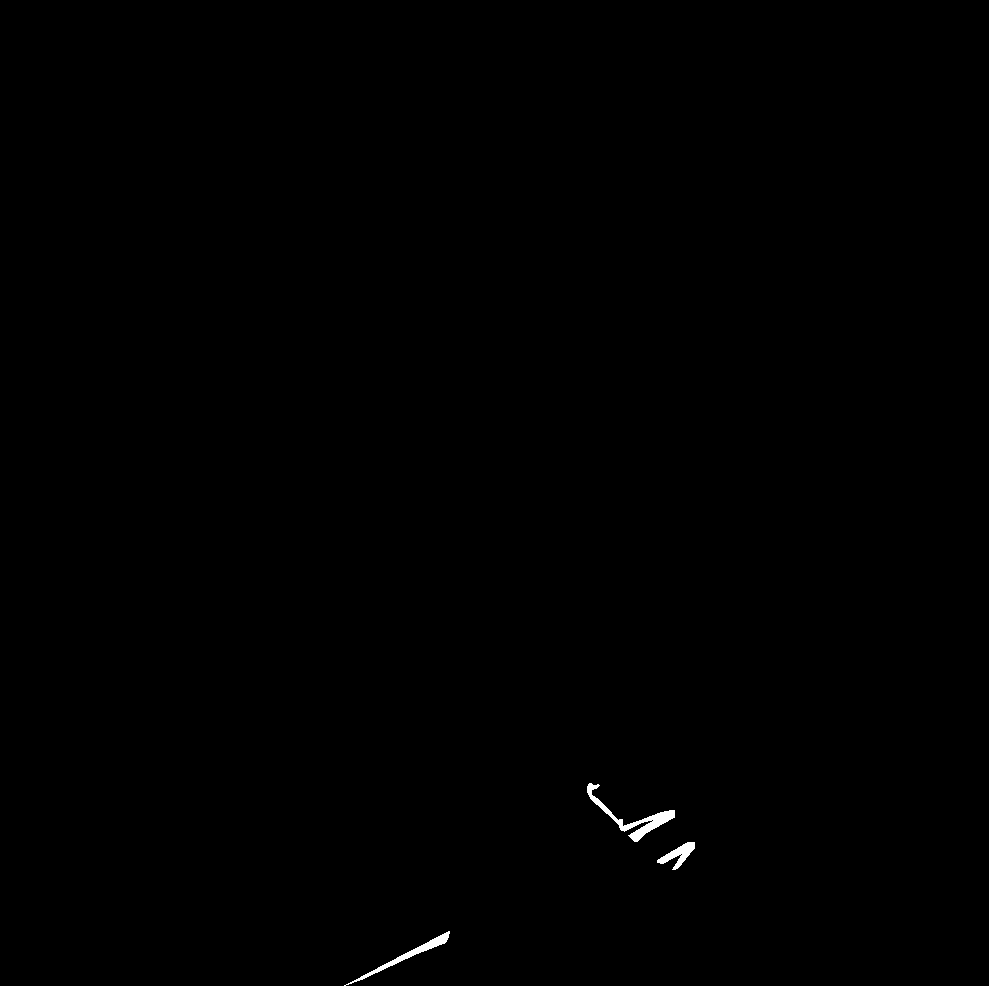}} &
   {\includegraphics[width=0.153\linewidth]{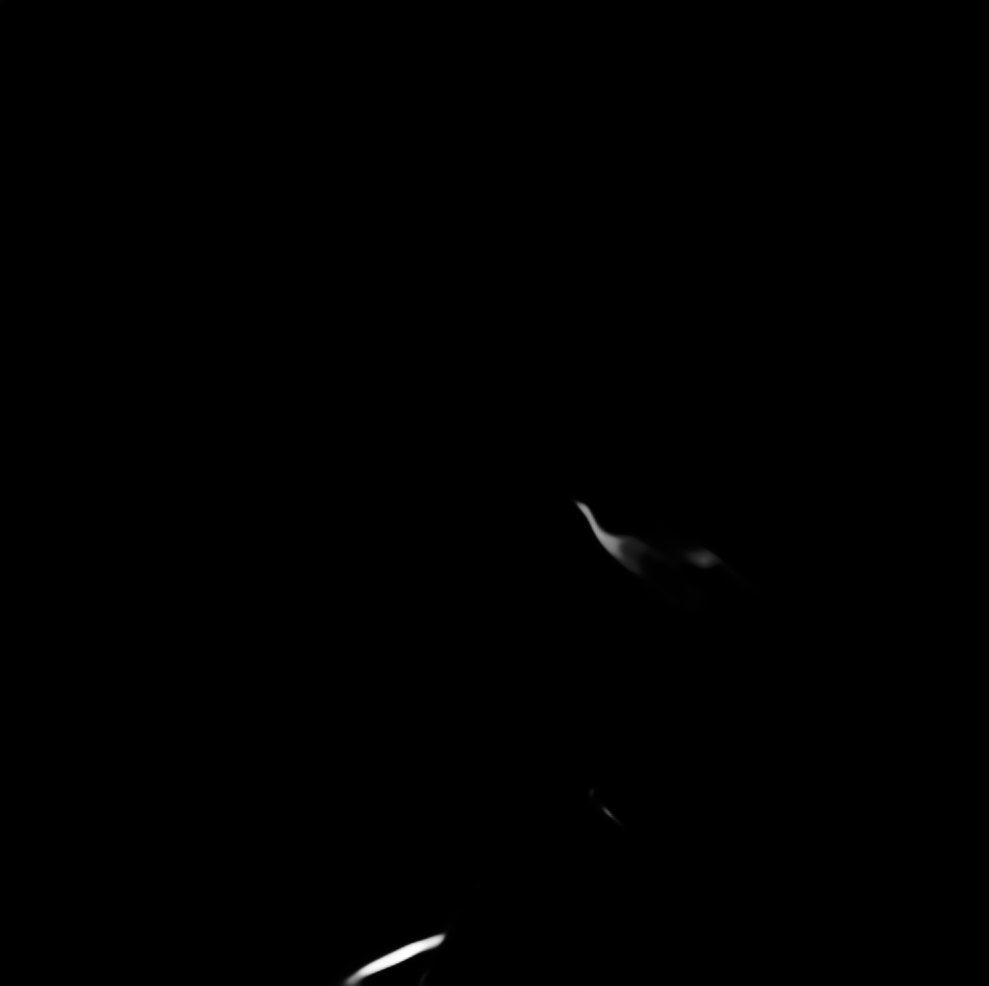}}&
   {\includegraphics[width=0.153\linewidth]{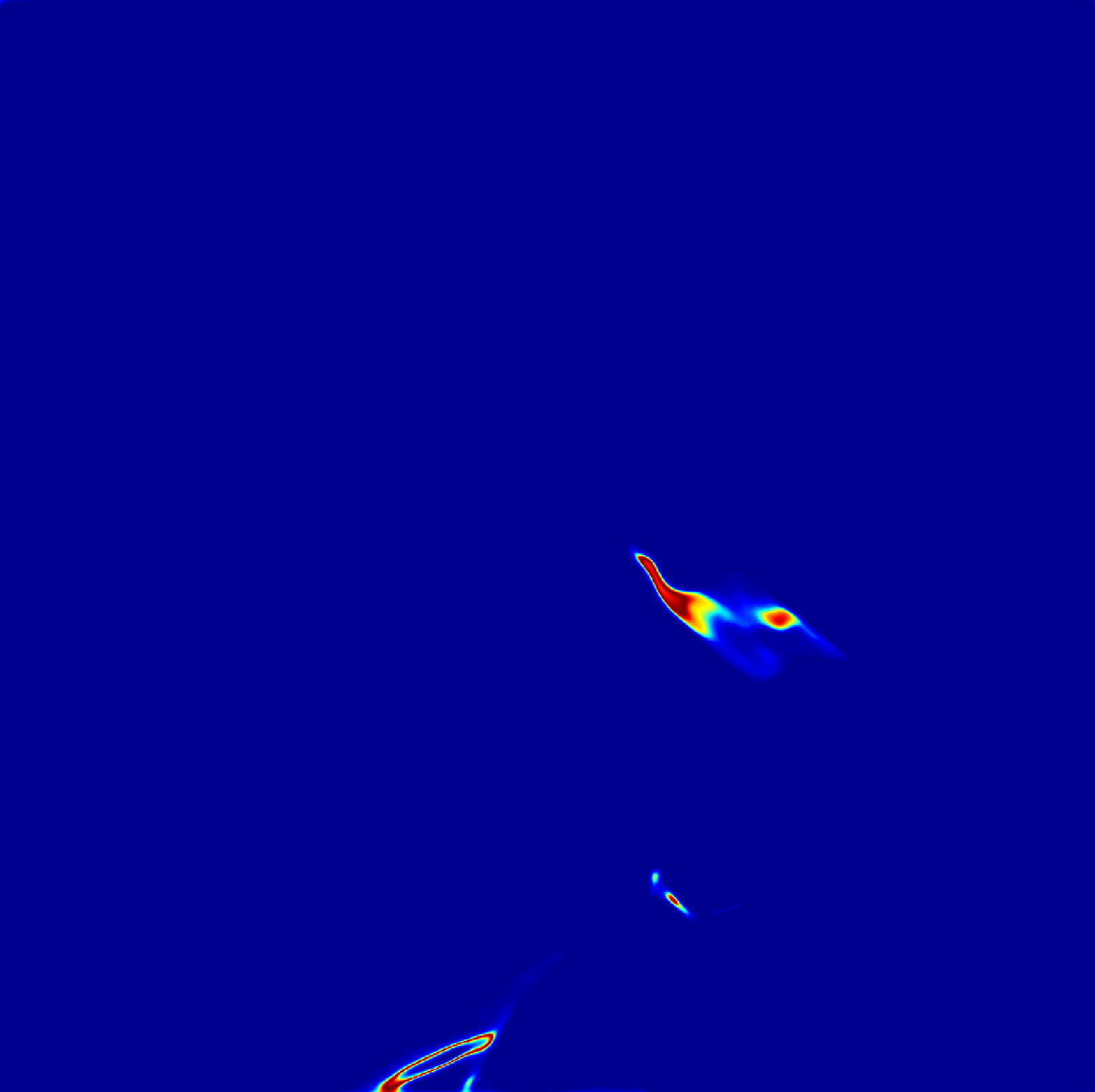}}&
   {\includegraphics[width=0.153\linewidth]{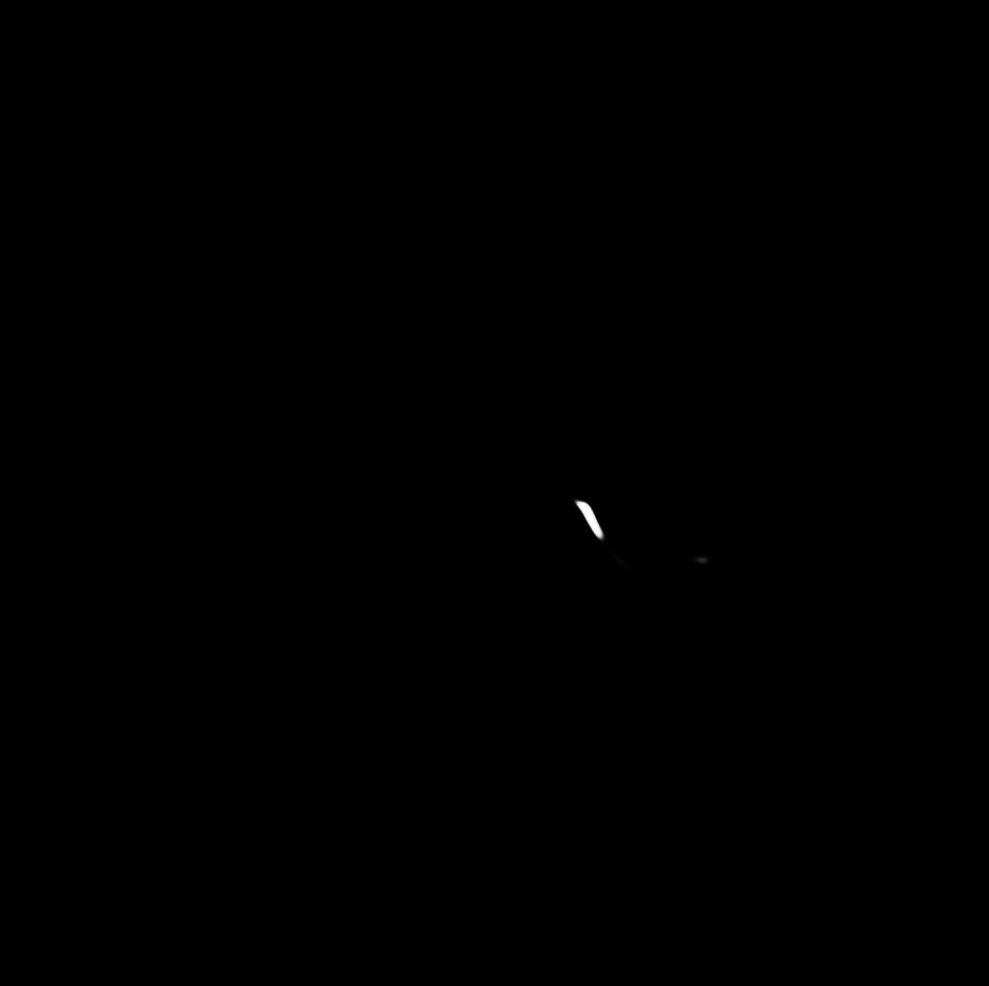}}&
   {\includegraphics[width=0.153\linewidth]{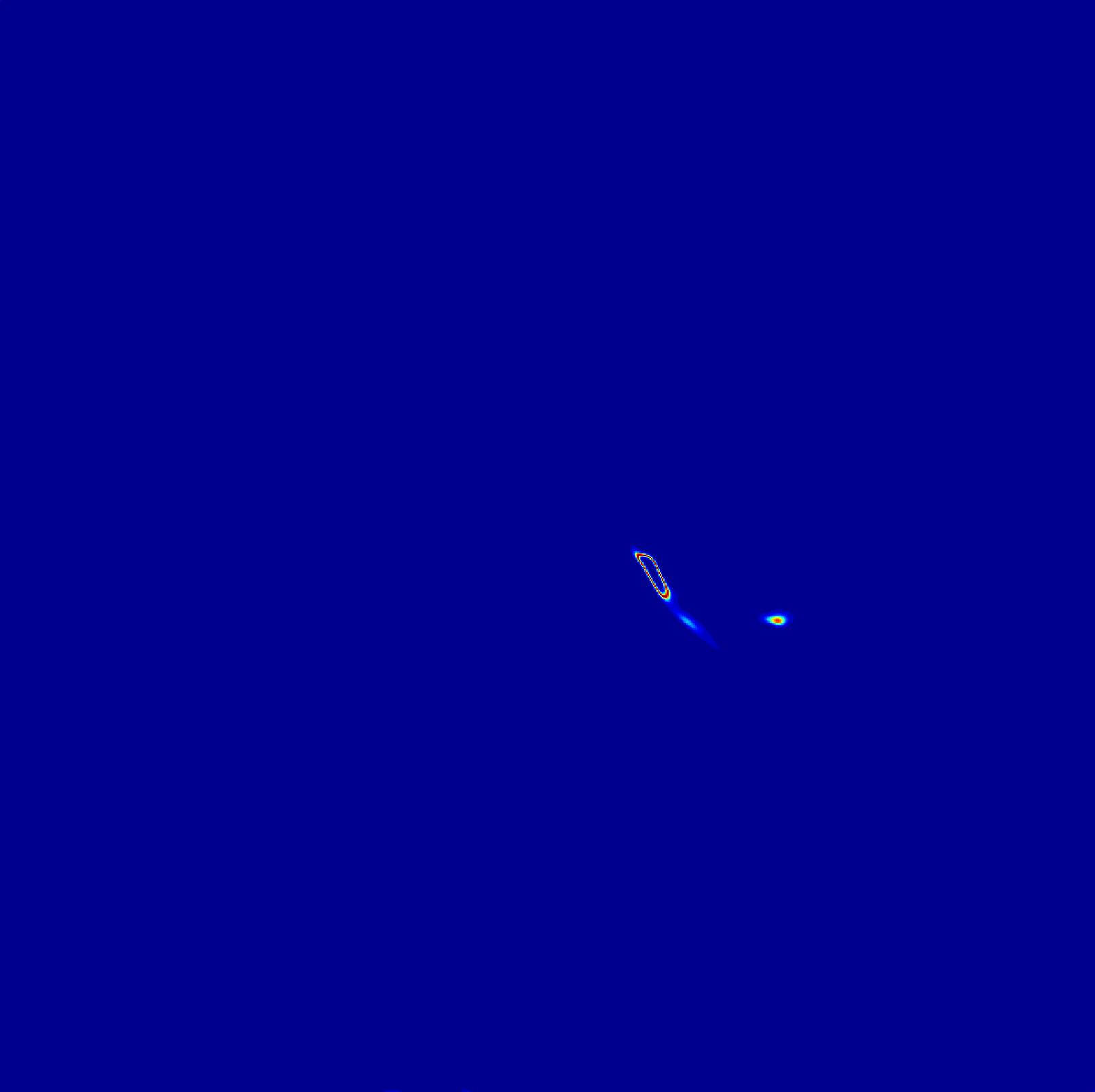}}\\
   {\includegraphics[width=0.153\linewidth]{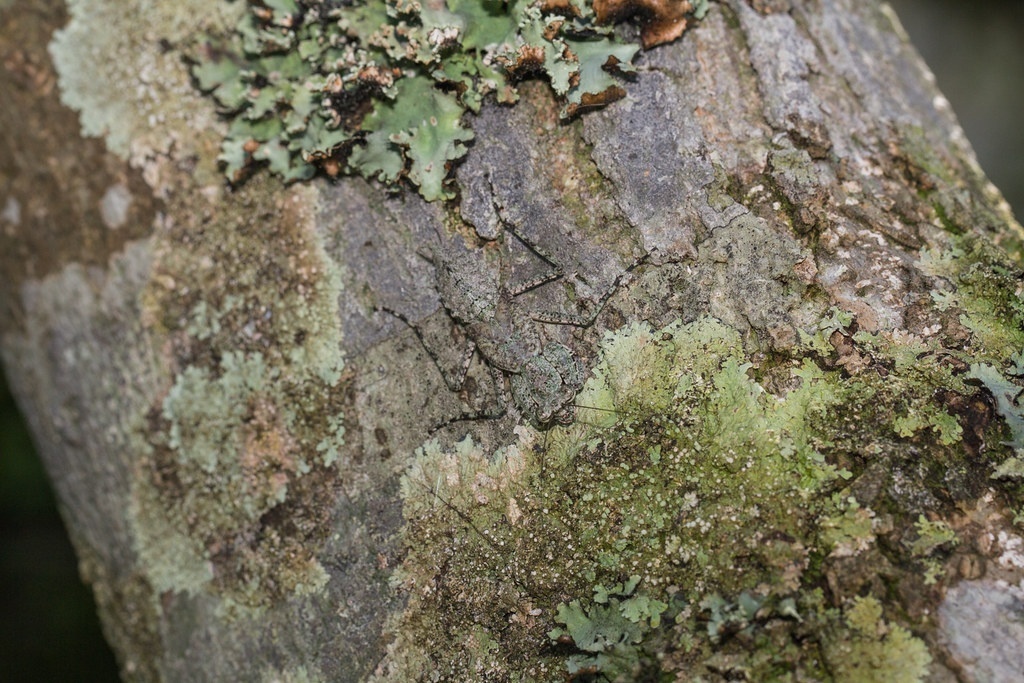}} &
   {\includegraphics[width=0.153\linewidth]{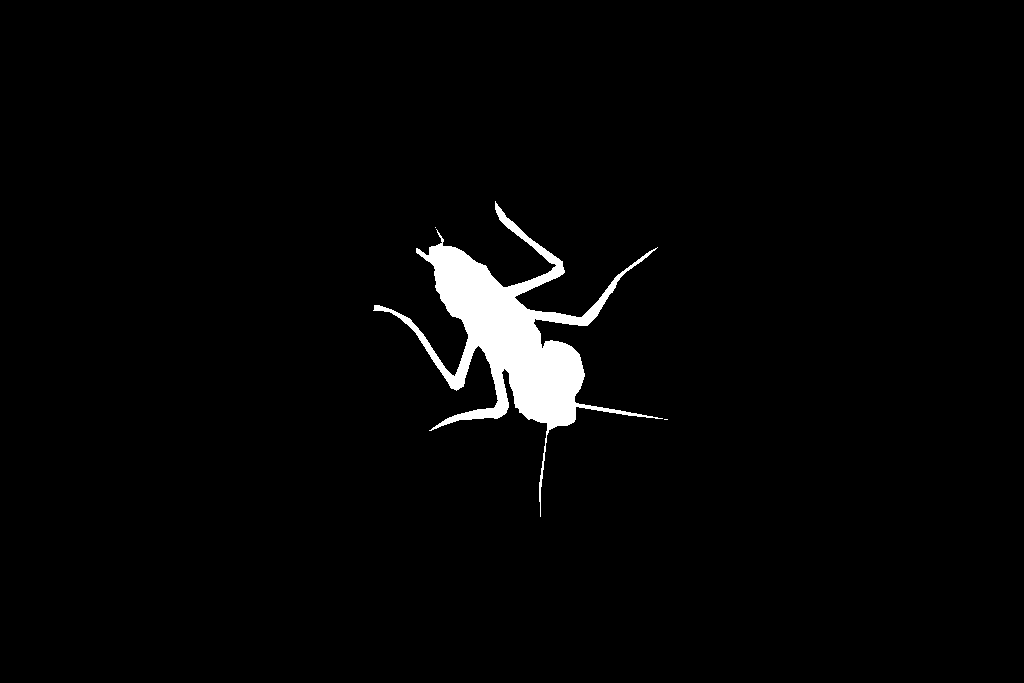}} &
   {\includegraphics[width=0.153\linewidth]{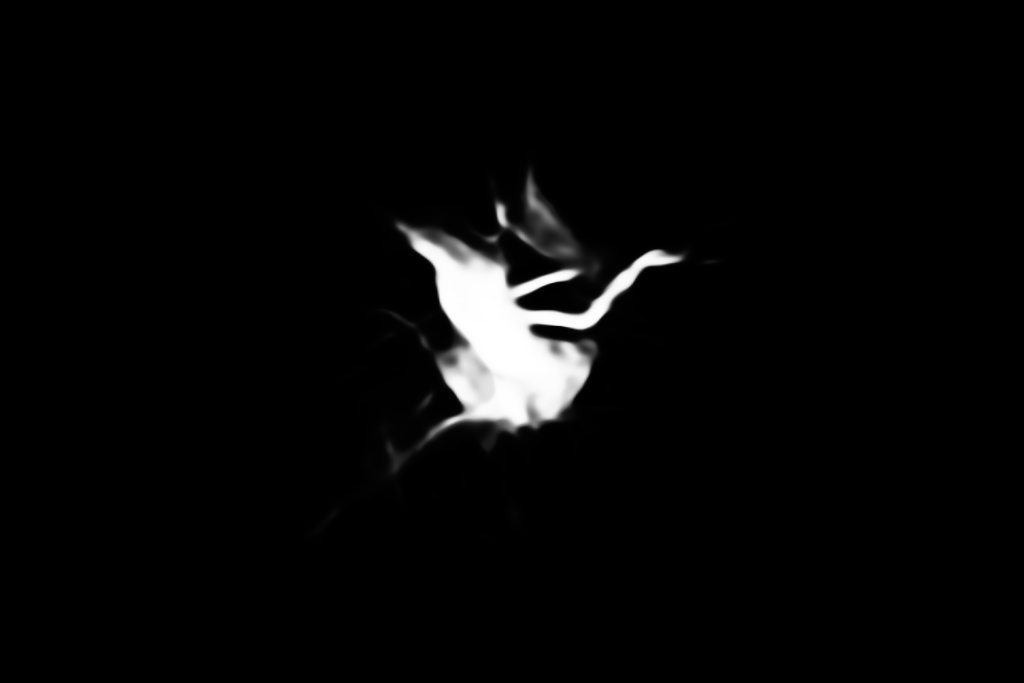}}&
   {\includegraphics[width=0.153\linewidth]{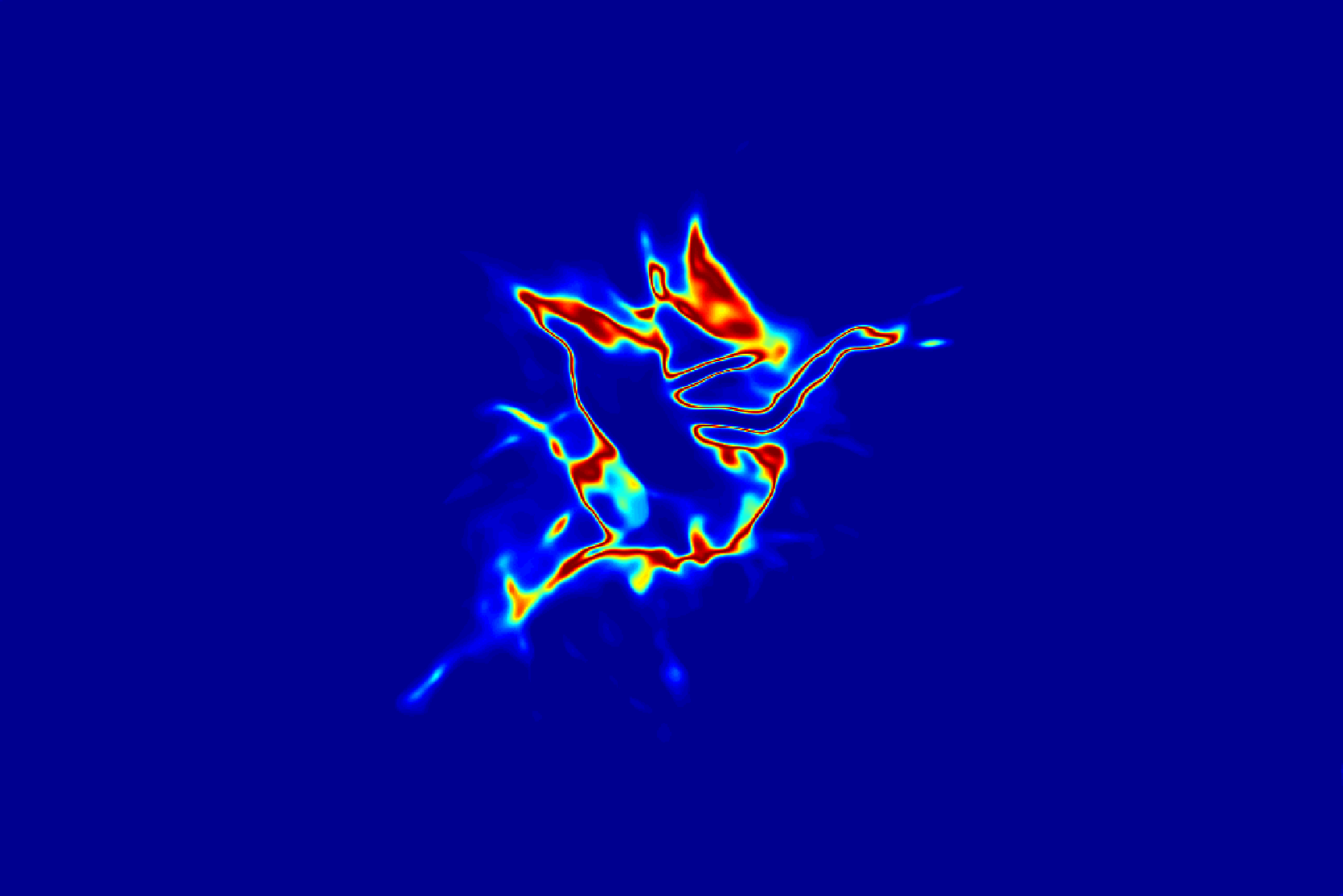}}&
   {\includegraphics[width=0.153\linewidth]{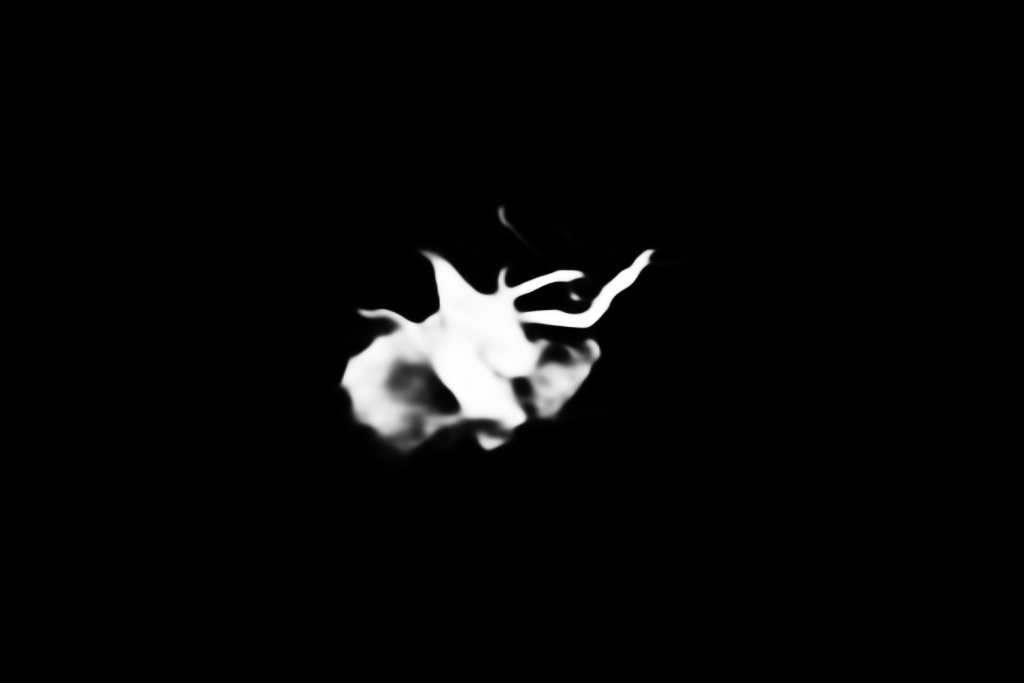}}&
   {\includegraphics[width=0.153\linewidth]{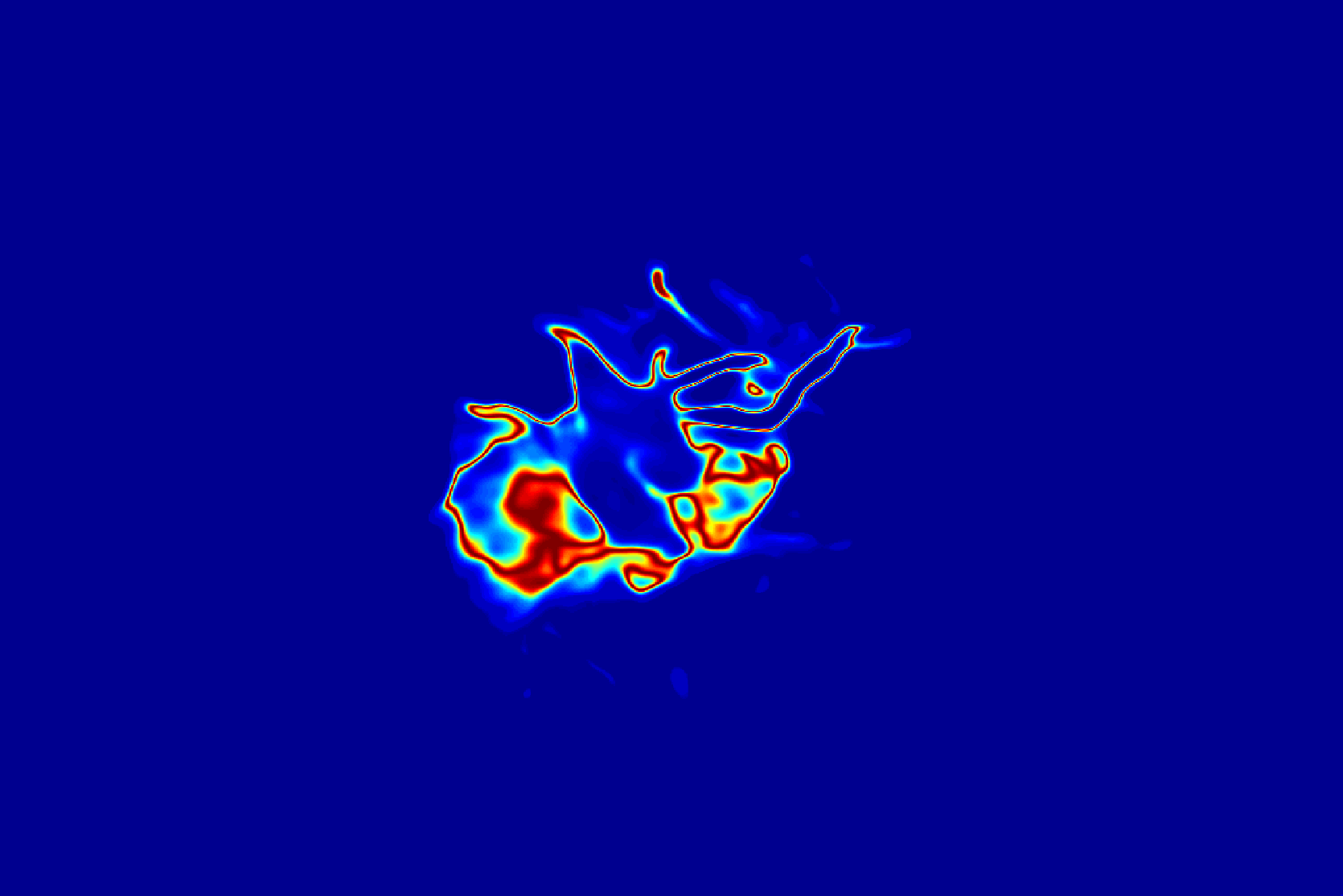}}\\
   {\includegraphics[width=0.153\linewidth]{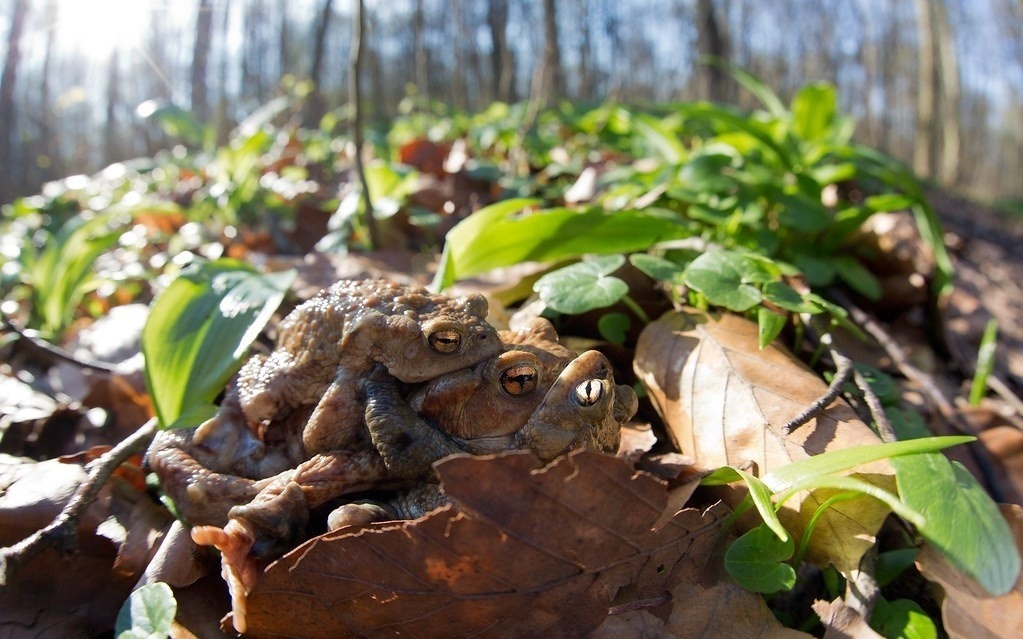}} &
   {\includegraphics[width=0.153\linewidth]{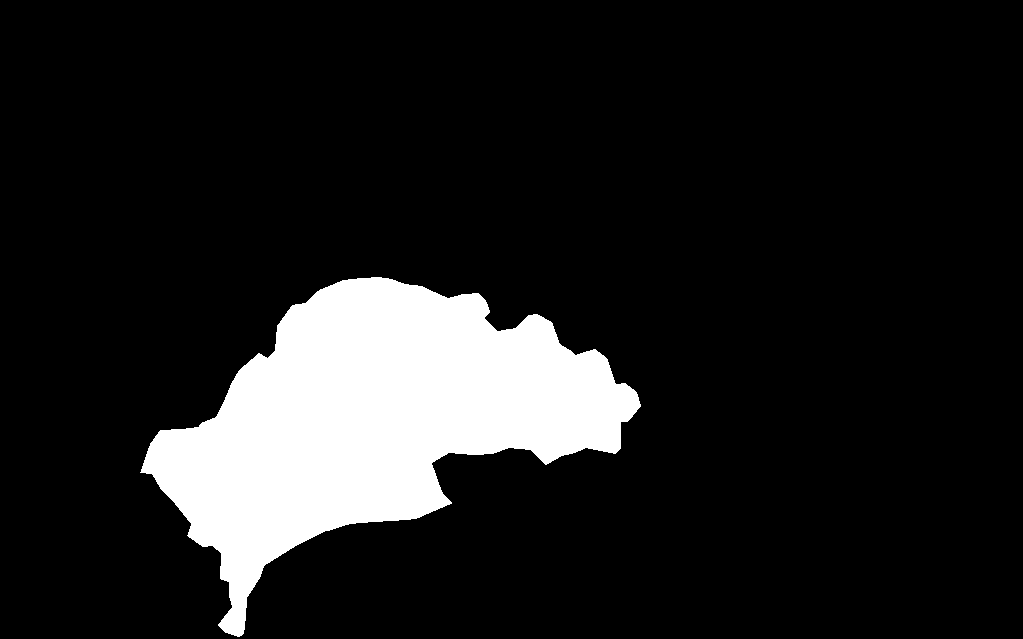}} &
   {\includegraphics[width=0.153\linewidth]{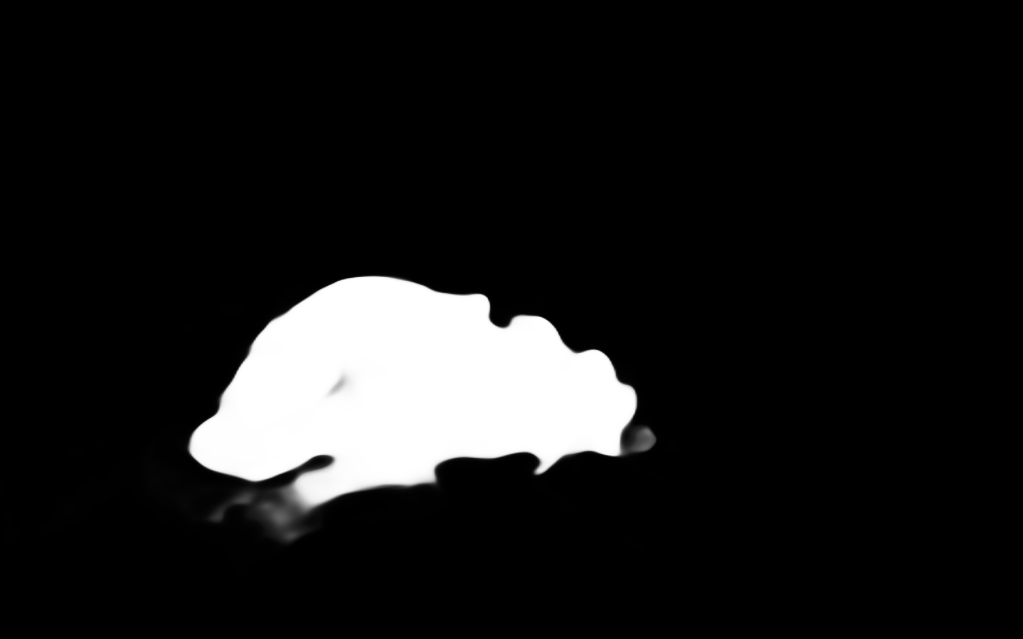}}&
   {\includegraphics[width=0.153\linewidth]{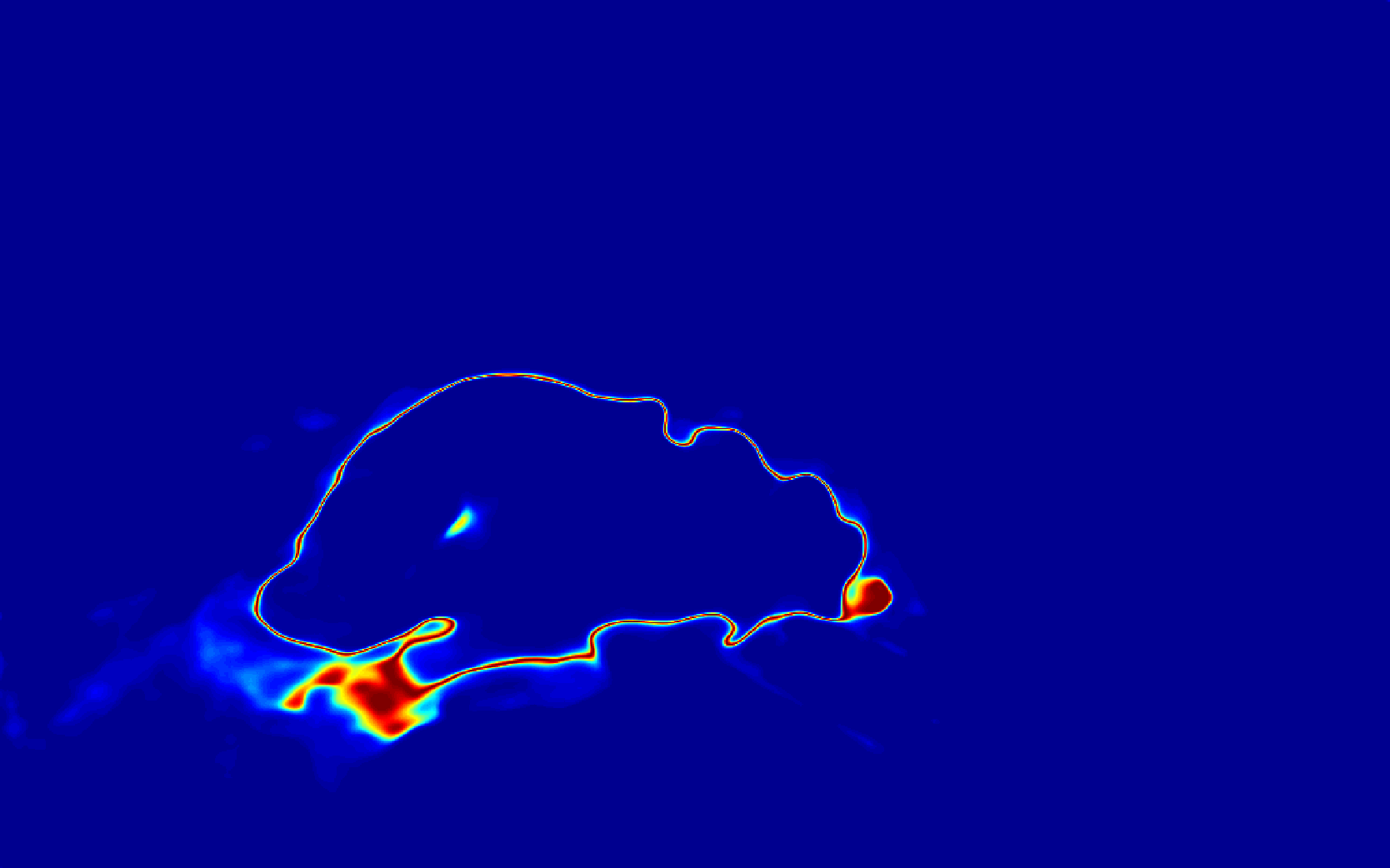}}&
   {\includegraphics[width=0.153\linewidth]{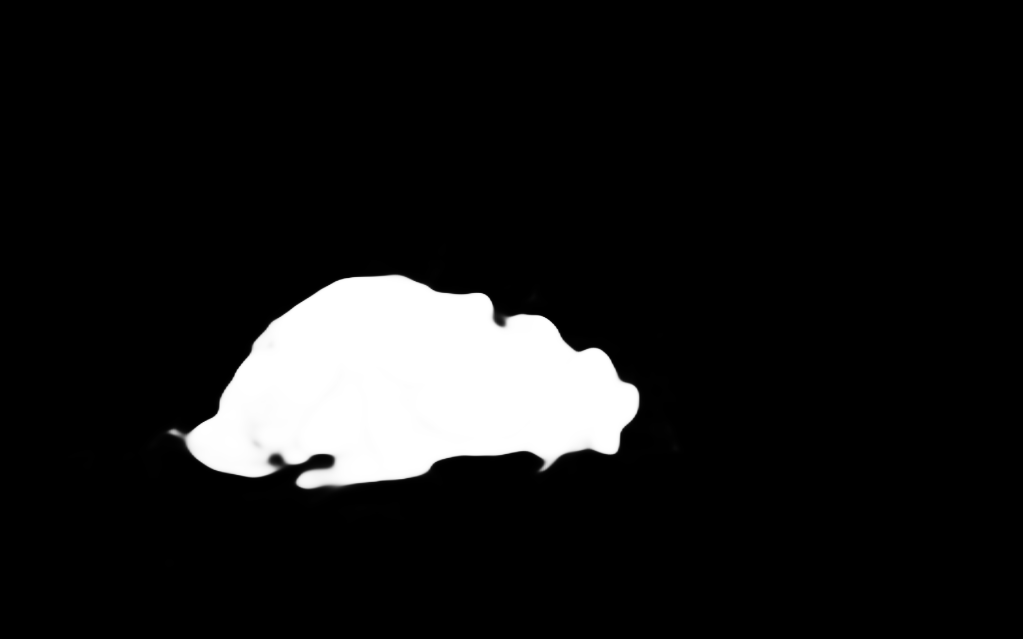}}&
   {\includegraphics[width=0.153\linewidth]{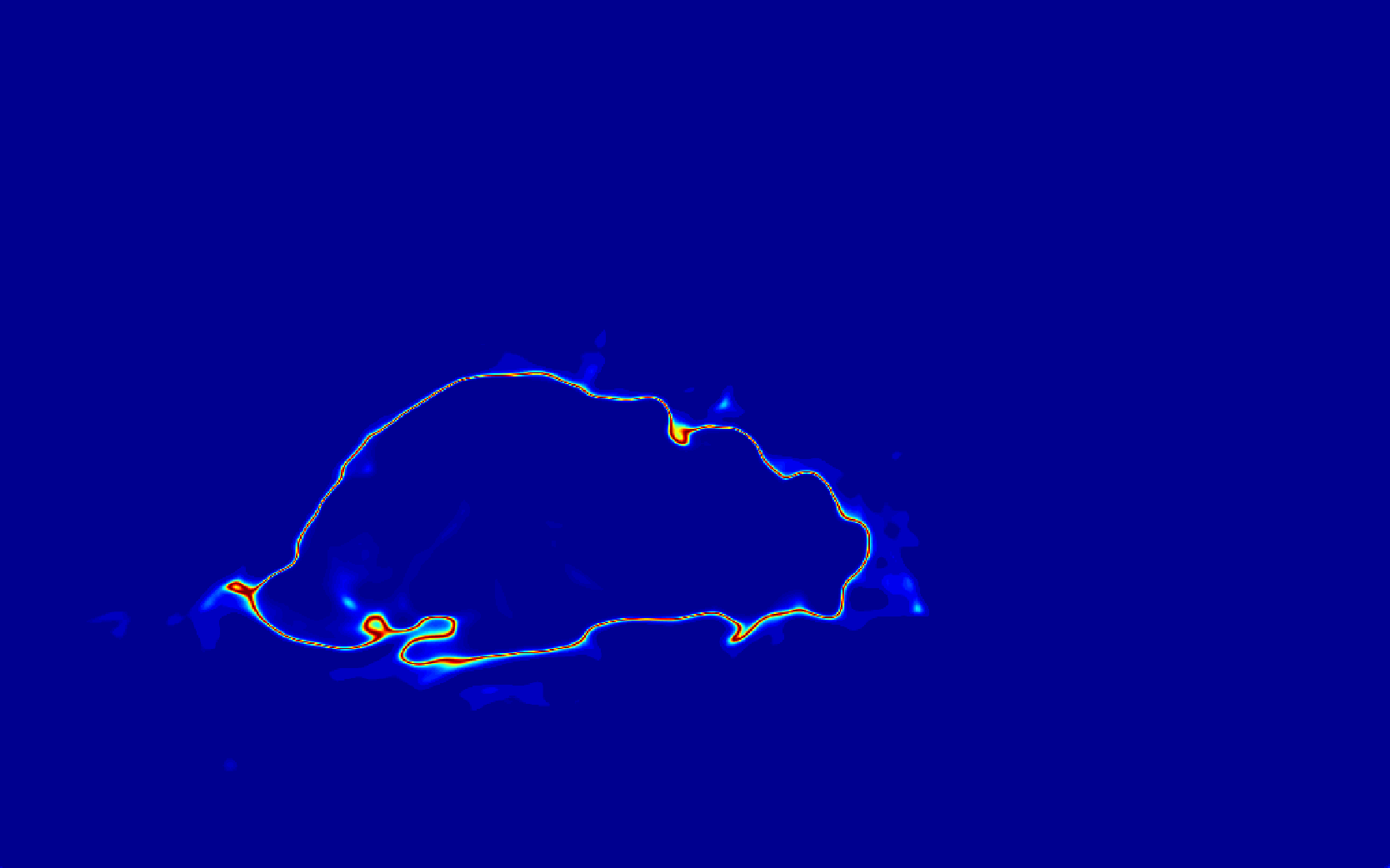}}\\
   \footnotesize{Image}&\footnotesize{GT}&\footnotesize{MD}&\footnotesize{$U_p$}&\footnotesize{CGAN}&\footnotesize{$U_p$}\\
   \end{tabular}
   \end{center}
   \caption{\footnotesize{Predictive uncertainty maps comparison between the MC-dropout based solution and CGAN based solution
   for \textbf{camouflaged object detection}.}
   }
\label{fig:three_uncertainty_cod}
\end{figure}

Further, we show uncertainty maps of the generative model based methods for both camouflaged object detection and salient object detection in Fig.~\ref{fig:predictive_generative_cod}, Fig.~\ref{fig:aleatoric_generative_cod}, Fig.~\ref{fig:epistemic_generative_cod}, Fig.~\ref{fig:predictive_generative_sod}, Fig.~\ref{fig:aleatoric_generative_sod} and Fig.~\ref{fig:epistemic_generative_sod}. Although the ABP \cite{ABP_aaai} and EBM \cite{LeCun06atutorial,coopnets} based solutions performs worse than the GAN \cite{NIPS2014_5423_gan} based model for deterministic performance evaluation as shown in Table \ref{tab:ablation_cod_generative} and Table \ref{tab:ablation_sod_generative}, the ABP \cite{ABP_aaai} and EBM \cite{LeCun06atutorial,coopnets} based solutions can generate better uncertainty map. The main reason is ABP samples from the true latent variable posterior distribution, and EBM samples from the true predictive distribution, making them suffer less from the posterior collapse issue \cite{he2018lagging} as VAE \cite{VAE1} or GAN \cite{NIPS2014_5423_gan} based framework.

\subsubsection{Ensemble vs Generative Model}
\begin{figure}[tp]
   \begin{center}
   \begin{tabular}{c@{ }c@{ }c@{ }c@{ }c@{ }c@{ }}
   {\includegraphics[width=0.153\linewidth]{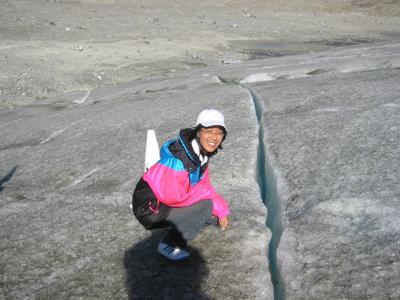}} &
   {\includegraphics[width=0.153\linewidth]{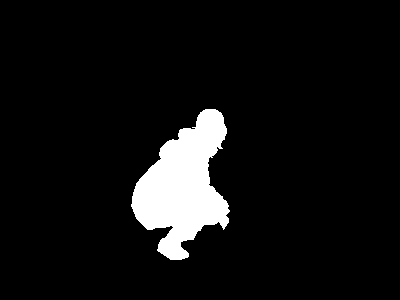}} &
   {\includegraphics[width=0.153\linewidth]{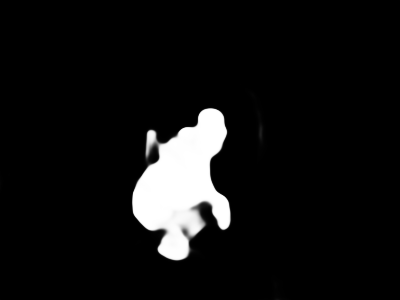}}&
   {\includegraphics[width=0.153\linewidth]{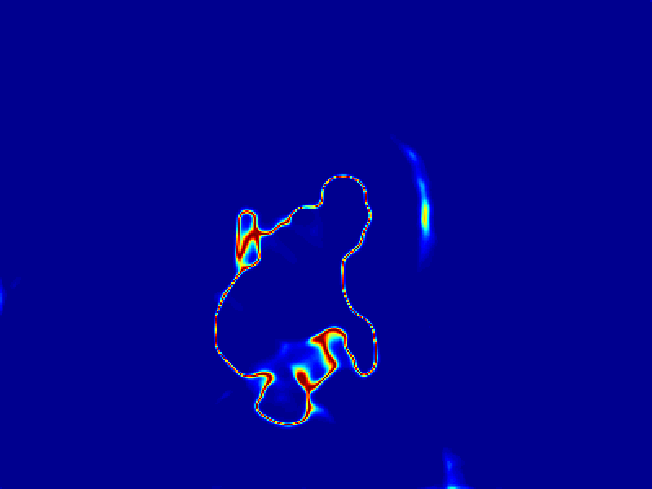}}&
   {\includegraphics[width=0.153\linewidth]{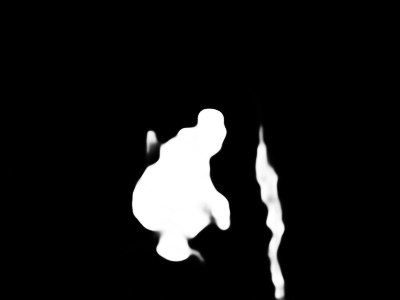}}&
   {\includegraphics[width=0.153\linewidth]{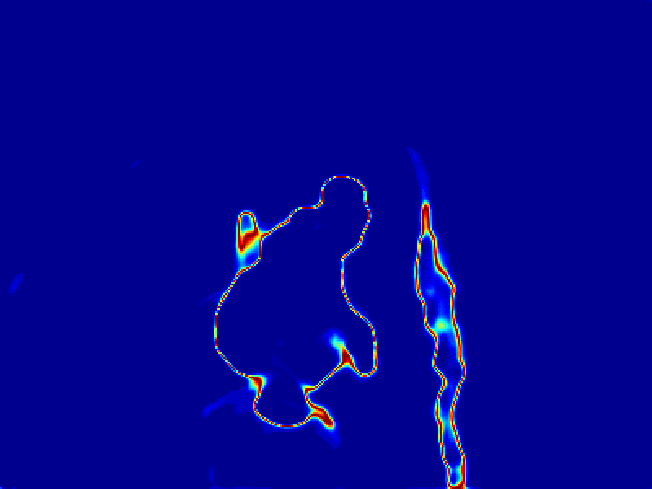}} \\
   {\includegraphics[width=0.153\linewidth]{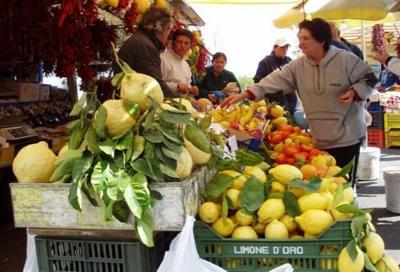}} &
   {\includegraphics[width=0.153\linewidth]{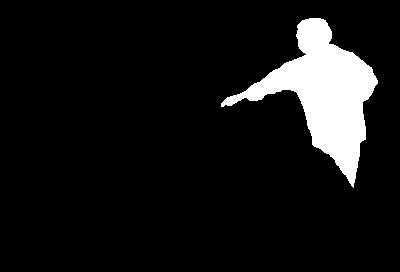}} &
   {\includegraphics[width=0.153\linewidth]{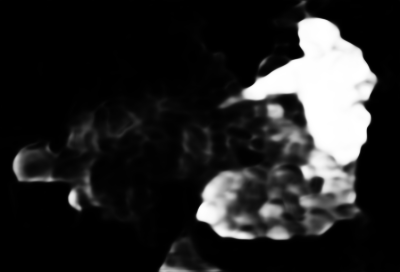}}&
   {\includegraphics[width=0.153\linewidth]{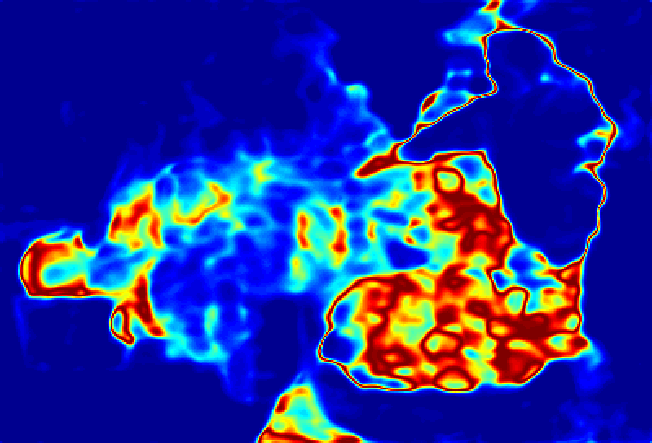}}&
   {\includegraphics[width=0.153\linewidth]{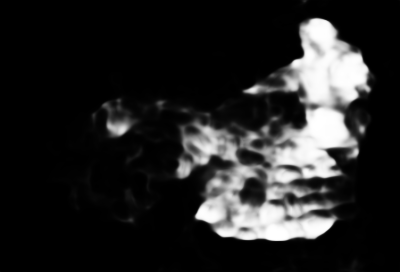}}&
   {\includegraphics[width=0.153\linewidth]{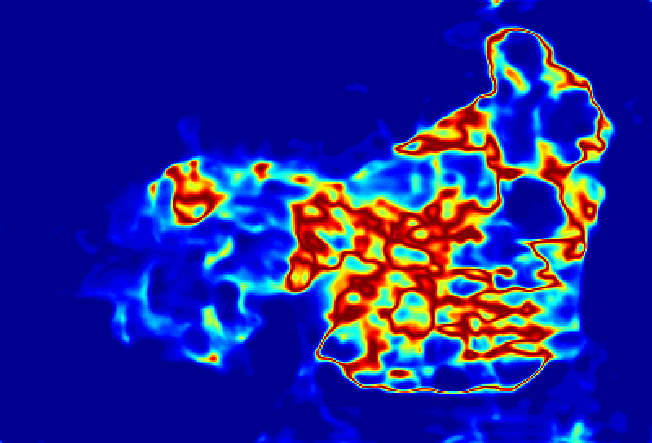}} \\
   {\includegraphics[width=0.153\linewidth]{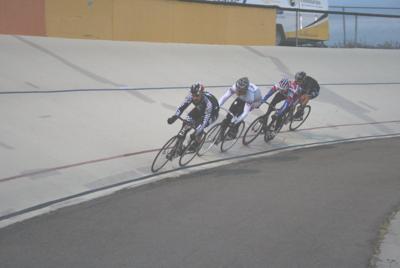}} &
   {\includegraphics[width=0.153\linewidth]{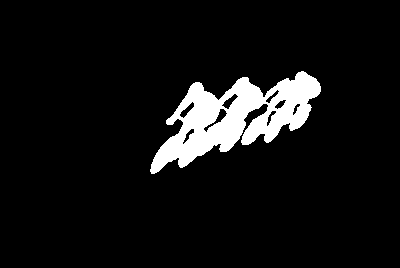}} &
   {\includegraphics[width=0.153\linewidth]{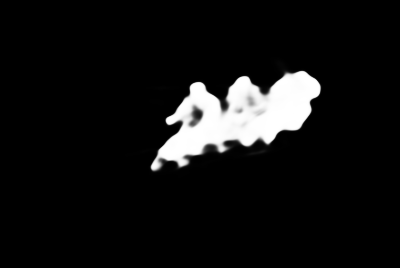}}&
   {\includegraphics[width=0.153\linewidth]{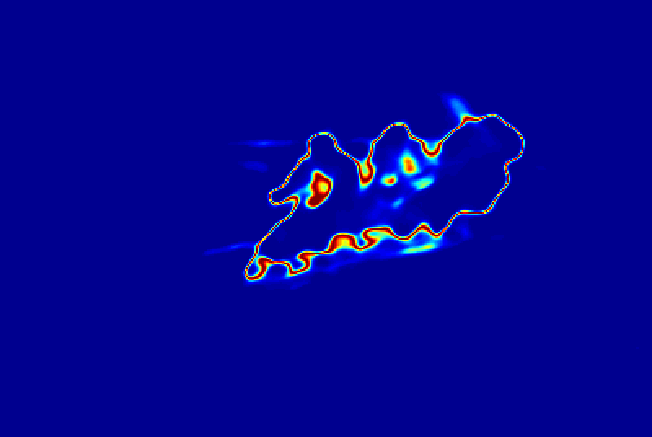}} &
   {\includegraphics[width=0.153\linewidth]{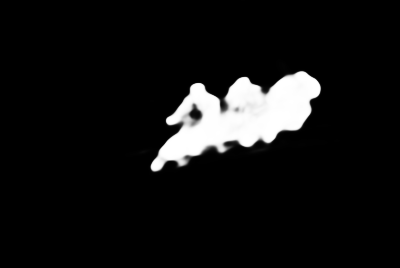}}&
   {\includegraphics[width=0.153\linewidth]{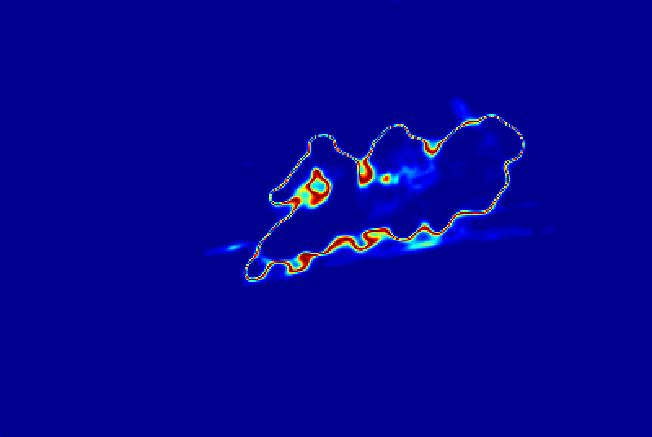}} \\
   \footnotesize{Image}&\footnotesize{GT}&\footnotesize{MD}&\footnotesize{$U_p$}&\footnotesize{CGAN}&\footnotesize{$U_p$}\\
   \end{tabular}
   \end{center}
   \caption{\footnotesize{Predictive uncertainty maps comparison between the MC-dropout based solution and CGAN based solution
   for \textbf{salient object detection}.}
   }
\label{fig:three_uncertainty_sod}
\end{figure}

We further compare the predictive uncertainty of the ensemble based uncertainty estimation method, MC-dropout \cite{Gal2016Dropout} in particular, with the generative model based uncertainty estimation method (CGAN), and show uncertainty maps for the camouflaged object detection and salient object detection tasks in Fig.~\ref{fig:three_uncertainty_cod} and Fig.~\ref{fig:three_uncertainty_sod} respectively. As the two types of uncertainty estimation techniques focus on different origins of uncertainties, we only show the predictive uncertainty to explain how the model can be aware about it's prediction when it makes mistakes.

\subsubsection{Bayesian latent variable based uncertainty estimation}
We discussed the Bayesian latent variable model in Section \ref{blvm_sec}, and present the Bayesian latent variable model for uncertainty estimation in Section \ref{implementation_details_sec}. We then propose BCVAE, BGAN, BABP and BEBM, representing the Bayesian conditional variational auto-encoder, Bayesian generative adversarial network, Bayesian alternating back-propagation and Bayesian energy-based model respectively. As an extension, we also introduce EBM as refinement and attach it to BCVAE, BAGN and BABP and achieve ECVAE, EGAN and EABP with several steps of Langevin sampling, where the prediction from BCVAE, BGAN or BABP serves as start point (warm start) for the prediction Langevin sampling as in Eq.~\ref{equ:ebm_Langevin}.

\begin{table}[t!]
  \centering
  \scriptsize
  \renewcommand{\arraystretch}{1.2}
  \renewcommand{\tabcolsep}{1.55mm}
  \caption{Generative model based solutions for \textbf{camouflaged object detection}. $\uparrow$ indicates the higher the score the better, and vice versa for $\downarrow$.}
  \begin{tabular}{l|cc|cc|cc|cc}
  \toprule
   Method &\multicolumn{2}{c|}{CAMO~\cite{le2019anabranch}}&\multicolumn{2}{c|}{CHAMELEON~\cite{Chameleon2018}}&\multicolumn{2}{c|}{COD10K~\cite{fan2020camouflaged}}&\multicolumn{2}{c}{NC4K~\cite{yunqiu2021ranking}} \\
    &$F_{\beta}\uparrow$&$\mathcal{M}\downarrow$&$F_{\beta}\uparrow$&$\mathcal{M}\downarrow$
    &$F_{\beta}\uparrow$&$\mathcal{M}\downarrow$
    &$F_{\beta}\uparrow$&$\mathcal{M}\downarrow$\\
  \hline
  Base & .757 & .079 & .848 & .029 & .731 & .035 & .803 & .048 \\
  CVAE & .758 & .081 & .848 & .030 & .731 & .034 & .802 & .048 \\
  CGAN & .762 & .080 & .852 & .026 & .730 & .034 & .807 & .048 \\
  ABP & .756 & .081 & .846 & .030 & .729 & .034 & .801 & .047 \\
  EBM & .777 & .076 & .844 & .031 & .721 & .038 & .796 & .050 \\
   \bottomrule
  \end{tabular}
  \label{tab:ablation_cod_generative}
\end{table}

\begin{table}[t!]
  \centering
  \scriptsize
  \renewcommand{\arraystretch}{1.2}
  \renewcommand{\tabcolsep}{1.55mm}
  \caption{Generative model based solutions for \textbf{salient object detection}, $\uparrow$ indicates the higher the score the better, and vice versa for $\downarrow$.}
  \begin{tabular}{l|cc|cc|cc|cc}
  \toprule
   Method &\multicolumn{2}{c|}{DUTS~\cite{imagesaliency}}&\multicolumn{2}{c|}{DUT~\cite{Manifold-Ranking:CVPR-2013}}&\multicolumn{2}{c|}{HKU-IS~\cite{MDF:CVPR-2015}}&\multicolumn{2}{c}{PASCAL~\cite{pascal_s_dataset}} \\
    &$F_{\beta}\uparrow$&$\mathcal{M}\downarrow$&$F_{\beta}\uparrow$&$\mathcal{M}\downarrow$
    &$F_{\beta}\uparrow$&$\mathcal{M}\downarrow$
    &$F_{\beta}\uparrow$&$\mathcal{M}\downarrow$\\
  \hline
   Base & .842 & .037 & .760 & .055 & .904 & .030 & .828 & .064 \\
  CVAE & .836 & .037 & .748 & .055 & .901 & .030 & .826 & .063 \\
  CGAN & .846 & .035 & .752 & .054 & .905 & .029 & .828 & .063 \\
  ABP & .829 & .040 & .740 & .059 & .889 & .034 & .818 & .068 \\
  EBM & .834 & .040 & .744 & .062 & .900 & .031 & .829 & .064 \\
   \bottomrule
  \end{tabular}
  \label{tab:ablation_sod_generative}
\end{table}

\section{Discussion}
Due to the close relationship of uncertainty estimation and model calibration, we discuss how the uncertainty estimation techniques can be useful for model calibration. Further, as an effective measure of model calibration, we explain how model performance on out-of-distribution detection can be an effective measure for uncertainty estimation. We then discuss the effective strategy of applying uncertainty estimation for model calibration by measuring it's performance on out-of-distribution samples.

\subsection{Uncertainty Estimation and Model Calibration}
\cite{improve_model_calibration_nips2020},
\cite{Guo2017OnCO} discovered that modern deep neural networks are prone to producing overconfident predictions. \cite{zhang_iclr} further proved that a deep neural network can fit random noise with high confidence. The over-confident behavior of current deep neural network is caused by the model's ignorance of what it does not know. A well-calibrated network should produce lower confidence prediction for high uncertain samples. Model calibration (confidence calibration) \cite{bayesian_calibration} is then defined as the problem of predicting probability estimates representative of the true correctness likelihood \cite{Guo2017OnCO}. 
There mainly exists two different directions for model calibration: 1) relaxing the supervision signal \cite{rethink_inception}; 2) smoothing the network prediction \cite{Guo2017OnCO}. The former one smooths the ground truth with a uniform distribution, while the latter one use temperature scaling to produce soften prediction.
From our perspective, both above strategies can to some extent produce a better calibrated network. The main disadvantage of those strategies is that they treat each sample equally, without considering about image-level or pixel-level uncertainty. Further, the final model calibration is usually obtained with decreased accuracy as sacrifice. We argue that a well-calibrated model should maintain high confidence prediction for confident samples/regions while lower their confidence for uncertain samples/regions. In this way, a pixel-level confidence map is desirable for model calibration to maintain the original model precision.

\subsection{Model Calibration and Out-of-Distribution Detection}

Out-of-distribution (OOD) detection aims to find samples whose distribution is different from the training data distribution \cite{baseline_ood,confi_cal18,open_c18}, and thus can be treated as a technique of modeling epistemic uncertainty \cite{kendall2017uncertainties}. Conventional definition of out-of-distribution samples for image-level classification or semantic segmentation can be summarized as category bias, where images with instances that do not belong to the existing categories are defined as out-of-distribution samples. As OOD highlights the distribution shift, another definition for OOD samples focus on dataset shift \cite{dataset_shift_mit}, where the joint distribution of inputs and outputs differs between the training and testing datasets. In this way, how the model perform on OOD samples can be treated as effective model calibration measure, as our basic assumption is that a well-calibrated model should has highly confident and accurate predictions for in-distribution samples, and less confident predictions for the out-of-distribution samples.

For category-aware segmentation, the OOD samples can be easily generated as samples containing objects of different categories. However, for category-agnostic segmentation, the OOD samples are different from the in-distribution samples in attribute-level. As we are not aware of the true joint data distribution $p(x,y)$, in practice, we can then generate OOD samples by finding samples with different appearance, context, or other attribute-level information. Then, how the model perform on these OOD samples can serve as effective measure for it's calibration level.


\subsection{Uncertainty Estimation and Out-of-Distribution Detection}
Existing deep model calibration mainly focus on image-level classification problems \cite{Guo2017OnCO,relaxed_softmax,pmlr-v80-kumar18a,rethink_inception,disturblabel} without considering sample-level or pixel-level uncertainty. We work on combining uncertainty estimation with model calibration to generate reasonable prediction with reliable uncertainty map for out-of-distribution samples.
The first step for our task is to estimate pixel-wise uncertainty with either the deep ensemble solutions \cite{Gal2016Dropout,emsemble,Durasov21_maskensemble, simple_scalable_uncertainty,ensemble_accurate_uncertainty,reverse_kl,Malinin2020Ensemble,hyperparameter_ensemble} or generative model solutions \cite{NIPS2014_5423_gan,VAE1,cvae,ABP_aaai,LeCun06atutorial,coopnets}.
The second step is to achieve model calibration with the produced uncertainty as guidance, which can be treated as a post-hoc technique or designed as online learning strategy.
The final step will be combining above two steps in a unified framework to achieve dense model calibration based on uncertainty estimation and evaluate model performance on out-of-distribution samples.

\section{Conclusion} 
In this paper, we investigate uncertainty estimation, especially ensemble based solutions for epistemic uncertainty estimation and generative model based solutions for aleatoric uncertainty estimation. We provide a detailed introduction and analysis for each solution along
with implementation details. We claim that uncertainty estimation is necessary for any machine learning model to explain its prediction, thus avoid over-confident predictions.

\ifCLASSOPTIONcaptionsoff
  \newpage
\fi



%


\bibliographystyle{ieeetr}
\bibliography{reference_dense}

\end{document}